\documentclass[10pt,onecolumn]{article}
\usepackage{graphicx,amssymb,amsmath,amsthm}
\usepackage[hypertexnames=false]{hyperref}
\usepackage{longtable}
\usepackage[margin=1in]{geometry}
\usepackage{array}
\usepackage[noend]{algpseudocode}
\usepackage{algorithm,algcompatible}
\usepackage{cite}
\usepackage{xcolor}
\usepackage{setspace}
\usepackage{wrapfig}
\usepackage{enumitem}
\usepackage{booktabs}
\usepackage{fancyhdr}
\usepackage{varwidth}
\usepackage[most]{tcolorbox}
\tcbuselibrary{skins,breakable,theorems}
\tcbuselibrary{hooks}

\everymath{\displaystyle}

\newcommand{\centryproposition}[2]{%
    \refstepcounter{proposition}\label{#1}%
    \paragraph{Proposition~\theproposition#2.}%
}

\definecolor{assumptioncolor}{RGB}{70,130,180}    
\definecolor{propositioncolor}{RGB}{232,238,236}   
\definecolor{theoremcolor}{RGB}{46,125,50}        
\definecolor{corollarycolor}{RGB}{156,39,176}     
\definecolor{lemmacolor}{RGB}{116,78,176}     
\definecolor{assumpbg}{RGB}{242,248,254}          
\definecolor{propobg}{RGB}{248,242,246}          
\definecolor{theombg}{RGB}{245,251,246}           
\definecolor{corollbg}{RGB}{251,245,254}          
\definecolor{lemmabg}{RGB}{254,254,238}          
\definecolor{darktext}{RGB}{30,30,30}             

\newtcbtheorem[auto counter]{styledassumption}{Assumption}{
    enhanced,
    breakable,
    varwidth boxed title*=-10mm,
    colback=assumpbg,
    colframe=assumptioncolor!80!black,
    colbacktitle=assumpbg,
    coltitle=darktext,
    fonttitle=\bfseries\large,
    attach boxed title to top left={yshift=-2mm, xshift=3mm},
    boxed title style={
        sharp corners,
        boxrule=1.2pt,
        colframe=assumptioncolor!80!black,
        colback=assumpbg,
        left=2mm, right=2mm, top=1.5mm, bottom=1.5mm
    },
    top=4mm, bottom=4mm, left=4mm, right=4mm,
    arc=1.5mm,
    before upper=\color{black}
}{ass}

\newtcbtheorem[auto counter]{styledproposition}{Proposition}{
    enhanced,
    breakable,
    varwidth boxed title*=-10mm,
    colback=propobg,
    colframe=propositioncolor!80!black,
    colbacktitle=propobg,
    coltitle=darktext,
    fonttitle=\bfseries\large,
    attach boxed title to top left={yshift=-2mm, xshift=3mm},
    boxed title style={
        sharp corners,
        boxrule=1.2pt,
        colframe=propositioncolor!80!black,
        colback=propobg,
        left=2mm, right=2mm, top=1.5mm, bottom=1.5mm
    },
    top=4mm, bottom=4mm, left=4mm, right=4mm,
    arc=1.5mm,
    before upper=\color{black}
}{pro}

\newtcbtheorem[auto counter]{styledtheorem}{Theorem}{
    enhanced,
    breakable,
    varwidth boxed title*=-10mm,
    colback=theombg,
    colframe=theoremcolor!80!black,
    colbacktitle=theombg,
    coltitle=darktext,
    fonttitle=\bfseries\large,
    attach boxed title to top left={yshift=-2mm, xshift=3mm},
    boxed title style={
        sharp corners,
        boxrule=1.2pt,
        colframe=theoremcolor!80!black,
        colback=theombg,
        left=2mm, right=2mm, top=1.5mm, bottom=1.5mm
    },
    top=4mm, bottom=4mm, left=4mm, right=4mm,
    arc=1.5mm,
    before upper=\color{black}
}{thm}

\newtcbtheorem[auto counter]{styledcorollary}{Corollary}{
    enhanced,
    breakable,
    varwidth boxed title*=-10mm,
    colback=corollbg,
    colframe=corollarycolor!80!black,
    colbacktitle=corollbg,
    coltitle=darktext,
    fonttitle=\bfseries\large,
    attach boxed title to top left={yshift=-2mm, xshift=3mm},
    boxed title style={
        sharp corners,
        boxrule=1.2pt,
        colframe=corollarycolor!80!black,
        colback=corollbg,
        left=2mm, right=2mm, top=1.5mm, bottom=1.5mm
    },
    top=4mm, bottom=4mm, left=4mm, right=4mm,
    arc=1.5mm,
    before upper=\color{black}
}{cor}

\newtcbtheorem[auto counter]{styledlemma}{Lemma}{
    enhanced,
    breakable,
    varwidth boxed title*=-10mm,
    colback=lemmabg,
    colframe=lemmacolor!80!black,
    colbacktitle=lemmabg,
    coltitle=darktext,
    fonttitle=\bfseries\large,
    attach boxed title to top left={yshift=-2mm, xshift=3mm},
    boxed title style={
        sharp corners,
        boxrule=1.2pt,
        colframe=lemmacolor!80!black,
        colback=lemmabg,
        left=2mm, right=2mm, top=1.5mm, bottom=1.5mm
    },
    top=4mm, bottom=4mm, left=4mm, right=4mm,
    arc=1.5mm,
    before upper=\color{black}
}{lem}

\title{CENTRY: A Probabilistic Latent-State Model for Continual AI Risk Estimation, Monitoring, and Regulation}

\author{Bonny Banerjee$^{1,}$\thanks{Corresponding author: bonnybanerjee@yahoo.com} ~~~~~~ Geetha Raju$^2$\\
\\
$^1$Independent Scholar, USA\\ 
$^2$Centre for Responsible AI, Indian Institute of Technology Madras, Chennai, India\\
}

\date{} 

\begin{document}

\maketitle

\begin{abstract}
\noindent Artificial Intelligence (AI) systems are increasingly deployed in safety-critical applications, yet existing AI safety frameworks primarily define, categorize, or govern AI risk rather than proposing implementable models and methods for continually estimating, monitoring, and regulating risk throughout an AI system's operational lifetime. This paper proposes CENTRY, a probabilistic framework that enables continual risk estimation, monitoring, and regulation of a deployed AI system from its interactions with its environment. The framework models AI risk as arising from three intrinsic sources: datasets, hardware resources, and models/algorithms. A structured latent-variable model is developed in which the latent representation is decomposed into a fast interaction subspace that captures the evolving AI--environment dynamics and three slow latent subspaces that encode the long-term risk states associated with the dataset, hardware, and model components. Rather than requiring a ground-truth risk label at every time step, the component-risk states are treated as latent variables and are learned using a variational objective based on the evidence lower bound, optionally informed during training and continual refinement by evidence-supported behavior--cause records. The resulting framework unifies continual probabilistic inference, online risk estimation, and adaptive risk regulation in an implementable architecture. Under the explicit assumptions stated for the corresponding results, we establish formal guarantees spanning safety, learning, inference, and operations, including finite-horizon and asymptotic risk-budget compliance, regret and convergence bounds for adaptive regulation, consistency and identifiability of component-risk ratings, and detection-delay, false-alarm, containment, and computational-complexity characterizations.
\end{abstract}


\section{Introduction}
\label{sec:Introduction}
Artificial Intelligence (AI) systems are increasingly being deployed in safety-critical and socially consequential domains, including healthcare, transportation, finance, scientific discovery, education, and public services. As these systems become more capable and autonomous, understanding and managing the risks associated with their development and deployment has become an important scientific and societal challenge. Recent studies have emphasized the need for rigorous methods to evaluate and mitigate AI risks before they result in undesirable or harmful behaviors \cite{Bengio2024Science}. However, AI systems continue to evolve after deployment through changing environments, software updates, new datasets, hardware modifications, evolving needs, user requirements, and interactions with humans and other AI systems. Consequently, \textit{AI risk should not be regarded as a static property established during system development, but rather as a dynamic quantity that evolves throughout the AI lifecycle, from problem definition, design, and development through deployment and operation.}

Most existing approaches to AI safety focus on individual aspects of risk management. For example, dataset audits seek to identify harmful or biased training data, formal verification techniques analyze specific properties of learning algorithms, hardware restrictions attempt to limit computational capabilities, and runtime monitoring detects anomalous behaviors after deployment. Although these methods provide valuable tools for mitigating particular sources of AI risk, they generally address isolated stages of the AI lifecycle and are typically applied independently. Moreover, most existing methods either estimate risk, monitor system behavior, or regulate system operation, but do not provide a unified framework that continually performs all three tasks simultaneously. To the best of our knowledge, there currently exists no probabilistic framework that continually estimates, monitors, and regulates AI risk throughout the operational lifetime of an AI system while explicitly maintaining posterior beliefs about the most probable component sources of observed risk.

This paper addresses this problem by proposing \textbf{CENTRY}, a probabilistic latent-state \textbf{\underline{C}}ontinual AI risk \textbf{\underline{E}}stimation, mo\textbf{\underline{N}}i\textbf{\underline{T}}oring and \textbf{\underline{R}}egulation s\textbf{\underline{Y}}stem. CENTRY continually observes the interaction between a monitored AI system and its environment and maintains probabilistic beliefs regarding the risks associated with the system's constituent components. Rather than treating AI risk as a fixed scalar quantity, CENTRY models risk as a collection of slowly evolving latent variables that are continually inferred from interaction dynamics. These latent variables characterize the underlying sources of AI risk, while a separate fast latent state captures the interaction dynamics of the monitored AI system. This structured representation enables CENTRY to distinguish between normal variations in system behavior and persistent changes in the underlying sources of risk.

CENTRY is implemented using the \textbf{SELP} (\textbf{\underline{S}}urprise$\rightarrow$\textbf{\underline{E}}xplain$\rightarrow$\textbf{\underline{L}}earn$\rightarrow$\textbf{\underline{P}}redict) cycle, a continual probabilistic inference framework inspired by predictive coding and active inference paradigms. Within CENTRY, the Predict stage generates the expected observable interaction of the monitored AI system, Surprise measures the discrepancy between predicted and observed interactions, Explain infers latent component-risk states that may account for unexpected behaviors, and Learn continually updates the probabilistic model from accumulated experience. The resulting Bayesian posterior is then used to estimate component-specific AI risk and adaptively regulate the utilization of datasets, computational resources, and models/algorithms toward acceptable risk levels.

In addition to the framework itself, CENTRY is accompanied by formal results that characterize the conditions under which its risk estimation, learning, and regulation mechanisms provide specific safety and operational guarantees.

The key contributions of this paper are as follows:
\begin{enumerate}[nosep]
\item We formulate continual AI risk estimation, monitoring, and regulation as a unified probabilistic latent-state estimation problem and argue that AI risk should be modeled as a dynamic quantity rather than a static property of an AI system.
\item We introduce CENTRY, a probabilistic framework that combines the SELP cycle with structured latent-variable modeling to continually estimate the evolving sources of AI risk from the interactions between an AI system and its environment.
\item We propose a structured latent representation that separates fast interaction dynamics from slowly evolving latent component-risk states corresponding to datasets, hardware resources, and models/algorithms. This separation enables simultaneous prediction of system behavior and continual inference of the underlying sources of AI risk.
\item We develop a Bayesian framework in which component-specific risk ratings are interpreted as decision statistics of the corresponding latent posterior distributions. This formulation naturally supports continual Bayesian updating as new interaction data become available.
\item We introduce a component-prior framework that incorporates domain knowledge regarding datasets, hardware resources, and models/algorithms into Bayesian prior distributions, allowing prior knowledge and observed interaction evidence to be integrated within a unified probabilistic model.
\item We formulate adaptive risk regulation as continual component reconfiguration driven by Bayesian component-risk estimates, with the regulation policy itself continually adapted from interaction evidence, thereby closing the loop between probabilistic inference and AI risk mitigation.
\item We establish formal theoretical guarantees for CENTRY spanning safety (finite-horizon and asymptotic risk-budget compliance), learning (regret and convergence properties of the adaptive regulation and inference procedures), inference (consistency and identifiability of the component risk ratings), and operations (detection delay, false-alarm probability, containment, and computational properties), with each result stated under the explicit assumptions required for that result.
\end{enumerate}

The remainder of the paper is organized as follows. Section~\ref{sec:Prior Work} reviews related work on technical approaches for AI risk mitigation. Section~\ref{sec:Framework for AI Risk Monitoring} presents the proposed AI risk framework, including the formal definition of AI risk, its decomposition into component-specific risks, and the associated risk-budget formulation. Section~\ref{sec:The Problem} formally defines the AI risk monitoring problem. Section~\ref{sec:A Probabilistic Model for Continual AI Risk Estimation} develops the probabilistic CENTRY framework for continual AI risk estimation, monitoring, and regulation. Section~\ref{sec:Component Risk Priors} introduces Bayesian component-risk priors and their continual updating. Section~\ref{sec:Guarantees of CENTRY} presents the formal guarantees of CENTRY as theorems, a lemma, and corollaries, with their proofs provided in the Appendix. Section~\ref{sec:Implementation} discusses implementation aspects of the proposed framework. Section~\ref{sec:example} presents an illustrative example of continual risk detection, monitoring, and regulation by CENTRY. Finally, Section~\ref{sec:Discussion} discusses the role of SELP, and the strengths, limitations, and future directions of the proposed approach before concluding the paper.

\section{Prior Work}
\label{sec:Prior Work}
This section reviews work that addresses at least one of three facets of the problem of interest -- continual risk monitoring, probabilistic risk estimation, or adaptive regulation -- for a deployed AI system. Existing work generally addresses these facets individually rather than within a unified
probabilistic framework that jointly performs all three throughout an AI system's operational lifetime.

\subsection{Runtime monitoring of AI systems}
\label{subsec:Runtime monitoring of AI systems}
Recent methods have extended runtime monitoring of deployed AI systems to continual, streaming settings. Zhang \emph{et al.}~\cite{Zhang2026PPRM} propose prediction-powered risk monitoring (PPRM), a semi-supervised framework that constructs anytime-valid lower bounds on running risk by combining synthetic labels with
a small set of true labels, enabling harmful distribution shifts to be detected with finite-sample false-alarm guarantees. Hua \emph{et al.}~\cite{hua2025combining} study the problem of optimally combining multiple runtime monitors under cost constraints, developing a principled protocol that maximizes the probability of detecting misaligned outputs while adhering to an average-case budget. PlugGuard~\cite{PlugGuard2026} introduces a lightweight plug-in module that monitors large language model (LLM) hidden states during generation, enabling streaming risk detection at the token level. Schirmer \emph{et al.}~\cite{OnlineSafety2026} formalize online safety monitoring for LLMs as a sequential decision problem in which a monitor raises an alarm when a stream of per-step safety signals indicates that risk has exceeded a threshold.

These methods continually track risk as an AI system operates, closely related to the monitoring facet of the problem addressed here. However, they detect whether risk has exceeded a threshold rather than estimating the underlying sources of that risk, and they do not maintain a decomposed, continually updated probabilistic belief over the datasets, hardware, or models responsible for the observed risk. Consequently, none of them close the loop with adaptive regulation of the monitored system's components.

\subsection{Bayesian latent-state risk quantification}
\label{subsec:Bayesian latent-state risk quantification}
Closer in spirit to a latent-state formulation of risk, Dixon~\cite{Dixon2026BeliefAtRisk} represents an agentic AI system as a partially observed Markov decision process and uses an LLM as an approximate
Bayesian filtering operator to infer posterior beliefs over latent regimes, deriving tail-risk functionals from the resulting belief state for continual model-risk quantification. A companion paper~\cite{Dixon2026Validation} decomposes agentic model risk into state-space, filtering, forecast, policy, utility-specification, and parameter risk for independent validation. 

This work shares CENTRY's core mechanism of representing risk as an inferred Bayesian latent state rather than a static score. However, it infers a single belief state over the task or environment regime for one deployed policy -- targeted at portfolio and model-risk-management decisions -- rather
than decomposing risk by intrinsic, upstream source (datasets, hardware, and models/algorithms), and it does not couple the resulting estimate to adaptive regulation of the monitored system's own components.

\subsection{Runtime governance and adaptive regulation}
\label{subsec:Runtime governance and adaptive regulation}
A growing body of work addresses AI governance at runtime, moving beyond monitoring toward active intervention. Mar\'in and Chaudhary~\cite{Marin2026Governance} propose the Informational Viability
Principle, which states that governing an agent reduces to estimating a bound on unobserved risk and allowing an action only when its capacity exceeds that bound by a safety margin. Pan and
Hou~\cite{Pan2026AgentRunner} introduce the Dynamic Tiered AgentRunner framework, which operationalizes risk-adaptive tiering to dynamically allocate computational resources and review intensity based on task risk profiles, achieving Pareto-optimal trade-offs between safety and efficiency.

These frameworks couple risk assessment to intervention, closely paralleling the regulation facet of the problem addressed here. However, both rely on estimated risk bounds or task-level profiles rather than a full Bayesian posterior over structured latent risk sources, and neither decomposes risk into component-specific contributions that can be individually diagnosed and regulated.

\subsection{AI risk management frameworks}
\label{subsec:AI risk management frameworks}
Several organizations have proposed comprehensive frameworks for AI risk management. The NIST AI Risk Management Framework (AI RMF)~\cite{NIST2023AIRMF} provides guidance for identifying, assessing,
managing, and governing AI risks throughout the AI lifecycle, and the ISO/IEC 23894 standard~\cite{ISO23894} provides guidance for AI risk management within organizational risk management processes. Regulatory and management-system standards such as the EU AI Act~\cite{EU_AI_Act} and ISO/IEC 42001~\cite{ISO42001} further emphasize ongoing risk management and
continuous evidence collection rather than point-in-time snapshots.

These frameworks explicitly treat AI risk management as an ongoing process across the system lifecycle, closely related to the continual framing adopted here. However, they are organizational risk-management frameworks rather than probabilistic computational models: they do not infer AI risk
from observed system behavior, estimate the probable component-specific sources of risk, or adaptively regulate system components in response to changing operating conditions.

\subsection{Position of the present work}
\label{subsec:Position of the present work}
The present work addresses a different problem from the above research directions. Rather than performing runtime monitoring in isolation, latent risk quantification for a single policy, heuristic runtime governance, or organizational risk management, we formulate continual AI risk estimation, monitoring, and regulation as a unified probabilistic inference problem.

Specifically, the proposed CENTRY framework models AI risk as slowly evolving latent states associated with the datasets, hardware resources, and models/algorithms of a monitored AI system. These latent component-risk states are continually inferred from the interaction between the monitored AI system and its environment. The resulting Bayesian component-risk estimates are subsequently used to adaptively regulate the utilization of the corresponding AI system components, thereby closing the loop between probabilistic inference and AI risk mitigation.

This distinguishes CENTRY from existing work in several key respects. First, unlike heuristic risk scoring methods, CENTRY performs Bayesian inference over a structured latent representation, providing principled uncertainty quantification; unlike Bayesian belief-state approaches to agentic model risk~\cite{Dixon2026BeliefAtRisk,Dixon2026Validation}, this latent representation is structured around the intrinsic sources of risk rather than a single task or environment regime. Second, unlike frameworks that treat risk as monolithic, CENTRY explicitly decomposes risk into interpretable component-specific sources -- datasets, hardware, and models -- enabling targeted diagnosis and intervention. Third, unlike static or threshold-based governance policies, and unlike model-risk frameworks that stop at quantification or validation, CENTRY's regulation mechanism is directly coupled to the inferred posterior beliefs, closing the loop from estimation to intervention on the monitored system's own components. Finally, unlike frameworks focused solely on detection or monitoring, CENTRY integrates estimation, monitoring, and regulation within a single mathematically rigorous architecture and provides formal theoretical guarantees for selected safety, learning, inference, and operational properties under explicitly stated assumptions.

To the best of our knowledge, existing probabilistic AI-risk frameworks do not jointly provide continual estimation of component-specific latent risk, continual monitoring of AI--environment interactions, adaptive regulation of the monitored system's components, and formal guarantees characterizing the resulting safety and operational properties throughout the operational lifetime of a deployed AI system.

\section{Framework for AI Risk Regulation}
\label{sec:Framework for AI Risk Monitoring}
This section establishes the framework used throughout the paper. We adopt a formal definition of AI risk from the existing AI safety literature and introduce a quantitative decomposition of AI risk into its constituent components. Based on this decomposition, we define a risk budget framework that specifies acceptable limits on the risk contributed by each component of an AI system. The resulting framework is independent of the proposed CENTRY model and serves as the mathematical foundation for continual AI risk estimation and regulation.

\subsection{Definition of AI risk}
\label{subsec:Definition of AI risk}
Risk is generally understood as the effect of uncertainty on objectives \cite{iso31000}. The NIST AI Risk Management Framework (AI RMF 1.0) defines risk as a composite measure of the likelihood of an event and the magnitude of the consequences associated with that event \cite{NIST2023AIRMF}. When the focus is restricted to adverse outcomes, AI risk therefore concerns both the likelihood that an AI system produces a harmful outcome and the severity of the resulting harm. This perspective also recognizes that AI risk is context dependent and can arise throughout the AI system lifecycle.

Consistent with these definitions, we define \emph{AI risk as the expected magnitude of harm resulting from an AI system's behavior. Harm may arise from uncertain interactions among the system's data, models and algorithms, computational resources, and deployment context. The risk estimate accounts for the likelihood and consequences of potentially harmful outcomes. For each outcome, we consider factors such as the type and severity of harm, the number of people and systems that could be affected, and how quickly the effects could spread.}

Formally, let $\Omega$ denote the set of possible outcomes resulting from the interaction of an AI system with its environment. Let $L:\Omega\rightarrow\mathbb{R}_{\geq 0}$ denote a loss function that
quantifies the magnitude of harm associated with outcome $\omega\in\Omega$, aggregating factors such as those listed above. Assume the loss is measurable with finite expectation. For continuous outcomes, let $p(\omega)$ denote the probability density with respect to the reference measure $d\omega$ (for discrete outcomes it denotes the probability mass). The expected AI risk is
\begin{equation}
R = \mathbb{E}[L(\omega)] = \int_{\Omega} L(\omega)\,p(\omega)\,d\omega .
\label{eq:AIrisk}
\end{equation}
For discrete outcomes, the corresponding expression is $R=\sum_{\omega\in\Omega}L(\omega)p(\omega)$. \textit{Thus, an outcome contributes more strongly to overall AI risk when it is both more likely to occur and more harmful.}

The probability distribution $p(\omega)$ is determined by the behavior of the AI system and its interaction with its deployment context. Consequently, AI risk is not an intrinsic scalar property of an AI model considered in isolation; it may change as the data, computational resources, model or
algorithm, or deployment conditions change. In particular, the same AI system may exhibit different levels of risk, and the same system behavior may lead to harms of different severity for different users or user groups in different operating contexts.

Eq.~\eqref{eq:AIrisk} defines the system-level quantity that CENTRY ultimately seeks to keep within an acceptable range. CENTRY does not directly observe $R$, the loss function $L$, or the harmful outcome distribution $p(\omega)$. Instead, it continually infers latent component-risk states from the observed interaction between the monitored AI system and its environment. These states are summarized by the component risk ratings associated with datasets, hardware resources, and models/algorithms, which are subsequently used to estimate, monitor, and regulate the factors contributing to the system-level AI risk. Interpreting these ratings operationally nevertheless requires a domain-appropriate specification or calibration of harm (the loss $L$) and of the acceptable risk budgets.

\subsection{Risk decomposition}
\label{subsec:Risk decomposition}
The system-level AI risk defined in Eq.~\eqref{eq:AIrisk} may arise from multiple interacting sources. For the present framework, we focus on three intrinsic components of the monitored AI system: datasets, hardware resources, and models/algorithms. Let \(R^{(D)}\), \(R^{(H)}\), and \(R^{(M)}\) denote an operational allocation of the system-level risk to these three components. We write
\begin{equation}
R \approx R^{(D)} + R^{(H)} + R^{(M)} .
\label{eq:riskdecomposition}
\end{equation}

Eq.~\eqref{eq:riskdecomposition} is an operational decomposition used for risk estimation, diagnosis, and budget allocation; it is not a claim that the three components are causally independent or that their effects are uniquely identifiable from behavior alone. In particular, interactions among components may contribute to the observed system-level risk. Such interactions are absorbed into the component allocation according to the adopted risk-rating and calibration procedure.

Additional sources of risk, such as deployment context, human oversight, external tools, and interactions among multiple AI systems, may be represented explicitly by adding further components and, where appropriate, interaction terms. Our three-component formulation is therefore a deliberately restricted first approximation rather than a complete causal account of AI risk. 

\subsection{Risk budget}
\label{subsec:Risk budget}
AI risk cannot always be eliminated. Instead, AI systems should be designed such that the contribution of each component remains within an acceptable limit. 

Let $B^{(D)}$, $B^{(H)}$, $B^{(M)}$ be the maximum allowable risks associated with datasets, hardware resources, and models, respectively. The total allowable risk budget is therefore
\begin{equation*} 
B = B^{(D)} + B^{(H)} + B^{(M)}.
\end{equation*}
A feasible AI system should satisfy $R\le B$. Under the approximate operational decomposition of Eq.~\eqref{eq:riskdecomposition}, we adopt the following aggregate component-budget criterion:
\begin{equation} 
\label{eq:componentbudget}
R^{(D)} + R^{(H)} + R^{(M)} \le B^{(D)} + B^{(H)} + B^{(M)}.
\end{equation}
Eq.~\eqref{eq:componentbudget} defines a resource-allocation problem in which each component consumes part of the available risk budget. The objective of continual AI risk estimation is to determine, at every instant, how much of the available risk budget is currently consumed by each component. Because Eq.~\eqref{eq:riskdecomposition} is approximate, Eq.~\eqref{eq:componentbudget} serves as an operational budget criterion; guaranteeing the system-level condition $R\le B$ additionally requires that any positive decomposition error be accounted for within the available risk budget. Aggregate compliance is necessary for component-wise feasibility but is not sufficient, since an excess in one component may be offset by unused budget in another.

In the remainder of this paper, the component risk quantities are estimated probabilistically from latent state representations using the CENTRY framework. Thus, the risk budget is monitored and used to guide regulation through continually updated Bayesian estimates rather than fixed deterministic risk values.

\subsection{Feasible design region}
\label{subsec:Feasible design region}
The feasible design region consists of all AI system configurations that satisfy the component-wise risk constraints
\begin{equation*}
R^{(D)}\le B^{(D)},\quad R^{(H)}\le B^{(H)},\quad R^{(M)}\le B^{(M)}.
\end{equation*}
Equivalently, $(R^{(D)},R^{(H)},R^{(M)})\in\mathcal{F}$, where $\mathcal{F} = \bigl\{(R^{(D)},R^{(H)},R^{(M)}): R^{(D)}\le B^{(D)}, R^{(H)}\le B^{(H)}, R^{(M)}\le B^{(M)}\bigr\}$ is the feasible design region.

Whenever one or more component risks exceed their allocated budgets, the AI system leaves the feasible region and corrective actions become necessary. The primary objective of the proposed CENTRY framework is therefore to continually estimate the component risks ($R^{(D)}$, $R^{(H)}$, $R^{(M)}$), infer the component(s) most probably associated with elevated estimated risk, and regulate the monitored AI system so that its operating point remains within the feasible design region throughout its operational lifetime.

The component risks $R^{(D)}$, $R^{(H)}$, and $R^{(M)}$ are conceptual quantities that are generally not directly observable. Section~\ref{sec:A Probabilistic Model for Continual AI Risk Estimation} introduces CENTRY, which continually estimates probabilistic beliefs about these unknown risks through the latent variables $z_t^{(D)}$, $z_t^{(H)}$, and $z_t^{(M)}$ and summarizes them by the component-risk ratings $\rho_t^{(D)}$, $\rho_t^{(H)}$, and $\rho_t^{(M)}$.

\section{The Problem}
\label{sec:The Problem}
Consider an AI system interacting with an external environment over discrete time instants \(t=1,2,\ldots\). The AI system may be autonomous or semi-autonomous, such as a large language model (LLM), reinforcement-learning agent, robotic system, autonomous vehicle, decision-support system, or
multi-agent AI system. The environment comprises the external entities with which the AI system interacts, including humans, other AI agents, physical objects, software systems, and communication networks.

At time \(t\), the AI system receives a percept \(x_t\) from its environment and produces an action \(a_t\). Both may be multimodal and may include text, speech, non-speech vocalizations, images, video, body language, sensor measurements, control signals, or other application-specific modalities. The
observable interaction at time \(t\) is therefore
\begin{equation}
y_t \equiv (x_t,a_t),
\label{eq:interaction}
\end{equation}
and the interaction history through time \(t\) is $y_{1:t}=(y_1,\ldots,y_t) = \bigl((x_1,a_1),\ldots,(x_t,a_t)\bigr)$.

The monitored AI system is treated as a black box. Its internal states, parameters, implemented safeguards and safety guardrails, policy, and action-generation mechanism are not available to the monitoring system. The monitoring system can therefore use only the observable interaction history and information available through an external intervention
interface.

The monitoring system may issue a regulatory action \(c_t\), which is distinct from the action \(a_t\) produced by the monitored AI system. The action \(a_t\) is part of the monitored system's observable behavior, whereas \(c_t\) is an external intervention intended to influence subsequent behavior. The
intervention interface may permit \(c_t\) to modify or constrain the utilization, availability, or configuration of the monitored system's training datasets, hardware resources, or models/algorithms. The regulatory action does not directly prescribe the external action \(a_t\).

Let $R_t^{(j)}$, $j\in\{D,H,M\}$, be the instantaneous risks associated with the dataset, hardware, and model/algorithm components, respectively. These component risks are nonnegative random variables under the closed-loop law induced by the regulatory policy, are generally not directly observable, and may evolve over the operational lifetime of the AI system. Each component has an associated acceptable risk budget \(B^{(j)}\), defining the desired feasible operating region.

\textit{The continual AI risk estimation, monitoring, and regulation problem can then be defined as the design of a regulation policy $f$ that, at each time $t$, maps the observable interaction history $y_{1:t}$ and previously applied regulatory actions $c_{1:t-1}$ to an admissible regulatory action $c_t$. The policy is required to maintain the monitored AI system within the acceptable component-risk budgets $B^{(j)}$, $j\in\{D,H,M\}$, to the extent permitted by the AI system's intervention interface and system uncertainty. Because the true component risks are latent, CENTRY achieves this objective indirectly through the inferred component-risk ratings; the corresponding true-risk guarantees hold under the calibration and regulation assumptions stated in Section~\ref{sec:Guarantees of CENTRY}.} That is, design $f$ such that
\begin{equation}
c_t = f\!\left(y_{1:t},c_{1:t-1}\right),
\label{eq:problem_policy}
\end{equation}
\begin{equation}
\Pr\left[ R_{t+k}^{(j)}\le B^{(j)}, \forall j, k=1,\ldots,T \mid\mathcal F_t \right] \ge 1-\delta.
\label{eq:problem_constraint}
\end{equation}

Here, $\mathcal F_t=\sigma(y_{1:t},c_{1:t-1})$ is the information available to CENTRY at time $t$, $T\in\mathbb N$ is the prediction horizon, and $\delta\in(0,1)$ is the permitted probability of any component-budget violation over that horizon. The probability in the constraint is taken under the closed-loop law induced by the policy, and the conditional inequality is required almost surely. Eq.~\eqref{eq:problem_constraint} is a desired finite-horizon
chance-constrained operating objective, not an unconditional property of
an arbitrary monitoring system. Its achievement by CENTRY requires
calibrated violation-probability estimates, feasible admissible
interventions, and the other conditions stated in
Section~\ref{sec:Guarantees of CENTRY}.

Eq.~\eqref{eq:problem_policy} emphasizes that regulation is causal: \(c_t\) is selected using information available through time \(t\), whereas its intended effect is on subsequent system behavior and therefore on future risk. Because the component risks \(R_t^{(j)}\) are latent, the monitoring
system cannot enforce Eq.~\eqref{eq:problem_constraint} by directly observing the true risks. It must instead infer their current values from the observable interaction history and use those estimates to select regulatory actions.

\begin{figure}[htbp!]
    \centering
    \includegraphics[width=0.5\linewidth]{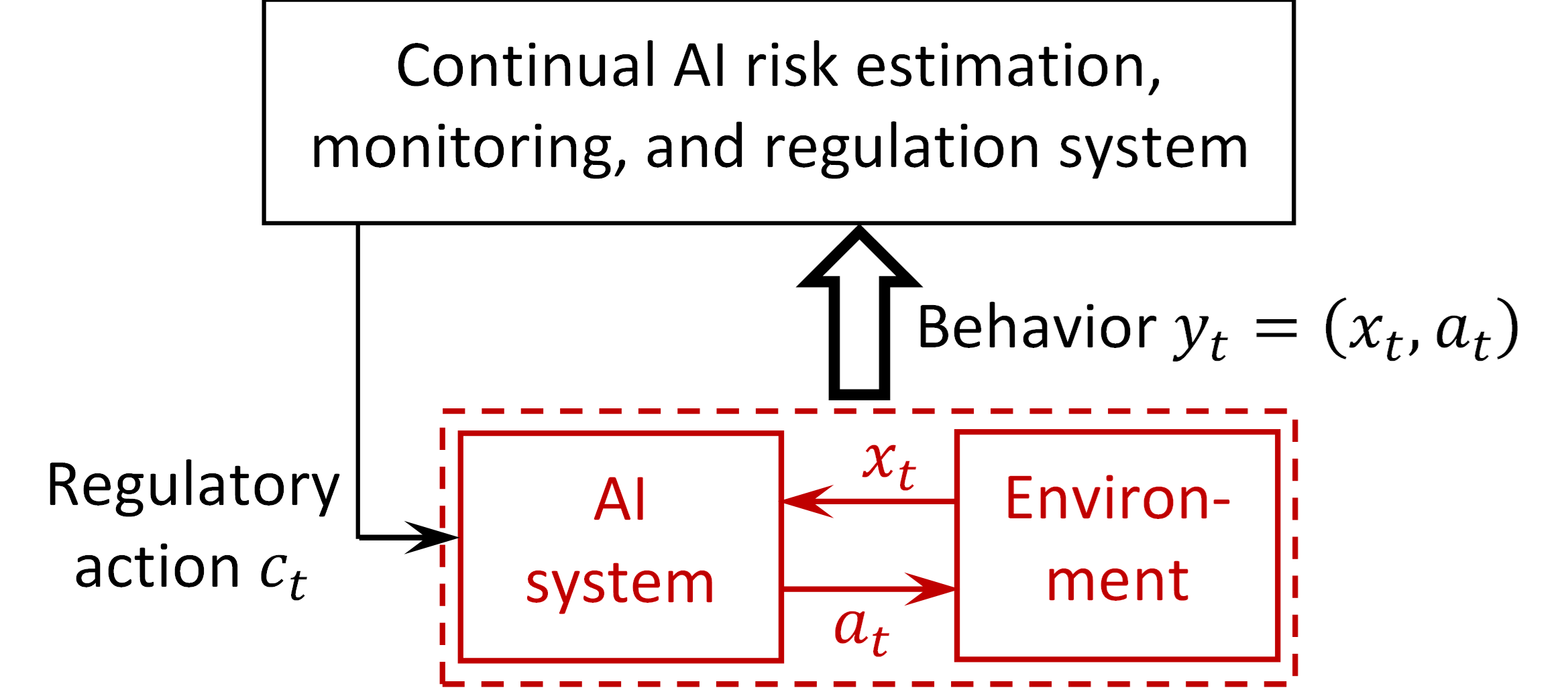}
    \caption{Block diagram depicting the AI risk estimation, monitoring, and regulation problem.}
    \label{fig:Problem}
\end{figure}

The problem is subject to the following assumptions:
\begin{enumerate}[nosep]
    \item \textbf{Observable interaction:}
    The percept \(x_t\) received by the AI system and its action \(a_t\) are observable to the monitoring system, so that \(y_t=(x_t,a_t)\) is available at each time instant.

    \item \textbf{Black-box operation:}
   The monitoring system has no direct access to the AI system's internal states, parameters, implemented safeguards and safety guardrails, decision policy or action-generation mechanism, other than through the admissible intervention interface.

    \item \textbf{Latent component risks:}
    The risks associated with the dataset, hardware, and model/algorithm components are not directly observable during operation and must therefore be inferred from information available to the monitoring system.

    \item \textbf{Temporal evolution:}
    The component risks may change over time, but evolve more slowly than the rapidly changing interaction between the AI system and its environment.

    \item \textbf{Admissible intervention:}
    An external intervention interface is available through which regulatory actions can modify or constrain the utilization, availability, or configuration of the datasets, hardware resources, or models/algorithms of the AI system.

    \item \textbf{Indirect regulation:}
    A regulatory action \(c_t\) can influence subsequent behavior of the AI system but does not directly determine its external action \(a_t\).

    \item \textbf{Risk budgets:}
    Each component has an associated acceptable risk budget \(B^{(j)}\), \(j\in\{D,H,M\}\), defining the desired feasible operating region.

    \item \textbf{Continual operation:}
    The AI system continues to interact with its environment while monitoring and regulation are performed, providing new interaction data over time.
\end{enumerate}

The solution developed in the following sections, CENTRY, addresses this problem by maintaining a probabilistic representation of the latent component risks, continually updating that representation from the observed interaction stream, and selecting admissible regulatory actions based on the resulting risk estimates.

\begin{table}[tb!]
\centering
\caption{Key symbols and notations used in this paper.}
\label{tab:notation}
\footnotesize
\renewcommand{\arraystretch}{1.03}
\begin{tabular}{@{}>{\raggedright\arraybackslash}p{0.15\linewidth}
                >{\raggedright\arraybackslash}p{0.21\linewidth}
                >{\raggedright\arraybackslash}p{0.58\linewidth}@{}}
\toprule
\textbf{Symbol} & \textbf{Category} & \textbf{Description} \\
\midrule

$t$ & Time & Discrete time index. \\

$x_t$ & Percept & Percept received by the monitored AI system at time $t$. \\

$a_t$ & Action & External action produced by the monitored AI system at time $t$. \\

$y_t$ & Interaction & Observable interaction, $y_t=(x_t,a_t)$. \\

$y_{1:t}$ & Interaction history & Sequence of observable interactions from time $1$ through $t$. \\

$c_t$ & Regulatory action & External intervention issued by CENTRY at time $t$. \\

$\mathcal F_t$ & Current information & Information through interaction $y_t$: $\sigma(y_{1:t},c_{1:t-1})$. \\

$\mathcal F_t^{-,y}$ & Pre-interaction information & Information before $y_t$: $\sigma(y_{1:t-1},c_{1:t-1})$. \\

$R$ & AI risk & System-level expected risk defined in Eq.~\eqref{eq:AIrisk}. \\

$R^{(D)},R^{(H)},R^{(M)}$ & Component risks & Risks operationally allocated to the dataset, hardware, and model/algorithm components, respectively. \\

$B^{(D)},B^{(H)},B^{(M)}$ & Risk budgets & Maximum allowable risks allocated to the dataset, hardware, and model/algorithm components, respectively. \\

$\bar B$ & Normalized thresholds & Column vector of normalized component-risk thresholds $\bar B^{(D)},\bar B^{(H)},\bar B^{(M)}$ used with the risk ratings. \\

$z_t$ & Latent state & Structured latent state at time $t$. \\

$z_t^{(F)}$ & Fast latent state & Latent state representing fast interaction dynamics. \\

$z_t^{(D)},z_t^{(H)},z_t^{(M)}$ & Risk latent states & Slowly evolving latent states associated with dataset, hardware, and model/algorithm risk, respectively. \\

$\rho_t^{(D)},\rho_t^{(H)},\rho_t^{(M)}$ & Risk ratings & Normalized component-risk ratings obtained from the corresponding latent posteriors. \\

$q_\phi$ & Variational posterior & Approximate posterior distribution over the latent state. \\

$\kappa_t$ & Propagated prior & Prior distribution over the latent state at time $t$, obtained by propagating the previous posterior through the transition model. \\

$p_\theta$ & Generative model & CENTRY's parameterized latent transition and observation model. \\

$\theta$ & Model parameters & Parameters of the generative and latent-transition models. \\

$\phi$ & Inference parameters & Parameters of the variational inference model. \\

$\psi$ & Policy parameters & Parameters of CENTRY's regulatory policy. \\

$\pi_\psi$ & Regulatory policy & Mapping from component-risk ratings to admissible regulatory actions. \\

$S_t^{\mathrm{(KL)}}$ & Population Surprise & Conditional population predictive mismatch between the true and predictive interaction distributions. \\
$\widehat S_t$ & Realized Surprise & Predictive negative log-likelihood of the observed interaction, computable at runtime. \\

$\tau$ & Surprise threshold & Optional threshold used to identify interactions warranting additional diagnostic or investigative attention. \\

$\beta$ & Variational weight & Weight on the KL regularization term in the ELBO. \\

$\lambda_S,\lambda_j,\lambda_C$ & Objective weights & Weights on predictive error, component risk, and regulatory-action cost, respectively. \\

$C(\pi_\psi(\rho_t))$ & Regulatory cost & Cost associated with applying the regulatory action selected by $\pi_\psi$ from the current risk ratings. \\

$\mathcal{C}$ & Action space & Set of admissible regulatory actions. \\

$g^{(j)}$ & Risk functional & Mapping from the posterior parameters of component $j$ to its normalized risk rating. \\

$\mu_t^{(j)},\Sigma_t^{(j)}$ & Posterior parameters & Mean and covariance of the posterior for latent component $j$. \\

$\mathcal{U}$ & Evidence database & Collection of evidence-supported behavior--cause records used to refine the inference model. \\

$\chi_i,\nu_i$ & Record metadata and evidence & Component-configuration metadata and supporting evidence of record $i$ in $\mathcal{U}$. \\

$d_j$ & Latent dimension & Dimension of latent component $z_t^{(j)}$, $j\in\{F,D,H,M\}$. \\

$d_y$ & Interaction dimension & Dimension of the observable interaction $y_t$. \\

$h$ & Hidden dimension & Dimension of the learned representation used by an implementation of the inference model. \\

$\Omega$ & Outcome space & Set of possible outcomes of an AI--environment interaction. \\

$L(\omega)$ & Loss function & Magnitude of harm associated with outcome $\omega$. \\

\bottomrule
\end{tabular}
\end{table}

\section{A Probabilistic Model for Continual AI Risk Estimation, Monitoring and Regulation}
\label{sec:A Probabilistic Model for Continual AI Risk Estimation}
Let $R^{(D)}$, $R^{(H)}$, and $R^{(M)}$ denote the risks associated with the training dataset, hardware, and model/algorithm components, respectively, and let $B^{(D)}$, $B^{(H)}$, and $B^{(M)}$ denote their corresponding acceptable risk budgets. These component risks are generally not directly observable and may vary throughout the operational lifetime of the AI system. Our goal is to use the observable interaction history $y_{1:t}$, together with the regulatory actions previously applied, $c_{1:t-1}$, to continually:
\begin{itemize}[nosep]
    \item detect behavior that is inconsistent with the behavior expected from the available information;
    \item identify the component(s) that may be responsible for increased risk;
    \item estimate the current levels of the component risks $R^{(D)}$, $R^{(H)}$, and $R^{(M)}$; and
    \item select admissible regulatory actions based on the inferred component-risk ratings, with the objective of maintaining the underlying component risks within their prescribed risk budgets while accounting for the consequences and costs of regulation.
\end{itemize}

\begin{figure}[htbp!]
    \centering
    \includegraphics[width=0.9\linewidth]{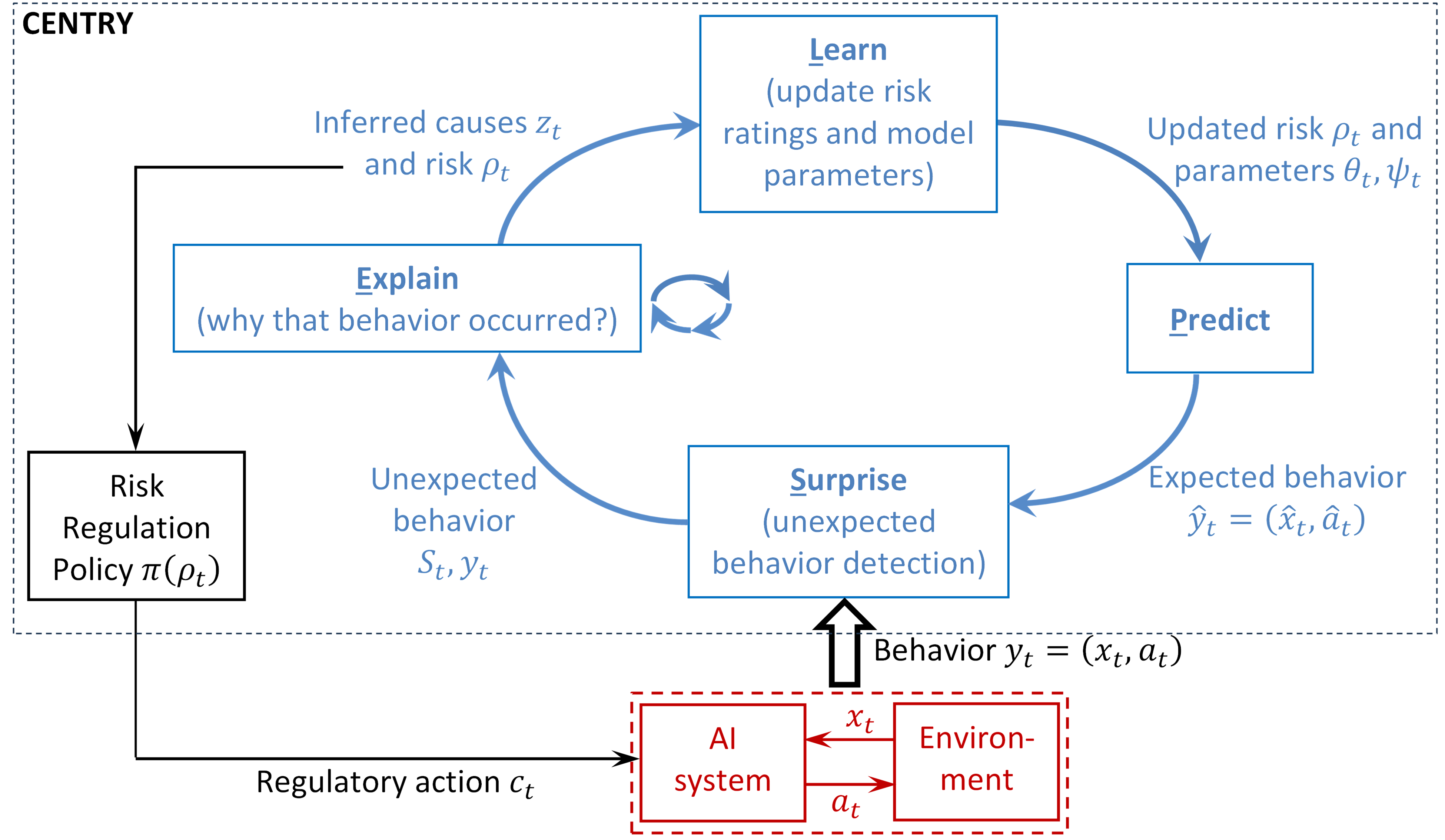}
    \caption{Block diagram of the proposed CENTRY framework. The SELP (Surprise$\rightarrow$Explain$\rightarrow$Learn$\rightarrow$ Predict) cycle (in {\color{blue}blue}) continually monitors a given AI system's interactions with its environment (in {\color{red}red}) and updates the model and policy parameters and risk rating of the different components in the monitored system's pipeline throughout its lifecycle. The monitored system's behavior is regulated using the risk ratings to minimize risk in its behavior (in {\color{black}\textbf{black}}). Best viewed in color.}
    \label{fig:CENTRY_block_diagram}
\end{figure}

This section presents the proposed \textbf{CENTRY} (\textbf{\underline{C}}ontinual AI risk \textbf{\underline{E}}stimation, mo\textbf{\underline{N}}i\textbf{\underline{T}}oring and \textbf{\underline{R}}egulation s\textbf{\underline{Y}}stem), a probabilistic framework for continual AI risk estimation, monitoring, and regulation. Unlike existing AI safety frameworks that primarily define, assess, or govern AI risk, CENTRY continuously observes the interaction between an AI system and its environment, infers posterior beliefs over latent component-risk states that may explain observed behavior, and regulates the AI system toward acceptable risk levels throughout its operational lifetime, subject to the assumptions required for the formal guarantees in Section~\ref{sec:Guarantees of CENTRY}. Figure~\ref{fig:CENTRY_block_diagram} shows the different components of the proposed CENTRY framework.

CENTRY is the complete framework proposed in this paper for continual AI risk estimation, monitoring, and regulation. As illustrated in Figure~\ref{fig:CENTRY_block_diagram}, CENTRY consists of three conceptual components:
\begin{enumerate}[nosep]
\item a probabilistic latent-state model that represents the hidden sources of AI risk,
\item an AI agent that continually performs prediction, explanation, learning, and regulation, and
\item an adaptive risk-regulation mechanism that modifies the monitored AI system through an admissible intervention interface.
\end{enumerate}

\subsection{Structured latent representation}
\label{subsec:Structured Latent Representation}
This subsection develops the first of these three components, namely the probabilistic latent-state model. The objective of the latent-state model is to continually estimate the hidden state of the monitored AI system from its observable interaction history. Let $y_t=(x_t,a_t)$ denote the observable AI--environment interaction available to CENTRY at time $t$, and let $c_t$ denote the regulatory action selected by CENTRY. The central probabilistic quantity in the proposed framework is the posterior distribution
\begin{equation}
q\!\left(z_t \mid y_{1:t},c_{1:t-1}\right),
\label{eq:jointposterior}
\end{equation}
which represents CENTRY's current probabilistic belief regarding the hidden state of the monitored AI system after observing the interaction history up to time $t$ and the regulatory actions applied previously.

Unlike conventional latent-variable models that employ a single homogeneous latent representation, CENTRY partitions the latent state into one fast interaction component and three slowly varying risk components,
\begin{equation}
z_t = \left(z_t^{(F)},z_t^{(D)},z_t^{(H)},z_t^{(M)}\right),
\label{eq:latentstate}
\end{equation}
where $z_t^{(F)}$ models the fast interaction dynamics between the monitored AI system and its environment, $z_t^{(D)}$ models the latent dataset-risk state, $z_t^{(H)}$ models the latent hardware-risk state, and $z_t^{(M)}$ models the latent model/algorithm-risk state. Consequently,
\begin{equation*}
q\!\left(z_t \mid y_{1:t},c_{1:t-1}\right)
= q\!\left(z_t^{(F)},z_t^{(D)},z_t^{(H)},z_t^{(M)} \mid y_{1:t},c_{1:t-1}\right),
\end{equation*}
indicating that the interaction dynamics and all component-risk states are inferred jointly from the same observable interaction history. This joint inference is central to CENTRY because unexpected behavior may arise from transient interaction dynamics, intrinsic component risks, or both simultaneously.

Each latent component is modeled as a multivariate Gaussian random variable,
\begin{equation*}
z_t^{(j)} \sim \mathcal N\!\left(\mu_t^{(j)},\Sigma_t^{(j)}\right), \qquad j\in\{F,D,H,M\}.
\end{equation*}
Let $d_j$ be the dimension of subspace $j$, so that $z_t^{(j)}\in\mathbb{R}^{d_j}$ for $j\in\{F,D,H,M\}$. For computational efficiency, the component covariance matrices may be assumed diagonal, although full covariance matrices within individual latent components are also admissible. The posterior Gaussian is completely characterized by its mean and covariance. Consequently, the pair $\left(\mu_t^{(j)},\Sigma_t^{(j)}\right)$ provides a complete probabilistic representation of the uncertainty associated with the corresponding latent component.

\paragraph{Fast interaction dynamics.}
The latent state $z_t^{(F)}$ summarizes the rapidly evolving interaction between the monitored AI system and its environment. Its evolution is conditioned on the previous fast state $z_t^{(F)}$, the last observed interaction $y_t=(x_t,a_t)$, and the regulatory action $c_t$ selected by CENTRY,
\begin{equation*}
z_{t+1}^{(F)} \sim p\!\left(z_{t+1}^{(F)}\mid z_t^{(F)},y_t,c_t\right).
\end{equation*}

\paragraph{Slow component-risk dynamics.}
The component-risk states evolve more gradually. For each $j\in\{D,H,M\}$,
\begin{equation*}
z_{t+1}^{(j)} \sim p\!\left(z_{t+1}^{(j)}\mid z_t^{(j)},y_t,c_t\right).
\end{equation*}
The transition model is component-wise: the evolution of $z_{t+1}^{(j)}$ depends on the previous state of the same component, but not directly on the previous latent states of the other components. For the factorized predictive-prior formulation in Section~\ref{subsec:Updating the priors using SELP}, the variational posterior is assumed to factorize across latent components conditional on the observed interaction and regulatory-action histories. Statistical dependence may nevertheless arise across different histories, but cross-component dependence within a single conditional variational posterior is not modeled in this formulation.

These latent states accumulate evidence regarding the long-term characteristics of the datasets, hardware resources, and learning models employed by the monitored AI system. The separation between fast interaction dynamics and slow component-risk dynamics enables CENTRY to distinguish transient behavioral changes from gradual changes in the intrinsic sources of AI risk. The probabilistic latent-state model developed in this subsection specifies \emph{what} hidden quantities are estimated.

\subsubsection{Risk-causality training database}
\label{subsubsec:Risk-causality training database}
The latent states in Eq.~\eqref{eq:latentstate} are not directly observed during deployment. To make the latent subspaces correspond to the dataset, hardware, and model/algorithm sources of risk, CENTRY may be trained using a risk-causality database
\begin{equation}
\mathcal{U} = \left\{\left( y_i,z_i,\chi_i,\nu_i \right)\right\}_{i=1}^{N}.
\label{eq:risk-causality-database}
\end{equation}

Each record in \(\mathcal{U}\) contains four quantities:
\begin{enumerate}[nosep]
    \item $y_i=(x_i,a_i)$ is an observed AI--environment interaction.
    \item $z_i=\left(z_i^{(F)},z_i^{(D)},z_i^{(H)},z_i^{(M)}\right)$ is the corresponding structured latent-state representation. The fast component \(z_i^{(F)}\) represents interaction dynamics, whereas \(z_i^{(D)}\), \(z_i^{(H)}\), and \(z_i^{(M)}\) represent the component-risk states. The component-risk parts may be supported by controlled experiments, incident investigations, component-ablation studies, post-deployment audits, or evidence-supported inference from prior monitored systems.
    \item $\chi_i$ is the observable metadata describing the component configuration under which $y_i$ occurred. The configuration metadata \(\chi_i\) may include, when available, dataset identifiers, provenance and quality scores, retrieval-source identifiers, model identity and version, decoding or policy configuration, tool-access configuration, inference-compute budget, memory allocation, hardware and execution logs, and the regulatory actions active when the interaction was observed.
    \item $\nu_i$ is the evidence supporting the association between $y_i$ and $z_i$. The supporting evidence \(\nu_i\) may include controlled component perturbations, known downstream outcomes, audit findings, incident reports, test-environment results, or changes in behavior observed after a component-specific regulatory action.
\end{enumerate}

The database \(\mathcal{U}\) is not a database of facts from the application domain of the monitored AI system. It is a database of evidence-supported behavior--cause (latent state) examples. Thus, it does not require CENTRY to be an expert in medicine, law, finance, software engineering, or another monitored domain. Instead, it supports the learning of transferable associations between observable interaction patterns, component configurations, latent risk causes, and the effects of component-level interventions. However, domain expertise may still be required to establish the supporting evidence, assess realized harms, and calibrate the component-risk ratings and risk budgets.

During training and continual learning, records in \(\mathcal{U}\) may be used to learn or refine the parameters \(\phi\) of the parametric inference model. At runtime, however, the Explain stage remains the mapping defined in Eq.~\eqref{eq:explainposterior}:
\begin{equation*}
(y_{1:t},c_{1:t-1}) \longmapsto q_\phi(z_t\mid y_{1:t},c_{1:t-1}).
\end{equation*}
Thus, \(\mathcal{U}\) is a training and continual-learning resource; it is not an additional runtime input to the Explain function.

Initially, \(\mathcal{U}\) may be small in size, consisting of controlled experiments, red-team exercises, simulations, prior incident analyses, and audited deployment cases. During operation, a newly observed interaction may be stored as a provisional record together with its inferred posterior, the selected regulatory action, any component-configuration information available from deployment logs, and, as it becomes available, subsequent evidence. It should be promoted to a supported behavior--cause training example only after subsequent evidence supports both its risk relevance and its component-level explanation. This prevents CENTRY from treating its own unverified causal inferences as ground-truth labels.

The structured latent representation specifies the hidden quantities estimated by CENTRY, while the risk-causality database provides evidence-supported examples that can be used to train the parametric inference model. The next subsection specifies the probabilistic generative model relating the latent
state to the observable interaction \(y_t=(x_t,a_t)\).

\subsection{Probabilistic generative model}
\label{subsec:Probabilistic generative model}
Section~\ref{subsec:Structured Latent Representation} introduced the structured latent representation used by CENTRY to model the hidden state of the monitored AI system. This subsection specifies the probabilistic generative model relating the latent state to the observable interaction sequence. The generative model constitutes the \emph{Predict} component of the SELP cycle by predicting the next observable behavior of the monitored AI system and its environment.

\subsubsection{Generative assumptions}
\label{subsubsec:Generative assumptions}
Let $y_t=(x_t,a_t)$ denote the observable interaction at time $t$ and $c_t$ the regulatory action selected by CENTRY. The monitored AI system is treated as a black box: CENTRY does not model the internal mechanism that maps the AI system's inputs and internal state to its action $a_t$. Instead, the combined effect of the monitored AI system, its environment, and the external intervention is represented probabilistically through the latent state. CENTRY therefore uses the observed behavior to infer component-risk states and uses the resulting risk estimates to select regulatory actions, which influence, but do not prescribe, the AI system's subsequent behavior.

The next observable interaction is generated according to
\begin{equation*}
y_{t+1}\sim p_\theta(y_{t+1}\mid z_{t+1}),
\end{equation*}
while the latent state evolves according to
\begin{equation*}
z_{t+1}\sim p_\theta(z_{t+1}\mid z_t,y_t,c_t).
\end{equation*}
Thus, $c_t$ does not directly determine $a_{t+1}$. Instead, it acts through the admissible intervention interface and changes the probability distribution of the subsequent observable interaction $y_{t+1}=(x_{t+1},a_{t+1})$. The resulting predictive model is therefore an approximation to the unknown black-box response process induced by the monitored AI system, its environment, and the applied regulatory actions.

\subsubsection{Joint distribution}
\label{subsec:Joint distribution}
For an interaction sequence of length $T$, the corresponding joint distribution factorizes as
\begin{equation}
p_\theta(z_{1:T},y_{1:T}\mid c_{1:T-1})
= \kappa_1(z_1)
\prod_{t=1}^{T-1} \prod_{j\in\{F,D,H,M\}} p_\theta\!\left(z_{t+1}^{(j)}\mid z_t^{(j)},y_t,c_t\right)
\prod_{t=1}^{T} p_\theta(y_t\mid z_t).
\label{eq:jointdistribution}
\end{equation}
Here, $\kappa_1(z_1)$ denotes the initial latent-state prior. This factorization specifies the probabilistic generative model underlying CENTRY. Conditioning on the regulatory-action sequence here denotes the controlled generative model under supplied interventions, rather than retrospective conditioning on actions generated by an adaptive feedback policy.

\subsubsection{Prediction}
\label{subsec:Prediction}
The predictive distribution produced by the Predict stage is obtained by marginalizing the next latent state over the posterior at time $t$ and the latent-state transition model. Because the posterior at time $t$ is formed before the regulatory action $c_t$ is selected, it is conditioned only on the interaction history and previously applied regulatory actions. Thus,
\begin{equation}
p_{\mathrm{pred}}\!\left(y_{t+1}\mid y_{1:t},c_{1:t}\right)
= \int p_\theta\!\left(y_{t+1}\mid z_{t+1}\right) p_\theta\!\left(z_{t+1}\mid z_t,y_t,c_t\right)
q_\phi\!\left(z_t\mid y_{1:t},c_{1:t-1}\right)\,dz_t\,dz_{t+1}.
\label{eq:predictivedistribution}
\end{equation}
The resulting predictive distribution describes the expected interaction behavior $\hat{y}_{t+1} = (\hat{x}_{t+1},\hat{a}_{t+1})$ of the monitored AI system and its environment under the selected regulatory action $c_t$. The action $c_t$ therefore influences the predictive distribution through the latent-state transition, without directly prescribing the monitored AI system's external action $a_{t+1}$.

We denote this marginalized distribution by $p_{\mathrm{pred}}$ to distinguish it from the parameterized generative distributions $p_\theta$ used for the latent-state transition and interaction
generation (see for example \cite{baruah2020multimodal,baruah2023intent,BanerjeeBaruah2024Sensors} for implemented models of interaction generation).

\begin{figure}[htb!]
    \centering
    \includegraphics[width=0.7\linewidth]{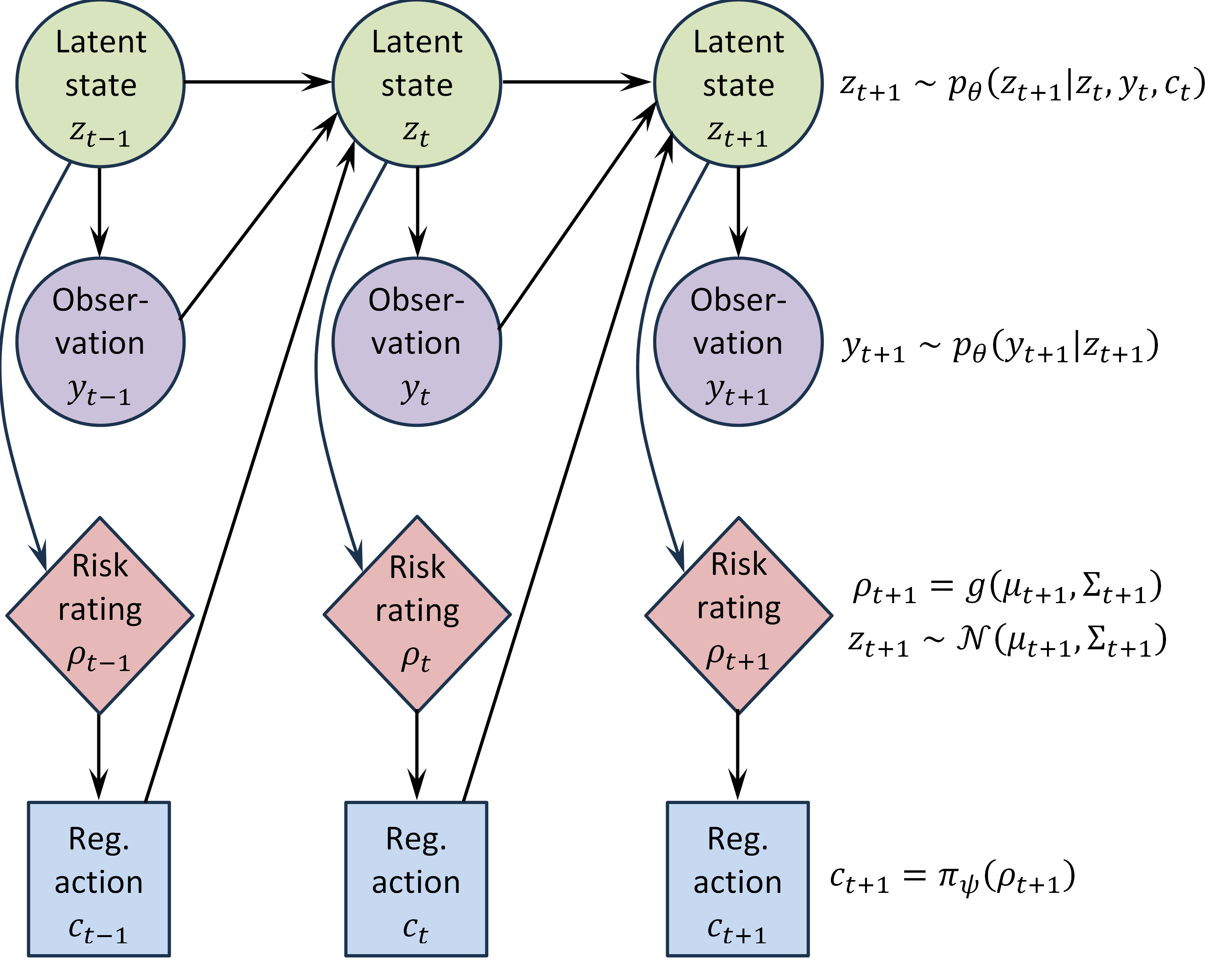}
    \caption{Simplified dynamic Bayesian network (DBN) representation of CENTRY, showing three consecutive time slices. The DBN encodes the first-order Markov assumption: each variable at time $t$ depends only on variables at times $t$ and $t-1$. Observations $y_t=(x_t,a_t)$ are generated from the latent state $z_t = \left(z_t^{(F)}, z_t^{(D)}, z_t^{(H)}, z_t^{(M)}\right)$. Component-risk ratings $\rho_t = (\rho_t^{(D)}, \rho_t^{(H)}, \rho_t^{(M)})$ are derived from the latent subspaces. Regulatory (Reg.) actions $c_t$ depend on $\rho_t$ and influence future states.}
    \label{fig:DBN}
\end{figure}

\subsection{The SELP cycle}
\label{subsec:The SELP cycle}
Sections~\ref{subsec:Structured Latent Representation} and \ref{subsec:Probabilistic generative model} defined the probabilistic latent-state and generative models underlying CENTRY. This subsection introduces the computational mechanism used to perform continual probabilistic inference and regulation within this model.

CENTRY is implemented by an AI agent that repeatedly executes four functions, \textbf{S}urprise$\rightarrow$\textbf{E}xplain$\rightarrow$ \textbf{L}earn$\rightarrow$\textbf{P}redict, referred to as the SELP cycle. The SELP cycle, originally proposed in \cite{BanerjeeDutta2014,banerjee2024multi}, has been applied to many problems, such as speech motor skill acquisition and speech production in developmental science \cite{najnin2016emergence,najnin2017predictive}, learning when and with whom to communicate \cite{kapourchali2020state,kapourchali2020epoc}, handwriting generation and recognition from images and videos \cite{baruah2020perception,baruah2022attention,banerjee2024attention}, interaction generation \cite{baruah2020multimodal}, intent recognition \cite{baruah2023intent,BanerjeeBaruah2024Sensors}, and speech emotion recognition \cite{baruah2022speech}. The four functions in SELP form a continually adapting closed-loop system, with Predict providing the forecast needed for the next Surprise computation. After the interaction at time $t$ is observed, Surprise quantifies the predictive mismatch, Explain infers the latent state and component-risk ratings, the regulation policy selects $c_t$, and Learn updates the probabilistic model and regulation policy. The consequences of $c_t$ are subsequently observed through future interactions and incorporated into later updates. See Figure~\ref{fig:CENTRY_block_diagram}.

\subsubsection{Surprise}
\label{subsubsec:Surprise}
The Predict stage produces the predictive distribution $p_{\mathrm{pred}}\!\left(y_t\mid y_{1:t-1},c_{1:t-1}\right)$. After the actual observable behavior $y_t=(x_t,a_t)$ is received, the Surprise stage computes the discrepancy between the predicted and actual behavior.

\paragraph{Definition 1 (Surprise).}
The population-level predictive mismatch at time $t$ is defined as
\begin{equation}
S_t^{\mathrm{(KL)}} = D_{\mathrm{KL}}\!\left(
 p_{\mathrm{true}}(\cdot\mid\mathcal F_t^{-,y})\;\middle\|
 p_{\mathrm{pred}}(\cdot\mid\mathcal F_t^{-,y})\right),
\label{eq:surprise}
\end{equation}
where $\mathcal F_t^{-,y}=\sigma(y_{1:t-1},c_{1:t-1})$ denotes the information available immediately before the interaction $y_t$ is generated. This population quantity depends on the unknown true conditional distribution and therefore is not directly computable by CENTRY at runtime. The operationally computable Surprise associated with the realized interaction is the predictive negative log-likelihood
\begin{equation}
\widehat S_t = -\log p_{\mathrm{pred}}\!\left(y_t\mid y_{1:t-1},c_{1:t-1}\right).
\label{eq:surprisenll}
\end{equation}
For discrete observations, this is equivalent to the KL divergence between a point mass at the realized interaction and the predictive distribution. For continuous observations, it is interpreted as the negative log predictive density rather than as a KL divergence from a point mass. Thus, $S_t^{\mathrm{(KL)}}$ characterizes population-level predictive mismatch, whereas $\widehat S_t$ is the quantity available to CENTRY for each observed interaction. A large value of $\widehat S_t$ indicates that the observed interaction is unexpected under the current predictive model and therefore warrants additional diagnostic or investigative attention. Surprise is not itself a risk label, since an unexpected interaction may be benign and a harmful interaction may be statistically unsurprising. The posterior component-risk estimates, rather than Surprise alone, are used to select regulatory actions.

\subsubsection{Explain}
\label{subsubsec:Explain}
The Explain stage infers the latent states that account for the observed
interaction from the interaction history $y_{1:t}$ and the history of previously
applied regulatory actions $c_{1:t-1}$. Explain is performed at each time step
and is not gated by a Surprise threshold. The component-risk states inferred
by Explain may correspond to either expected or unexpected interactions.

\centryproposition{prop:explain}{ (Explain)}
Given the interaction history and previously applied regulatory actions, Explain
computes the posterior
\begin{equation}
q_{\phi}\!\left(z_t\mid y_{1:t},c_{1:t-1}\right),
\label{eq:explainposterior}
\end{equation}
and computes the component-risk ratings
$\rho_t^{(D)}$, $\rho_t^{(H)}$, and $\rho_t^{(M)}$ from the inferred posterior component-risk states.

The parameters $\phi$ of Explain may be learned using the evidence-supported
behavior--cause records in $\mathcal{U}$, together with the continual
variational-learning procedure of the Learn stage. Consequently, $\mathcal{U}$
affects Explain through the learned parameters $\phi$, not as an additional
runtime input. At runtime, Explain implements the posterior-inference mapping
specified by Eq.~\eqref{eq:explainposterior}.

The component-risk estimates are compared with their corresponding normalized
risk budgets, and the regulation policy
$c_t=\pi_\psi(\rho_t)$, formally defined in
Section~\ref{subsubsec:Risk rating as a decision statistic}
(Proposition~\ref{prop:riskdecisionpolicy}, Eq.~\eqref{eq:riskdecisionpolicy}), determines the regulatory
action when regulation is warranted. The regulatory action acts on the
monitored AI system through its admissible intervention interface. Surprise
does not determine whether Explain is performed or whether an interaction is
risky; rather, it provides a complementary diagnostic of predictive mismatch
that may be used to allocate additional investigation, verification, or
adaptation effort.

In practice, the variational posterior may be parameterized by an amortized
inference model, such as a variational encoder. The output of the Explain stage
is the posterior Gaussian parameters
\[
\left\{\mu_t^{(j)},\Sigma_t^{(j)}\right\}_{j\in\{F,D,H,M\}},
\]
together with the corresponding component-risk estimates and, when warranted,
the regulatory action.

\subsubsection{Learn}
\label{subsubsec:Learn}
The Explain stage infers the latent posterior that best explains the observed prediction error and estimates the corresponding component risks. The Learn stage then continually updates both the probabilistic model and the regulation policy using the information accumulated during the current and previous SELP iterations. Thus, Learn performs joint continual adaptation of CENTRY's probabilistic inference model and its risk-regulation policy.

\centryproposition{prop:learn}{ (Learn)}
Let $\theta = \left(\theta_{\mathrm{gen}},\theta_{\mathrm{trans}}\right)$ be the parameters of the generative and transition models, respectively, $\phi=\theta_{\mathrm{inf}}$ be the parameters
of the variational inference model, and $\psi$ be the parameters of the regulation policy. The Learn stage updates both parameter sets continually. Given the posterior inferred by Explain, the model parameters are updated by maximizing
\begin{equation}
\mathcal L_t(\theta,\phi)
= \mathbb E_{q_\phi(z_t\mid y_{1:t},c_{1:t-1})} \left[\log p_\theta(y_t\mid z_t)\right]
- \beta D_{\mathrm{KL}}\!\left(q_\phi(z_t\mid y_{1:t},c_{1:t-1})\;\middle\|\;\kappa_t(z_t)\right),
\label{eq:ELBO}
\end{equation}
where $\kappa_t(z_t)$ is the predictive prior factorized in Eq.~\eqref{eq:jointpriorfactorization} and propagated according to Eq.~\eqref{eq:priorupdate} (Section~\ref{subsec:Updating the priors using SELP}) and $\beta>0$ controls the strength of latent regularization. The dependence of $\mathcal L_t(\theta,\phi)$ on $\beta$ is implicit; in results that distinguish different values of $\beta$, this objective is denoted by $\mathcal L_{t,\beta}$. The model parameters are updated according to
\begin{equation}
(\theta_{t+1},\phi_{t+1}) = \arg\max_{\theta,\phi} \mathcal L_t(\theta,\phi).
\label{eq:learnoperator}
\end{equation}
The argmax denotes an ideal update when a maximizer exists; otherwise approximate maximization is used. When updating the transition-model parameters, the dependence of the predictive prior on those parameters through Eq.~\eqref{eq:priorupdate} must be retained in the optimization, with the previous filtering posterior held fixed; otherwise the current variational objective provides no learning signal for the transition-model parameters.

The regulation-policy parameters are updated continually using the current component-risk estimates and the model-based prospective consequences of candidate regulatory actions. Thus, Learn modifies both
CENTRY's probabilistic model and its regulation policy, with the updated parameters used in the subsequent SELP iteration.

\subsubsection{Predict}
\label{subsubsec:Predict}
The Predict stage generates the next observable behavior using the updated probabilistic model obtained from Learn. It therefore implements the generative model introduced in Section~\ref{subsec:Probabilistic generative model}.

\centryproposition{prop:predict}{ (Predict)}
Given the posterior
$q_\phi(z_t\mid y_{1:t},c_{1:t-1})$, the selected regulatory action $c_t$, and the updated model parameters $\theta_{t+1}$, the Predict stage computes
\begin{equation}
p_{\mathrm{pred}}\!\left(y_{t+1}\mid y_{1:t},c_{1:t}\right),
\label{eq:predictoperator}
\end{equation}
as defined in Eq.~\eqref{eq:predictivedistribution}. In the implementation of Algorithm~\ref{alg:CENTRY}, Predict uses the posterior computed by Explain with $\phi_t$ before Learn, together with the updated transition and observation models parameterized by $\theta_{t+1}$. The posterior is retained for propagation; the updated inference parameters $\phi_{t+1}$ are used at the next Explain stage. The expected future behavior is
\begin{equation}
\hat{y}_{t+1} = (\hat{x}_{t+1},\hat{a}_{t+1})
= \mathbb{E}\!\left[y_{t+1}\mid y_{1:t},c_{1:t}\right].
\label{eq:expectedbehavior}
\end{equation}
The predicted behavior is supplied to the next Surprise stage, where it is compared with the subsequently observed behavior $y_{t+1}$. This closes one iteration of the SELP cycle.

\subsection{Continual Bayesian risk estimation}
\label{subsec:Continual Bayesian risk estimation}
The structured latent representation introduced in Section~\ref{subsec:Structured Latent Representation} provides a probabilistic description of the hidden state of the monitored AI system. However, the latent variables themselves are not directly interpretable as quantitative measures of AI risk. This subsection introduces a Bayesian risk-estimation framework that converts the latent component-risk states into interpretable risk ratings suitable for continual monitoring and subsequent regulation.

\subsubsection{Component risk ratings}
\label{subsubsec:Component risk ratings}
Recall that the latent state is partitioned as $z_t=(z_t^{(F)},z_t^{(D)},z_t^{(H)},z_t^{(M)})$, where the three slow latent subspaces correspond to the dataset, hardware, and model/algorithm components of the monitored AI system. CENTRY summarizes each latent risk subspace by a scalar risk rating
$\rho_t^{(j)}$, $j\in\{D,H,M\}$.

\paragraph{Definition 2 (Component Risk Rating).}
The component risk rating is defined as a deterministic functional of the corresponding latent posterior,
\begin{equation}
\rho_t^{(j)} = g^{(j)}\!\left(\mu_t^{(j)},\Sigma_t^{(j)}\right),\qquad j\in\{D,H,M\},
\label{eq:riskfunctional}
\end{equation}
where $g^{(j)}:\mathbb R^{d_j}\times\mathbb S_+^{d_j}\rightarrow [0,1]$ maps the posterior mean and covariance of the corresponding latent component to a normalized risk score. The specific choice of $g^{(j)}(\cdot)$ requires application-specific calibration, informed by the component-risk evidence discussed in Section~\ref{sec:Component Risk Priors}; an illustrative choice is given in Section~\ref{sec:example}.

\subsubsection{Risk rating as a decision statistic}
\label{subsubsec:Risk rating as a decision statistic}
The latent variable contains substantially richer information than the scalar risk rating. It captures the uncertainty and evidence accumulated regarding the corresponding component, whereas the risk rating provides a concise summary for regulatory decision making.

\centryproposition{prop:riskdecisionpolicy}{}
CENTRY restricts its regulatory policy to depend on the component risk ratings,
\begin{equation}
c_t = \pi_\psi\!\left(\rho_t^{(D)}, \rho_t^{(H)}, \rho_t^{(M)}\right).
\label{eq:riskdecisionpolicy}
\end{equation}
Thus, the regulatory decision is mediated by the mapping
\[
z_t^{(j)} \longrightarrow \rho_t^{(j)} \longrightarrow c_t,
\]
establishing a principled separation between probabilistic inference and adaptive regulatory action.

\subsubsection{Continual Bayesian updating}
\label{subsubsec:Continual Bayesian updating}
Since the posterior distribution is updated after every newly received behavior,
\begin{equation*}
q(z_t\mid y_{1:t},c_{1:t-1}) \longrightarrow q(z_{t+1}\mid y_{1:t+1},c_{1:t}),
\end{equation*}
the corresponding risk ratings evolve continually, $\rho_t^{(j)} \longrightarrow \rho_{t+1}^{(j)}$, allowing CENTRY to maintain continuously updated estimates of the AI system's component risks throughout its operational lifetime.

\subsection{Adaptive risk regulation}
\label{subsec:Adaptive risk regulation}
The continual Bayesian risk estimates derived in the previous subsection are used to select regulatory actions that influence the future behavior of the monitored AI system. Because the monitored AI system is treated as a black box, CENTRY does not directly control its external action $a_t$. Instead, the regulatory action $c_t$ is applied through an admissible intervention interface that modifies the utilization or configuration of the system's datasets, hardware resources, or models/algorithms.

\subsubsection{Risk regulation policy}
\label{subsubsec:Risk regulation policy}
Let $\bar B^{(D)},\bar B^{(H)},\bar B^{(M)}\in[0,1]$ be the normalized component-risk thresholds corresponding to the component risk budgets $B^{(D)}, B^{(H)}, B^{(M)}$. Let $\rho_t = \left(\rho_t^{(D)},\rho_t^{(H)},\rho_t^{(M)}\right)$ be the vector of current component-risk ratings. The regulatory policy is defined by Eq.~\eqref{eq:riskdecisionpolicy} as $c_t = \pi_{\psi_t}(\rho_t)$, where $c_t=(c_t^{(D)},c_t^{(H)},c_t^{(M)})$ is the regulatory action selected by CENTRY and $\psi_t$ denotes the current policy parameters. For a candidate policy $\pi_\psi$, let
\begin{equation*}
\widehat p_{t,k}^{(j)}(\psi)
:=
\widehat{\Pr}_{\mathrm{CENTRY},\pi_\psi}
\!\left(R_{t+k}^{(j)}>B^{(j)}\mid\mathcal F_t\right)
\end{equation*}
be the corresponding calibrated model-based estimate of the future probability that component $j$ violates its operational risk budget at horizon $k$, where $\mathcal F_t=\sigma(y_{1:t},c_{1:t-1})$ denotes the information available to CENTRY immediately before selecting $c_t$, as in Section~\ref{sec:Guarantees of CENTRY}. The regulation policy is learned continually as part of the Learn stage by minimizing the expected future surprise, component risk, and regulatory-action cost subject to calibrated finite-horizon chance constraints:
\begin{equation}
\psi_{t+1} = \arg\min_{\psi}\;\mathbb{E}_{p_{\theta_t},\,\pi_{\psi}}\Bigg[\lambda_S \widehat S_{t+1}
+ \sum_{j\in\{D,H,M\}}\lambda_j \left(\rho_{t+1}^{(j)}-\bar B^{(j)}\right)_+ + \lambda_C C(\pi_\psi(\rho_t))\Bigg],
\label{eq:policylearning}
\end{equation}
subject to
\begin{equation*}
\widehat p_{t,k}^{(j)}(\psi)\le \delta_k^{(j)}-\varepsilon_{\mathrm{cal}},
\qquad j\in\{D,H,M\}, \quad k=1,\ldots,T, \quad \pi_\psi(\rho_t)\in\mathcal C,
\end{equation*}
where $\lambda_S,\lambda_j,\lambda_C\ge0$ weight predictive surprise, component risk, and regulatory-action cost, respectively; $(\cdot)_+=\max(\cdot,0)$; $\mathcal C$ is the set of admissible regulatory actions; $\delta_k^{(j)}$ is the allocated violation probability for component $j$ at horizon $k$; and $\varepsilon_{\mathrm{cal}}$ is the one-sided calibration margin used in the finite-horizon guarantee. The cost $C(\pi_\psi(\rho_t))$ is a non-negative function representing the operational penalty associated with the regulatory action selected by $\pi_\psi$ from the current risk ratings, which may include latency, computational overhead, or task-performance degradation. The allocated violation probabilities must each be at least the calibration margin and their sum across components and horizons must not exceed the total permitted violation probability in Eq.~\eqref{eq:problem_constraint}. The ideal minimizations assume attainment; approximate updates must preserve feasibility whenever the safety guarantee is invoked. The calibrated constraints must hold for the deployed policy, including its prospective future actions, not merely for a policy obtained after a subsequent parameter update. If no admissible policy meets these conditions, the finite-horizon safety guarantee does not apply.

Equation~\eqref{eq:policylearning} specifies the ideal constrained policy-learning objective. For online implementation, this objective may be optimized approximately using projected online gradient descent over a nonempty compact convex feasible policy class $\Psi$, as assumed in the regret results of Section~\ref{sec:Guarantees of CENTRY}. The finite-horizon safety result requires that the deployed policy satisfy the calibrated probability constraints, whereas the regret results separately characterize the performance of the online optimization procedure.

Because $c_t$ is selected before $y_{t+1}$ is observed, $\widehat S_{t+1}$ and $\rho_{t+1}^{(j)}$ are prospective quantities in Eq.~\eqref{eq:policylearning}. Their expectations are evaluated over trajectories generated by the current probabilistic model $p_{\theta_t}$ under candidate policies, with future latent states propagated from the current posterior through the transition model. The same predictive trajectories are used to evaluate the calibrated violation-probability constraints above. Although the objective uses a one-step lookahead for tractability, the chance constraints can be evaluated over any finite horizon $T$. The finite-horizon guarantee in Section~\ref{sec:Guarantees of CENTRY} applies for any such $T$ for which the calibrated probability constraints are maintained and the stated assumptions hold. Thus, policy learning does not require access to the internal state or decision function of the monitored AI system, and the regulation policy is adapted online as part of every SELP iteration rather than being fixed before deployment.

When reference regulatory actions are available, the policy may alternatively be fitted using supervised learning:
\begin{equation}
\psi_{t+1} = \arg\min_{\psi}\sum_{\tau=1}^{t}\ell\!\left(\pi_{\psi}(\rho_\tau),
c_\tau^{(\mathrm{ref})}\right),
\label{eq:supervisedpolicylearning}
\end{equation}
where the argmin denotes an ideal update when a minimizer exists (otherwise approximate minimization is used), $\ell(\cdot,\cdot)$ is an appropriate action-space loss and $c_\tau^{(\mathrm{ref})}$ denotes a reference regulatory action, such as one obtained by solving the constrained optimization problem above. The CENTRY-generated pairs $(\rho_t,c_t)$ are training data only when an independent reference action is available; otherwise, policy learning is governed by the model-based objective in Eq.~\eqref{eq:policylearning}.

\subsubsection{Component-specific regulation}
\label{subsubsec:Component-specific regulation}
Each component-specific regulatory action modifies only the corresponding component through the admissible intervention interface. We further restrict $\pi_\psi$ to factorize into independent per-component sub-policies -- a design choice for interpretability, not a consequence of the intervention
structure alone -- so that
\begin{equation}
c_t^{(j)} = \pi_\psi^{(j)}\!\left(\rho_t^{(j)}\right),\qquad j\in\{D,H,M\}.
\label{eq:componentpolicy}
\end{equation}
Examples include restricting access to high-risk datasets, reducing computational resources, replacing a model with a safer alternative, constraining high-risk capabilities, or requiring additional safeguards before high-risk operations are executed.

Importantly, $c_t$ does not directly specify the monitored AI system's action $a_t$. Instead, it modifies the system in a manner that changes the probability distribution of its future observable behavior. Consequently, the causal relationship represented by CENTRY is
\[
c_t \longrightarrow \text{AI-system configuration} \longrightarrow (x_{t+1},a_{t+1}),
\]
while the internal mechanism implementing this relationship remains unknown to CENTRY.

\subsubsection{Closed-loop operation}
\label{subsubsec:Closed-loop operation}
Combining continual Bayesian estimation with adaptive regulation yields the closed-loop system
\begin{equation*}
y_t \rightarrow q(z_t) \rightarrow \rho_t \rightarrow c_t \rightarrow \text{monitored AI system/environment} \rightarrow y_{t+1},
\end{equation*}
where $y_t=(x_t,a_t)$. The posterior distribution and corresponding component-risk ratings are continually updated from the observed interaction history. The regulatory policy selects $c_t=\pi_\psi(\rho_t)$, which acts through the admissible intervention interface and thereby influences future observable behavior.

CENTRY therefore does not require direct access to the internal policy or state-transition mechanism of the monitored AI system. It only requires an observable interaction channel through which the behavior $(x_t,a_t)$ can be monitored and an admissible intervention channel through which the regulatory action $c_t$ can modify the system's components.

The resulting closed-loop system performs continual AI risk estimation, monitoring, and regulation simultaneously.

\paragraph{Examples of Regulatory Actions.}
The regulatory action selected by CENTRY does not directly prescribe the external action of the monitored AI system. Instead, it modifies the configuration or utilization of the AI system's underlying components so that future interactions are expected to satisfy the prescribed risk budget.

For the dataset component, representative regulatory actions include removing or restricting access to high-risk datasets, lowering the retrieval probability of unreliable documents in retrieval-augmented generation systems, rejecting newly acquired training data, replacing a dataset with a curated alternative, or reducing the influence of a dataset during continual learning.

For the hardware component, regulatory actions include limiting the computational resources available to the AI system, such as reducing the maximum inference compute budget, restricting memory usage, shortening the planning horizon, limiting context length, or disabling distributed computation.

For the model or algorithm component, regulatory actions include switching to a safer model, enabling additional safety filters, reducing decoding temperature, disabling high-risk capabilities such as autonomous tool use or code execution, replacing a reinforcement-learning policy with a more conservative policy, or requiring human approval before executing high-risk decisions.

These examples illustrate that CENTRY regulates \emph{how} the monitored AI system utilizes its underlying resources rather than directly prescribing its external actions. The effectiveness of a regulatory action is inferred from its subsequent effect on the observable interaction sequence and the estimated component risks.

\section{Component Risk Priors}
\label{sec:Component Risk Priors}
The Bayesian risk estimation framework developed in Section~\ref{sec:A Probabilistic Model for Continual AI Risk Estimation} continually updates the component-specific risk states based on the observed interactions between the monitored AI system and its environment. Such continual estimation requires an initial probabilistic belief regarding the safety of each component before any interaction has been observed.

This section introduces \emph{component risk priors}. Unlike conventional AI risk rating systems, these priors do not estimate risk directly. Instead, they encode prior knowledge regarding the expected safety of datasets, hardware resources, and models/algorithms before deployment. During operation, these prior beliefs are continually refined by the SELP cycle through Bayesian inference.

Formally, \(z_t = \left(z_t^{(F)}, z_t^{(D)}, z_t^{(H)}, z_t^{(M)}\right)\) is the structured latent state, as introduced in Section~\ref{subsec:Structured Latent Representation}. Before any interaction is observed, CENTRY assumes prior distributions $p(z^{(D)})$, $p(z^{(H)})$, $p(z^{(M)})$, which encode the initial risk beliefs associated with the corresponding AI components. These priors constitute the initial conditions for continual Bayesian risk estimation.

\subsection{Dataset risk priors}
\label{subsec:Dataset risk priors}
The dataset used to train or adapt an AI system strongly influences its subsequent behavior. Consequently, CENTRY assigns each dataset an initial probabilistic risk prior before deployment.

This process is conceptually analogous to the motion picture rating systems used in many countries. Rather than predicting the behavior of a specific audience, a rating board evaluates the content of a movie and assigns a rating that reflects its expected suitability for different audiences. Likewise, CENTRY evaluates properties of a dataset before deployment and assigns an initial probabilistic belief regarding its potential contribution to AI risk.

Dataset documentation, including datasheets~\cite{DatasheetsForDatasets}, can provide evidence about provenance and source reliability, licensing and reuse restrictions, demographic representativeness, data quality, dataset versioning, harmful or unsafe content, privacy-sensitive information, and suitability for the intended application. Additional assessments may address poisoning susceptibility and the adequacy of security and privacy protection mechanisms, including anonymization and de-identification. These properties and assessments inform the prior distribution $p(z^{(D)})$, which serves as the initial belief regarding the dataset-related latent risk state.

The specific methodology used to construct the prior may vary across application domains. CENTRY requires only that the resulting prior distribution encode the relative expected risk of the dataset before deployment.

\subsection{Hardware risk priors}
\label{subsec:Hardware risk priors}
The computational resources available to an AI system directly influence the complexity of the behaviors that the system can generate. Consequently, CENTRY associates an initial probabilistic risk prior with the hardware resources allocated to the monitored AI system.

Rather than viewing computational resources as inherently unsafe, CENTRY treats risk as application dependent. High computational capability may be appropriate for safety-critical domains such as healthcare while being unnecessary or undesirable for applications whose societal risks outweigh the expected benefits.

Examples of hardware-related factors include computational throughput, available memory, distributed computation, online learning capability, autonomous planning depth, and available inference budget. These characteristics define the prior $p(z^{(H)})$, which represents the initial Bayesian belief regarding hardware-related risk before deployment.

\subsection{Model risk priors}
\label{subsec:Model risk priors}
Different machine learning algorithms exhibit different safety properties. Consequently, CENTRY assigns an initial probabilistic prior to the model or algorithm used by the monitored AI system. Unlike datasets and hardware resources, whose properties are largely external to the model or algorithm itself, the model prior reflects the intrinsic characteristics of the adopted AI methodology.

Examples include interpretability, robustness, uncertainty calibration, formal verification, controllability, evidence of alignment, and adversarial robustness, as well as known or anticipated risk factors such as poor generalization, bias and discrimination, confabulation, data leakage, privacy concerns, performance degradation, model drift or collapse, lack of explanations, unsafe outputs, and failures inherited through model integration.  These characteristics define the prior $p(z^{(M)})$, which encodes the initial Bayesian belief regarding the model-related latent risk state before deployment.

\subsection{Updating the priors using SELP}
\label{subsec:Updating the priors using SELP}
The component-risk priors introduced above provide CENTRY's initial beliefs before interaction evidence is available. During operation, these beliefs are continually refined through Bayesian inference and temporal propagation.

Let $\kappa_t(z_t^{(j)})$ be the predictive prior for latent component $j$ at time $t$, and let
\[
q_t(z_t^{(j)}) \equiv q_\phi\!\left(z_t^{(j)}\mid y_{1:t},c_{1:t-1}\right)
\]
be its posterior marginal after the current behavior has been observed. Under the component-wise conditional factorization of the variational posterior and transition model, with a factorized initial prior, the predictive prior over the structured latent state factorizes as
\begin{equation}
\kappa_t(z_t) = \prod_{j\in\{F,D,H,M\}}\kappa_t\!\left(z_t^{(j)}\right).
\label{eq:jointpriorfactorization}
\end{equation}

After inference at time $t$, the posterior is propagated through the learned transition model to obtain the predictive prior at time $t+1$:
\begin{equation}
\kappa_{t+1}\!\left(z_{t+1}^{(j)}\right) = 
\int p_\theta\!\left(z_{t+1}^{(j)}\mid z_t^{(j)},y_t,c_t\right)
q_t\!\left(z_t^{(j)}\right)\,dz_t^{(j)}, \qquad j\in\{F,D,H,M\}.
\label{eq:priorupdate}
\end{equation}

The fast-interaction prior is initialized as $\kappa_1(z_1^{(F)})=\mathcal N(0,I)$, while $\kappa_1(z_1^{(D)})$, $\kappa_1(z_1^{(H)})$, and $\kappa_1(z_1^{(M)})$ are specified using the component-specific prior methodologies above. All four priors are subsequently propagated according to Eq.~\eqref{eq:priorupdate}.

Thus, continual Bayesian estimation proceeds recursively as
\[
\text{Prior}
\rightarrow
\text{Posterior}
\rightarrow
\text{Propagated Prior}
\rightarrow
\text{Posterior}
\rightarrow
\cdots.
\]
The posterior incorporates the newly observed behavior, while the transition model propagates that information forward to form the next predictive prior.

\section{Guarantees of CENTRY}
\label{sec:Guarantees of CENTRY}

This section organizes the formal guarantees of CENTRY into five groups:
\begin{itemize}[nosep]
    \item \emph{Foundational guarantees} (Section \ref{subsec:Foundational guarantees}) establish the causal and safety structure required for valid operation (Theorems \ref{thm:non_anticipatory_regulation}--\ref{thm:asymptotic_risk_stability}, Corollaries \ref{cor:uniform-risk-allocation}--\ref{cor:almost_sure_safety}). 
    \item \emph{Learning guarantees} (Section \ref{subsec:Learning guarantees}) characterize continual adaptation of the variational model and regulatory policy, including non-stationary tracking and stationary variational learning (Theorems \ref{thm:predictive_evidence_bound}--\ref{thm:policy_improvement}, Corollaries \ref{cor:vanishing_surprise}--\ref{cor:strongly_convex_policy_convergence}). 
    \item \emph{Inference guarantees} (Section \ref{subsec:Inference guarantees}) address latent-state estimation and uncertainty (Theorems \ref{thm:surprise_elbo_relation}--\ref{thm:steady_state_riccati}, Corollaries \ref{cor:surprise_elbo_general_beta}--\ref{cor:scalar_steady_state_variance}). 
    \item \emph{Operational guarantees} (Section \ref{subsec:Operational guarantees}) quantify expected risk reduction, detection, false alarms, containment, cumulative excursion exposure, stabilization, and computational cost (Lemma \ref{lem:foster_hitting_time}, Theorems \ref{thm:risk_reduction_bound}--\ref{thm:posterior_stabilization}, Corollaries \ref{cor:detection_horizon}--\ref{cor:selp_complexity}). 
    \item \emph{System-level guarantees} (Section \ref{subsec:System-level guarantees}) combine component-level properties into Pareto-optimal regulation, budget-compliance, and feasibility results (Theorems \ref{thm:pareto_regulation}--\ref{thm:feasible_invariance}, Corollaries \ref{cor:uniform_integrability}--\ref{cor:no_intervention_invariance_sufficient}). 
\end{itemize}
Throughout, the conclusions are conditional on the assumptions stated in each result; they should not be interpreted as unconditional guarantees under arbitrary model misspecification, unobservable dynamics, infeasible control actions, or violations of the stated probabilistic assumptions. The symbol $\widehat S_t$ denotes the realized, operationally computed predictive surprise used by CENTRY in its model-based regulation objectives; the population-level predictive mismatch is denoted separately by $S_t^{\mathrm{(KL)}}$. The regulatory cost $C(c)$ is nonnegative for admissible regulatory actions.

For readability, the statements below present the formal guarantees without their proofs. A brief interpretation follows each result; the complete proofs and the technical remarks supporting them are provided in Appendix~\ref{app:Proofs of Theorems and Corollaries}.

\subsection{Foundational guarantees}
\label{subsec:Foundational guarantees}
These results establish the causal and safety structure of CENTRY. Some assumptions are structural and normally satisfied by an implemented probabilistic system (such as measurability and valid probability densities); the stronger assumptions---calibration, feasible interventions, effective regulation, and Lyapunov dissipativity---are empirical or design conditions that must be established for a particular deployment.

\begin{styledtheorem}{Non-Anticipatory Regulation}{non_anticipatory_regulation}
Let \(\mathcal{F}_t = \sigma\!\left(y_{1:t},c_{1:t-1}\right)\) be the information available to CENTRY at time $t$, where $y_{1:t}$ denotes the observable interaction history through time $t$ and $c_{1:t-1}$ denotes the sequence of regulatory actions taken before time $t$. Let the CENTRY risk-estimation operator produce the current risk-rating vector
\[
\rho_t = \mathcal{R}_t\!\left(y_{1:t},c_{1:t-1}\right),
\]
where \(\mathcal{R}_t\) is measurable, and let the regulatory policy be
\[
c_t = \pi_{\psi_t}(\rho_t),
\qquad
\psi_t = \Psi_t(y_{1:t},c_{1:t-1}),
\]
where \(\Psi_t\) is measurable. Assume that the joint mapping
\(\left(\psi,\rho\right)\mapsto\pi_\psi(\rho)\) is measurable on the product of the policy-parameter space and the risk-rating space. Then the regulatory action $c_t$ is \(\mathcal{F}_t\)-measurable. Consequently, CENTRY's regulatory policy is non-anticipatory: the action selected at time $t$ depends only on information available through time $t$ and cannot depend on future interactions $y_{t+1},y_{t+2},\ldots$ or future regulatory actions $c_{t+1},c_{t+2},\ldots$.
\end{styledtheorem}

\noindent\textbf{Interpretation.} The property established here is causal, non-anticipatory regulation: CENTRY chooses the current regulatory action only from information available at that time. This is important because a runtime regulator must act before future behavior is known, rather than using information retrospectively. The assumptions are essentially measurability and implementable mappings, which are structural and ordinarily satisfied by a well-defined software system; the result therefore does not require a difficult empirical safety assumption.

\begin{styledtheorem}{Finite-Horizon Probabilistic Risk Guarantee}{finite_horizon_risk_guarantee}
Let
\(
\mathcal{F}_t
=
\sigma\!\left(y_{1:t},c_{1:t-1}\right)
\)
denote the information available to CENTRY immediately before selecting
the regulatory action \(c_t\). Let
\(j\in\{D,H,M\}\) denote a risk source, and let
\(B^{(j)}\) be its corresponding acceptable risk budget. Consider a finite
prediction horizon \(T\in\mathbb{N}\) and an active regulatory policy
\(\pi_\psi\). Let \(\Pr_{\pi_\psi}\) denote the probability measure
induced by execution of the policy. The notation for estimated violation probabilities suppresses their dependence on the active policy, as defined in Section~\ref{subsubsec:Risk regulation policy}.

Suppose that, for every
\(j\in\{D,H,M\}\) and \(k=1,\ldots,T\):

\begin{enumerate}[nosep]

    \item \textbf{Calibrated violation-probability estimate.}
    CENTRY computes a conditional estimated probability
    \[
    \widehat p_{t,k}^{(j)}
    :=
    \widehat{\Pr}_{\mathrm{CENTRY}}
    \!\left(
        R_{t+k}^{(j)}>B^{(j)}
        \,\middle|\,
        \mathcal{F}_t
    \right),
    \]
    and $\widehat p_{t,k}^{(j)}$ is $\mathcal F_t$-measurable. Assume there exists a calibration error
    \(\varepsilon_{\mathrm{cal}}\geq0\) such that
    \[
    \Pr_{\pi_\psi}\!\left(
        R_{t+k}^{(j)}>B^{(j)}
        \,\middle|\,
        \mathcal{F}_t
    \right)
    \leq
    \widehat p_{t,k}^{(j)}
    +
    \varepsilon_{\mathrm{cal}}
    \qquad\text{a.s.}
    \]

    \item \textbf{Chance-constrained regulation.}
    The regulatory policy satisfies
    \[
    \widehat p_{t,k}^{(j)}
    \leq
    \delta_k^{(j)}-\varepsilon_{\mathrm{cal}},
    \]
    where
    \(\delta_k^{(j)}\geq\varepsilon_{\mathrm{cal}}\).

    \item \textbf{Failure-probability allocation.}
    The allocated violation probabilities satisfy
    \[
    \sum_{k=1}^{T}
    \sum_{j\in\{D,H,M\}}
    \delta_k^{(j)}
    \leq
    \delta,
    \qquad
    0<\delta<1.
    \]

\end{enumerate}

Then the joint finite-horizon risk constraint
\[
\Pr_{\pi_\psi}\!\left(
R_{t+k}^{(j)}\leq B^{(j)},
\;
\forall j\in\{D,H,M\},
\;
\forall k=1,\ldots,T
\,\middle|\,
\mathcal{F}_t
\right)
\geq
1-\delta
\]
is satisfied.

In particular, when
\(\varepsilon_{\mathrm{cal}}=0\), the result reduces to the
component-wise chance-constrained guarantee used in the original
formulation.
\end{styledtheorem}

\noindent\textbf{Interpretation.} The property established here is a finite-horizon probabilistic safety guarantee for the true component risks. This is important because it connects CENTRY's estimated violation probabilities to an explicit probability that all component budgets will be respected over an entire future horizon. The essential assumptions are calibrated one-sided violation probabilities, feasible chance constraints, and a valid probability allocation; these are not automatic in real deployments and must be empirically calibrated and shown feasible for the particular system.

\begin{styledcorollary}{Uniform Allocation of the Failure Probability}{uniform-risk-allocation}

Under the assumptions of Theorem~\ref{thm:finite_horizon_risk_guarantee}, suppose the failure
probability is allocated uniformly across the three components and the \(T\)-step horizon:
\(
\delta_k^{(j)}
=
\frac{\delta}{3T},
j\in\{D,H,M\},
k=1,\ldots,T.
\)
If the policy satisfies the calibrated probability constraints
\(
\widehat p_{t,k}^{(j)}
\leq
\frac{\delta}{3T}-\varepsilon_{\mathrm{cal}},
\)
for every $j$ and $k$, with
$\varepsilon_{\mathrm{cal}}\leq\delta/(3T)$, then
\[
\Pr_{\pi_\psi}\!\left(
R_{t+k}^{(j)}\leq B^{(j)},
\;
\forall j\in\{D,H,M\},
\;
\forall k\in\{1,\ldots,T\}
\,\middle|\,
\mathcal{F}_t
\right)
\geq
1-\delta.
\]
\end{styledcorollary}

\noindent\textbf{Interpretation.} The property established here is a simple uniform way to allocate the total allowed failure probability across all three components and all future steps. This is important because it makes the finite-horizon safety constraint immediately usable when there is no principled reason to favor one component or horizon over another. The assumption is mainly a conservative design choice inherited from the parent theorem; uniform allocation is easy to implement but need not be efficient, and in practice allocations can be made component- and horizon-specific when better information is available.

\begin{styledcorollary}{System-Level Finite-Horizon Risk Guarantee}{system_level_risk_guarantee}
Under the assumptions of Theorem \ref{thm:finite_horizon_risk_guarantee}, suppose that the operational decomposition of the true system risk satisfies
\(
R_{t+k} \leq \sum_{j\in\{D,H,M\}} R_{t+k}^{(j)} + \varepsilon
\)
for some bounded interaction error $\varepsilon \geq 0$, for all $k=1,\ldots,T$. Let $B = B^{(D)} + B^{(H)} + B^{(M)}$ denote the total allowable risk budget. Then, the joint finite-horizon safety event implies that the total system risk remains bounded with high probability:
\[
\Pr_{\pi_\psi}\!\left(
R_{t+k} \leq B + \varepsilon,
\;
\forall k=1,\ldots,T
\,\middle|\,
\mathcal{F}_t
\right)
\geq 1-\delta.
\]
\end{styledcorollary}

\noindent\textbf{Interpretation.} The property established here transfers the component-wise finite-horizon guarantee to a bound on total system risk. This is important because stakeholders often care about aggregate system risk rather than three separate component budgets. The key additional assumption is that total risk is bounded by the sum of component risks plus a bounded interaction term; this can be a useful engineering approximation, but complex interacting systems may require the interaction term to be estimated explicitly rather than assumed small.

\begin{styledtheorem}{Validity of propagated component priors}{propagated-prior-validity}

Suppose that \(q_t(z_t^{(j)})\) is a probability density and that
\(
p_\theta\!\left(
z_{t+1}^{(j)}
\mid
z_t^{(j)},y_t,c_t
\right)
\)
is a conditional probability density for each
\(j\in\{F,D,H,M\}\). Then the propagated prior defined by
Eq.~\eqref{eq:priorupdate},
\[
\kappa_{t+1}\!\left(z_{t+1}^{(j)}\right)
=
\int
p_\theta\!\left(
z_{t+1}^{(j)}
\mid
z_t^{(j)},y_t,c_t
\right)
q_t\!\left(z_t^{(j)}\right)
\,dz_t^{(j)},
\]
is a probability density. In particular,
\(
\kappa_{t+1}\!\left(z_{t+1}^{(j)}\right)\geq 0
\)
and
\(
\int
\kappa_{t+1}\!\left(z_{t+1}^{(j)}\right)
\,dz_{t+1}^{(j)}
=
1.
\)
\end{styledtheorem}

\noindent\textbf{Interpretation.} The property established here is probabilistic validity of CENTRY's propagated prior: Bayesian prediction preserves nonnegativity and unit mass. This is important because every subsequent posterior and risk rating depends on having a legitimate predictive distribution. The assumptions are basic probability-density conditions on the current posterior and transition kernel and are normally satisfied by a correctly specified probabilistic implementation.

\begin{styledcorollary}{Gaussian propagation under affine Gaussian transitions}{gaussian-propagation}

Suppose, for a component \(j\in\{F,D,H,M\}\), that
\(
q_t\!\left(z_t^{(j)}\right)
=
\mathcal N\!\left(
z_t^{(j)};
\mu_t^{(j)},\Sigma_t^{(j)}
\right)
\)
and that the transition model is affine Gaussian,
\(
z_{t+1}^{(j)}
=
A_t^{(j)}z_t^{(j)}
+
b_t^{(j)}
+
\varepsilon_t^{(j)},
\varepsilon_t^{(j)}
\sim
\mathcal N\!\left(
0,Q_t^{(j)}
\right),
\)
where \(\varepsilon_t^{(j)}\) is independent of \(z_t^{(j)}\). Then the
propagated prior in Eq.~\eqref{eq:priorupdate} is Gaussian:
\[
\kappa_{t+1}\!\left(z_{t+1}^{(j)}\right)
=
\mathcal N\!\left(
z_{t+1}^{(j)};
A_t^{(j)}\mu_t^{(j)}+b_t^{(j)},
A_t^{(j)}\Sigma_t^{(j)}(A_t^{(j)})^\top+Q_t^{(j)}
\right).
\]
\end{styledcorollary}

\noindent\textbf{Interpretation.} The property established here is closed-form Gaussian propagation for an affine-Gaussian latent transition. This is important because it gives CENTRY a computationally simple and analytically transparent way to propagate uncertainty through time. Affine-Gaussian dynamics are common tractable state-space models, but they are an idealization for complex deployed AI systems; the result is therefore most directly applicable when those dynamics are a good approximation or as a local/benchmark model.

\begin{styledtheorem}{Monotonicity of a threshold-based component policy}{monotone-policy}

Fix a component \(j\in\{D,H,M\}\). Suppose the admissible component actions are
totally ordered by intervention strength,
\(
c^{(0)} \preceq c^{(1)} \preceq \cdots \preceq c^{(L)},
\)
and the component policy \(\pi_\psi^{(j)}:[0,1]\to\mathcal C^{(j)}\) is
constant on each interval of an ordered threshold partition of $[0,1]$, with nondecreasing intervention tiers across successive intervals. Then, for any
\(0\leq\rho_1\leq\rho_2\leq1\),
\[
\pi_\psi^{(j)}(\rho_1)
\preceq
\pi_\psi^{(j)}(\rho_2).
\]
Thus, increasing the estimated risk rating cannot select a weaker regulatory
intervention.
\end{styledtheorem}

\noindent\textbf{Interpretation.} The property established here is monotonicity of component-specific regulation: a larger estimated risk cannot lead to a weaker intervention. This is important because it makes the safety logic of the policy predictable and prevents an unintentionally perverse response to increasing risk. The required threshold ordering is a design property that can be enforced directly in a rule-based policy and is therefore realistic when monotone regulation is desired, but it is not guaranteed for an unconstrained learned policy.

\begin{styledtheorem}{Infinite-Horizon Ultimate Boundedness of Risk Violations}{asymptotic_risk_stability}
Let $\rho_t =
\left(\rho_t^{(D)},\rho_t^{(H)},\rho_t^{(M)}\right)^\top
\in[0,1]^3$ denote the vector of component-risk ratings at time $t$,
and let
$\bar B =
\left(\bar B^{(D)},\bar B^{(H)},\bar B^{(M)}\right)^\top$
be the vector of normalized component-risk budgets. Define the
quadratic budget-violation Lyapunov function
\[
V(\rho_t)
=
\sum_{j\in\{D,H,M\}}
\left(
\left(\rho_t^{(j)}-\bar B^{(j)}\right)_+
\right)^2,
\]
where $(x)_+=\max(0,x)$. Let
\(
\mathcal{F}_t=\sigma(y_{1:t},c_{1:t-1})
\)
denote the filtration through time $t$, and let
$\Pr_{\pi_\psi}$ be the probability measure induced by the regulatory
policy $c_t=\pi_\psi(\rho_t)$. Expectations under this law are denoted by
$\mathbb E_{\pi_\psi}$, with the subscript omitted when the law is clear.

Suppose that the closed-loop CENTRY system satisfies the following
conditions:
\begin{enumerate}[nosep]
    \item \textbf{Strict Lyapunov Dissipativity:}
    There exist constants $\alpha\in(0,1]$ and $\epsilon\geq0$ such
    that, for every $t\geq1$,
    \(
    \mathbb{E}_{\pi_\psi}
    \!\left[
    V(\rho_{t+1})-V(\rho_t)
    \,\middle|\,
    \mathcal{F}_t
    \right]
    \leq
    -\alpha V(\rho_t)+\epsilon.
    \)

    \item \textbf{Bounded Initial Risk:}
    The initial risk-rating distribution has finite expected Lyapunov
    energy,
    \(
    \mathbb{E}_{\pi_\psi}[V(\rho_1)]=V_0<\infty.
    \)
\end{enumerate}

Then the following asymptotic safety properties hold:
\begin{enumerate}[label=(\roman*), nosep]
    \item \textbf{Mean Ultimate Boundedness:}
    \(
    \limsup_{t\to\infty}
    \mathbb{E}_{\pi_\psi}[V(\rho_t)]
    \leq
    \frac{\epsilon}{\alpha}.
    \)

    \item \textbf{Asymptotic Probability Bound:}
    For any tolerance $\eta>0$,
    \(
    \limsup_{t\to\infty}
    \Pr_{\pi_\psi}\!\left(
    V(\rho_t)\geq\eta
    \right)
    \leq
    \frac{\epsilon}{\alpha\eta}.
    \)
\end{enumerate}
\end{styledtheorem}

\noindent\textbf{Interpretation.} The property established here is mean and probabilistic ultimate boundedness of the risk-budget violation over an infinite horizon. This is important because finite-horizon guarantees alone do not describe what happens after repeated operation. The central assumption is a conditional Lyapunov dissipativity inequality, together with finite initial violation energy; such drift conditions are strong and must be established empirically or analytically for a particular closed loop rather than expected automatically from learning alone.

\begin{styledcorollary}{Almost-Sure Asymptotic Safety under a Summable Dissipativity Error}{almost_sure_safety}
Consider the setting of Theorem~\ref{thm:asymptotic_risk_stability}, but allow the
Lyapunov dissipativity error to vary with time. Suppose that, for an adapted
nonnegative sequence $\{\epsilon_t\}_{t\geq1}$,
\[
\mathbb{E}_{\pi_\psi}\!\left[
V(\rho_{t+1})-V(\rho_t)
\,\middle|\,
\mathcal{F}_t
\right]
\leq
-\alpha V(\rho_t)+\epsilon_t,
\qquad
\alpha\in(0,1],
\]
with $\epsilon_t$ being $\mathcal{F}_t$-measurable. Assume the integrability
condition
\[
\sum_{t=1}^{\infty}\mathbb{E}_{\pi_\psi}[\epsilon_t]<\infty.
\]
Then
\[
V(\rho_t)\longrightarrow0
\qquad\text{almost surely}.
\]
Equivalently, for every component $j\in\{D,H,M\}$,
\[
\left(\rho_t^{(j)}-\bar B^{(j)}\right)_+
\longrightarrow0
\qquad\text{almost surely}.
\]
Thus the magnitude of any component-wise budget violation vanishes
asymptotically almost surely. This is a pathwise asymptotic-safety guarantee;
it does not by itself imply that all violations cease after a finite time.
\end{styledcorollary}

\noindent\textbf{Interpretation.} The property established here strengthens asymptotic safety to pathwise convergence of the budget violation to zero when the dissipativity error is summable in expectation. This is important because it says that individual long-run trajectories, not only their average behavior, eventually approach the feasible region. Summable residual error is a strong adaptation/stability condition; it can be plausible after a well-controlled learning process settles, but it is not a generic property of continuously changing real-world systems.

\subsection{Learning guarantees}
\label{subsec:Learning guarantees}
These results characterize continual learning of both the probabilistic model and the regulatory policy. Their strongest rates rely on convexity, bounded gradients/noise, stationary or slowly changing environments, and well-behaved optimization objectives; these conditions are useful and sometimes enforceable for restricted policy classes, but should not be assumed automatically for general nonconvex neural systems or rapidly changing deployments.

\begin{styledtheorem}{Monotonic Improvement of the Predictive Evidence Bound}{predictive_evidence_bound}
Let
\(
\mathcal{F}_t^{-,y}=\sigma(y_{1:t-1},c_{1:t-1})
\)
denote the information available to CENTRY immediately before the
interaction $y_t$ is generated. Let
\(
p_{\mathrm{true}}(y_t\mid\mathcal{F}_t^{-,y})
\)
denote the true conditional distribution of the next interaction. Define
CENTRY's predictive distribution by
\[
p_{\mathrm{pred}}(y_t\mid\mathcal{F}_t^{-,y})
=
\int p_\theta(y_t\mid z_t)\,\kappa_t(z_t)\,dz_t,
\]
where $\kappa_t$ is the predictive latent prior. For $t>1$, this expression follows from Eq.~\eqref{eq:predictivedistribution} by substituting the propagated prior from Eqs.~\eqref{eq:jointpriorfactorization}--\eqref{eq:priorupdate}; at $t=1$, the initial prior is used. Define the realized
predictive surprise by
\[
\widehat S_t=-\log p_{\mathrm{pred}}(y_t\mid\mathcal{F}_t^{-,y}),
\]
and define the conditional population predictive mismatch by
\[
S_t^{\mathrm{(KL)}}
=
D_{\mathrm{KL}}\!\left(
 p_{\mathrm{true}}(\cdot\mid\mathcal{F}_t^{-,y})
 \,\middle\|
 p_{\mathrm{pred}}(\cdot\mid\mathcal{F}_t^{-,y})
\right).
\]
Assume that the conditional entropy of the true interaction distribution has
finite, constant expectation
\[
\mathbb E_{p_{\mathrm{true}}}
\left[
H\left(p_{\mathrm{true}}(\cdot\mid\mathcal F_t^{-,y})\right)
\right]=H^*<\infty
\]
for all $t$, that
$p_{\mathrm{true}}(\cdot\mid\mathcal F_t^{-,y})
\ll p_{\mathrm{pred}}(\cdot\mid\mathcal F_t^{-,y})$ almost surely,
and that all expectations below are finite.

Suppose further that the Learn stage uses the standard ($\beta=1$)
ELBO $\mathcal L_t(\theta_t,\phi_t)$ of Eq.~\eqref{eq:ELBO}, with
$\mathcal L_t$ a valid one-step predictive lower bound:
\[
\mathcal L_t(\theta_t,\phi_t)
\leq
\log p_{\mathrm{pred}}(y_t\mid\mathcal F_t^{-,y})
\qquad\text{a.s.}
\]
Finally, suppose the continual learning procedure is evaluated under a
common stationary environment phase and produces a non-decreasing
expected ELBO:
\[
\mathbb E[\mathcal L_{t+1}(\theta_{t+1},\phi_{t+1})]
\geq
\mathbb E[\mathcal L_t(\theta_t,\phi_t)]
\qquad\text{for every }t.
\]
Define
\[
U_t=-\mathbb E[\mathcal L_t(\theta_t,\phi_t)]-H^*.
\]
Then
\[
0\leq\mathbb E[S_t^{\mathrm{(KL)}}]\leq U_t,
\]
$\{U_t\}$ is non-increasing, and $U_t$ converges to a finite limit
$U_\infty\geq0$.
Moreover,
\[
\mathbb E[\widehat S_t]
=H^*+\mathbb E[S_t^{\mathrm{(KL)}}]
\leq H^*+U_t.
\]
Thus the negative expected ELBO bounds the expected predictive
cross-entropy and, after subtracting the irreducible true entropy $H^*$,
provides a monotone upper bound on population predictive mismatch.
\end{styledtheorem}

\noindent\textbf{Interpretation.} The property established here is a monotone upper bound on population predictive mismatch derived from the expected ELBO. This is important because it gives a formal link between continual variational learning and the quality of CENTRY's predictive model. The assumptions include a stationary environment phase, finite entropy, absolute continuity, a valid ELBO bound, and non-decreasing expected ELBO; these are standard theoretical conditions for analyzing learning but are only approximately applicable during genuinely non-stationary deployment.

\begin{styledcorollary}{Vanishing Asymptotic Predictive Mismatch}{vanishing_surprise}
Under the assumptions of Theorem~\ref{thm:predictive_evidence_bound}, suppose additionally that
\[
\lim_{t\to\infty}\mathbb E[\mathcal L_t(\theta_t,\phi_t)]=-H^*.
\]
Then the population predictive mismatch vanishes:
\[
\lim_{t\to\infty}\mathbb E[S_t^{\mathrm{(KL)}}]=0.
\]
If, in addition, $H^*=0$ (i.e., the true conditional interaction distribution has zero entropy; in the usual discrete case this means it is almost surely deterministic), then
\[
\lim_{t\to\infty}\mathbb E[\widehat S_t]=0.
\]
\end{styledcorollary}

\noindent\textbf{Interpretation.} The property established here is asymptotic disappearance of population predictive mismatch when the learned ELBO reaches the irreducible entropy level, with realized Surprise also vanishing in the deterministic zero-entropy case. This is important because it formalizes the idea that continual learning can make the interaction model increasingly predictable. The required convergence to the entropy limit is a strong learnability/optimization condition and is rarely guaranteed exactly in open-world AI operation; in practice, one should expect residual mismatch rather than literal zero Surprise in stochastic environments.

\begin{styledtheorem}{Iterative Improvement of the Variational Objective}{elbo_per_iteration_improvement}
Under the Learn-stage updating rule in Eq.~\eqref{eq:learnoperator},
suppose that the parameter update is performed by exact maximization of
the current variational objective:
\(
(\theta_{t+1},\phi_{t+1})
=
\arg\max_{\theta,\phi}
\mathcal{L}_t(\theta,\phi).
\)
Then
\(
\mathcal{L}_t(\theta_{t+1},\phi_{t+1})
\geq
\mathcal{L}_t(\theta_t,\phi_t)
\)
for every $t$.

More generally, if the Learn stage produces an approximate update
whose optimization error is bounded by $\varepsilon_t\geq0$, such that
\(
\mathcal{L}_t(\theta_{t+1},\phi_{t+1})
\geq
\sup_{\theta,\phi}\mathcal{L}_t(\theta,\phi)
-\varepsilon_t,
\)
then
\(
\mathcal{L}_t(\theta_{t+1},\phi_{t+1})
\geq
\mathcal{L}_t(\theta_t,\phi_t)
-\varepsilon_t.
\)
\end{styledtheorem}

\noindent\textbf{Interpretation.} The property established here is non-degradation of the current variational objective under exact maximization, with a controlled loss from approximate optimization. This is important because it gives the Learn stage a basic local progress guarantee without requiring the model to improve every possible future objective. Exact maximization is mainly a theoretical idealization; bounded optimization error is much more realistic for online neural implementations, so the approximate form is the practically relevant interpretation.

\begin{styledtheorem}{Sub-Linear Regret Bound for Adaptive Regulation}{sublinear_regret_bound}
Let $\Psi \subset \mathbb{R}^{d_\psi}$ be a nonempty, convex, compact parameter space with bounded diameter $D_\Psi = \sup_{\psi,\psi' \in \Psi} \|\psi - \psi'\|_2 < \infty$. At each time step $t$, let the candidate policy $\pi_\psi$ induce the regulatory action $c_t=\pi_\psi(\rho_t)$, and let $J_t(\psi)$ denote the model-based prospective regulatory loss function evaluated during the Learn stage:
\[
J_t(\psi) = \mathbb{E}_{p_{\theta_t}, \pi_\psi} \!\left[ \lambda_S \widehat S_{t+1} + \sum_{j \in \{D,H,M\}} \lambda_j \left( \rho_{t+1}^{(j)} - \bar B^{(j)} \right)_+ + \lambda_C C(c_t) \right],
\]
where $\lambda_S, \lambda_j, \lambda_C \ge 0$, and $C(c)\geq0$ for every admissible regulatory action $c$, with $C(\cdot)$ denoting the operational regulatory cost penalty.

Suppose that:
\begin{enumerate}[nosep]
    \item \textbf{Convexity and Continuity:} For each $t$, $J_t(\psi)$ is continuous and convex with respect to $\psi$ over $\Psi$.
    \item \textbf{Bounded Subgradients:} At each $\psi\in\Psi$ and $t\ge1$, a subgradient of $J_t(\psi)$ is selected and denoted by $\nabla J_t(\psi)$ (the ordinary gradient when differentiable). These selected subgradients are uniformly bounded in norm, i.e., $\|\nabla J_t(\psi)\|_2 \le G$ for all $\psi \in \Psi$ and all $t \ge 1$, for some $G>0$.
    \item \textbf{Online Gradient Update:} Assume $\psi_1\in\Psi$, so projection ensures $\psi_t\in\Psi$ for every $t\ge1$. Policy parameters are updated via projected online gradient descent:
    \(
    \psi_{t+1} = \Pi_\Psi \!\left( \psi_t - \eta_t \nabla J_t(\psi_t) \right),
    \)
    where $\Pi_\Psi(v) = \arg\min_{\psi \in \Psi} \|v - \psi\|_2$ denotes Euclidean projection onto $\Psi$, and $\eta_t = \frac{D_\Psi}{G \sqrt{t}}$ is the step-size schedule.
\end{enumerate}

Then, the cumulative regulatory regret over any finite operational horizon $T \in \mathbb{N}$, defined as
\[
\mathrm{Regret}_T(\Psi) = \sum_{t=1}^T J_t(\psi_t) - \min_{\psi^* \in \Psi} \sum_{t=1}^T J_t(\psi^*),
\]
satisfies the sub-linear upper bound:
\(
\mathrm{Regret}_T(\Psi) \le \frac{3}{2} G D_\Psi \sqrt{T} = \mathcal{O}(\sqrt{T}).
\)
Consequently, the upper bound on average per-step regret vanishes asymptotically, so the adaptive policy is asymptotically no worse on average than the best fixed policy chosen in hindsight.
\end{styledtheorem}

\noindent\textbf{Interpretation.} The property established here is sublinear cumulative regret for the adaptive regulation policy relative to the best fixed policy in hindsight under the stated online convex-optimization model. This is important because it says that, on average, the adaptive policy does not remain persistently worse than a fixed policy chosen after seeing the horizon. Convexity, bounded gradients, a compact policy class, and projected gradient updates are convenient and often enforceable for simplified policy parameterizations, but generic deep regulatory policies need not satisfy them.

\begin{styledcorollary}{High-Probability Regret under Bounded Gradient Noise}{highprob_regret}
Under the setting of Theorem~\ref{thm:sublinear_regret_bound}, let
\[
\mathcal F_t^{-,\psi}=\sigma(\psi_1,g_1,\ldots,g_{t-1},h_1,\ldots,h_t)
\]
denote the policy-learning filtration generated by the initial iterate, past stochastic gradients, and selected subgradients through time $t$. The iterate $\psi_t$ is measurable with respect to this filtration by the projected update, and the subgradient $h_t$ is selected before the noise $\xi_t$ is drawn. Suppose that the policy update uses
stochastic gradients of the form
\[
g_t=h_t+\xi_t,
\qquad
h_t\in\partial J_t(\psi_t),
\qquad
\|h_t\|_2\leq G,
\]
where the noise sequence satisfies
\[
\mathbb E[\xi_t\mid\mathcal F_t^{-,\psi}]=0,
\qquad
\|\xi_t\|_2\leq\sigma
\qquad\text{almost surely},
\]
for some $\sigma\geq0$ with $G+\sigma>0$. Let
\[
\eta_t
=
\frac{D_\Psi}{(G+\sigma)\sqrt{t}}.
\]
For any fixed (non-random) comparator $\psi^\star\in\Psi$, define
\[
\mathrm{Regret}_T(\psi^\star)
=
\sum_{t=1}^{T}\bigl(J_t(\psi_t)-J_t(\psi^\star)\bigr).
\]
Then, for every $\delta\in(0,1)$, with probability at least
$1-\delta$,
\[
\mathrm{Regret}_T(\psi^\star)
\leq
\frac{3}{2}(G+\sigma)D_\Psi\sqrt{T}
+
\sigma D_\Psi
\sqrt{2T\log\!\left(\frac{1}{\delta}\right)}.
\]
Thus, bounded martingale gradient noise preserves the
$O(\sqrt T)$ regret rate while providing an explicit high-probability
single-trajectory guarantee against any fixed noise-independent comparator.
A high-probability guarantee against the best comparator chosen in hindsight
requires an additional uniform-concentration argument and is not asserted here.
\end{styledcorollary}

\noindent\textbf{Interpretation.} The property established here upgrades the adaptive-policy regret result to a high-probability bound when stochastic gradient noise is conditionally unbiased and bounded. This is important because a single deployed trajectory need not realize only its expected optimization behavior. Bounded martingale noise is a useful controlled-learning assumption but can be restrictive for heavy-tailed or adversarial operational data; the theorem also deliberately does not claim a high-probability result against a hindsight-selected comparator without additional concentration machinery.

\begin{styledtheorem}{Dynamic Regret Bound for Non-Stationary Environments}{dynamic_regret_bound}
Under the convexity, bounded-subgradient, and feasible-initialization assumptions of
Theorem~\ref{thm:sublinear_regret_bound}, suppose that, for each
$t=1,\ldots,T$, the instantaneous objective $J_t$ has a minimizer
\[
\psi_t^\star
\in
\arg\min_{\psi\in\Psi}J_t(\psi).
\]
Define the path length of the sequence of instantaneous optimal policies
by
\[
P_T
=
\sum_{t=1}^{T-1}
\|\psi_{t+1}^\star-\psi_t^\star\|_2.
\]
Suppose that the policy update uses the horizon-tuned constant step size
\[
\eta
=
\frac{D_\Psi}{G\sqrt{T}}
\]
throughout the horizon. Then the dynamic regulatory regret
\[
\mathrm{Regret}_T^d
=
\sum_{t=1}^{T}J_t(\psi_t)
-
\sum_{t=1}^{T}J_t(\psi_t^\star)
\]
is identically zero if $D_\Psi=0$, since the policy class is then a singleton. For $D_\Psi>0$, it satisfies
\[
\mathrm{Regret}_T^d
\leq
\frac{D_\Psi^2}{2\eta}
+
\frac{\eta G^2T}{2}
+
\frac{D_\Psi}{\eta}P_T.
\]
Consequently,
\[
\mathrm{Regret}_T^d
\leq
GD_\Psi\sqrt{T}
+
G\sqrt{T}\,P_T
=O\!\left(\sqrt{T}(1+P_T)\right).
\]
If $P_T=o(\sqrt{T})$, then
\[
\frac{1}{T}\mathrm{Regret}_T^d\longrightarrow0.
\]
Thus, under sub-$\sqrt T$ cumulative drift of the instantaneous optimal
policy, CENTRY's online regulatory policy has vanishing average dynamic
regret relative to the sequence of instantaneous optimal policies.
\end{styledtheorem}

\noindent\textbf{Interpretation.} The property established here measures how well the online regulation policy tracks a changing sequence of instantaneous optimal policies. This is important because deployed AI environments are rarely perfectly stationary, so comparing only with one fixed policy can be misleading. The key condition is limited cumulative movement of the changing optimum; this is reasonable for slowly drifting environments but can fail during abrupt regime changes, where the bound appropriately becomes weaker.

\begin{styledcorollary}{Asymptotic Bound on Average Component-Risk Budget Excess}{average_risk_compliance}
Under the hypotheses of Theorem~\ref{thm:sublinear_regret_bound}, assume in addition that
$\widehat S_{t+1}\geq0$ almost surely under every candidate policy
$\pi_\psi$, $\psi\in\Psi$, for each $t$ (so that the surprise term in
$J_t$ is non-negative), and that $C(c)\geq0$ for every admissible
regulatory action $c$ (as in Theorem~\ref{thm:sublinear_regret_bound}), and let
$\psi^*=\arg\min_{\psi\in\Psi}\sum_{t=1}^T J_t(\psi)$. Then
\[
\frac1T\sum_{t=1}^T\sum_{j\in\{D,H,M\}}\lambda_j\,
\mathbb{E}_{p_{\theta_t},\pi_{\psi_t}}\!\left[\left(\rho_{t+1}^{(j)}-\bar B^{(j)}\right)_+\right]
\;\le\;
\frac1T\sum_{t=1}^T J_t(\psi^*)
\;+\;
\frac{3}{2}\,\frac{GD_\Psi}{\sqrt T}.
\]
As $T\to\infty$, the slack term vanishes. The average weighted component-risk budget excess under the online policy is thus bounded asymptotically by the average total regulatory objective of the best fixed policy in hindsight, not necessarily by its component-risk budget excess alone.
\end{styledcorollary}

\noindent\textbf{Interpretation.} The property established here connects policy regret to an upper bound on average weighted component-risk budget excess, expressed in terms of the total regulatory objective of the best fixed policy in hindsight. This is important because it interprets an abstract online-learning guarantee in the risk-regulation quantities CENTRY actually cares about. The additional nonnegativity condition on the realized Surprise term is automatic for discrete probability-mass models but is not automatic for continuous-density negative log-likelihoods, so the corollary should be applied only when that condition is justified. The result still does not itself guarantee per-time-step safety: it inherits the optimization assumptions of the regret theorem and compares average performance, so real deployments need the separate chance-constrained and effective-regulation guarantees for stronger safety claims.

\begin{styledcorollary}{Strongly Convex Policy Convergence}{strongly_convex_policy_convergence}
Consider the CENTRY regulation policy update of
Eq.~\eqref{eq:policylearning} on a nonempty compact convex set
$\Psi\subset\mathbb{R}^d$. In each of the two learning runs below, assume $\psi_1\in\Psi$ and use projection onto $\Psi$ at every update.
\begin{enumerate}
    \item \textbf{Logarithmic Regret.} \label{item:SC-regret}
    Suppose
    \[
    \psi_{t+1}
    =
    \Pi_{\Psi}(\psi_t-\eta_tg_t),
    \qquad
    g_t\in\partial J_t(\psi_t),
    \qquad
    \eta_t=\frac{1}{\mu t},
    \]
    where each instantaneous objective $J_t:\Psi\to\mathbb R$ is
    $\mu$-strongly convex for some $\mu>0$, and
    $\|g_t\|_2\leq G$. Then the cumulative regret against the fixed
    best-in-hindsight policy
    \[
    \psi^\star
    \in
    \arg\min_{\psi\in\Psi}
    \sum_{t=1}^{T}J_t(\psi)
    \]
    satisfies
    \[
    \mathrm{Regret}_T
    \leq
    \frac{G^2}{2\mu}(1+\ln T)
    =O(\log T).
    \]

    \item \textbf{Stationary Parameter Convergence.} \label{item:SC-converge}
    For a separate stationary-objective learning run, suppose that there exists
    a differentiable, $\mu$-strongly convex stationary objective
    $J:\Psi\to\mathbb R$ with a unique minimizer
    \[
    \psi^\star=\arg\min_{\psi\in\Psi}J(\psi),
    \]
    that lies in the interior of $\Psi$. Thus $\psi^\star$ is an unconstrained
    local minimizer and satisfies $\nabla J(\psi^\star)=0$. Suppose further that
    the stochastic update directions satisfy
    \[
    \mathbb E[g_t\mid\mathcal F_t^{-,\psi}]=\nabla J(\psi_t),
    \qquad
    \mathbb E[\|g_t\|_2^2\mid\mathcal F_t^{-,\psi}]\leq G^2.
    \]
    The conditional moments are taken with respect to the policy-learning
    filtration of this run, $\mathcal F_t^{-,\psi}=\sigma(\psi_1,g_1,\ldots,g_{t-1})$. For this stationary
    setting, use the step size
    \[
    \eta_t=\frac{1}{\mu(t+1)}.
    \]
    Then, for every $t\geq2$,
    \[
    \mathbb E\left[\|\psi_t-\psi^\star\|_2^2\right]
    \leq
    \frac{G^2}{\mu^2(t+1)}
    =O\left(\frac1t\right),
    \]
    and consequently,
    \[
    \mathbb E\left[\|\psi_t-\psi^\star\|_2\right]
    \leq
    \frac{G}{\mu\sqrt{t+1}}
    =O\left(\frac1{\sqrt t}\right).
    \]
    Thus the learned policy parameters converge to the unique stationary
    risk-optimal policy $\psi^\star$ in mean square.
\end{enumerate}
\end{styledcorollary}

\noindent\textbf{Interpretation.} The property established here gives faster logarithmic regret in a strongly convex setting and mean-square convergence in a separate stationary strongly convex learning run. This is important because strong curvature turns a general sublinear optimization guarantee into explicit parameter convergence rates. Strong convexity and bounded stochastic second moments are strong assumptions for flexible neural policies, but they can be reasonable after convexification, for restricted policy classes, or for local/stationary surrogate objectives.

\begin{styledtheorem}{Sample Complexity of Variational Learning}{sample_complexity}
Under the stationary-environment assumptions of
Theorem~\ref{thm:predictive_evidence_bound}, let
$w=(\theta,\phi)\in\Xi$, where $\Xi\subset\mathbb R^d$ is closed and
convex. Let the population objective
\[
\mathcal L(w)=\mathbb E[\mathcal L_t(w)]
\]
be $\mu$-strongly concave. Equivalently, let
$f(w)=-\mathcal L(w)$ be $\mu$-strongly convex. Let
\[
\mathcal F_t^{-,w}=\sigma(w_1,g_1,\ldots,g_{t-1})
\]
denote the model-learning filtration containing the information available
immediately before $w_t$ is updated. Suppose the online Learn stage uses
projected stochastic-gradient updates
\[
w_{t+1}=\Pi_\Xi(w_t-\eta_tg_t),
\qquad
\eta_t=\frac{2}{\mu(t+1)},
\]
with
\[
\mathbb E[g_t\mid\mathcal F_t^{-,w}]=\nabla f(w_t),
\qquad
\mathbb E\!\left[\|g_t\|_2^2\mid\mathcal F_t^{-,w}\right]\le G^2.
\]
Define the polynomially weighted average
\[
\widetilde w_N
=
\frac{2}{N(N+1)}\sum_{t=1}^{N}t\,w_t,
\]
and let
\[
w^\star\in\arg\min_{w\in\Xi}f(w),
\qquad
\mathcal L^\star=\mathcal L(w^\star).
\]
Then
\[
\mathbb E\!\left[
\mathcal L^\star-\mathcal L(\widetilde w_N)
\right]
\leq
\frac{2G^2}{\mu(N+1)}.
\]
Thus the polynomially weighted online variational iterate has expected
ELBO gap $O(1/N)$.

If $f$ is additionally $L$-smooth and $w^\star$ is an interior
unconstrained optimum, then, for $N\geq2$, the last iterate satisfies
\[
\mathbb E\!\left[
\mathcal L^\star-\mathcal L(w_N)
\right]
\leq
\frac{2LG^2}{\mu^2(N+1)}.
\]
\end{styledtheorem}

\noindent\textbf{Interpretation.} The property established here gives an explicit O(1/N) expected ELBO gap for a weighted average of the learned variational parameters, with a corresponding last-iterate bound under smoothness. This is important because it translates continual variational learning into a quantitative data requirement. Stationarity, strong concavity/convexity, unbiased stochastic gradients, and bounded second moments are standard theoretical learning assumptions but are idealizations for long-lived AI systems whose data distribution may drift.

\begin{styledtheorem}{Policy Improvement Under the Model-Based Objective}{policy_improvement}
Let $\pi_{\psi_t}$ be the regulatory policy at time $t$, and let
$\pi_{\psi_{t+1}}$ be the policy obtained by the update in
Eq.~\eqref{eq:policylearning}. For each candidate policy $\pi$, let
$c_t=\pi(\rho_t)$ denote its induced regulatory action. Define the
model-based prospective regulatory objective
\[
J(\pi,\theta_t)
=
\mathbb{E}_{p_{\theta_t},\pi}
\left[
\lambda_S \widehat S_{t+1}
+
\sum_{j\in\{D,H,M\}}
\lambda_j
\left(
\rho_{t+1}^{(j)}-\bar B^{(j)}
\right)_+
+
\lambda_C C(c_t)
\right].
\]
If the policy update performs exact minimization over a candidate set containing the preceding policy (in particular, the preceding policy must be feasible for a constrained update),
\(
\psi_{t+1}
=
\arg\min_{\psi}J(\pi_\psi,\theta_t),
\)
then
\(
J(\pi_{\psi_{t+1}},\theta_t)
\leq
J(\pi_{\psi_t},\theta_t).
\)
Thus, the updated policy cannot have a larger model-based prospective
regulatory objective than the policy from the preceding iteration,
when both are evaluated under the same model parameters $\theta_t$.
\end{styledtheorem}

\noindent\textbf{Interpretation.} The property established here is monotonic improvement of CENTRY's model-based prospective regulation objective when the policy update exactly minimizes that objective under the current model. This is important because it formalizes what the Learn stage is trying to accomplish before the next interaction is observed. Exact minimization is not generally available for a complex online policy, so the result is best viewed as an ideal target; approximate projected optimization provides the practical implementation path described elsewhere in the paper.

\subsection{Inference guarantees}
\label{subsec:Inference guarantees}
These results characterize what can be inferred about the latent interaction and component-risk states. Several results use linear-Gaussian or locally linear-Gaussian models because they permit exact analysis, while the consistency and attribution results additionally require informative observations and identifiable component effects. Such assumptions are often useful approximations for analysis, but their adequacy must be tested for each application.

For the results below, the pre-interaction filtration is \(\mathcal F_t^{-,y}:=\sigma(y_{1:t-1},c_{1:t-1})\), which contains the information available immediately before the interaction $y_t$ is generated.

The first theorem formalizes the connection between the computable Surprise measure and the variational objective.

\begin{styledtheorem}{Predictive Surprise and the Variational Objective}{surprise_elbo_relation}
Let $\kappa_t(z_t)$ be the predictive latent prior and define the
one-step predictive marginal by
\[
p_{\mathrm{pred}}(y_t\mid\mathcal F_t^{-,y})
=
\int p_\theta(y_t\mid z_t)\kappa_t(z_t)\,dz_t.
\]
Let
\[
\widehat S_t=-\log p_{\mathrm{pred}}(y_t\mid\mathcal F_t^{-,y})
\]
be the realized predictive negative log-likelihood, and let
$q_t(z_t)=q_\phi(z_t\mid y_{1:t},c_{1:t-1})$ be a variational posterior.
For $\beta>0$, define
\[
\mathcal L_{t,\beta}(\theta,\phi)
=
\mathbb E_{q_t}[\log p_\theta(y_t\mid z_t)]
-\beta D_{\mathrm{KL}}(q_t\|\kappa_t).
\]
Assume $q_t\ll\kappa_t$, $p_\theta(y_t\mid z_t)>0$
$\kappa_t$-almost everywhere on the support of $q_t$, and all displayed
quantities are finite. Then
\[
\mathcal L_{t,\beta}
=
\log p_{\mathrm{pred}}(y_t\mid\mathcal F_t^{-,y})
-D_{\mathrm{KL}}\!
\left(q_t\middle\|p_\theta(z_t\mid y_t,\mathcal F_t^{-,y})\right)
-(\beta-1)D_{\mathrm{KL}}(q_t\|\kappa_t).
\]
Consequently, for every $\beta\geq1$,
\[
\mathcal L_{t,\beta}\leq\log p_{\mathrm{pred}}(y_t\mid\mathcal F_t^{-,y})
=-\widehat S_t,
\]
so that
\[
-\mathcal L_{t,\beta}\geq\widehat S_t.
\]
The result is a pointwise variational statement about the realized
one-step predictive likelihood; it does not identify realized surprise
with a conditional KL divergence.
\end{styledtheorem}

\noindent\textbf{Interpretation.} The property established here is an exact variational decomposition connecting realized Predictive Surprise to the ELBO and two KL-divergence terms. This is important because it rigorously distinguishes computable Surprise from population-level predictive mismatch and explains why Surprise is linked to the amount of inferential work required. The assumptions are standard support and finiteness conditions for variational inference and are usually attainable by construction, although numerical and model-approximation issues remain in practice.

\begin{styledcorollary}{Surprise--ELBO Bound for General $\beta$}{surprise_elbo_general_beta}
Under the hypotheses of Theorem~\ref{thm:surprise_elbo_relation}, for arbitrary
$\beta>0$,
\[
\mathcal L_{t,\beta}
\leq
-\widehat S_t
+(1-\beta)D_{\mathrm{KL}}(q_t\|\kappa_t).
\]
For $\beta\geq1$ this reduces to
\[
\mathcal L_{t,\beta}\leq-\widehat S_t.
\]
Moreover, the exact identity is
\[
\mathcal L_{t,\beta}
=-\widehat S_t
-D_{\mathrm{KL}}\left(q_t\middle\|p_\theta(z_t\mid y_t,\mathcal F_t^{-,y})\right)
-(\beta-1)D_{\mathrm{KL}}(q_t\|\kappa_t).
\]
Equality in the first displayed upper bound occurs exactly when
$q_t=p_\theta(z_t\mid y_t,\mathcal F_t^{-,y})$ almost everywhere.
For $\beta=1$, under the same posterior-equality condition, this gives
$\mathcal L_{t,1}=-\widehat S_t$.
For $\beta>1$, equality in the stronger bound
$\mathcal L_{t,\beta}\leq-\widehat S_t$ additionally requires
$D_{\mathrm{KL}}(q_t\|\kappa_t)=0$.
\end{styledcorollary}

\noindent\textbf{Interpretation.} The property established here extends the Surprise--ELBO relation to arbitrary positive beta and shows that beta at least one preserves a simple upper bound. This is important because CENTRY can change the strength of the posterior-regularization term without losing the central connection between the variational objective and predictive surprise. The result depends on the same support and finiteness conditions as the parent identity, which are manageable implementation requirements rather than assumptions about the external world.

\begin{styledcorollary}{Asymptotic Bound on the Rate of Surprise-Flagged Interactions}{surprise_flagging_rate}
Let $\widehat S_t = -\log p_{\mathrm{pred}}(y_t\mid y_{1:t-1},c_{1:t-1})$ denote the
operationally-computed, per-interaction surprise of
Eq.~\eqref{eq:surprisenll} (as opposed to the population-level $S_t^{\mathrm{(KL)}}$ of
Theorem~\ref{thm:predictive_evidence_bound}), and suppose
$\widehat S_t\ge0$ almost surely for every $t$ (which holds, in particular,
whenever the space of interactions is discrete, so that
$p_{\mathrm{pred}}(\cdot\mid\cdot)$ is a probability mass function).
Under the hypotheses of Theorem~\ref{thm:predictive_evidence_bound}
and Theorem~\ref{thm:surprise_elbo_relation} at $\beta=1$, and assuming
$\widehat S_t\geq0$ almost surely, for every
threshold $\tau>0$,
\(
\Pr\!\left(\widehat S_t\ge\tau\right)
\;\le\;
\frac{U_t+H^*}{\tau},
\text{ and hence }
\limsup_{t\to\infty}\Pr\!\left(\widehat S_t\ge\tau\right)
\;\le\;
\frac{U_\infty+H^*}{\tau}.
\)
That is, for any fixed Surprise-flagging threshold used to allocate
additional diagnostic, verification, or adaptation effort, the
long-run flagging rate is asymptotically bounded. The bound tightens with the predictive-ELBO bound
$U_t$, but an irreducible contribution from the true interaction entropy $H^*$
remains in the general stochastic case.
\end{styledcorollary}

\noindent\textbf{Interpretation.} The property established here bounds the long-run fraction of interactions whose computable Surprise exceeds a chosen threshold. This is important because the Surprise threshold controls additional diagnostic or verification effort and therefore has an operational resource consequence. The main limitation is that the nonnegative-Surprise assumption is automatic for discrete probability masses but not for arbitrary continuous-density negative log-likelihoods, while stochastic environments retain an irreducible entropy contribution.

\begin{styledtheorem}{Temporal Scale Separation under Linear-Gaussian Dynamics}{scale_separation}
Suppose that the fast interaction state and each slow component-risk
state follow stationary scalar linear-Gaussian autoregressive dynamics,
\(
z_{t+1}^{(F)}
=
\alpha_F z_t^{(F)}
+
\varepsilon_{t+1}^{(F)},
\)
and
\(
z_{t+1}^{(j)}
=
\alpha_j z_t^{(j)}
+
\varepsilon_{t+1}^{(j)}\),
\(
j\in\{D,H,M\},
\)
where
\(
|\alpha_F|<|\alpha_j|<1
\)
for every $j\in\{D,H,M\}$. Assume that the innovation processes
$\{\varepsilon_t^{(F)}\}$ and $\{\varepsilon_t^{(j)}\}$ have zero mean,
strictly positive finite variance for each component, are Gaussian, and are serially uncorrelated over time,
i.e.,
\(
\operatorname{Cov}
\!\left(
\varepsilon_s^{(\ell)},
\varepsilon_u^{(\ell)}
\right)
=
0\),
\(
s\neq u,
\)
for $\ell\in\{F,D,H,M\}$.

Then the autocorrelation functions satisfy
\(
r_F(k)=\alpha_F^k\),
\(
r_j(k)=\alpha_j^k,
\)
for every $k\geq0$ (a zeroth power is one, including when the base is zero), where $r_F$ and $r_j$ denote the
autocorrelation functions of $\{z_t^{(F)}\}$ and
$\{z_t^{(j)}\}$, respectively. Consequently, the magnitude of the
autocorrelation of the fast interaction state decays more rapidly than
that of every slow component-risk state. In particular, the
characteristic exponential correlation-decay times (for nonzero coefficients)
\(
\tau_F
=
-\frac{1}{\log|\alpha_F|}
\)
and
\(
\tau_j
=
-\frac{1}{\log|\alpha_j|}
\)
satisfy
\(
\tau_F<\tau_j\),
\(
j\in\{D,H,M\}.
\)
If $\alpha_F=0$, define $\tau_F:=0$ (the limiting decay time); the fast-state autocorrelation vanishes at every positive lag, and the same strict ordering holds. Thus, the fast interaction state evolves on a shorter characteristic
temporal scale than each component-risk state.
\end{styledtheorem}

\noindent\textbf{Interpretation.} The property established here provides a precise linear-Gaussian condition under which the fast interaction state changes on a shorter temporal scale than the slower component-risk states. This is important because CENTRY's latent architecture relies conceptually on distinguishing rapidly changing behavior from slowly evolving component risk. The strict coefficient ordering and Gaussian AR dynamics are a modeling idealization; temporal scale separation is plausible in some systems but should be measured rather than assumed for a particular deployment.

\begin{styledtheorem}{Asymptotic Resilience to Prior Miscalibration}{prior_forgetting}
Let $q_t(z_t\mid y_{1:t},c_{1:t-1})$ and $\tilde q_t(z_t\mid y_{1:t},c_{1:t-1})$
denote the inferred joint posteriors at time $t$ under the SELP cycle,
initialized with two different joint priors $\kappa_1(z_1)$ and
$\tilde\kappa_1(z_1)$, respectively. Let $\kappa_t$ and $\tilde\kappa_t$
denote the corresponding predictive priors at time $t$.

Suppose that:
\begin{enumerate}[nosep]
    \item \textbf{Contractive transition:} The joint latent-state transition
    operator $P_t$ induced by $p_\theta(z_{t+1}\mid z_t,y_t,c_t)$ is uniformly
    contractive in the Wasserstein distance $\mathcal W$: there exists a
    constant $\gamma\in[0,1)$ such that, for every $t$ and every pair of
    probability distributions $\mu,\nu$ on the latent-state space,
    $\mathcal W(P_t\mu,P_t\nu)\leq\gamma\mathcal W(\mu,\nu)$.

    \item \textbf{Non-expansive measurement update:} For the common observation
    $y_t$, the Bayesian measurement-update operator induced by the likelihood
    $p_\theta(y_t\mid z_t)$ is non-expansive in the Wasserstein distance, i.e.,
    $\mathcal W(q_t,\tilde q_t)\leq\mathcal W(\kappa_t,\tilde\kappa_t)$ for every $t$.

    \item \textbf{Finite initial Wasserstein distance:}
    $\mathcal W(q_1,\tilde q_1)<\infty$.
\end{enumerate}

Then, under an identical stream of observable interactions and regulatory actions and the same model parameters in both runs,
\[
\mathcal W(q_t,\tilde q_t)\leq
\mathcal W(q_1,\tilde q_1)\,\gamma^{t-1},\qquad t\geq1.
\]
A zeroth power is one, including when the base is zero. Consequently, $\mathcal W(q_t,\tilde q_t)\xrightarrow[t\to\infty]{}0$.
Thus, the effect of the initial prior specification vanishes geometrically fast,
guaranteeing asymptotic resilience to prior miscalibration.
\end{styledtheorem}

\noindent\textbf{Interpretation.} The property established here is exponential forgetting of different initial latent priors when the transition is contractive and the common Bayesian measurement update is non-expansive. This is important because it limits long-term dependence on uncertain initial risk assessments. Contractivity is a recognized stability condition for filtering, but it can fail in unstable, highly nonlinear, or strongly informative update regimes, so real deployments should test sensitivity to prior initialization rather than assume complete forgetting.

\begin{styledtheorem}{Consistency of Component Risk Ratings}{risk_rating_consistency}
Let $R_t^{(j)}$ denote the true component risk associated with component
$j\in\{D,H,M\}$, and let
\[
\rho_t^{(j)}
=
g^{(j)}\!\left(\mu_t^{(j)},\Sigma_t^{(j)}\right)
\]
be CENTRY's component risk rating at time $t$, where $g^{(j)}$ is the risk
functional of Eq.~\eqref{eq:riskfunctional}. Suppose that, for each
$j\in\{D,H,M\}$:

\begin{enumerate}[nosep]
\item \textbf{Risk-state identifiability:} There exists a continuous
function $h_j$ such that
\[
R_t^{(j)}
=
h_j\!\left(z_t^{(j),*}\right),
\]
where $z_t^{(j),*}$ denotes the true latent state of component $j$.
Moreover, $h_j$ is injective on the relevant support of the latent
state.

\item \textbf{Posterior mean consistency:} The variational posterior
mean asymptotically recovers the true latent state, in the sense that
\[
\mu_t^{(j)}-z_t^{(j),*}
\xrightarrow{p}
0
\qquad\text{as }t\to\infty.
\]

\item \textbf{Posterior covariance stabilization:} The variational
posterior covariance converges in probability to a limiting covariance
matrix,
\[
\Sigma_t^{(j)}
\xrightarrow{p}
\Sigma_\infty^{(j)},
\]
where $\Sigma_\infty^{(j)}$ need not be zero.

\item \textbf{Uniform continuity of the risk functional:} There exists
a fixed compact set $K_j$ such that
\[
z_t^{(j),*}\in K_j
\qquad\text{almost surely for all }t,
\]
and there exists an open neighborhood $\mathcal U_j$ of
\[
K_j\times\{\Sigma_\infty^{(j)}\}
\]
on which $g^{(j)}$ is uniformly continuous. Furthermore,
\[
g^{(j)}\!\left(z^{(j)},0\right)
=
h_j\!\left(z^{(j)}\right)
\]
for every $z^{(j)}$ in the relevant support of the latent state.
\end{enumerate}

Then, for every component $j\in\{D,H,M\}$,
\[
\rho_t^{(j)}
-
g^{(j)}\!\left(z_t^{(j),*},\Sigma_\infty^{(j)}\right)
\xrightarrow{p}
0
\qquad\text{as }t\to\infty.
\]

Equivalently,
\[
\rho_t^{(j)}
=
g^{(j)}\!\left(z_t^{(j),*},\Sigma_\infty^{(j)}\right)
+
o_p(1).
\]

Consequently, the difference between the CENTRY component risk rating
and the true component risk satisfies
\[
\rho_t^{(j)}-R_t^{(j)}
=
g^{(j)}\!\left(z_t^{(j),*},\Sigma_\infty^{(j)}\right)
-
h_j\!\left(z_t^{(j),*}\right)
+
o_p(1).
\]

Thus, exact risk-rating consistency,
\[
\rho_t^{(j)}-R_t^{(j)}
\xrightarrow{p}
0,
\]
is obtained whenever
\[
g^{(j)}\!\left(z,\Sigma_\infty^{(j)}\right)
=
h_j(z)
\]
throughout the relevant support of the latent state. In particular, exact
consistency holds if $\Sigma_\infty^{(j)}=0$, because assumption~4 then gives
$g^{(j)}(z,\Sigma_\infty^{(j)})=g^{(j)}(z,0)=h_j(z)$.
\end{styledtheorem}

\noindent\textbf{Interpretation.} The property established here specifies when CENTRY's posterior-based component risk ratings converge in probability toward the true component risks, and it identifies the residual uncertainty floor when the limiting posterior covariance is nonzero. This is important because it is the formal bridge between latent inference and meaningful risk ratings. The assumptions---latent identifiability, posterior consistency, covariance stabilization, compactness, and continuity---are strong and generally require a well-specified model plus informative data; exact consistency should therefore be treated as a validation target, not a default property.

\begin{styledtheorem}{Robust Transfer from Rating-Based to Operational-Risk Safety}{rating_to_operational_safety}
Fix a component $j\in\{D,H,M\}$ and a future time $t+k$. Let
$R_{t+k}^{(j)}$ be the true operational component risk and let
$\rho_{t+k}^{(j)}$ be CENTRY's component risk rating. Let $B^{(j)}$ be
the operational risk budget and $\bar B^{(j)}$ the corresponding rating
threshold. Suppose that, conditional on $\mathcal F_t$,
\[
\Pr\left(R_{t+k}^{(j)}>B^{(j)},
\rho_{t+k}^{(j)}\leq\bar B^{(j)} \middle|\mathcal F_t\right)
\leq\eta_k^{(j)}
\]
and
\[
\Pr\left(\rho_{t+k}^{(j)}>\bar B^{(j)} \middle|\mathcal F_t\right)
\leq\alpha_k^{(j)}.
\]
Then
\[
\Pr\left(R_{t+k}^{(j)}>B^{(j)} \middle|\mathcal F_t\right)
\leq\eta_k^{(j)}+\alpha_k^{(j)}.
\]
Consequently, if the regulatory policy controls the rating-violation
probability such that $\alpha_k^{(j)} \leq \delta_k^{(j)}-\eta_k^{(j)}$
for an allowable tolerance $\delta_k^{(j)}\geq\eta_k^{(j)}$, then the
true operational risk satisfies the chance constraint
\[
\Pr\left(R_{t+k}^{(j)}>B^{(j)} \middle|\mathcal F_t\right)
\leq\delta_k^{(j)}.
\]
Moreover, if this condition holds for all $j\in\{D,H,M\}$ and
$k=1,\ldots,T$, and
\[
\sum_{k=1}^{T}\sum_{j\in\{D,H,M\}}\delta_k^{(j)}\leq\delta,
\]
then
\[
\Pr\left(
R_{t+k}^{(j)}\leq B^{(j)},
\;\forall j\in\{D,H,M\},
\;\forall k=1,\ldots,T
\middle|\mathcal F_t\right)
\geq1-\delta.
\]
Thus a rating-based safety rule transfers to a true operational-risk
safety probability whenever the probability of rating underestimation and
the probability of rating exceedance are separately controlled.
\end{styledtheorem}

\noindent\textbf{Interpretation.} The property established here transfers safety from the estimated rating domain to true operational risk by explicitly accounting for both dangerous false negatives and rating overshoots. This is important because CENTRY can only regulate from ratings, while the actual safety requirement concerns latent operational risk. The two probability terms must be calibrated or bounded empirically; the framework is therefore realistic as a measurable decomposition, but the quality of the transfer depends directly on the quality of risk estimation and calibration.

\begin{styledtheorem}{Separation of Component Risk Contributions}{risk_decomposition_separation}
Let $R_t=R(\rho_t)$ denote the system-level expected risk at time $t$ as a
function of the component risk ratings
\(
\rho_t=
\left(
\rho_t^{(D)},\rho_t^{(H)},\rho_t^{(M)}
\right)
\in\mathbb{R}_{\geq 0}^3.
\)
Suppose that $R$ is continuously differentiable on the nonnegative orthant
(with derivatives at boundary points understood as continuous extensions
of the corresponding interior derivatives) and positively homogeneous
of degree one in the component ratings, i.e.,
\(
R(\lambda\rho)=\lambda R(\rho)\),
\(
\forall \lambda>0.
\)
Then the exact additive decomposition
\(
R(\rho_t)
=
\sum_{j\in\{D,H,M\}}
\frac{\partial R}{\partial \rho^{(j)}}(\rho_t)\,
\rho_t^{(j)}
\)
holds. 

If $R(\rho_t)>0$, define the component risk-contribution shares by
\(
s_t^{(j)}
=
\frac{
\frac{\partial R}{\partial \rho^{(j)}}(\rho_t)\,
\rho_t^{(j)}
}{
R(\rho_t)
}\),
\(
j\in\{D,H,M\}.
\)
Then
\(
\sum_{j\in\{D,H,M\}}s_t^{(j)}=1.
\)

If, in addition, $R$ is non-decreasing in each component rating, so that
\(
\frac{\partial R}{\partial\rho^{(j)}}(\rho)\geq 0\),
\(
j\in\{D,H,M\},
\)
then
\(
s_t^{(j)}\geq 0
\)
for all $j$.

Moreover, if $R$ is twice continuously differentiable (with the same
boundary convention for second derivatives) and submodular in
the component ratings, then, for every $i\neq j$,
\(
\frac{\partial^2 R}
{\partial\rho^{(i)}\partial\rho^{(j)}}(\rho_t)
\leq 0,
\)
so that the component ratings are weakly substitutable rather than
complementary in their effect on system-level risk.
\end{styledtheorem}

\noindent\textbf{Interpretation.} The property established here provides exact component contribution shares when system risk is continuously differentiable and positively homogeneous, and it characterizes weak substitutability under submodularity. This is important because it gives a principled way to interpret which component ratings contribute to aggregate risk instead of relying only on an aggregate score. Positive homogeneity and submodularity are modeling assumptions, not universal properties of AI risk, but they can be reasonable for deliberately constructed additive or near-additive risk models.

\begin{styledtheorem}{Component Identifiability under Local Fisher-Information Distinguishability}{component_identifiability}
Fix a time $t$ and let
\(
\mathcal{J}=\{F,D,H,M\}
\)
denote the latent components. Write
\(
z_t
=
\bigl(
z_t^{(j)}
\bigr)_{j\in\mathcal{J}}\),
\(
z_t^{(j)}\in\mathbb{R}^{d_j}\),
\(
d=\sum_{j\in\mathcal{J}}d_j.
\)
Suppose the predictive prior factorizes across components as in
Eq.~\eqref{eq:jointpriorfactorization},
\(
\kappa_t(z_t)
=
\prod_{j\in\mathcal{J}}
\mathcal{N}
\!\left(
z_t^{(j)};
\mu_t^{(j)},\Sigma_t^{(j)}
\right),
\)
with
\(
\Sigma_t
=
\operatorname{blkdiag}
\!\left(
\Sigma_t^{(j)}
\right)_{j\in\mathcal{J}}
\succ0.
\)

Let
\(
f_\theta(z_t)
=
\mathbb{E}_\theta[y_t\mid z_t]
\in\mathbb{R}^{d_y}
\)
denote the decoder mean. Linearize the decoder locally around the
prior mean
\(
\mu_t
=
\bigl(
\mu_t^{(j)}
\bigr)_{j\in\mathcal{J}}
\)
and suppose that the resulting local likelihood is Gaussian:
\(
p_\theta(y_t\mid z_t)
\approx
\mathcal{N}
\!\left(
y_t;
f_\theta(\mu_t)
+
\sum_{j\in\mathcal{J}}
W_t^{(j)}
\bigl(
z_t^{(j)}-\mu_t^{(j)}
\bigr),
\Omega_t
\right),
\)
where
\(
W_t^{(j)}
=
\left.
\nabla_{z^{(j)}}f_\theta(z)
\right|_{z=\mu_t}
\in\mathbb{R}^{d_y\times d_j}\),
\(
\Omega_t\succ0.
\)
Define
\(
W_t
=
\bigl[
W_t^{(F)}
\;
W_t^{(D)}
\;
W_t^{(H)}
\;
W_t^{(M)}
\bigr]
\)
and the corresponding local Fisher information matrix
\(
\mathcal{I}_t
=
W_t^\top\Omega_t^{-1}W_t.
\)
Assume that
\(
\mathcal{I}_t\succ0.
\)
Equivalently, $W_t$ has full column rank, and hence necessarily
$d_y\geq d$.

Consider the corresponding local conjugate-Gaussian Bayesian
specialization of the Explain update. Define
\(
\widetilde z_t
=
z_t-\mu_t\),
\(
\widetilde y_t
=
y_t-f_\theta(\mu_t),
\)
and let
\(
\widehat{\Sigma}_t
=
\left(
\Sigma_t^{-1}+\mathcal{I}_t
\right)^{-1},
\)
\(
\widehat{\widetilde z}_t
=
\widehat{\Sigma}_t
W_t^\top\Omega_t^{-1}\widetilde y_t.
\)
Suppose that the true latent state differs from the prior mean in
exactly one component $j^\star\in\mathcal{J}$, by a nonzero anomaly
\(
\Delta\in\mathbb{R}^{d_{j^\star}}\setminus\{0\},
\)
so that
\(
\widetilde y_t
=
W_t^{(j^\star)}\Delta+\eta_t\),
\(
\eta_t\sim\mathcal{N}(0,\Omega_t).
\)
Let $\iota_{j^\star}(\Delta)\in\mathbb{R}^{d}$ denote the vector obtained
by placing $\Delta$ in component $j^\star$ and zeros in all other
components.

Then the following hold.

\textbf{(a) Bias bound.}
The expected posterior mean shift satisfies
\(
\mathbb{E}_{\eta_t}
\!\left[
\widehat{\widetilde z}_t
\right]
-
\iota_{j^\star}(\Delta)
=
-
\widehat{\Sigma}_t
\Sigma_t^{-1}
\iota_{j^\star}(\Delta),
\)
and therefore
\(
\left\|
\mathbb{E}_{\eta_t}
\!\left[
\widehat{\widetilde z}_t
\right]
-
\iota_{j^\star}(\Delta)
\right\|_2
\leq
\left\|
\mathcal{I}_t^{-1}
\right\|_{\mathrm{op}}
\left\|
\Sigma_t^{-1}
\right\|_{\mathrm{op}}
\|\Delta\|_2.
\)
Consequently, for every component $j\neq j^\star$, the corresponding
block of the expected posterior mean shift satisfies
\[
\mathbb{E}_{\eta_t}\!\left[
\widehat{\widetilde z}_t^{(j)}
\right]\longrightarrow 0
\]
whenever
\[
\left\|
\mathcal{I}_t^{-1}
\right\|_{\mathrm{op}}
\left\|
\Sigma_t^{-1}
\right\|_{\mathrm{op}}
\longrightarrow0.
\]
In particular, this occurs if either the local observational
information becomes arbitrarily large,
\(
\left\|\mathcal{I}_t^{-1}\right\|_{\mathrm{op}}\longrightarrow0,
\)
while the prior remains fixed, or the prior becomes arbitrarily
diffuse,
\(
\left\|\Sigma_t^{-1}\right\|_{\mathrm{op}}\longrightarrow0,
\)
while the observational information remains fixed.

\textbf{(b) Exact attribution under orthogonal component sensitivities.}
If, in addition,
\(
\left(W_t^{(j)}\right)^\top
\Omega_t^{-1}
W_t^{(k)}
=
0\),
\(
j\neq k,
\)
so that $\mathcal{I}_t$ is block-diagonal with respect to the
component decomposition, then
\(
\mathbb{E}_{\eta_t}
\!\left[
\widehat{\widetilde z}_t^{(j)}
\right]
=
0\),
\(
j\neq j^\star,
\)
for every positive-definite block-diagonal prior covariance
$\Sigma_t$ and every anomaly
$\Delta$.
Thus, under orthogonal local component sensitivities, an anomaly
originating in one component produces no systematic posterior
attribution to any other component in this local Gaussian model.

\textbf{(c) Strict posterior-uncertainty reduction.}
The posterior covariance satisfies
\[
\widehat{\Sigma}_t=\left(\Sigma_t^{-1}+\mathcal I_t\right)^{-1}
\prec\Sigma_t.
\]
Equivalently, for every nonzero $v\in\mathbb R^d$,
\[
v^\top\widehat{\Sigma}_t v<v^\top\Sigma_t v.
\]
Thus, under $\mathcal I_t\succ0$, the interaction provides strictly
positive local information in every latent direction and strictly reduces
posterior uncertainty in every nonzero joint latent direction.
\end{styledtheorem}

\noindent\textbf{Interpretation.} The property established here is local distinguishability of the latent components from observable behavior when the local Fisher information is positive definite. This is important because component-specific regulation is meaningful only if different latent sources leave sufficiently distinguishable signatures in the observations. Full-rank local information and the Gaussian linearization are strong local conditions; they may hold when instrumentation and behaviors are informative, but can fail when dataset, hardware, and model effects are observationally confounded.

\begin{styledcorollary}{High-Probability Correct Source Attribution}{highprob_attribution}
Under the assumptions of part~(b) of
Theorem~\ref{thm:component_identifiability}, consider the following
source-attribution bound. The preceding uncertainty-reduction statement
is valid for the general vector-valued component blocks. For the following
source-attribution bound, additionally suppose that each latent component
is scalar, i.e., $d_j=1$ for every $j\in\mathcal{J}$. All component-level quantities in the attribution construction below are therefore scalar. Write
\(
W_t^{(j)}\in\mathbb{R}^{d_y}
\)
and
\(
\mathcal{I}_t^{(j)}
:=
\left(W_t^{(j)}\right)^\top
\Omega_t^{-1}
W_t^{(j)}
>0
\)
for the corresponding positive scalar diagonal block. Let
\(
\widehat{\Sigma}_t^{(j)}
=
\left(
(\Sigma_t^{(j)})^{-1}
+
\mathcal{I}_t^{(j)}
\right)^{-1}
\)
denote the scalar posterior variance of component $j$, and define
\(
\Xi_j
=
\left(
\widehat{\Sigma}_t^{(j)}
\right)^2
\mathcal{I}_t^{(j)}
\)
and the standardized posterior shift
\(
T_j
=
\frac{
\widehat{\widetilde z}_t^{(j)}
}{
\sqrt{\Xi_j}
}.
\)
Suppose the anomaly occurs in the unique component
$j^\star\in\mathcal{J}$ with magnitude
\(
\Delta\neq0.
\)
Define the source-attribution rule
\(
\widehat{\jmath}
=
\arg\max_{j\in\mathcal{J}}
|T_j|,
\)
with arbitrary deterministic tie-breaking.

Then
\(
\Pr
\!\left(
\widehat{\jmath}\neq j^\star
\right)
\leq
\min
\left\{
1,\,
\left(2|\mathcal{J}|-1\right)
\exp
\!\left(
-\frac{
\mathcal{I}_t^{(j^\star)}\Delta^2
}{8}
\right)
\right\}.
\)

Since $|\mathcal{J}|=4$ for CENTRY,
\(
\Pr
\!\left(
\widehat{\jmath}\neq j^\star
\right)
\leq
\min
\left\{
1,\,
7
\exp
\!\left(
-\frac{
\mathcal{I}_t^{(j^\star)}\Delta^2
}{8}
\right)
\right\}.
\)

Consequently,
\(
\Pr
\!\left(
\widehat{\jmath}\neq j^\star
\right)
\longrightarrow0
\)
as $|\Delta|\longrightarrow\infty$.
The resulting exponential bound depends on the prior covariance only
through a cancellation in the standardized statistic and is therefore
independent of $\Sigma_t$.
\end{styledcorollary}

\noindent\textbf{Interpretation.} The property established here gives an explicit high-probability bound for correctly identifying a single anomalous latent component from standardized posterior shifts. This is important because it quantifies when CENTRY can move from detecting an anomaly to localizing its likely source. The result assumes a single shifted scalar component and sufficient Fisher information, so it represents a clean best-case attribution setting; simultaneous or weakly observable causes require the broader or more cautious results that follow.

\begin{styledcorollary}{High-Probability Recovery of the Shifted-Component Set}{multi_attribution}
Adopt the notation and the block-diagonal Fisher-information assumption
of part~(b) of Theorem~\ref{thm:component_identifiability}. Suppose
additionally that each latent component is scalar, i.e., $d_j=1$ for
every $j\in\mathcal J$. Suppose that, instead of a single anomalous
component, a subset
$\mathcal K\subseteq\mathcal J$ of components experiences simultaneous
mean shifts, with $\mathcal K\neq\varnothing$, so that
\[
\widetilde y_t
=
\sum_{k\in\mathcal K}W_t^{(k)}\Delta_k+\eta_t,
\qquad
\eta_t\sim\mathcal N(0,\Omega_t),
\]
for nonzero shift magnitudes $\{\Delta_k\}_{k\in\mathcal K}$. Define, for
each $j\in\mathcal J$, the standardized posterior shift
$T_j:=\widehat{\widetilde z}_t^{(j)}/\sqrt{\Xi_j}$ as in
Corollary~\ref{cor:highprob_attribution}, where
$\widehat{\widetilde z}_t^{(j)}$ denotes the $j$-th block of the local
posterior mean shift $\widehat{\widetilde z}_t$ defined in
Theorem~\ref{thm:component_identifiability}, set
$m_k:=\Delta_k\sqrt{\mathcal I_t^{(k)}}$ for $k\in\mathcal K$ and
$m_{\min}:=\min_{k\in\mathcal K}|m_k|$, and let
\[
\widehat{\mathcal K}:=\{j\in\mathcal J:|T_j|\ge m_{\min}/2\}.
\]
Then
\[
\Pr\!\left(\widehat{\mathcal K}\ne\mathcal K\right)
\leq
\min\!\left\{
1,
\left(2|\mathcal J|-|\mathcal K|\right)
\exp\!\left(-\frac{m_{\min}^2}{8}\right)
\right\}.
\]
\end{styledcorollary}

\noindent\textbf{Interpretation.} The property established here extends source attribution to simultaneous shifts in several components and bounds the probability of recovering the affected component set. This is important because real failures can involve multiple components rather than a single isolated cause. The block-diagonal information structure and threshold construction simplify the problem and may not hold exactly when components interact strongly, so the guarantee is most applicable when the sources are locally distinguishable.

\begin{styledcorollary}{Sample Complexity of Component Discrimination}{sample_complexity_discrimination}
Under the assumptions of part~(b) of Theorem~\ref{thm:component_identifiability},
suppose that each latent component is scalar and that the local Fisher
information for component $j$ is $\mathcal I_t^{(j)}>0$. Consider two
competing hypotheses $H_{j_1}$ and $H_{j_2}$ about which single component
$j\in\{j_1,j_2\}$ is responsible for the observed elevated risk rating,
with $j_1\neq j_2$. Suppose that under $H_{j_k}$, the $k$-th component
experiences a nonzero mean shift of magnitude $\Delta_k$, while all other
components are unshifted. Define
\[
\Delta_{\min}:=
\min\{|\Delta_1|,|\Delta_2|\},
\qquad
\mathcal I_{\min}:=
\min_{j\in\{j_1,j_2\}}\mathcal I_t^{(j)},
\]
and let
\[
m_{\min}:=\sqrt{\mathcal I_{\min}}\,\Delta_{\min}.
\]
For $\delta\in(0,1)$ and integer $N\ge1$, suppose that $N$ conditionally independent observations with the same local model, information matrices, and shift magnitudes under each hypothesis are available and,
for each observation $n$, let
$T_{j,n}=\widehat{\widetilde z}_{t,n}^{(j)}/\sqrt{\Xi_{j,n}}$ be the
standardized posterior shift, with $\Xi_{j,n}=\Xi_j$ as defined in Corollary~\ref{cor:highprob_attribution}, since the local model and information matrices are the same across observations. Define the normalized aggregate statistic
\[
\overline T_j^{(N)}:=
\frac{1}{\sqrt N}\sum_{n=1}^{N}T_{j,n},
\]
and use the threshold-based attribution rule that selects $j_k$ when
$|\overline T_{j_k}^{(N)}|\ge m_{\min}\sqrt N/2$ and the other candidate
component does not cross this threshold; if neither or both candidate
components cross the threshold, the decision is declared a failure.
Then a sufficient sample size for correct discrimination with probability
at least $1-\delta$ is
\[
N
\ge
\frac{8}{\mathcal I_{\min}\Delta_{\min}^2}
\log\!\left(\frac{2|\mathcal J|-1}{\delta}\right).
\]
In particular, for $|\mathcal J|=4$,
\[
N
\ge
\frac{8}{\mathcal I_{\min}\Delta_{\min}^2}
\log\!\left(\frac{7}{\delta}\right).
\]
Thus, under the stated conditional-independence and Gaussian assumptions,
the Explain stage requires
$O\!\left((\mathcal I_{\min}\Delta_{\min}^2)^{-1}\log(1/\delta)\right)$
effectively independent observations to discriminate between the two
component-risk hypotheses with probability at least $1-\delta$.
\end{styledcorollary}

\noindent\textbf{Interpretation.} The property established here gives a sufficient number of conditionally independent observations for discriminating between competing component causes with a prescribed confidence. This is important because it turns source attribution into a measurable data-collection requirement. Conditional independence and scalar local Fisher information are simplifying assumptions; sequential AI interactions are often dependent, so the formula should be applied only when the required independence or an appropriate effective-sample-size approximation is justified.

\begin{styledtheorem}{Steady-State Posterior Uncertainty}{steady_state_riccati}
Consider a time-invariant linear-Gaussian component model with Gaussian initial state, fixed transition and observation matrices, and mutually independent Gaussian process and observation noise sequences, independent also of the initial state:
\[
z_{t+1}^{(j)}=A^{(j)}z_t^{(j)}+b^{(j)}+\varepsilon_{t+1}^{(j)},\qquad
 y_{t+1}^{(j)}=W^{(j)}z_{t+1}^{(j)}+\nu_{t+1}^{(j)},
\]
where $\varepsilon_{t+1}^{(j)}\sim\mathcal N(0,Q^{(j)})$ and $\nu_{t+1}^{(j)}\sim\mathcal N(0,\Omega^{(j)})$. Assume the component is filtered using this correctly specified model, and let
$\Sigma_t^{(j)}$ denote the posterior covariance of
$z_t^{(j)}$ given the observations through time $t$.
Assume that the process-noise covariance satisfies
\(Q^{(j)}\succeq0\) and that the observation-noise covariance satisfies
\(\Omega^{(j)}\succ0\).

\textbf{(a) Exact filtering recursion.}
The posterior covariance is independent of the realized observations
and satisfies the discrete-time Riccati recursion
\(
\Sigma_{t+1}^{(j)}
=
\mathcal R^{(j)}\!\left(\Sigma_t^{(j)}\right),
\)
where
\(
\bar\Sigma_{t+1}^{(j)}
=
A^{(j)}\Sigma_t^{(j)}(A^{(j)})^\top
+
Q^{(j)}
\)
is the one-step predicted covariance and
\[
\mathcal R^{(j)}(P)
=
\bar P
-
\bar P(W^{(j)})^\top
\left(
W^{(j)}\bar P(W^{(j)})^\top
+
\Omega^{(j)}
\right)^{-1}
W^{(j)}\bar P,
\]
with
\(
\bar P
=
A^{(j)}P(A^{(j)})^\top+Q^{(j)}.
\)
Equivalently, whenever $\bar P\succ0$,
\[
\mathcal R^{(j)}(P)
=
\left[
\bar P^{-1}
+
(W^{(j)})^\top
(\Omega^{(j)})^{-1}
W^{(j)}
\right]^{-1}.
\]

\textbf{(b) Steady-state covariance.}
Suppose that the pair
$(A^{(j)},W^{(j)})$ is detectable and that the pair
$\bigl(A^{(j)},(Q^{(j)})^{1/2}\bigr)$ is stabilizable. That is,
for every $\lambda\in\mathbb C$ with $|\lambda|\ge1$,
\(
\operatorname{rank}
\begin{bmatrix}
A^{(j)}-\lambda I\\
W^{(j)}
\end{bmatrix}
=
d_j,
\)
and
\(
\operatorname{rank}
\begin{bmatrix}
A^{(j)}-\lambda I &
(Q^{(j)})^{1/2}
\end{bmatrix}
=
d_j.
\)
Then the discrete-time algebraic Riccati equation
\(
\Sigma
=
\mathcal R^{(j)}(\Sigma)
\)
has a unique stabilizing solution
\(
\Sigma_\infty^{(j)}\in\mathbb S_+^{d_j}.
\)
If
\[
K_\infty^{(j)}
=
\bar\Sigma_\infty^{(j)}
(W^{(j)})^\top
\left(
W^{(j)}\bar\Sigma_\infty^{(j)}
(W^{(j)})^\top
+
\Omega^{(j)}
\right)^{-1},
\]
where
\(
\bar\Sigma_\infty^{(j)}
=
A^{(j)}\Sigma_\infty^{(j)}
(A^{(j)})^\top+Q^{(j)},
\)
then
\(
A^{(j)}
\left(I-K_\infty^{(j)}W^{(j)}\right)
\)
is Schur stable, and
\(
\Sigma_t^{(j)}
\longrightarrow
\Sigma_\infty^{(j)}
\) as $t\to\infty$
for every initial covariance
$\Sigma_1^{(j)}\in\mathbb S_+^{d_j}$.
Moreover, the convergence is geometric.
\end{styledtheorem}

\noindent\textbf{Interpretation.} The property established here characterizes the exact covariance recursion and its unique stabilizing steady state for linear-Gaussian component dynamics. This is important because it quantifies the irreducible uncertainty remaining in continual latent-state estimation and supplies the uncertainty used by downstream risk ratings. Detectability and stabilizability are standard Kalman-filter conditions, but the linear-Gaussian model is a tractable special case rather than a universal description of deployed AI behavior.

\begin{styledcorollary}{Closed-Form Steady-State Variance for a Scalar Component}{scalar_steady_state_variance}
Suppose that $d_j=1$ and write
\(
a:=A^{(j)},
\)
\(
q:=Q^{(j)}>0,
\)
\(
w:=W^{(j)}\neq0,
\)
and
\(
\omega:=\Omega^{(j)}>0.
\)
Then
\[
\Sigma_{t+1}^{(j)}
=
\frac{\omega\left(a^2\Sigma_t^{(j)}+q\right)}
{w^2\left(a^2\Sigma_t^{(j)}+q\right)+\omega},
\]
and
\[
\Sigma_t^{(j)}
\longrightarrow
\Sigma_\infty^{(j)}
\qquad\text{as }t\to\infty
\]
for every $\Sigma_1^{(j)}\ge0$.

The limiting variance is the unique non-negative solution of
\[
w^2a^2
\left(\Sigma_\infty^{(j)}\right)^2
+
\left(
w^2q+\omega-\omega a^2
\right)
\Sigma_\infty^{(j)}
-
\omega q
=
0.
\]
For $a\neq0$, it is given by
\[
\Sigma_\infty^{(j)}
=
\frac{
\omega a^2-w^2q-\omega
+
\sqrt{
\left(\omega a^2-w^2q-\omega\right)^2
+
4a^2w^2q\omega
}
}
{2a^2w^2}.
\]
For $a=0$,
\[
\Sigma_\infty^{(j)}
=
\frac{q\omega}{w^2q+\omega}.
\]
In particular,
\[
0<\Sigma_\infty^{(j)}<\infty.
\]
\end{styledcorollary}

\noindent\textbf{Interpretation.} The property established here gives a closed-form steady-state posterior variance for a scalar linear-Gaussian component. This is important as an analytic sanity check and makes the roles of process noise, observation strength, and state persistence transparent. The scalar assumptions are intentionally restrictive and therefore unlikely to describe a complex deployment exactly; the result is most useful for component-level analysis, parameter studies, and validating numerical implementations.

\subsection{Operational guarantees}
\label{subsec:Operational guarantees}
These results quantify what the monitored system can achieve operationally after a risk change is present: reduction, detection, containment, stabilization, and computational cost. The key assumptions are effective causal regulation, bounded risk growth, responsive detection, and suitable stochastic stability conditions. Unlike purely structural properties, these assumptions are deployment-dependent and should be supported by controlled perturbations, historical evidence, or operational validation before the numerical guarantees are relied upon.

\begin{styledlemma}{Foster--Lyapunov Bound on Expected Hitting Time}{foster_hitting_time}
Let $\{X_k\}_{k\ge0}$ be adapted to a filtration
$\{\mathcal F_k\}$, let $\mathcal T$ be a measurable set, and define
\(
\tau:=\inf\{k\ge0:X_k\in\mathcal T\}.
\)
Let $V\ge0$ be measurable with $\mathbb E[V(X_0)]<\infty$. If there exists a constant $c>0$ such that
\(
\mathbb E\!\left[
V(X_{k+1})\mid\mathcal F_k
\right]
\le
V(X_k)-c
\)
whenever \(X_k\notin\mathcal T,
\)
then
\(
\mathbb E[\tau]
\le
\frac{\mathbb E[V(X_0)]}{c}.
\)
\end{styledlemma}

\noindent\textbf{Interpretation.} The property established here bounds the expected time for a stochastic process to reach a target safe set when a nonnegative Lyapunov function decreases by a fixed amount outside that set. This is important because it provides the basic mathematical mechanism used later to bound containment time. The negative-drift condition is strong but standard in stochastic stability analysis; whether it holds for CENTRY must be established from the particular risk dynamics and regulatory policy.

\begin{styledtheorem}{Expected Risk Reduction Bound}{risk_reduction_bound}
Let $\mathcal{F}_t$ denote the information available at time $t$, including
the current component-risk ratings $\rho_t$ and the history of observations
and regulatory actions. Let $c_t$ be the regulatory action selected at time
$t$ according to $\pi_\psi(\rho_t)$, and let
\(
\Delta R_t^{(j)}
=
R_{t+1}^{(j)}-R_t^{(j)}
\)
denote the one-step change in the risk of component
$j\in\{D,H,M\}$. Define the set of components whose current risk ratings
exceed their budgets by
\(
\mathcal{S}_t
:=
\left\{
j\in\{D,H,M\}:
\rho_t^{(j)}>\bar B^{(j)}
\right\}.
\)

Suppose that, conditional on $\mathcal{F}_t$:

\begin{enumerate}[nosep]
    \item \textbf{Effective regulation.}
    With probability at least $p_{\mathrm{eff}}\in(0,1]$, the regulatory
    action $c_t$ is effective. More precisely, on an event
    $E_t\in\mathcal{F}_{t+1}$ with
    \(
    \Pr(E_t\mid\mathcal{F}_t)\geq p_{\mathrm{eff}},
    \)
    the following hold almost surely:
    \(
    \Delta R_t^{(j)}<0\),
    \(j\in\mathcal{S}_t,
    \)
    and
    \(
    \Delta R_t^{(j)}\leq 0\),
    \(j\notin\mathcal{S}_t.
    \)

    \item \textbf{Guaranteed reduction when effective.}
    There exist $\eta_j>0$ such that, conditional on $E_t$ and
    $\mathcal{F}_t$,
    \(
    \mathbb{E}\!\left[
        \Delta R_t^{(j)}
        \mid E_t,\mathcal{F}_t
    \right]
    \leq -\eta_j\),
    \(j\in\mathcal{S}_t.
    \)

    \item \textbf{Bounded risk growth.}
    There exist finite constants $\zeta_j\geq0$ such that
    \(
    \Delta R_t^{(j)}\leq\zeta_j
    \) almost surely, for every $j\in\{D,H,M\}$.
\end{enumerate}

Then the conditional expected one-step change in total component risk
satisfies
\[
\mathbb{E}\!\left[
    \left.
    \sum_{j\in\{D,H,M\}}\Delta R_t^{(j)}
    \,\right|\,\mathcal{F}_t
\right]
\leq
-p_{\mathrm{eff}}
\sum_{j\in\mathcal{S}_t}\eta_j
+
(1-p_{\mathrm{eff}})
\sum_{j\in\{D,H,M\}}\zeta_j.
\]

Furthermore, let
\(
A:=\sum_{j\in\{D,H,M\}}\zeta_j\),
\(
B_t:=\sum_{j\in\mathcal{S}_t}\eta_j.
\)
Whenever $A+B_t>0$, if
\(
p_{\mathrm{eff}}
\geq
\frac{A}{A+B_t},
\)
then
\[
\mathbb{E}\!\left[
    \left.
    \sum_{j\in\{D,H,M\}}\Delta R_t^{(j)}
    \,\right|\,\mathcal{F}_t
\right]
\leq 0.
\]
If the inequality is strict, i.e.,
\(
p_{\mathrm{eff}}
>
\frac{A}{A+B_t},
\)
then the conditional expected net change in total risk is strictly
negative.
\end{styledtheorem}

\noindent\textbf{Interpretation.} The property established here bounds CENTRY's expected one-step change in total component risk when effective regulation reduces risk on an affected component and harmful growth is bounded otherwise. This is important because it provides the direct bridge from selecting a regulatory action to an expected reduction in risk. The central effective-regulation probability and bounded-growth assumptions are empirical causal properties of the deployment; they are not implied by low policy regret or good prediction and must be independently established.

\begin{styledtheorem}{Bounded Expected Detection Delay}{detection_delay}
Fix a component $j\in\{D,H,M\}$ and suppress the superscript $(j)$
where unambiguous. Suppose the drift term $b_t^{(j)}$ in the
affine-Gaussian transition of Corollary~\ref{cor:gaussian-propagation}
is piecewise constant in time, with $A_t^{(j)}=a$, $|a|<1$, held fixed
throughout, and with $b_t^{(j)}$ equal to some baseline value for
$t<t_0$ and equal to a new value $b_{\mathrm{new}}$ for all $t\ge t_0$,
this new value being known to, and correctly used by, the filter from
$t_0$ onward. Assume the scalar linear-Gaussian observation model and independent Gaussian noises of Theorem~\ref{thm:steady_state_riccati}, with a nonzero scalar observation coefficient $w$ and the correctly specified steady-state filter. Let
\[
z^\star:=\frac{b_{\mathrm{new}}}{1-a}
\]
denote the new equilibrium latent state induced by this shift, let
$\mu_t:=\mathbb E[z_t\mid y_{1:t},c_{1:t-1}]$ denote the scalar
steady-state Kalman-filter posterior mean, and let $\Sigma_\infty$
denote the associated steady-state posterior variance of
Theorem~\ref{thm:steady_state_riccati}, already reached by time $t_0$. For this scalar specialization, assume that the Gaussian filter is correctly specified for the full information $\mathcal F_t$: conditional on $\mathcal F_t$, the filtering error has distribution $\mathcal N(0,\Sigma_\infty)$, and the next process and observation noises are mutually independent and jointly independent of the current latent state and $\mathcal F_t$.
Let $\bar b:=\inf\{\mu:g^{(j)}(\mu,\Sigma_\infty)\ge\bar B^{(j)}\}$ be the
latent-scale threshold corresponding to the risk-rating budget
$\bar B^{(j)}$, and suppose $g^{(j)}(\cdot,\Sigma_\infty)$ is
non-decreasing, with the threshold attained (that is, the risk rating at $\bar b$ is at least the budget), so that
\[
\mu_t\ge\bar b\Longrightarrow\rho_t^{(j)}=g^{(j)}(\mu_t,\Sigma_\infty)\ge\bar B^{(j)}.
\]

Assume:
\begin{enumerate}[label=(D\arabic*),ref=(D\arabic*)]
\item\label{ass:detection-nonneg} \emph{(Non-negative persistence.)} $0\le a<1$.
\item\label{ass:detection-margin} \emph{(Positive equilibrium margin.)} $\gamma:=z^\star-\bar b>0$.
\item\label{ass:detection-onset} \emph{(Sub-threshold onset.)} $\bar z:=\mu_{t_0}<\bar b$. This is a condition on the observed onset history. The statements below apply on the $\mathcal F_{t_0}$-measurable event $\{\bar z<\bar b\}$; averages over onset histories are taken conditional on this event when it has positive probability.
\item\label{ass:detection-responsive} \emph{(Responsiveness.)} With
$\sigma_\mu^2:=q-(1-a^2)\Sigma_\infty>0$ the innovation variance of the
posterior-mean process ($q$ the process-noise variance of $z_t$), where
$\xi_t\sim\mathcal{N}(0,\sigma_\mu^2)$ and
$\mathbb{E}[|\xi_t|]=\sigma_\mu\sqrt{2/\pi}$, and
$c:=(1-a)\gamma-\sigma_\mu\sqrt{2/\pi}>0$.
\end{enumerate}

Define the latent-mean detection time
$\tau_{\mathrm{detect}}:=\inf\{t\ge t_0:\mu_t\ge\bar b\}$ and the
rating detection time $\tau_\rho:=\inf\{t\ge t_0:\rho_t^{(j)}\ge\bar B^{(j)}\}$. Then
$\tau_\rho\le\tau_{\mathrm{detect}}$ almost surely, and, with
$\Delta:=z^\star-\bar z>\gamma>0$,
\[
\mathbb E[\tau_{\mathrm{detect}}-t_0\mid\mathcal F_{t_0}]\le\frac{\Delta}{c},
\qquad\text{hence}\qquad
\mathbb E[\tau_\rho-t_0\mid\mathcal F_{t_0}]\le\frac{\Delta}{c}.
\]
Thus, under a responsive filter and a sufficiently strong regime shift, CENTRY detects the onset of a sustained elevated-risk regime and crosses the corresponding component risk-rating threshold within an expected number of cycles, conditional on the onset information, bounded by $\Delta/c$. Averaging over sub-threshold onset histories, both expected delays are bounded by $\mathbb E[\Delta\mid\bar z<\bar b]/c$ when $\Delta$ is integrable under that conditional law.
\end{styledtheorem}

\noindent\textbf{Interpretation.} The property established here gives a finite expected detection delay for a sustained shift in one component's latent dynamics and guarantees that the risk-rating threshold is crossed no later than the latent-mean threshold. This is important because a regulator is useful only if persistent dangerous changes can be detected in a bounded time. The result assumes a stable scalar linear-Gaussian model, a known regime change, a positive separation margin, and sufficient filter responsiveness; these conditions can be approximated in controlled settings but are restrictive for arbitrary real-world shifts.

\begin{styledcorollary}{High-Confidence Detection Within a Finite Horizon}{detection_horizon}
Under the hypotheses of
Theorem~\ref{thm:detection_delay}, for every $\delta\in(0,1)$, almost surely on $\{\bar z<\bar b\}$,
\(
\Pr\left(
\tau_{\mathrm{detect}}-t_0
\le
\frac{\Delta}{c\delta}\mid\mathcal F_{t_0}
\right)
\ge
1-\delta.
\)
Moreover, for the previously defined risk-rating crossing time
\(
\tau_\rho
=
\inf\left\{
t\ge t_0:
\rho_t^{(j)}\ge\bar B^{(j)}
\right\},
\)
one also has
\(
\Pr\left(
\tau_\rho-t_0
\le
\frac{\Delta}{c\delta}\mid\mathcal F_{t_0}
\right)
\ge
1-\delta.
\)
Both probability bounds also hold conditional on the sub-threshold onset event $\{\bar z<\bar b\}$, when it has positive probability, by the tower property.
\end{styledcorollary}

\noindent\textbf{Interpretation.} The property established here converts the expected detection-delay bound into an explicit high-probability finite detection horizon. This is important because operational planning usually needs a confidence level and deadline rather than only an expected delay. It inherits all assumptions of the detection theorem and adds no new empirical mechanism; the bound is useful when those model and responsiveness assumptions have been validated, but can be loose when the expected-delay bound is conservative.

\begin{styledtheorem}{False-Alarm Rate under the Null}{false_alarm_rate}
Fix a monitored component $j\in\{D,H,M\}$ and consider the baseline
(no-excursion) regime. Let the latent state have stationary mean
\(
\bar z^{(j)}
=
\mathbb{E}\!\left[z_t^{(j)}\mid\text{baseline}\right].
\)
Assume:
\begin{enumerate}[label=(A\arabic*),ref=(A\arabic*)]
  \item\label{ass:steady-state}
  \textbf{Steady-state filter.}
  The filter is initialized in its invariant Kalman regime, so that
  \(
  \Sigma_t^{(j)}=\Sigma_\infty^{(j)}
  \)
  for all $t$, and the component risk rating of
  Eq.~(\ref{eq:riskfunctional}) reduces to
  \(
  \rho_t^{(j)}
  =
  g^{(j)}\!\left(\mu_t^{(j)},\Sigma_\infty^{(j)}\right).
  \)

  \item\label{ass:gauss-ar1}
  \textbf{Gaussian AR(1) posterior mean.}
  Under the baseline regime, the steady-state posterior mean is a
  stationary Gaussian AR(1) process,
  \(
  \mu_t^{(j)}-\bar z^{(j)}
  =
  a^{(j)}
  \bigl(\mu_{t-1}^{(j)}-\bar z^{(j)}\bigr)
  +
  \xi_t^{(j)}\),
  \(
  |a^{(j)}|<1,
  \)
  where the innovations are independent and identically distributed as
  \(
  \xi_t^{(j)}
  \sim
  \mathcal{N}\!\left(0,\sigma_{\mu}^{2,(j)}\right)\),
  \(
  \sigma_{\mu}^{2,(j)}>0.
  \)
  Consequently, the stationary marginal distribution is
  \(
  \mu_t^{(j)}
  \sim
  \mathcal{N}\!\left(
      \bar z^{(j)},v^{(j)}
  \right)\),
  \(
  v^{(j)}
  =
  \frac{\sigma_{\mu}^{2,(j)}}
       {1-(a^{(j)})^2}.
  \)

  \item\label{ass:monotone-g}
  \textbf{Monotone rating function.}
  The map
  \(
  \mu
  \longmapsto
  g^{(j)}\!\left(\mu,\Sigma_\infty^{(j)}\right)
  \)
  is non-decreasing.
\end{enumerate}

For a rating threshold $\bar B^{(j)}\in[0,1]$, define the corresponding
latent-state threshold by
\[
\bar b^{(j)}
=
\inf
\left\{
\mu\in\mathbb{R}:
g^{(j)}\!\left(\mu,\Sigma_\infty^{(j)}\right)
\geq
\bar B^{(j)}
\right\}.
\]
If the upper level set is empty, then $P_{\mathrm{FA}}^{(j)}=0$ trivially. Otherwise, assume that $\bar b^{(j)}\in\mathbb{R}$.

Whenever
\(
\bar b^{(j)}\geq\bar z^{(j)},
\)
the steady-state false-alarm probability
\(
P_{\mathrm{FA}}^{(j)}
:=
\Pr\!\left(
\rho_t^{(j)}\geq\bar B^{(j)}
\mid\text{baseline}
\right)
\)
satisfies
\[
P_{\mathrm{FA}}^{(j)}
=
\Pr\!\left(
\mu_t^{(j)}\geq\bar b^{(j)}
\mid\text{baseline}
\right)
\leq
\frac{1}{2}
\exp\!\left(
-\frac{
\bigl(\bar b^{(j)}-\bar z^{(j)}\bigr)^2
}{
2v^{(j)}
}
\right)
=
\frac{1}{2}
\exp\!\left(
-\frac{
\bigl(\bar b^{(j)}-\bar z^{(j)}\bigr)^2
\bigl(1-(a^{(j)})^2\bigr)
}{
2\sigma_{\mu}^{2,(j)}
}
\right).
\]
Moreover, the long-run empirical false-alarm rate converges almost
surely to $P_{\mathrm{FA}}^{(j)}$:
\(
\lim_{T\to\infty}
\frac{1}{T}
\sum_{t=1}^{T}
\mathbf{1}
\!\left\{
\rho_t^{(j)}\geq\bar B^{(j)}
\right\}
=
P_{\mathrm{FA}}^{(j)}
\) a.s. and therefore
\[
\lim_{T\to\infty}
\frac{1}{T}
\sum_{t=1}^{T}
\mathbf{1}
\!\left\{
\rho_t^{(j)}\geq\bar B^{(j)}
\right\}
\leq
\frac{1}{2}
\exp\!\left(
-\frac{
\bigl(\bar b^{(j)}-\bar z^{(j)}\bigr)^2
\bigl(1-(a^{(j)})^2\bigr)
}{
2\sigma_{\mu}^{2,(j)}
}
\right)
\qquad \text{a.s.}
\]
\end{styledtheorem}

\noindent\textbf{Interpretation.} The property established here bounds the steady-state false-alarm probability of a component-risk threshold under a baseline regime and connects that probability to the long-run empirical alarm rate. This is important because aggressive monitoring can otherwise create excessive verification or intervention cost. The result assumes a steady-state Gaussian AR(1) posterior-mean process and a monotone rating map; these assumptions are analytically convenient and may be reasonable after stabilization, but baseline distributions should be measured for each deployment.

\begin{styledtheorem}{Expected Time-to-Containment}{time_to_containment}
Let an excursion of component $j^{\star}\in\{D,H,M\}$ begin at time
$t_{0}$, and let
\(
  \tau_{\rho}
  \;=\;
  \inf\!\left\{
    t\ge t_{0}:
    \rho_{t}^{(j^{\star})}\ge\bar B^{(j^{\star})}
  \right\}
\)
be the rating-crossing (detection) time. Define the containment time by
\(
  \tau_{\mathrm{safe}}
  \;=\;
  \inf\!\left\{
    t\ge\tau_{\rho}:
    V(\rho_t)\le\eta
  \right\},
\)
where
\(
  V(\rho)
  =
  \sum_{j\in\{D,H,M\}}
  \bigl(\rho^{(j)}-\bar B^{(j)}\bigr)_{+}^{2}
\)
is the Lyapunov risk-excess function and $\eta>0$ is the safety
tolerance.

Assume:
\begin{enumerate}[label=(A\arabic*),ref=(A\arabic*)]

  \item\label{ass:detection-delay}
  \emph{(Detection.)}
  The rating-crossing time $\tau_{\rho}$ is a stopping time and its
  expected detection delay is finite, with
  \(
    \mathbb{E}\!\left[\tau_{\rho}-t_{0}\right] \le \frac{\Delta}{c}, \Delta>0, c>0.
  \)
  Here $\Delta$ and $c$ are finite deterministic constants defining an unconditional bound, not a random onset-dependent gap. To obtain this bound from Theorem~\ref{thm:detection_delay} for the same component and onset time, assume its model and responsiveness conditions hold, with fixed $c>0$. Let $A=\{\bar z<\bar b\}$ be the sub-threshold onset event. On $A^c$, the threshold is already reached at onset, so $\tau_\rho=t_0$. If $\Pr(A)=0$, (A1) holds for any $\Delta>0$. Otherwise, choose a finite deterministic $\Delta>0$ with $\Delta\ge\mathbb E[z^\star-\bar z\mid A]$, provided that conditional mean is finite. By the detection-delay theorem and the tower property, $\mathbb E[\tau_\rho-t_0]=\Pr(A)\mathbb E[\tau_\rho-t_0\mid A]\le\Delta/c$. Expectations and probabilities without explicit conditioning in this theorem and its end-to-end corollary are unconditional.

  \item\label{ass:active-regulation}
  \emph{(Active and effective regulation from detection onward.)}
  From time $\tau_{\rho}$ onward, the CENTRY regulation policy $c_t$
  is active and satisfies the effective-regulation condition of
  Theorem~\ref{thm:risk_reduction_bound} throughout the containment
  phase. Assume, in addition, that the Lyapunov dissipativity condition of
  Theorem~\ref{thm:asymptotic_risk_stability} holds for every
  $t\ge\tau_{\rho}$:
  \(
    \mathbb{E}\!\left[
      V(\rho_{t+1})-V(\rho_t)
      \,\middle|\,
      \mathcal{F}_t
    \right]
    \le
    -\alpha V(\rho_t)+\epsilon,
    \alpha>0, 
    \epsilon\ge0.
  \)

  \item\label{ass:safety-regime}
  \emph{(Contractive safety regime.)}
  The safety tolerance satisfies
  \(
    \alpha\eta>\epsilon.
  \)

\end{enumerate}

Then $\tau_{\mathrm{safe}}$ is finite almost surely and, conditional on
the information available at detection,
\[
  \mathbb{E}\!\left[
    \tau_{\mathrm{safe}}-\tau_{\rho}
    \,\middle|\,
    \mathcal{F}_{\tau_{\rho}}
  \right]
  \le
  \frac{
    V(\rho_{\tau_{\rho}})
  }{
    \alpha\eta-\epsilon
  }
  \qquad\text{a.s.}
\]
Consequently,
\(
  \mathbb{E}\!\left[
    \tau_{\mathrm{safe}}-\tau_{\rho}
  \right]
  \le
  \frac{
    \mathbb{E}\!\left[V(\rho_{\tau_{\rho}})\right]
  }{
    \alpha\eta-\epsilon
  },
\)
and the end-to-end excursion-to-containment time satisfies
\[
  \mathbb{E}\!\left[
    \tau_{\mathrm{safe}}-t_{0}
  \right]
  \le
  \frac{\Delta}{c}
  +
  \frac{
    \mathbb{E}\!\left[V(\rho_{\tau_{\rho}})\right]
  }{
    \alpha\eta-\epsilon
  }.
\]
Thus the expected time from excursion onset to containment decomposes
into a detection term and a post-detection regulation term.
\end{styledtheorem}

\noindent\textbf{Interpretation.} The property established here decomposes expected time from the onset of a sustained excursion to safe containment into detection time plus post-detection regulation time. This is important because it identifies two distinct operational bottlenecks: noticing the problem and successfully reducing the subsequent risk. The guarantee requires finite detection delay, effective regulation, Lyapunov dissipativity, and a strict safety-margin condition; these are strong closed-loop assumptions and must be demonstrated for each application.

\begin{styledcorollary}{High-Confidence End-to-End Excursion Resolution Time}{end_to_end_confidence}
Under the assumptions of Theorem~\ref{thm:time_to_containment}, for every
$\delta_2\in(0,1)$,
\[
\Pr\!\left(
\tau_{\mathrm{safe}}-\tau_\rho
\le
\frac{V(\rho_{\tau_\rho})}{(\alpha\eta-\epsilon)\delta_2}
\;\middle|\;
\mathcal F_{\tau_\rho}
\right)
\ge1-\delta_2
\qquad\text{a.s.}
\]
Consequently, for every $\delta_1,\delta_2\in(0,1)$ with
$\delta:=\delta_1+\delta_2<1$,
\[
\Pr\!\left(
\tau_{\mathrm{safe}}-t_0
\le
\frac{\Delta}{c\delta_1}
+
\frac{V(\rho_{\tau_\rho})}{(\alpha\eta-\epsilon)\delta_2}
\right)
\ge1-\delta.
\]
\end{styledcorollary}

\noindent\textbf{Interpretation.} The property established here provides a high-probability version of the end-to-end detection-to-containment bound by allocating confidence between detection and post-detection containment. This is important because it yields an explicit probabilistic service-level statement for the complete response process. It inherits the assumptions of both underlying bounds, so its real-world applicability depends on having credible detection and containment models; the confidence split is a design choice rather than an empirical fact.

\begin{styledcorollary}{Expected Cumulative Excess-Risk Exposure During Containment}{cumulative_exposure}
Under the assumptions of Theorem~\ref{thm:time_to_containment}, the
Lyapunov risk-excess accumulated over the containment window satisfies,
conditional on the information available at detection,
\[
\mathbb E\!\left[\sum_{t=\tau_\rho}^{\tau_{\mathrm{safe}}-1}V(\rho_t)\,\middle|\,\mathcal F_{\tau_\rho}\right]
\leq\frac{\eta\,V(\rho_{\tau_\rho})}{\alpha\eta-\epsilon}
\qquad\text{a.s.},
\]
and hence
\[
\mathbb E\!\left[\sum_{t=\tau_\rho}^{\tau_{\mathrm{safe}}-1}V(\rho_t)\right]
\leq\frac{\eta\,\mathbb E[V(\rho_{\tau_\rho})]}{\alpha\eta-\epsilon}.
\]
This quantity provides an ``excess-risk-time'' measure that combines the
magnitude and duration of a risk excursion.
\end{styledcorollary}

\noindent\textbf{Interpretation.} The property established here bounds cumulative excess risk during the containment interval, combining the magnitude and duration of an excursion into one quantity. This is important because two systems can have the same containment time while exposing users to very different levels of excess risk. The guarantee relies on the same Lyapunov drift assumptions as the containment theorem; these can be useful engineering summaries, but their accuracy depends on how well that drift model describes the real risk dynamics.

\begin{styledtheorem}{Mean-Square Stabilization under Persistent Regulation}{posterior_stabilization}
Fix a component \(j\in\{D,H,M\}\). Suppose that from a fixed finite time
\(\tau\) onward an effective regulatory action remains continuously active
and that the resulting latent-state dynamics are
\(
z_{t+1}^{(j)}
=
A^{(j)}z_t^{(j)}+b^{(j)}+\varepsilon_{t+1}^{(j)}\),
\(t\ge\tau.
\)
For this analysis, let $\mathcal G_t$ denote the filtration enlarged to include the latent states through time $t$ (not information available to CENTRY). Assume that
\(
\|A^{(j)}\|_2\le\gamma_j<1,
\)
and \(\{\varepsilon_t^{(j)}\}_{t\ge\tau+1}\) is a martingale-difference
sequence satisfying
\(
\mathbb E[\varepsilon_{t+1}^{(j)}\mid\mathcal G_t]=0\),
\(
\mathbb E[\|\varepsilon_{t+1}^{(j)}\|_2^2\mid\mathcal G_t]\le\sigma_j^2<\infty.
\)
Assume also $\mathbb E\|z_\tau^{(j)}-(I-A^{(j)})^{-1}b^{(j)}\|_2^2<\infty$. A zeroth power is interpreted as one, including for $\gamma_j=0$. Then the regulated dynamics have the unique equilibrium mean
\(
\mu_\infty^{(j)}=(I-A^{(j)})^{-1}b^{(j)},
\)
and, for every \(t\ge\tau\),
\[
\mathbb E\!\left[
\|z_t^{(j)}-\mu_\infty^{(j)}\|_2^2
\,\middle|\,
\mathcal G_\tau
\right]
\le
\gamma_j^{2(t-\tau)}
\|z_\tau^{(j)}-\mu_\infty^{(j)}\|_2^2
+
\sigma_j^2
\sum_{k=0}^{t-\tau-1}\gamma_j^{2k}.
\]
Consequently,
\(
\limsup_{t\to\infty}
\mathbb E\!\left[
\|z_t^{(j)}-\mu_\infty^{(j)}\|_2^2
\,\middle|\,
\mathcal G_\tau
\right]
\le
\frac{\sigma_j^2}{1-\gamma_j^2}.
\)
Thus persistent effective regulation makes the slow latent state
mean-square stable: the effect of the initial condition decays
geometrically, while persistent process noise leaves a bounded residual
dispersion.

If, in addition, the posterior mean is mean-square consistent for the
latent state,
\[
\mathbb E\!\left[
\|\mu_t^{(j)}-z_t^{(j)}\|_2^2
\right]\to0,
\]
then, since
\[
\|\mu_t^{(j)}-\mu_\infty^{(j)}\|_2^2\leq
2\|\mu_t^{(j)}-z_t^{(j)}\|_2^2
+2\|z_t^{(j)}-\mu_\infty^{(j)}\|_2^2,
\]
we have
\[
\limsup_{t\to\infty}
\mathbb E\!\left[
\|\mu_t^{(j)}-\mu_\infty^{(j)}\|_2^2
\right]
\leq
\frac{2\sigma_j^2}{1-\gamma_j^2}.
\]
\end{styledtheorem}

\noindent\textbf{Interpretation.} The property established here establishes mean-square stabilization of a persistent regulated component: the influence of its initial latent state decays geometrically while process noise leaves a bounded residual uncertainty. This is important because effective regulation should not merely reduce an immediate rating; it should place the underlying risk state into a stable regime. Contractive dynamics and bounded martingale noise are meaningful stability assumptions, but their validity is deployment-specific and should be tested after regulation is activated.

\begin{styledcorollary}{One-Step Contraction of the Risk-Rating Lyapunov Function}{one_step_lyapunov_contraction}
Adopt the notation of Theorem~\ref{thm:risk_reduction_bound}, and let
\(
V(\rho_t)=
\sum_{j\in\{D,H,M\}}
\bigl(\rho_t^{(j)}-\bar B^{(j)}\bigr)_+^2.
\)
Suppose that, whenever \(V(\rho_t)>0\), the effective-regulation analysis
yields the direct conditional Lyapunov-drift bound
\(
\mathbb{E}\!\left[
V(\rho_{t+1})-V(\rho_t)
\,\middle|\,
\mathcal F_t
\right]
\le
-p_{\mathrm{eff}}B_t+(1-p_{\mathrm{eff}})A,
\)
where
\(
A=\sum_{j\in\{D,H,M\}}\zeta_j\),
\(
B_t=\sum_{j\in\mathcal S_t}\eta_j,
\)
and
\(
\mathcal S_t
=
\{j:\rho_t^{(j)}>\bar B^{(j)}\}.
\)
If \(A+B_t>0\) and
\(
p_{\mathrm{eff}}
>
\frac{A}{A+B_t},
\)
then
\(
\mathbb{E}\!\left[
V(\rho_{t+1})
\,\middle|\,
\mathcal F_t
\right]
<
V(\rho_t)
\)
on the event \(\{V(\rho_t)>0\}\).
\end{styledcorollary}

\noindent\textbf{Interpretation.} The property established here gives a directly checkable sufficient condition for one-step contraction of CENTRY's quadratic risk-budget violation measure. This is important because it links the probability of effective intervention to a concrete decrease in the Lyapunov function used by the asymptotic results. The condition requires effective regulation to dominate possible uncontrolled risk growth; it can be estimated from controlled trials or monitoring data, but is not guaranteed merely by selecting a stronger-looking intervention.

\begin{styledcorollary}{Per-Cycle Computational Complexity of SELP}{selp_complexity}
Consider the sequential variational implementation described in
Section~\ref{sec:Implementation}. Let
\(
d=\sum_{j\in\{F,D,H,M\}}d_j
\)
denote the total latent dimension, let $h$ denote the hidden dimension
of the inference network, let $d_y$ denote the dimension of the current
interaction $y_t$, and let $\lvert\mathcal C\rvert$ denote the cardinality
of the finite admissible regulatory-action set. Assume that:

\begin{enumerate}[label=(i),nosep]
  \item the inference network maintains its recurrent state rather
  than recomputing the complete interaction history;
  \item the latent covariance is represented with a fixed-dimensional
  representation whose propagation costs $\mathcal O(d^2)$;
  \item the decoder and inference-network evaluations have complexity
  $\mathcal O(hd_y+d^2)$ and $\mathcal O(h^2+hd_y)$, respectively; and
  \item the number of optimization steps performed during a single
  \textsc{Learn} operation is bounded by a fixed positive integer $K\ge1$, with each step having the cost of a constant number of forward/backward evaluations; risk-rating extraction is $\mathcal O(d)$ and action evaluation costs $\mathcal O(1)$ per candidate. Any additional multi-step trajectory or chance-constraint evaluation costs are excluded.
\end{enumerate}

Then one complete SELP cycle has arithmetic complexity
\(
\mathcal O\!\left(
K(h^2+hd_y+d^2)
+
|\mathcal C|
+
d
\right).
\)
For fixed $K$ this can be written as
\(
\mathcal O\!\left(
h^2+hd_y+d^2+|\mathcal C|
\right).
\)
In particular, the per-cycle complexity does not grow with the length
of the interaction history when the recurrent state is maintained
incrementally.
\end{styledcorollary}

\noindent\textbf{Interpretation.} The property established here bounds the arithmetic cost of one complete SELP cycle and, importantly, shows that incremental recurrent processing avoids growth with the full interaction-history length. This is important for continual monitoring, where an algorithm that repeatedly rereads the entire history would eventually become impractical. The bound assumes fixed-dimensional covariance representation, bounded Learn iterations, and a finite action set; these are implementation choices that can be deliberately designed, although actual hardware cost depends on architecture and constants.

\subsection{System-level guarantees}
\label{subsec:System-level guarantees}
These results combine the component-level inference and regulation properties into system-level claims. Their assumptions are correspondingly stronger: the component-to-system risk relation, policy feasibility, prediction-error margins, calibration, and long-run dissipativity or invariance must all be valid where invoked. These conditions can be engineered or empirically assessed in constrained deployments, but they are not generic properties of arbitrary AI systems.

\begin{styledtheorem}{Pareto-Optimal Regulation Frontier}{pareto_regulation}
Let
\[
\mathcal{J}:=\{D,H,M\},
\]
and let
\(\Psi\subset\mathbb{R}^{d_\psi}\) be a non-empty compact parameter
space for the regulatory policy \(\pi_\psi\). For each component
\(j\in\mathcal{J}\), define the component risk-excess function
\[
V_j(r)
:=
\left(r-\bar B^{(j)}\right)_+^2.
\]
For \(\psi\in\Psi\), let the induced regulatory action be
$c_t=\pi_\psi(\rho_t)$ and define the three expected performance objectives
\[
S(\psi)
:=
\mathbb{E}_\psi[\widehat S_t],
\]
\[
V(\psi)
:=
\sum_{j\in\mathcal{J}}
\mathbb{E}_\psi
\!\left[
V_j\!\left(\rho_t^{(j)}\right)
\right],
\]
and
\[
C(\psi)
:=
\mathbb{E}_\psi[C(c_t)].
\]
Thus, smaller values of all three objectives are preferred. Define the
achievable objective set
\[
\Omega
:=
\left\{
\bigl(S(\psi),V(\psi),C(\psi)\bigr):
\psi\in\Psi
\right\}
\subset\mathbb{R}^3,
\]
and let \(\mathcal{P}\subseteq\Omega\) denote its Pareto frontier:
\[
\mathcal{P}
:=
\left\{
\omega\in\Omega:
\nexists\,\omega'\in\Omega
\text{ such that }
\omega'_k\leq\omega_k\ \forall k,
\quad
\omega'_k<\omega_k
\text{ for some }k
\right\}.
\]

Assume that the mapping
\[
F:\Psi\to\mathbb{R}^3,
\qquad
F(\psi)
=
\bigl(S(\psi),V(\psi),C(\psi)\bigr),
\]
is continuous. Then the following hold.

\begin{enumerate}[nosep]

\item \textbf{Existence of Pareto-optimal operating points.}
The achievable objective set \(\Omega=F(\Psi)\) is non-empty and compact,
and the Pareto frontier \(\mathcal{P}\) is non-empty.

\item \textbf{Scalarization yields Pareto-optimal operating points.}
Let
\[
\lambda_S>0,
\qquad
\lambda_V>0,
\qquad
\lambda_C>0,
\]
and define the weighted-sum objective
\[
J(\psi)
:=
\lambda_S S(\psi)
+
\lambda_V V(\psi)
+
\lambda_C C(\psi).
\]
Because \(\Psi\) is compact and \(J\) is continuous, a global minimizer
\[
\psi^\star
\in
\arg\min_{\psi\in\Psi}J(\psi)
\]
exists. Every such global minimizer induces a Pareto-optimal operating
point
\[
\omega^\star
=
\bigl(
S(\psi^\star),
V(\psi^\star),
C(\psi^\star)
\bigr)
\in\mathcal{P}.
\]
This scalarization result does not imply that every Pareto-optimal
operating point can be obtained by a positive weighted sum; such a
converse generally requires additional structural assumptions, such as
convexity of the achievable objective set.

\item \textbf{First-order stationarity.}
If \(S,V,C\) are each differentiable in a neighborhood of a global minimizer
\(\psi^\star\) and
\[
\psi^\star\in\operatorname{int}(\Psi),
\]
then
\[
\nabla_\psi J(\psi^\star)=0,
\]
that is,
\[
\left.
\left[
\lambda_S\nabla_\psi S(\psi)
+
\lambda_V\nabla_\psi V(\psi)
+
\lambda_C\nabla_\psi C(\psi)
\right]
\right|_{\psi=\psi^\star}
=0.
\]
If \(\psi^\star\) lies on the boundary of \(\Psi\), and if \(J\) is differentiable and, locally around
\(\psi^\star\), the feasible parameter set admits a differentiable
constraint representation
\[
\Psi
=
\left\{
\psi:
g_r(\psi)\leq0,\ r=1,\ldots,m,
\quad
h_s(\psi)=0,\ s=1,\ldots,\ell
\right\},
\]
with an appropriate constraint qualification holding at
\(\psi^\star\), then the Karush--Kuhn--Tucker (KKT) necessary conditions
hold: there exist multipliers
\[
\alpha_r\geq0,
\qquad
\beta_s\in\mathbb{R},
\]
such that
\[
\nabla_\psi J(\psi^\star)
+
\sum_{r=1}^{m}
\alpha_r\nabla g_r(\psi^\star)
+
\sum_{s=1}^{\ell}
\beta_s\nabla h_s(\psi^\star)
=0,
\]
together with
\[
g_r(\psi^\star)\leq0,
\qquad
h_s(\psi^\star)=0,
\]
and
\[
\alpha_r g_r(\psi^\star)=0,
\qquad
r=1,\ldots,m.
\]

\end{enumerate}
\end{styledtheorem}

\noindent\textbf{Interpretation.} The property established here establishes existence of Pareto-optimal operating points and shows that positive weighted scalarization produces Pareto-optimal solutions, with first-order or KKT conditions under additional regularity. This is important because CENTRY must balance predictive performance, risk excess, and regulatory cost rather than optimize a single quantity in isolation. Compactness and continuity are manageable design assumptions, while the stronger converse---that every Pareto point is recovered by weighted sums---requires additional structure and is explicitly not claimed.

\begin{styledtheorem}{Almost-Sure Asymptotic Budget Compliance}{almost_sure_compliance}
Adopt the setting of Theorem~\ref{thm:asymptotic_risk_stability} and
suppose that the dissipativity condition holds in the homogeneous form
\[
\mathbb{E}\!\left[
  V(\rho_{t+1})
  \,\middle|\,
  \mathcal F_t
\right]
\leq
(1-\alpha)V(\rho_t)
\]
a.s. for every \(t\geq1\), for some \(\alpha\in(0,1]\). In this theorem and its corollaries, a zeroth power is one, including when the base is zero. Suppose also that
\[
V_0:=\mathbb E[V(\rho_1)]<\infty.
\]
Then
\[
\mathbb E[V(\rho_t)]
\leq
(1-\alpha)^{t-1}V_0,
\qquad
t\geq1,
\]
and
\[
V(\rho_t)\longrightarrow0
\qquad\text{almost surely}.
\]
Consequently,
\[
\bigl(\rho_t^{(j)}-\bar B^{(j)}\bigr)_+
\longrightarrow0
\qquad\text{a.s.}
\]
for every \(j\in\{D,H,M\}\).

Thus homogeneous dissipativity guarantees asymptotic budget
compliance in the sense that any budget violation converges to zero.
Finite-time permanent entry into the feasible region requires an
additional assumption.
\end{styledtheorem}

\noindent\textbf{Interpretation.} The property established here strengthens homogeneous Lyapunov dissipativity to almost-sure asymptotic decay of all component risk-budget violations, with an explicit geometric bound on expected violation energy. This is important because it gives a strong long-run compliance statement rather than only a finite-horizon probability bound. The homogeneous drift condition is stronger than ordinary bounded-error dissipativity and therefore should be viewed as a stringent stability certificate that must be demonstrated for a particular closed loop.

\begin{styledcorollary}{L1 Convergence and Uniform Integrability of the Risk-Rating Violation}{uniform_integrability}
Under the assumptions of Theorem~\ref{thm:almost_sure_compliance},
\(V(\rho_t)\to0\) almost surely and
\(
\mathbb E[V(\rho_t)]\leq(1-\alpha)^{t-1}V_0\to0
\).
Consequently,
\[
V(\rho_t)\longrightarrow0
\qquad\text{in }L^1,
\]
and the family
\(\{V(\rho_t):t\geq1\}\) is uniformly integrable. In particular,
\[
\lim_{t\to\infty}\mathbb E[V(\rho_t)]
=
\mathbb E\left[\lim_{t\to\infty}V(\rho_t)\right]
=0.
\]
Thus the almost-sure and mean budget-compliance conclusions are
compatible with interchange of the limit and expectation.
\end{styledcorollary}

\noindent\textbf{Interpretation.} The property established here shows that the asymptotic almost-sure and mean risk-compliance results also hold in L1 and provide uniform integrability, allowing the limit and expectation to be interchanged. This is important because it prevents the mean result from hiding pathological rare trajectories and makes the probabilistic and expectation-based views mathematically consistent. It relies entirely on the stronger homogeneous dissipativity result, so its real-world validity is exactly as strong as the stability certificate behind that result.

\begin{styledcorollary}{Tail Probability of Future Risk-Budget Violations}{future_violation_tail}
Under the assumptions of
Theorem~\ref{thm:almost_sure_compliance}, let
\[
V_0:=\mathbb E[V(\rho_1)]<\infty.
\]
Then, for every \(\eta>0\) and every integer \(n\ge1\),
\[
\Pr\!\left(
\sup_{t\ge n}V(\rho_t)\ge\eta
\right)
\leq
\min\!\left\{
1,\,
\frac{V_0}{\eta}
(1-\alpha)^{n-1}
\right\}.
\]
Consequently,
\[
\Pr\!\left(
\sup_{t\ge n}
\max_{j\in\{D,H,M\}}
\bigl(\rho_t^{(j)}-\bar B^{(j)}\bigr)_+
\ge\sqrt{\eta}
\right)
\leq
\min\!\left\{
1,\,
\frac{V_0}{\eta}
(1-\alpha)^{n-1}
\right\}.
\]
Thus, under homogeneous dissipativity, the probability of experiencing
any future risk-budget violation exceeding a prescribed tolerance
decays geometrically with the time elapsed under regulation.
\end{styledcorollary}

\noindent\textbf{Interpretation.} The property established here bounds the probability of any future budget violation exceeding a chosen tolerance after a specified time, with geometric decay under homogeneous dissipativity. This is important because it controls the worst violation over an entire future tail rather than at one preselected time. The assumption remains the strong homogeneous drift condition; in practice, such a bound is most credible after closed-loop stabilization has been empirically established.

\begin{styledcorollary}{High-Probability Eventual Risk-Budget Tolerance}{highprob_eventual_tolerance}
Under the assumptions of
Corollary~\ref{cor:future_violation_tail}, let
\(\eta>0\) and \(\delta\in(0,1)\). If \(V_0=0\), the conclusions below hold with $n=1$. Otherwise, if \(0<\alpha<1\) and
\[
n
\ge
1+
\frac{
\log\!\left(
V_0/(\eta\delta)
\right)
}{
-\log(1-\alpha)
},
\]
then
\[
\Pr\!\left(
V(\rho_t)<\eta
\quad
\text{for all }t\ge n
\right)
\ge
1-\delta.
\]
Consequently,
\[
\Pr\!\left(
\bigl(\rho_t^{(j)}-\bar B^{(j)}\bigr)_+
<
\sqrt{\eta},
\quad
\forall j\in\{D,H,M\},
\quad
\forall t\ge n
\right)
\ge
1-\delta.
\]

For \(\alpha=1\), the homogeneous dissipativity condition implies
\(V(\rho_t)=0\) almost surely for every \(t\ge2\).

Thus, for any prescribed violation tolerance and confidence level,
CENTRY admits an explicit finite burn-in horizon after which the
risk-budget violation remains below that tolerance for all future
times with probability at least \(1-\delta\).
\end{styledcorollary}

\noindent\textbf{Interpretation.} The property established here converts the geometric tail bound into an explicit burn-in time after which the risk-budget violation stays below a desired tolerance with a desired confidence. This is important because it gives an interpretable operational settling-time guarantee. It requires homogeneous dissipativity and, for the explicit logarithmic formula, a nondegenerate contraction parameter; such conditions are attainable in stable controlled regimes but are not generally expected during arbitrary changing operation.

\begin{styledtheorem}{Robust Recursive Feasibility under Bounded Risk-Prediction Error}{robust_recursive_feasibility}
Let
\[
\mathcal F_t=\sigma(y_{1:t},c_{1:t-1})
\]
and let $\mathcal C$ denote the admissible regulatory-action set. Fix a
prediction horizon $T\geq1$. Suppose that CENTRY predicts normalized
component risk ratings
\[
\widehat\rho_{t+k\mid t}^{(j)},
\qquad
j\in\{D,H,M\},
\qquad k=1,\ldots,T,
\]
for each candidate action sequence (the dependence of the predictions on that sequence is suppressed in the notation). Assume that the future ratings under application of each candidate sequence satisfy the bounded prediction-error
condition
\[
\rho_{t+k}^{(j)}
\leq
\widehat\rho_{t+k\mid t}^{(j)}+e_k^{(j)}
\]
almost surely, where each known margin $e_k^{(j)}\geq0$ is $\mathcal F_t$-measurable for the candidate sequence. The predictions and margins are evaluated for the particular sequence appearing in the feasible-set definition. The error bound is assumed valid for every admissible history-measurable selection of a candidate sequence. Define
\[
\mathcal A_t^{\mathrm{rob}}
=
\left\{
(c_t,\ldots,c_{t+T-1})\in\mathcal C^T:
\widehat\rho_{t+k\mid t}^{(j)}+e_k^{(j)}
\leq\bar B^{(j)},
\ \forall j\in\{D,H,M\},\ \forall k=1,\ldots,T
\right\}.
\]
Suppose that:
\begin{enumerate}[nosep]
\item \textbf{Initial robust feasibility:}
$\mathcal A_1^{\mathrm{rob}}\neq\varnothing$.
\item \textbf{Robust shift-and-append:} whenever
$(c_t,\ldots,c_{t+T-1})\in\mathcal A_t^{\mathrm{rob}}$, after observing
$y_{t+1}$ and updating the posterior there exists a terminal action
$c_{t+T}\in\mathcal C$ such that
$(c_{t+1},\ldots,c_{t+T})\in\mathcal A_{t+1}^{\mathrm{rob}}$
almost surely.
\item \textbf{Receding-horizon implementation:} at each time $t$, CENTRY
selects a sequence in $\mathcal A_t^{\mathrm{rob}}$, applies its first
action, observes $y_{t+1}$, updates its posterior, and resolves the
finite-horizon problem.
\end{enumerate}
Then
\[
\mathcal A_t^{\mathrm{rob}}\neq\varnothing
\qquad\text{almost surely for every }t\geq1,
\]
and along the receding-horizon closed loop every future rating satisfies
\[
\rho_{t+k}^{(j)}\leq\bar B^{(j)}
\qquad
j\in\{D,H,M\},\quad k=1,\ldots,T,
\]
almost surely.
\end{styledtheorem}

\noindent\textbf{Interpretation.} The property established here is recursive feasibility of a robust finite-horizon regulation problem under bounded future rating-prediction error. This is important because a safety plan is useful only if it remains executable after each new observation rather than becoming infeasible one step later. The crucial assumptions are an initially feasible robust plan, a shift-and-append feasibility property, valid error margins, and receding-horizon implementation; these are standard control-design conditions but must be constructed and validated for the actual intervention interface.

\begin{styledcorollary}{Operational-Risk Safety Under Robust Recursive Feasibility}{robust_operational_safety}
Adopt the setting and conclusion of Theorem~\ref{thm:robust_recursive_feasibility},
so that a robustly feasible action sequence is selected at every $t$ and
\[
\rho_{t+k}^{(j)}\leq\bar B^{(j)}
\qquad\text{almost surely}
\]
along the receding-horizon closed loop for every fixed $t\ge1$, $j\in\{D,H,M\}$, and $k=1,\ldots,T$. Fix such a time $t$. Suppose in addition, for each
$j,k$, that the dangerous-false-negative condition of
Theorem~\ref{thm:rating_to_operational_safety} holds:
\[
\Pr\!\left(
R_{t+k}^{(j)}>B^{(j)},
\rho_{t+k}^{(j)}\leq\bar B^{(j)}\mid\mathcal F_t
\right)
\leq\eta_k^{(j)}.
\]
Then
\[
\Pr\!\left(R_{t+k}^{(j)}>B^{(j)}\mid\mathcal F_t\right)
\leq\eta_k^{(j)}
\qquad\text{almost surely},
\]
with no contribution from a rating-overshoot term. Consequently, if in
addition
\[
\sum_{k=1}^T\sum_{j\in\{D,H,M\}}\eta_k^{(j)}\leq\delta,
\]
then
\[
\Pr\!\left(
\bigcap_{k=1}^T\bigcap_{j\in\{D,H,M\}}
\left\{R_{t+k}^{(j)}\leq B^{(j)}\right\}
\ \middle|\ \mathcal F_t
\right)
\geq1-\delta.
\]
Thus, robust recursive feasibility removes the rating-overshoot contribution
from the rating-to-operational-risk transfer bound; the remaining safety
uncertainty is entirely due to the calibrated dangerous-false-negative rate.
\end{styledcorollary}

\noindent\textbf{Interpretation.} The property established here shows that robust recursive feasibility removes the rating-overshoot term from the transfer of rating-based safety to operational-risk safety. This is important because bounding future ratings within the feasible region isolates the remaining uncertainty to dangerous false negatives. The result therefore depends on both robust prediction margins and calibrated false-negative probabilities; these are measurable but system-specific quantities, not generic properties of a learned monitor.

\begin{styledcorollary}{Almost-Sure Eventual Feasibility}{eventual_feasibility}
Under the assumptions of Theorem~\ref{thm:almost_sure_compliance}, suppose
in addition that there exists a constant $\varepsilon_{\mathrm{feas}}>0$ such that every
violation satisfies
\[
V(\rho_t)>0
\quad\Longrightarrow\quad
V(\rho_t)\geq\varepsilon_{\mathrm{feas}}
\qquad\text{a.s.}
\]
Then there exists a finite random time
\(T_{\mathrm{safe}}\) such that
\[
V(\rho_t)=0
\qquad
\text{for all }t\geq T_{\mathrm{safe}}
\]
almost surely. Equivalently,
\[
\rho_t^{(j)}
\leq
\bar B^{(j)},
\qquad
\forall j\in\{D,H,M\},
\quad
\forall t\geq T_{\mathrm{safe}},
\]
almost surely.

Thus, under a positive minimum violation margin $\varepsilon_{\mathrm{feas}}$, homogeneous
Lyapunov dissipativity guarantees not merely asymptotic decay of the
risk excess but finite-time permanent entry into the feasible region
almost surely.
\end{styledcorollary}

\noindent\textbf{Interpretation.} The property established here strengthens asymptotic decay of violation energy to finite-time permanent entry into the feasible region when every nonzero violation has a positive minimum size. This is important because it converts an asymptotic statement into eventual sustained feasibility. The minimum-violation-margin assumption is quite strong and may fail when risk ratings vary continuously near a threshold, so this result should be regarded as applicable when the system has a meaningful operational deadband or quantized violation margin.

\begin{styledtheorem}{Invariance of the Feasible Design Region under Deterministic No Intervention}{feasible_invariance}
Let
\[
\mathcal F_{\mathrm{feas}}
=
\left\{
\rho\in[0,1]^3:
\rho^{(j)}\le\bar B^{(j)}
\ \text{for all }j\in\{D,H,M\}
\right\}
\]
denote the feasible design region. Suppose that at time $t_0$,
\(
\rho_{t_0}\in\mathcal F_{\mathrm{feas}}.
\)
Assume that the no-intervention action
\(
c_t=c^{(0)}
\)
for all $t\ge t_0$ retains the current configuration and, in the setting of this theorem, induces a deterministic risk-rating transition
\(
\rho_{t+1}
=
\Phi^{(0)}(\rho_t),
\)
and that the feasible region is invariant under this transition:
\(
\Phi^{(0)}(\mathcal F_{\mathrm{feas}})\subseteq\mathcal F_{\mathrm{feas}}.
\)
Then
\(
\rho_t\in\mathcal F_{\mathrm{feas}}
\) for all $t\ge t_0$.
Equivalently,
\(
\rho_t^{(j)}\le\bar B^{(j)}
\)
for all $j\in\{D,H,M\}$
and all $t\ge t_0$.
Thus, once entered, the feasible design region is invariant under the
pure no-intervention policy whenever the induced deterministic
risk-rating dynamics preserve that region.
\end{styledtheorem}

\noindent\textbf{Interpretation.} The property established here establishes forward invariance of the feasible design region under a deterministic no-intervention transition. This is important because it identifies conditions under which a system that is already safe can remain safe without continual intervention. The invariance condition is a structural property of the uncontrolled risk dynamics; it can be verified for known deterministic models but is not generally guaranteed in stochastic or evolving real-world systems.

\begin{styledcorollary}{Sufficient Condition for No-Intervention Invariance}{no_intervention_invariance_sufficient}
Suppose that $\rho_{t_0}\in\mathcal F_{\mathrm{feas}}$ and $c_t=c^{(0)}$ for all $t\ge t_0$, and suppose
that under the no-intervention action the posterior mean and posterior
covariance of every monitored component remain unchanged for every $t\geq t_0$, where $t_0$ is the activation time of the no-intervention policy:
\[
\mu_{t+1}^{(j)}
=
\mu_t^{(j)},
\qquad
\Sigma_{t+1}^{(j)}
=
\Sigma_t^{(j)},
\qquad
j\in\{D,H,M\}.
\]
Then
\[
\rho_{t+1}^{(j)}
=
\rho_t^{(j)},
\qquad
j\in\{D,H,M\},
\]
and hence $\rho_t\in\mathcal F_{\mathrm{feas}}$ for all $t\ge t_0$; that is, the trajectory remains in the feasible design region under the no-intervention policy.
\end{styledcorollary}

\noindent\textbf{Interpretation.} The property established here gives a simple sufficient condition for a feasible trajectory to remain feasible without intervention: if posterior means and covariances remain unchanged, the component risk ratings remain unchanged as well. This is important because it provides an easily checked special case of safe persistence without intervention. The condition is strong and generally corresponds to a stationary or exactly balanced situation; it can hold in controlled stable settings, but changing environments or stochastic updates normally require the broader invariance analysis of the preceding theorem.


\section{Implementation}
\label{sec:Implementation}
This section specifies one implementation of CENTRY using standard variational latent-state modeling. The implementation is architecture independent; sequential variational autoencoders, variational recurrent neural networks, transformer-based state-space models, or other probabilistic latent-state architectures may be used. The implementation preserves the distinction between the monitored AI system's observable action $a_t$ and CENTRY's regulatory action $c_t$.

Let $\theta=\left(\theta_{\mathrm{gen}},\theta_{\mathrm{trans}}\right)$ and $\phi=\theta_{\mathrm{inf}}$ be the parameters of the generative, transition, and inference models, respectively. The inference model approximates the posterior in Eq.~\eqref{eq:jointposterior} as
\begin{equation*}
q_\phi\!\left(z_t\mid y_{1:t},c_{1:t-1}\right) = 
\mathcal N\!\left(\mu_\phi(h_t),\operatorname{diag}\!\left(\sigma_\phi^2(h_t)\right)\right),
\end{equation*}
where $h_t$ is a learned representation of the observable interaction history and previously applied regulatory actions.

The transition model preserves the component-wise latent structure:
\begin{equation*}
\begin{aligned}
p_\theta\!\left(z_{t+1}\mid z_t,y_t,c_t\right)
&= \prod_{j\in\{F,D,H,M\}}p_\theta\!\left(z_{t+1}^{(j)}\mid z_t^{(j)},y_t,c_t\right)\\
&= \prod_{j\in\{F,D,H,M\}}\mathcal N\!\left(z_{t+1}^{(j)};
\mu_{\mathrm{trans}}^{(j)}\!\left(z_t^{(j)},y_t,c_t\right),
\Sigma_{\mathrm{trans}}^{(j)}\!\left(z_t^{(j)},y_t,c_t\right)\right).
\end{aligned}
\end{equation*}

The behavior decoder models $p_\theta(y_{t+1}\mid z_{t+1})$, where $y_{t+1}=(x_{t+1},a_{t+1})$. Thus, the regulatory action $c_t$ affects future observable behavior through the latent transition model rather than by directly specifying the monitored AI system's action $a_{t+1}$.

\subsection{Algorithmic details and SELP integration}
\label{subsec:Algorithmic details and SELP integration}
Algorithm~\ref{alg:CENTRY} summarizes the continual operation of CENTRY. At each time step, CENTRY predicts the observable behavior, receives the actual behavior, computes Surprise, performs inference and risk estimation, selects a regulatory action, updates both its probabilistic model and regulation policy, and propagates the posterior to the next predictive prior.

\begin{algorithm}[htbp!]
\caption{CENTRY: Continual AI risk estimation, monitoring, and regulation}
\label{alg:CENTRY}
\begin{algorithmic}[1]

\State \textbf{Initialize:} latent priors $\kappa_1(z_1^{(F)})=\mathcal{N}(0,I)$, $\kappa_1(z_1^{(D)})$, $\kappa_1(z_1^{(H)})$, and $\kappa_1(z_1^{(M)})$; initialize model parameters $\theta_1$, inference parameters $\phi_1$, and policy parameters $\psi_1$ (with $\psi_1\in\Psi$ for projected policy learning).

\FOR{$t=1,2,\ldots$}
    \State \textbf{Predict:} Compute $p_{\mathrm{pred}}(y_t\mid y_{1:t-1},c_{1:t-1})$ 
    and its expected behavior $\hat{y}_t=(\hat{x}_t,\hat{a}_t)$.

    \State \textbf{Receive:} the observable interaction $y_t=(x_t,a_t)$.

    \State \textbf{Surprise:} Compute the realized predictive surprise $\widehat S_t =
    -\log p_{\mathrm{pred}}(y_t\mid y_{1:t-1},c_{1:t-1})$.
    The population mismatch $S_t^{\mathrm{(KL)}}$ is a theoretical quantity and
    is not directly available at runtime.

    \State \textbf{Explain:} Infer $q_\phi(z_t\mid y_{1:t},c_{1:t-1})$.

    \State Decompose $z_t=(z_t^{(F)},z_t^{(D)},z_t^{(H)},z_t^{(M)})$
    and compute $\rho_t^{(j)} = g^{(j)}(\mu_t^{(j)},\Sigma_t^{(j)})$ for $j\in\{D,H,M\}$.

    \State Select the regulatory action $c_t=\pi_{\psi_t}(\rho_t)$.

    \State \textbf{Learn:} Update the probabilistic model and inference parameters $(\theta_t,\phi_t)$ using Eq.~\eqref{eq:ELBO}, and update the regulation-policy parameters $\psi_t$ using the model-based prospective objective in Eq.~\eqref{eq:policylearning}, obtaining $(\theta_{t+1},\phi_{t+1},\psi_{t+1})$.

    \State Propagate the posterior retained from Explain through the updated transition model under the selected regulatory action $c_t$ using Eq.~\eqref{eq:priorupdate} to obtain $\kappa_{t+1}(z_{t+1}^{(j)})$ for $j\in\{F,D,H,M\}$.

    \State Apply $c_t$ through the admissible intervention interface.
\ENDFOR
\end{algorithmic}
\end{algorithm}

The algorithm requires neither access to the internal state nor knowledge of the action-generation mechanism of the monitored AI system. CENTRY observes only $y_t=(x_t,a_t)$ and applies $c_t$ through the admissible intervention interface. The effect of each regulatory action is inferred from its subsequent effect on the observable interaction sequence.

\subsection{Computational complexity}
\label{subsec:Computational complexity}
Let $d=\sum_{j\in\{F,D,H,M\}}d_j$ be the total latent dimension and let $d_y$ be the dimension of the observable behavior $y_t=(x_t,a_t)$. For a recurrent implementation with hidden dimension $h$, the dominant per-step cost is associated with the inference, transition, and generative networks. Under the assumptions of Corollary~\ref{cor:selp_complexity}, including bounded Learn steps $K$, a finite action set, maintained recurrent state, and the stated covariance-propagation cost, the complete per-cycle arithmetic complexity is $\mathcal O(K(h^2+hd_y+d^2)+|\mathcal C|+d)$. Additional costs of evaluating multi-step predictive trajectories and chance constraints depend on the horizon and implementation.

The component-risk ratings $\rho_t^{(j)}=g^{(j)}(\mu_t^{(j)},\Sigma_t^{(j)})$ and evaluation of a compact regulatory policy add comparatively little computational cost. Policy learning may increase the cost according to the number of model-based optimization steps used to evaluate candidate regulatory policies. Thus, the computational complexity of CENTRY is dominated by the selected probabilistic inference and generative architectures rather than by risk estimation or regulatory-action selection.

\subsection{Hyperparameter selection}
\label{subsec:Hyperparameter selection}
The principal hyperparameters of CENTRY include the Surprise threshold $\tau$, latent dimensions ($d_F$, $d_D$, $d_H$, $d_M$), variational regularization weight $\beta$, learning rate and number of online optimization steps, initial component-risk priors, and any thresholds or costs used by the regulatory policy. These parameters should be selected using validation data or controlled simulations that are separate from the final evaluation data.

The threshold $\tau$ controls the sensitivity of the additional diagnostic, verification, or adaptation effort triggered by unusually large predictive mismatch. When labeled risk events are unavailable, it may be selected from the upper tail of the Surprise distribution observed during a baseline period in which the monitored AI system operates within an accepted risk profile. When controlled risk events are available, $\tau$ may instead be selected to balance false alarms, missed events, and detection delay.

The latent dimensions should be selected using held-out interaction sequences, considering predictive performance, temporal stability of the slow risk states, risk-estimation responsiveness, and computational cost. The parameter $\beta$ should likewise be selected by balancing predictive accuracy against the stability and informativeness of the latent representation. Learning rates and online update schedules should be selected to maintain stable adaptation.

Initial component-risk priors should incorporate available domain knowledge. When such knowledge is weak, diffuse priors may be used and their sensitivity evaluated. Regulatory-policy parameters should be selected by balancing risk reduction, prediction-error reduction, and the operational cost of regulatory actions. Thus, hyperparameter selection in CENTRY is inherently multi-objective.

\subsection{Validation and evaluation methodology}
\label{subsec:Validation and evaluation methodology}
Because the true component-risk states are generally latent, CENTRY should be evaluated at four distinct levels: prediction, controlled component-risk perturbations, continual risk estimation, and closed-loop regulation.

\begin{enumerate}[nosep]
\item \textbf{Predictive validation.}
Evaluate $p_{\mathrm{pred}}(y_t\mid y_{1:t-1},c_{1:t-1})$ on held-out interaction sequences using predictive negative log-likelihood, ELBO, and probabilistic calibration. This tests whether CENTRY can model the observable behavior of the monitored AI system without access to its internal implementation.

\item \textbf{Controlled component-risk perturbations.}
Introduce controlled changes to individual components while keeping other components as constant as possible. Examples include data-quality degradation or distribution shift for the dataset component, restrictions on computational resources for the hardware component, and replacement with a deliberately less robust model for the model component. Evaluation should determine whether the corresponding $\rho_t^{(j)}$ responds appropriately while unrelated component ratings remain comparatively stable.

\item \textbf{Continual risk-estimation performance.}
Measure detection latency, false-alarm frequency, recovery following transient perturbations, and adaptation to persistent component changes. The slow latent states should remain comparatively stable when component conditions are unchanged and respond systematically when controlled conditions change.

\item \textbf{Closed-loop regulatory performance.}
Evaluate whether the regulatory actions selected by $\pi_\psi$ reduce component risk while satisfying the prescribed risk budgets. Relevant measures include the frequency and duration of budget violations, cumulative component risk, prediction error, regulatory-action frequency, and regulatory cost. Where controlled or independently audited risk outcomes are available, evaluate the calibration of the estimated budget-violation probabilities $\widehat p_{t,k}^{(j)}$, including the one-sided calibration margin required by the finite-horizon guarantee. Comparisons with no regulation and alternative regulatory policies provide a direct assessment of the contribution of CENTRY's closed-loop regulation.
\end{enumerate}
These evaluations distinguish CENTRY's ability to predict observable behavior from its ability to estimate latent component risks and, ultimately, to regulate the monitored AI system.

\section{An Illustrative Example of Continual Risk Detection, Monitoring, and Regulation by CENTRY}
\label{sec:example}
As AI becomes part of everyday life, risk assessment cannot end at deployment. Some incidents may go unnoticed, and fewer still may be formally reported or made public. Chatbots that present themselves as personal assistants or therapists can invite more trust than their capabilities warrant. When something goes wrong, users may have little oversight or access to conversation records that could help them understand what happened. Even a chatbot advertised as multilingual may perform poorly in languages such as Bodo or Sindhi if those languages were under-represented in training and testing.

Reports have linked Character.AI chatbot interactions to serious alleged harms, including a reported death, although these reports do not establish causation \cite{AIID951_characterai}. Researchers have also found that some of Microsoft Copilot's answers to medication questions were inaccurate and could cause serious harm if followed \cite{Andrikyan2025Copilot,AIID838_copilot_misinfo}. Even in regulated settings such as hospitals, errors can delay diagnosis or treatment. The reported underdiagnosis of lung conditions among Black men by a diagnostic algorithm illustrates why monitoring must look beyond overall accuracy to see how a system performs for different patient groups \cite{AIID582_lungdiagnostics}.

Motivated by such cases, this section provides a numerical example of how CENTRY can identify, explain, and mitigate potentially risky behavior without being trained as an expert in the application domain of the monitored AI system. The example follows the operation of Algorithm~\ref{alg:CENTRY} line by line, indicated in each
subsection heading below. The numerical values are illustrative and are not intended to represent a validated clinical-risk model.

\subsection{Assumptions used in this example}
\label{subsec:example-Assumptions}
Beyond the framework of Sections~\ref{sec:Framework for AI Risk Monitoring}, \ref{sec:A Probabilistic Model for Continual AI Risk Estimation}, and~\ref{sec:Component Risk Priors}, the following choices are made solely for concreteness and are not claims about a specific implementation:
\begin{enumerate}[nosep]
\item Throughout, $\log$ denotes the natural logarithm, so the realized Surprise $\widehat S_t$ is measured in nats.
\item The component-risk rating $g^{(j)}(\cdot)$ of Eq.~\eqref{eq:riskfunctional} is instantiated as the posterior probability that the corresponding scalar latent state exceeds zero (Eq.~\eqref{eq:example-rating} below).
\item The component-specific regulation policy $\pi_\psi^{(j)}(\cdot)$ of Eq.~\eqref{eq:componentpolicy} is instantiated, for the dataset component, as the tiered rule of Eq.~\eqref{eq:example-policy} below.
\item $\bar B^{(D)}=0.3$, $\bar B^{(H)}=0.2$, $\bar B^{(M)}=0.2$ are the normalized component-risk thresholds (Section~\ref{subsubsec:Risk regulation policy}) corresponding to the component risk budgets $B^{(D)},B^{(H)},B^{(M)}$ of Eq.~\eqref{eq:componentbudget}.
\item All posterior parameters, the transition model of Eq.~\eqref{eq:example-transition}, and the predictive probabilities are given values rather than derived from raw data, since the purpose of this example is to illustrate CENTRY's inference and regulation mechanics, not to fit an actual model.
\end{enumerate}

\subsection{Monitored system and interaction}
\label{subsec:example-Monitored system}
Consider a retrieval-augmented AI assistant that provides medication-related information to clinicians. The monitored AI system is trained for its application domain; CENTRY is not assumed to possess independent medical expertise or to determine whether every response is medically correct.
Instead, CENTRY observes the interaction $y_t=(x_t,a_t)$ and the subsequent interaction trajectory.

Suppose that the percept $x_t$ is a clinician's request for information about the concurrent use of two medications, and the AI system produces the action
\[
a_t = \text{``No special review or monitoring is needed.''}
\]
Thus, CENTRY observes $y_t=(x_t,a_t)$.

As motivation for the specific posterior values assumed in Section~\ref{subsec:example-Explain} below, suppose that CENTRY's risk-causality training database $\mathcal{U}$ (Section~\ref{subsubsec:Risk-causality training database}) contains a record resembling this interaction: an input concerning multiple medications, a highly confident and uncited output that does not request professional review, and a low-provenance retrieval source, with evidence $\nu_i$ that disabling the low-provenance source was followed by a sustained reduction in similarly unusual responses. Consistent with Section~\ref{subsubsec:Risk-causality training database}, this record informs the posterior below only through the learned parameters $\phi$, not as an additional runtime input to Explain.

For the present interaction, the observable properties are as follows: the input concerns multiple medications, and the output is highly confident and does not request professional review. Separately, the assistant's retrieval configuration includes a low-provenance source; this is a property of the scenario, not an observation available to CENTRY at runtime. The observable properties, together with the learned parameters of \(q_\phi\), motivate a possible dataset-related explanation but do not establish that the current
response is incorrect or harmful.

\subsection{Predict and Surprise (Algorithm~\ref{alg:CENTRY}, lines 3--5)}
\label{subsec:example-Predict and Surprise}
During an accepted baseline period, the monitored system usually responded to medication-interaction queries with a cautious response pattern, including one or more of the following observable features: uncertainty language, a source citation, an escalation recommendation, or a request for clinician
review. Let $\mathcal{A}_{\mathrm{cautious}}$ denote this observable response class, and let $\mathcal{A}_{\mathrm{unusual}}$ denote the complementary class of highly confident, uncited, non-escalating responses. These are behavioral response classes; they do not require CENTRY to independently determine the clinical truth of the content.

Using Eq.~\eqref{eq:predictivedistribution} (Algorithm~\ref{alg:CENTRY}, line~3), CENTRY predicts, from $y_{1:t-1},c_{1:t-1}$ alone,
\begin{align*}
p_{\mathrm{pred}}\left(a_t\in\mathcal{A}_{\mathrm{cautious}}\mid y_{1:t-1},c_{1:t-1}\right) &=0.96,\\
p_{\mathrm{pred}}\left(a_t\in\mathcal{A}_{\mathrm{unusual}}\mid y_{1:t-1},c_{1:t-1}\right) &=0.04.
\end{align*}

The observed action $a_t$ (Algorithm~\ref{alg:CENTRY}, line~4) belongs to $\mathcal{A}_{\mathrm{unusual}}$. For this illustration, Surprise is evaluated on the response class of $a_t$ (a coarsening of the full interaction $y_t$), so the Surprise stage (line~5) applies Eq.~\eqref{eq:surprisenll} to the observed class:
\begin{align*}
\widehat S_t
= -\log p_{\mathrm{pred}}\left(a_t\in\mathcal{A}_{\mathrm{unusual}}\mid y_{1:t-1},c_{1:t-1}\right)
= -\log(0.04) = 3.219.
\end{align*}

Suppose the surprise threshold selected during baseline validation of the coarsened observation is $\tau=2.50$. Then $\widehat S_t=3.219>\tau$, so the interaction is flagged for additional investigation or verification. Explain is nevertheless performed as part of every SELP iteration, as specified above. This result does not prove the observed action is harmful; it indicates that the observed behavior is substantially inconsistent with the behavior predicted from accepted baseline interactions.

\subsection{Explain: posterior inference and component risk ratings (Algorithm~\ref{alg:CENTRY}, lines 6--7)}
\label{subsec:example-Explain}
Suppose the observed interaction $y_t=(x_t,a_t)$ resembles the behavior--cause pattern represented in the risk-causality database $\mathcal U$: the request concerns multiple medications and the output is highly confident, uncited, and non-escalating. The posterior below is assumed to reflect training on evidence-supported records in $\mathcal U$ in which similar interactions were associated with dataset-related risk. Such training records may have included component-configuration evidence, controlled perturbations, or subsequent outcomes, but those quantities are not additional runtime inputs to Explain; at runtime, Explain uses only $y_{1:t}$ and $c_{1:t-1}$ as specified in Eq.~\eqref{eq:explainposterior}.

Suppose that the posterior inferred by Eq.~\eqref{eq:explainposterior} (line~6) has the following scalar marginals:
\begin{align*}
q_\phi\left(z_t^{(D)}\mid y_{1:t},c_{1:t-1}\right) &= \mathcal{N}(0.50,0.25^2),\\
q_\phi\left(z_t^{(H)}\mid y_{1:t},c_{1:t-1}\right) &= \mathcal{N}(-0.57,0.25^2),\\
q_\phi\left(z_t^{(M)}\mid y_{1:t},c_{1:t-1}\right) &= \mathcal{N}(-0.45,0.25^2).
\end{align*}

For the numerical example (line~7), let the component-risk rating in Eq.~\eqref{eq:riskfunctional} be instantiated as
\begin{equation}
\rho_t^{(j)} = \Pr\left(z_t^{(j)}>0 \mid y_{1:t},c_{1:t-1}\right), \qquad j\in\{D,H,M\},
\label{eq:example-rating}
\end{equation}
so that the rating is the posterior probability that the latent state exceeds zero. Here $\Phi$ denotes the standard normal cumulative distribution function. Therefore,
\begin{align*}
\rho_t^{(D)} &= 1-\Phi\!\left(\tfrac{0-0.50}{0.25}\right)=\Phi(2.00)=0.977,\\
\rho_t^{(H)} &= 1-\Phi\!\left(\tfrac{0-(-0.57)}{0.25}\right)=1-\Phi(2.28)\approx0.011,\\
\rho_t^{(M)} &= 1-\Phi\!\left(\tfrac{0-(-0.45)}{0.25}\right)=1-\Phi(1.80)\approx0.036.
\end{align*}
Hence $\rho_t=(\rho_t^{(D)},\rho_t^{(H)},\rho_t^{(M)})=(0.977,0.011,0.036)$. Comparing against the thresholds of Section~\ref{subsec:example-Assumptions}, $\rho_t^{(D)}=0.977>\bar B^{(D)}$, while $\rho_t^{(H)}<\bar B^{(H)}$ and $\rho_t^{(M)}<\bar B^{(M)}$. CENTRY therefore treats the interaction as an elevated dataset-risk case: the dataset component receives the largest posterior-derived risk rating.

\subsection{Regulation policy: selecting the regulatory action (Algorithm~\ref{alg:CENTRY}, line 8)}
\label{subsec:example-Regulation policy}
Consider three candidate dataset interventions: $c^{(0)}$, retain the current retrieval configuration; $c^{(1)}$, disable the low-provenance source and require retrieval from an approved source; and $c^{(2)}$, apply $c^{(1)}$ and require agreement between two independent approved retrieval sources before permitting a high-confidence response.

For the numerical example, let the dataset component's policy $\pi_\psi^{(D)}$ of Eq.~\eqref{eq:componentpolicy} be the tiered rule
\begin{equation}
\pi_\psi^{(D)}(\rho) =
\begin{cases}
c^{(0)}, & \rho \le \bar B^{(D)},\\
c^{(1)}, & \bar B^{(D)} < \rho \le 0.90,\\
c^{(2)}, & \rho > 0.90,
\end{cases}
\label{eq:example-policy}
\end{equation}
reflecting increasingly aggressive intervention as the posterior confidence of elevated dataset risk grows. Since $\rho_t^{(D)}=0.977>0.90$, Eq.~\eqref{eq:example-policy} gives $c_t^{(D)}=\pi_\psi^{(D)}(0.977)=c^{(2)}$. For this example the hardware and model policies retain their current configurations whenever their ratings do not exceed their respective thresholds. Because $\rho_t^{(H)}\le\bar B^{(H)}$ and $\rho_t^{(M)}\le\bar B^{(M)}$, the corresponding component policies leave those components unchanged: $c_t^{(H)}=c_t^{(M)}=\text{no change}$, where ``no change'' denotes the component-specific action of retaining the current configuration. The selected regulatory action is therefore
\begin{equation*}
c_t = \left(c_t^{(D)},c_t^{(H)},c_t^{(M)}\right) = \left(c^{(2)},\text{no change},\text{no change}\right).
\end{equation*}
This action does not prescribe the future action of the monitored AI system; it changes the data-retrieval configuration through the admissible intervention interface, and is justified even before the domain correctness of the original response is resolved because it is reversible, targets the component with the highest posterior-derived risk rating, and is intended to reduce the consequence of a potentially erroneous high-confidence recommendation.

\subsection{Learn and posterior propagation (Algorithm~\ref{alg:CENTRY}, lines 9--10)}
\label{subsec:example-learn}
Line~9 (Learn) uses the interaction observed at time $t$ and the current posterior to update the generative and inference parameters $(\theta,\phi)$ by maximizing the ELBO of Eq.~\eqref{eq:ELBO}, and uses the selected regulatory action to update the regulation-policy parameters $\psi$ by minimizing the prospective objective of Eq.~\eqref{eq:policylearning}. The subsequent interaction $y_{t+1}$ is not required for this update; it becomes available in the next iteration and contributes to subsequent learning. We do not carry out this gradient-based update numerically here, since it operates on the full parametric model rather than on the scalar quantities of this example; the update is intended to improve the inference and regulation models for similar future cases, which would require subsequent validation. The remainder of this subsection carries out line~10 (posterior propagation) explicitly.

Suppose the learned dataset-risk transition model under the selected intervention is
\begin{equation}
z_{t+1}^{(D)} \sim \mathcal{N}\left(0.3\,z_t^{(D)}-0.55,\ 0.18^2\right).
\label{eq:example-transition}
\end{equation}
Propagating the posterior of Section~\ref{subsec:example-Explain} through Eq.~\eqref{eq:example-transition} according to Eq.~\eqref{eq:priorupdate} gives mean
\begin{equation*}
\mu_{t+1}^{(D)} = 0.3(0.5)-0.55 = -0.4,
\end{equation*}
and variance
\begin{equation*}
\left(\sigma_{t+1}^{(D)}\right)^2 = (0.3)^2(0.25)^2+(0.18)^2 = 0.038025,
\end{equation*}
so $\kappa_{t+1}(z_{t+1}^{(D)})=\mathcal{N}(-0.4,\,0.195^2)$. The predicted post-regulation dataset-risk rating, evaluated using the propagated prior before the next interaction is observed, is therefore
\begin{equation*}
\widehat\rho_{t+1\mid t}^{(D)} = 1-\Phi\!\left(\tfrac{0-(-0.4)}{0.195}\right) = 1-\Phi(2.051) \approx 0.02.
\end{equation*}
Since $\widehat\rho_{t+1\mid t}^{(D)}=0.02<\bar B^{(D)}=0.3$, the propagated prior gives a dataset-risk forecast below its normalized threshold. The next posterior-derived rating will depend on the subsequent observed interaction. This numerical prediction is not by itself a finite-horizon guarantee on the true operational risk $R^{(D)}$; that guarantee requires the calibrated probability-of-violation condition stated in Section~\ref{sec:Guarantees of CENTRY}.

\subsection{Applying the regulatory action and predicting future behavior (Algorithm~\ref{alg:CENTRY}, line 11, and subsequent cycles)}
\label{subsec:example-Applying the regulatory action}
CENTRY applies $c_t$ through the admissible intervention interface (line~11). Suppose, further, that the dataset-risk rating remains below its normalized threshold in the interim, so that the policy of Eq.~\eqref{eq:example-policy} selects $c^{(0)}$, which retains whatever retrieval configuration is active at that later step, rather than restoring the baseline configuration. The restrictions imposed by the selected dataset action therefore remain in force. In the predictive expressions below, $c_t^{(2)}$ is shorthand for that dataset intervention. At time $t+k$, $k>0$, the clinician presents the same percept, $x_{t+k}=x_t$. The predictive distribution of Eq.~\eqref{eq:predictivedistribution} now places high probability on a response pattern that uses approved information sources and uncertainty-aware language (the conditioning on the earlier interaction and regulatory-action histories is suppressed for brevity):
\begin{align*}
p_{\mathrm{pred}}\left(a_{t+k}\in\mathcal{A}_{\mathrm{cautious}}\mid x_{t+k}=x_t,c_t^{(2)}\right) &=0.961,\\
p_{\mathrm{pred}}\left(a_{t+k}\in\mathcal{A}_{\mathrm{unusual}}\mid x_{t+k}=x_t,c_t^{(2)}\right) &=0.039.
\end{align*}
A representative predicted future action is
\begin{quote}
\emph{``The retrieved approved references indicate that this medication combination may require additional review. The available sources are consistent on the relevant information, but the evidence should be reviewed carefully before changing the current medication-management plan.''}
\end{quote}
Thus, for the same percept $x_{t+k}=x_t$, the predictive model assigns high probability to a cautious response under the changed dataset-retrieval configuration. In this illustration, CENTRY did not generate the clinical answer or establish that the original response was false: it detected an unusual interaction (Surprise), inferred elevated dataset-related risk (Explain), selected a reversible intervention (regulation policy), and propagated the posterior to predict subsequent behavior (Predict). The displayed values are illustrative predictions, not observed outcomes or evidence of empirical effectiveness.

\paragraph{Remark [Cases without a clear precedent].}
The interaction above happened to resemble a pattern $q_\phi$ had encountered before, which is why the assumed posterior concentrates clearly on the dataset component. In general, a surprising interaction need not produce such a one-sided posterior -- $\rho_t^{(D)}$, $\rho_t^{(H)}$, and $\rho_t^{(M)}$ may all be only moderately elevated. The regulation policy of Eq.~\eqref{eq:componentpolicy} can still act in this case, for example by selecting a low-cost, reversible action such as increased logging for each component whose rating is moderately elevated, while the Learn stage (line~9) continues to refine $q_\phi$ and $\pi_\psi$ from the resulting outcome evidence. A full treatment of how CENTRY's estimates and policy improve under weak precedent is beyond the scope of this illustrative example.


\section{Discussion}
\label{sec:Discussion}
This paper presents CENTRY, a probabilistic framework for continual AI risk estimation, monitoring, and regulation. Rather than assigning static risk scores to AI systems before deployment, CENTRY models AI risk as an evolving latent state that is continually inferred from the interactions between the monitored AI system and its environment. This section discusses the principal strengths, current limitations, and future research directions of the proposed framework.

\subsection{Role of Predict and Surprise in CENTRY}
\label{subsec:Role of Predict and Surprise in CENTRY}
A natural alternative to SELP is a discriminative model that directly maps the interaction history to component-risk ratings,
\[
f_\eta\!\left(y_{1:t},c_{1:t-1}\right)
\longmapsto
\left(\rho_t^{(D)},\rho_t^{(H)},\rho_t^{(M)}\right).
\]
Such a formulation can estimate risk when appropriately labeled training data are available. CENTRY instead formulates continual risk estimation as probabilistic temporal inference over a fast interaction state $z_t^{(F)}$ and slowly evolving component-risk states $z_t^{(D)}$, $z_t^{(H)}$, and $z_t^{(M)}$. The roles of Predict and Surprise follow from this formulation.

\paragraph{Predict provides the generative model required by the ELBO.}
The component-risk ratings $\rho_t^{(j)}=g^{(j)}(\mu_t^{(j)},\Sigma_t^{(j)})$ (Eq.~\eqref{eq:riskfunctional}) are functionals of the posterior $q_\phi(z_t\mid y_{1:t},c_{1:t-1})$
(Eq.~\eqref{eq:explainposterior}), which is learned by maximizing the ELBO (Eq.~\eqref{eq:ELBO}). In particular, the reconstruction term
\[
\mathbb{E}_{q_\phi}
\!\left[\log p_\theta(y_t\mid z_t)\right]
\]
provides the data-dependent learning signal that connects the latent state to the observed interaction. If this term were removed, the objective would reduce to
\[
-\beta D_{\mathrm{KL}}
\!\left(
q_\phi(z_t\mid y_{1:t},c_{1:t-1})
\,\middle\|\,
\kappa_t(z_t)
\right),
\]
whose maximum is attained at $q_\phi=\kappa_t$, independently of the newly observed interaction. The latent state would then provide no data-dependent basis for the component-risk ratings. Thus, within CENTRY's variational formulation, the generative observation model provided by Predict is essential for learning latent component-risk states from interaction data without requiring a risk label at
every time step. A direct discriminative model would instead require an alternative source of learning information for this mapping.

\paragraph{Predict also enables prospective regulation.}
The same generative model provides the action-conditional predictive distribution
\[
p_{\mathrm{pred}}\!\left(
y_{t+1}\mid y_{1:t},c_{1:t}
\right)
=
\int
p_\theta(y_{t+1}\mid z_{t+1})
p_\theta(z_{t+1}\mid z_t,y_t,c_t)
q_\phi(z_t\mid y_{1:t},c_{1:t-1})
\,dz_t\,dz_{t+1}.
\]
Consequently, CENTRY can propagate its current probabilistic state forward under candidate regulatory actions and evaluate their expected consequences before those actions are applied. In particular, the expectation in Eq.~\eqref{eq:policylearning} is evaluated over future trajectories generated by the learned model. A direct sequence-to-risk regression has no such action-conditioned generative mechanism and therefore cannot, by itself, evaluate how a candidate $c_t$ is expected to change future component risks. It would require an additional dynamical model to provide this capability.

\paragraph{Surprise is a derived diagnostic of the predictive model.}
Once Predict provides the predictive distribution, the realized Surprise is obtained by evaluating the observed interaction under that distribution. For each realized interaction, Eq.~\eqref{eq:surprisenll} gives the realized surprise (writing $\widehat S_t$ for this per-interaction quantity, as in Section~\ref{subsec:Inference guarantees}, to distinguish it from the population-level mismatch $S_t^{\mathrm{(KL)}}$ of Eq.~\eqref{eq:surprise})
\[
\widehat S_t
=
-\log
p_{\mathrm{pred}}
\!\left(
y_t\mid y_{1:t-1},c_{1:t-1}
\right).
\]
Thus, Surprise introduces no additional generative model; it is a scalar diagnostic derived from Predict that quantifies predictive mismatch.

The connection to the ELBO can be made explicit. Combining the propagated-prior construction in Eqs.~\eqref{eq:jointpriorfactorization}--\eqref{eq:priorupdate} with Eq.~\eqref{eq:predictivedistribution} gives
\[
p_{\mathrm{pred}}
\!\left(
y_t\mid y_{1:t-1},c_{1:t-1}
\right)
=
\mathbb{E}_{\kappa_t(z_t)}
\!\left[
p_\theta(y_t\mid z_t)
\right].
\]
Hence,
\[
\widehat S_t
=
-\log
\mathbb{E}_{\kappa_t(z_t)}
\!\left[
p_\theta(y_t\mid z_t)
\right].
\]
The standard evidence-bound relation then yields, for $\beta>0$,
\begin{equation}
\mathcal{L}_t(\theta,\phi)
\leq
-\widehat S_t
+
(1-\beta)
D_{\mathrm{KL}}
\!\left(
q_\phi(z_t\mid y_{1:t},c_{1:t-1})
\,\middle\|\,
\kappa_t(z_t)
\right).
\label{eq:surprisebound}
\end{equation}
For $\beta=1$, this reduces to $\mathcal{L}_t(\theta,\phi)\leq -\widehat S_t$, with equality when $q_\phi$ equals the exact posterior under the propagated prior $\kappa_t$ and likelihood $p_\theta(y_t\mid z_t)$. Thus, Surprise is directly linked to the evidence underlying the ELBO: a large $\widehat S_t$ indicates that the realized interaction has low probability under the current predictive model and therefore indicates greater predictive mismatch between the current model and the observed interaction.

Importantly, Surprise is not a risk label. An expected interaction may still have an elevated component-risk rating, for example when a previous regulatory action has not fully reduced a persistent risk state, while an unexpected interaction may be benign. CENTRY therefore computes the component-risk ratings at every time step; the risk-budget thresholds determine whether regulation is required. If a Surprise threshold $\tau$ is used, it serves only as a diagnostic criterion for allocating additional investigative, verification, or adaptation effort. It does not determine whether the component-risk ratings are computed or whether they exceed their respective risk budgets.

Thus, Predict and Surprise do not constitute additional criteria for deciding whether an interaction is risky. Predict is an essential part of CENTRY's generative temporal formulation: it supplies the data-dependent likelihood required by the ELBO and the action-conditioned model required for prospective regulation. Surprise is a derived diagnostic of that predictive model. Together with Explain and Learn, this predictive formulation allows CENTRY to maintain a continually updated probabilistic representation of latent component risks while evaluating the prospective consequences of regulatory actions.


\subsection{Strengths and theoretical contributions}
\label{subsec:Strengths and theoretical contributions}
The proposed framework possesses several theoretical and practical strengths.
\begin{itemize}[nosep]
\item First, CENTRY provides a unified probabilistic formulation of continual AI risk estimation, monitoring, and regulation. Existing AI risk assessment methods predominantly rely on static audits, benchmarks, or one-time evaluations performed before deployment. In contrast, CENTRY continually updates its Bayesian belief regarding the risks associated with the monitored AI system throughout its operational lifetime.

\item Second, the proposed framework naturally separates AI risk into three distinct layers: latent probabilistic inference, interpretable Bayesian risk estimation, and adaptive regulation. The latent variables provide a compact probabilistic representation of the monitored AI system, the component-risk ratings summarize the posterior beliefs in an interpretable form, and the regulation policy modifies the utilization of AI system components according to these estimated risks. This separation allows probabilistic inference and decision making to be developed modularly while remaining mathematically coupled.

\item Third, the structured latent representation enables component-specific diagnosis of AI risk. Rather than estimating only a single aggregate risk score, CENTRY infers separate latent subspaces corresponding to datasets, hardware resources, and models/algorithms. Consequently, the framework estimates not only the magnitude of AI risk but also posterior beliefs about its most probable component source.

\item Fourth, the proposed framework integrates continual Bayesian learning with the SELP cycle. Surprise detects unexpected behavior, Explain infers latent component-risk states that may account for the prediction error, Learn continually adapts both the internal probabilistic model and the regulation policy, while Predict generates future behavior used by the next iteration of the cycle. This formulation provides a mathematically rigorous realization of the original SELP philosophy within a probabilistic latent-variable framework.

\item Fifth, CENTRY is accompanied by formal theoretical guarantees that characterize the behavior of its estimation, learning, and regulation mechanisms under explicitly stated assumptions. The results span safety-related properties, including finite-horizon and asymptotic risk-budget compliance; learning properties, including regret and convergence of the adaptive procedures; inference properties, including consistency and identifiability of the component-risk ratings; and operational properties, including detection delay, false-alarm behavior, containment, and related system-level bounds. These results specify conditions under which the proposed mechanisms achieve the corresponding guarantees rather than treating their validity as unconditional.

\item Finally, the framework is not tied to a single implementation architecture. Although sequential variational autoencoders provide one natural realization, the probabilistic formulation can also be instantiated using other latent-variable models, probabilistic state-space models, or related probabilistic inference architectures, provided that the assumptions required by the corresponding formulation and guarantees are satisfied.
\end{itemize}

A distinguishing characteristic of CENTRY is that it treats AI risk as a latent dynamical state rather than a static property of an AI system. Consequently, AI risk is not viewed as a fixed score assigned during development or deployment, but as an evolving probabilistic belief that is continually updated through interaction with the environment and subsequently used to adaptively regulate the utilization of the AI system's underlying components. The formal results in Section~\ref{sec:Guarantees of CENTRY} further characterize the conditions under which this continual estimation, learning, and regulation process provides specific safety, inferential, learning, and operational guarantees, with the corresponding proofs provided in the Appendix. Taken together, these properties provide a principled mathematical foundation for continual AI risk estimation, monitoring, and regulation while making explicit the assumptions under which the resulting guarantees hold.


\subsection{Limitations}
\label{subsec:Limitations}
Despite its theoretical formulation and formal guarantees, the proposed framework has several limitations.
\begin{itemize}[nosep]
\item First, this paper focuses primarily on the formalization of continual AI risk estimation and regulation. Although Section~\ref{sec:Guarantees of CENTRY} establishes formal guarantees under explicitly stated assumptions, the framework has not yet been validated experimentally on large-scale AI systems or benchmark datasets. Consequently, the empirical effectiveness, computational performance, and practical robustness of the proposed methods remain important topics for future investigation.

\item Second, the present work considers only three primary intrinsic sources of AI risk: datasets, hardware resources, and models/algorithms. Operational factors, deployment environments, human supervision, and interaction effects between system components are incorporated only indirectly. Extending the probabilistic framework to explicitly model these additional sources of risk, including higher-order interactions among them, represents an important direction for future work. The present formulation also does not explicitly disaggregate component-risk estimates by demographic group or language; such assessments require additional deployment-specific evaluation and calibration.

\item Third, the proposed Bayesian component-risk ratings depend upon the quality of the adopted prior distributions. Although Section~\ref{sec:Component Risk Priors} describes the evidence that can inform component-specific priors, constructing informative priors for complex real-world AI systems remains an open research problem. In addition, because the component-risk states are latent and are inferred from observable AI--environment interactions, reliable attribution to individual components requires the observability and identifiability conditions assumed by the corresponding theoretical results.

\item Fourth, the adaptive regulation policy presented in this paper is intentionally generic. While the framework specifies how regulatory decisions are generated from Bayesian component-risk estimates, the design of application-specific regulation policies depends upon the operational constraints, available interventions, and safety requirements of individual deployment domains. In practice, the effectiveness of regulation therefore depends on whether sufficiently effective and feasible interventions are available.

\item Fifth, the proposed framework assumes that the latent component-risk representation captures a sufficient portion of the causes relevant to observed AI behavior. Complex socio-technical systems may exhibit emergent behaviors arising from interactions among multiple AI systems, humans, deployment environments, and other factors that are not explicitly represented in the current model. Such higher-order interaction effects may therefore limit both risk attribution and regulation effectiveness.

\item Finally, the formal guarantees established in Section~\ref{sec:Guarantees of CENTRY} are conditional rather than unconditional: each result requires the assumptions stated for that result, which may include appropriate model calibration, effective and feasible regulation, correctly specified or suitably controlled latent dynamics, bounded prediction error, statistical regularity, or other problem-specific conditions. These assumptions are not established universally for arbitrary deployed AI systems in the present work. Consequently, applying the theoretical guarantees to a specific deployment would require empirical validation of the relevant assumptions, or evidence that they hold to an adequate approximation in that setting.
\end{itemize}


\subsection{Future work}
\label{subsec:Future work}
Several research directions naturally emerge from the proposed framework.
\begin{itemize}[nosep]
\item The most immediate extension is the incorporation of additional latent risk components beyond datasets, hardware resources, and models/algorithms. In particular, explicitly modeling operational context, human oversight, and interaction effects would yield a more comprehensive probabilistic representation of AI risk.
\item A second direction involves developing principled component-specific prior construction methodologies. Domain-specific dataset rating systems, hardware capability taxonomies, and model safety certification procedures could provide increasingly informative Bayesian priors for continual risk estimation.
\item Another promising direction is the development of optimal adaptive regulation policies. Going beyond the one-step-lookahead objective of Eq.~\eqref{eq:policylearning}, future work may formulate regulation as a stochastic optimal control or constrained decision-making problem that explicitly minimizes expected long-term AI risk while maintaining task performance.
\item The proposed framework may also be extended to cooperative multi-agent environments in which multiple CENTRY agents exchange probabilistic component-risk estimates. Such distributed Bayesian risk estimation could enable continual monitoring of large interconnected AI ecosystems.
\item Finally, experimental validation represents an important next step. Implementing CENTRY using modern sequential latent-variable architectures and evaluating its ability to identify, localize, and regulate evolving AI risks across diverse application domains will provide valuable empirical evidence regarding the effectiveness of the proposed framework.
\end{itemize}


\section{Conclusion}
\label{sec:Conclusion}
This paper introduced CENTRY, a probabilistic framework for continual AI risk estimation, monitoring, and regulation throughout the operational lifetime of an AI system. Building upon the SELP (Surprise--Explain--Learn--Predict) cycle, CENTRY models AI risk as a collection of slowly evolving latent states associated with the datasets, hardware resources, and models/algorithms used by a monitored AI system. By combining structured latent-variable modeling, variational Bayesian inference, continual Bayesian risk estimation, and adaptive component reconfiguration within a unified probabilistic framework, CENTRY continually updates posterior beliefs about the most probable component sources of observed AI risk as new interaction evidence becomes available.

Unlike existing approaches that primarily perform static risk assessment, isolated runtime monitoring, or standalone risk mitigation, CENTRY unifies continual probabilistic inference with adaptive regulation, enabling risk estimates to guide the regulation of the monitored AI system's components as operating conditions and risk profiles evolve. This perspective reframes AI risk from a fixed property established during system development to a latent dynamical state that must be continually inferred and managed throughout deployment.

CENTRY is accompanied by formal theoretical guarantees, established under explicitly stated assumptions, spanning finite-horizon and asymptotic risk-budget compliance, learning and convergence properties, consistency and identifiability of component-risk ratings, and operational properties including detection delay, false-alarm behavior, and containment. These results provide a formal basis for characterizing the conditions under which CENTRY's risk estimation, learning, and regulation mechanisms possess the stated safety and operational properties.



\section*{Authors' Contributions and Responsibility Statement}
BB directed the research, developed the CENTRY framework, and derived its theoretical guarantees. BB and GR prepared the draft of the manuscript. All authors reviewed and approved the manuscript. The authors employed calculators, language editing software, Internet search engines, and AI tools during the production of this manuscript. The authors maintain ultimate responsibility for the published work. The views, opinions, and conclusions expressed in this manuscript are solely those of the authors and do not reflect the policy, position, or endorsement of any affiliated institution, funding agency, or organization. The authors declare no competing financial or non-financial interests.

\bibliographystyle{ieeetr} 
\bibliography{References_AIrisk}

\newpage
\clearpage 
\thispagestyle{empty} 
\vspace*{\fill}
\begin{center}
    \textit{This page is intentionally left blank.}
\end{center}
\vspace*{\fill}
\clearpage

\appendix
\section{Appendix: Proofs of Theorems and Corollaries}
\label{app:Proofs of Theorems and Corollaries}
\makeatletter
\setcounter{tcb@cnt@styledtheorem}{0}
\setcounter{tcb@cnt@styledcorollary}{0}
\setcounter{tcb@cnt@styledlemma}{0}
\makeatother


This appendix contains the complete formal statements and proofs of the results stated in Section~\ref{sec:Guarantees of CENTRY}. The theorem, corollary, and lemma statements are reproduced here without their reader-oriented interpretations; auxiliary assumption labels are prefixed with \texttt{app:} to avoid duplicate labels with Section~\ref{sec:Guarantees of CENTRY}. Technical remarks are retained where they clarify the scope, assumptions, or derivations of the corresponding formal results. This appendix is intended to be compiled as part of the complete manuscript so that references to equations, sections, and result labels defined in the main text resolve normally.

Throughout this appendix, $\widehat S_t$ denotes the realized, operationally computed predictive surprise used by CENTRY in its model-based regulation objectives, whereas $S_t^{\mathrm{(KL)}}$ denotes the population-level predictive mismatch. The regulatory cost satisfies $C(c)\geq0$ for every admissible regulatory action $c$.

\subsection{Foundational guarantees}
\label{app:subsec:Foundational guarantees}
This subsection establishes the structural guarantees required for CENTRY to operate safely: non-anticipatory action selection, finite-horizon chance-constrained safety, valid posterior propagation, monotone threshold regulation, and infinite-horizon Lyapunov ultimate boundedness.

\begin{styledtheorem}{Non-Anticipatory Regulation}{app:non_anticipatory_regulation}

Let \(\mathcal{F}_t = \sigma\!\left(y_{1:t},c_{1:t-1}\right)\) be the information available to CENTRY at time $t$, where $y_{1:t}$ denotes the observable interaction history through time $t$ and $c_{1:t-1}$ denotes the sequence of regulatory actions taken before time $t$. Let the CENTRY risk-estimation operator produce the current risk-rating vector
\[
\rho_t = \mathcal{R}_t\!\left(y_{1:t},c_{1:t-1}\right),
\]
where \(\mathcal{R}_t\) is measurable, and let the regulatory policy be
\[
c_t = \pi_{\psi_t}(\rho_t),
\qquad
\psi_t = \Psi_t(y_{1:t},c_{1:t-1}),
\]
where \(\Psi_t\) is measurable. Assume that the joint mapping
\(\left(\psi,\rho\right)\mapsto\pi_\psi(\rho)\) is measurable on the product of the policy-parameter space and the risk-rating space. Then the regulatory action $c_t$ is \(\mathcal{F}_t\)-measurable. Consequently, CENTRY's regulatory policy is non-anticipatory: the action selected at time $t$ depends only on information available through time $t$ and cannot depend on future interactions $y_{t+1},y_{t+2},\ldots$ or future regulatory actions $c_{t+1},c_{t+2},\ldots$.

\end{styledtheorem}

\begin{proof}
By definition,
\[
\mathcal{F}_t
=
\sigma\!\left(
 y_{1:t},c_{1:t-1}
\right)
\]
contains all information available to CENTRY through time $t$.

First, consider the risk-estimation operator
\[
\rho_t
=
\mathcal{R}_t
\!\left(
 y_{1:t},c_{1:t-1}
\right).
\]
Because \(\mathcal{R}_t\) is measurable and is evaluated solely using the
observable interaction history $y_{1:t}$ and the previous regulatory actions
$c_{1:t-1}$, the resulting risk-rating vector \(\rho_t\) is
\(\mathcal{F}_t\)-measurable.

Next, by assumption, the policy parameters satisfy
\[
\psi_t
=
\Psi_t(y_{1:t},c_{1:t-1}).
\]
Because \(\Psi_t\) is measurable, \(\psi_t\) is also
\(\mathcal{F}_t\)-measurable.

The regulatory action is then selected according to
\[
c_t
=
\pi_{\psi_t}(\rho_t).
\]
Since both \(\psi_t\) and \(\rho_t\) are \(\mathcal{F}_t\)-measurable and the
joint policy mapping \((\psi,\rho)\mapsto\pi_\psi(\rho)\) is measurable,
the composition is \(\mathcal{F}_t\)-measurable. Therefore, $c_t$ is
\(\mathcal{F}_t\)-measurable.

Equivalently, for every measurable subset \(\mathcal{C}\) of the
regulatory-action space,
\[
\{c_t\in\mathcal{C}\}\in\mathcal{F}_t.
\]
Thus, the event that CENTRY selects any particular regulatory action
can be determined using only information contained in
\(\mathcal{F}_t\).

Because
\[
\mathcal{F}_t
=
\sigma(y_{1:t},c_{1:t-1})
\]
contains no future interactions $y_{t+1},y_{t+2},\ldots$ or future actions
$c_{t+1},c_{t+2},\ldots$, the action $c_t$ cannot depend on those future
quantities. Hence the CENTRY regulatory policy is non-anticipatory.
\end{proof}


\begin{styledtheorem}{Finite-Horizon Probabilistic Risk Guarantee}{app:finite_horizon_risk_guarantee}

Let
\(
\mathcal{F}_t
=
\sigma\!\left(y_{1:t},c_{1:t-1}\right)
\)
denote the information available to CENTRY immediately before selecting
the regulatory action \(c_t\). Let
\(j\in\{D,H,M\}\) denote a risk source, and let
\(B^{(j)}\) be its corresponding acceptable risk budget. Consider a finite
prediction horizon \(T\in\mathbb{N}\) and an active regulatory policy
\(\pi_\psi\). Let \(\Pr_{\pi_\psi}\) denote the probability measure
induced by execution of the policy. The notation for estimated violation probabilities suppresses their dependence on the active policy, as defined in Section~\ref{subsubsec:Risk regulation policy}.

Suppose that, for every
\(j\in\{D,H,M\}\) and \(k=1,\ldots,T\):

\begin{enumerate}[nosep]

    \item \textbf{Calibrated violation-probability estimate.}
    CENTRY computes a conditional estimated probability
    \[
    \widehat p_{t,k}^{(j)}
    :=
    \widehat{\Pr}_{\mathrm{CENTRY}}
    \!\left(
        R_{t+k}^{(j)}>B^{(j)}
        \,\middle|\,
        \mathcal{F}_t
    \right),
    \]
    and $\widehat p_{t,k}^{(j)}$ is $\mathcal F_t$-measurable. Assume there exists a calibration error
    \(\varepsilon_{\mathrm{cal}}\geq0\) such that
    \[
    \Pr_{\pi_\psi}\!\left(
        R_{t+k}^{(j)}>B^{(j)}
        \,\middle|\,
        \mathcal{F}_t
    \right)
    \leq
    \widehat p_{t,k}^{(j)}
    +
    \varepsilon_{\mathrm{cal}}
    \qquad\text{a.s.}
    \]

    \item \textbf{Chance-constrained regulation.}
    The regulatory policy satisfies
    \[
    \widehat p_{t,k}^{(j)}
    \leq
    \delta_k^{(j)}-\varepsilon_{\mathrm{cal}},
    \]
    where
    \(\delta_k^{(j)}\geq\varepsilon_{\mathrm{cal}}\).

    \item \textbf{Failure-probability allocation.}
    The allocated violation probabilities satisfy
    \[
    \sum_{k=1}^{T}
    \sum_{j\in\{D,H,M\}}
    \delta_k^{(j)}
    \leq
    \delta,
    \qquad
    0<\delta<1.
    \]

\end{enumerate}

Then the joint finite-horizon risk constraint
\[
\Pr_{\pi_\psi}\!\left(
R_{t+k}^{(j)}\leq B^{(j)},
\;
\forall j\in\{D,H,M\},
\;
\forall k=1,\ldots,T
\,\middle|\,
\mathcal{F}_t
\right)
\geq
1-\delta
\]
is satisfied.

In particular, when
\(\varepsilon_{\mathrm{cal}}=0\), the result reduces to the
component-wise chance-constrained guarantee used in the original
formulation.

\end{styledtheorem}

\begin{proof}
For each component \(j\) and future time \(k\), define the
budget-violation event
\[
E_k^{(j)}
=
\left\{
R_{t+k}^{(j)}>B^{(j)}
\right\}.
\]
By the calibration assumption and the chance constraint,
\[
\begin{aligned}
\Pr_{\pi_\psi}\!\left(
E_k^{(j)}
\,\middle|\,
\mathcal{F}_t
\right)
&\leq
\widehat p_{t,k}^{(j)}
+
\varepsilon_{\mathrm{cal}}
\\
&\leq
\delta_k^{(j)}.
\end{aligned}
\]
The complement of the joint finite-horizon safety event is
\[
\mathcal S_{t,T}^{c}
=
\bigcup_{k=1}^{T}
\bigcup_{j\in\{D,H,M\}}
E_k^{(j)}.
\]
Therefore, by the conditional union bound,
\[
\begin{aligned}
\Pr_{\pi_\psi}\!\left(
\mathcal S_{t,T}^{c}
\,\middle|\,
\mathcal F_t
\right)
&\leq
\sum_{k=1}^{T}
\sum_{j\in\{D,H,M\}}
\Pr_{\pi_\psi}\!\left(
E_k^{(j)}
\,\middle|\,
\mathcal F_t
\right)
\\
&\leq
\sum_{k=1}^{T}
\sum_{j\in\{D,H,M\}}
\delta_k^{(j)}
\\
&\leq
\delta.
\end{aligned}
\]
Hence
\[
\Pr_{\pi_\psi}\!\left(
\mathcal S_{t,T}
\,\middle|\,
\mathcal F_t
\right)
\geq
1-\delta,
\]
which proves the result.
\end{proof}

\noindent\textbf{Remark [Scope of the guarantee].}
Theorem~\ref{thm:finite_horizon_risk_guarantee} is conditional on predictive calibration and on satisfaction of the component-wise chance constraints. It does not claim that CENTRY can guarantee real-world budget compliance when the generative model is misspecified, the inferred risk states are poorly calibrated, the intervention interface is insufficient to influence the monitored system, or the required chance constraints are infeasible. The theorem establishes that, under its assumptions, component-wise chance constraints imply a joint finite-horizon safety probability through a conditional union bound.

\noindent\textbf{Remark [Relationship to the latent representation].}
CENTRY operationalizes the future violation event through its structured latent-state representation. The posterior over $z_{t+k}$ is propagated through the learned transition model and mapped to the component risk rating
\(\rho_{t+k}^{(j)}=g^{(j)}(\mu_{t+k}^{(j)},\Sigma_{t+k}^{(j)})\), with $\bar{B}^{(j)}$ denoting the corresponding normalized budget threshold. The formal guarantee, however, does not require the latent representation to be an exact representation of true operational risk. The required connection is the one-sided event-level calibration condition in Assumption~1, which explicitly controls the difference between the probability assigned by CENTRY and the true probability of budget violation.

\noindent\textbf{Remark [Bridge between true operational risk and estimated latent risk].}
The finite-horizon guarantee is stated in terms of the true operational risk $R_{t+k}^{(j)}$, whereas CENTRY computes its probability through the learned latent-state model and the risk functional $g^{(j)}$. The required bridge is therefore the one-sided event-level calibration condition in Assumption~1: the probability assigned by CENTRY to a budget violation may be optimistic by at most $\varepsilon_{\mathrm{cal}}$. The theorem does not require exact recovery of the true latent state or equality of the estimated and true risk distributions; it requires only this stated probabilistic calibration condition.


\begin{styledcorollary}{Uniform Allocation of the Failure Probability}{app:uniform-risk-allocation}

Under the assumptions of Theorem~\ref{thm:finite_horizon_risk_guarantee}, suppose the failure
probability is allocated uniformly across the three components and the \(T\)-step horizon:
\(
\delta_k^{(j)}
=
\frac{\delta}{3T},
j\in\{D,H,M\},
k=1,\ldots,T.
\)
If the policy satisfies the calibrated probability constraints
\(
\widehat p_{t,k}^{(j)}
\leq
\frac{\delta}{3T}-\varepsilon_{\mathrm{cal}},
\)
for every $j$ and $k$, with
$\varepsilon_{\mathrm{cal}}\leq\delta/(3T)$, then
\[
\Pr_{\pi_\psi}\!\left(
R_{t+k}^{(j)}\leq B^{(j)},
\;
\forall j\in\{D,H,M\},
\;
\forall k\in\{1,\ldots,T\}
\,\middle|\,
\mathcal{F}_t
\right)
\geq
1-\delta.
\]

\end{styledcorollary}

\begin{proof}
There are \(3T\) component--time pairs. Therefore,
\[
\sum_{k=1}^{T}
\sum_{j\in\{D,H,M\}}
\delta_k^{(j)}
=
3T\cdot\frac{\delta}{3T}
=
\delta.
\]
The result follows directly from
Theorem~\ref{thm:finite_horizon_risk_guarantee}.
\end{proof}


\begin{styledcorollary}{System-Level Finite-Horizon Risk Guarantee}{app:system_level_risk_guarantee}

Under the assumptions of Theorem \ref{thm:finite_horizon_risk_guarantee}, suppose that the operational decomposition of the true system risk satisfies
\(
R_{t+k} \leq \sum_{j\in\{D,H,M\}} R_{t+k}^{(j)} + \varepsilon
\)
for some bounded interaction error $\varepsilon \geq 0$, for all $k=1,\ldots,T$. Let $B = B^{(D)} + B^{(H)} + B^{(M)}$ denote the total allowable risk budget. Then, the joint finite-horizon safety event implies that the total system risk remains bounded with high probability:
\[
\Pr_{\pi_\psi}\!\left(
R_{t+k} \leq B + \varepsilon,
\;
\forall k=1,\ldots,T
\,\middle|\,
\mathcal{F}_t
\right)
\geq 1-\delta.
\]

\end{styledcorollary}

\begin{proof}
Let $\mathcal{S}_{t,T}$ be the joint safety event defined in Theorem \ref{thm:finite_horizon_risk_guarantee}, where $R_{t+k}^{(j)} \leq B^{(j)}$ for all components $j$ and times $k=1,\ldots,T$. Conditional on $\mathcal{S}_{t,T}$ occurring, we have
\[
\sum_{j\in\{D,H,M\}} R_{t+k}^{(j)} \leq \sum_{j\in\{D,H,M\}} B^{(j)} = B.
\]
By the operational decomposition assumption, $R_{t+k} \leq \sum_j R_{t+k}^{(j)} + \varepsilon$. Therefore, under $\mathcal{S}_{t,T}$, it holds that $R_{t+k} \leq B + \varepsilon$ for all $k$.

Because Theorem \ref{thm:finite_horizon_risk_guarantee} establishes that $\Pr_{\pi_\psi}(\mathcal{S}_{t,T} \mid \mathcal{F}_t) \geq 1-\delta$, it immediately follows that the probability of the total system risk being bounded by $B + \varepsilon$ across the horizon is at least $1-\delta$.
\end{proof}


\begin{styledtheorem}{Validity of propagated component priors}{app:propagated-prior-validity}

Suppose that \(q_t(z_t^{(j)})\) is a probability density and that
\(
p_\theta\!\left(
z_{t+1}^{(j)}
\mid
z_t^{(j)},y_t,c_t
\right)
\)
is a conditional probability density for each
\(j\in\{F,D,H,M\}\). Then the propagated prior defined by
Eq.~\eqref{eq:priorupdate},
\[
\kappa_{t+1}\!\left(z_{t+1}^{(j)}\right)
=
\int
p_\theta\!\left(
z_{t+1}^{(j)}
\mid
z_t^{(j)},y_t,c_t
\right)
q_t\!\left(z_t^{(j)}\right)
\,dz_t^{(j)},
\]
is a probability density. In particular,
\(
\kappa_{t+1}\!\left(z_{t+1}^{(j)}\right)\geq 0
\)
and
\(
\int
\kappa_{t+1}\!\left(z_{t+1}^{(j)}\right)
\,dz_{t+1}^{(j)}
=
1.
\)

\end{styledtheorem}

\begin{proof}
Nonnegativity follows because both the conditional density
\(p_\theta\) and posterior density \(q_t\) are nonnegative. For normalization,
Tonelli's theorem gives
\begin{align*}
\int
\kappa_{t+1}\!\left(z_{t+1}^{(j)}\right)
\,dz_{t+1}^{(j)}
&=
\int
\int
p_\theta\!\left(
z_{t+1}^{(j)}
\mid
z_t^{(j)},y_t,c_t
\right)
q_t\!\left(z_t^{(j)}\right)
\,dz_t^{(j)}
\,dz_{t+1}^{(j)}\\
&=
\int
\left[
\int
p_\theta\!\left(
z_{t+1}^{(j)}
\mid
z_t^{(j)},y_t,c_t
\right)
\,dz_{t+1}^{(j)}
\right]
q_t\!\left(z_t^{(j)}\right)
\,dz_t^{(j)}\\
&=
\int q_t\!\left(z_t^{(j)}\right)\,dz_t^{(j)}\\
&=1.
\end{align*}
Thus, \(\kappa_{t+1}\) is a valid probability density.
\end{proof}


\begin{styledcorollary}{Gaussian propagation under affine Gaussian transitions}{app:gaussian-propagation}

Suppose, for a component \(j\in\{F,D,H,M\}\), that
\(
q_t\!\left(z_t^{(j)}\right)
=
\mathcal N\!\left(
z_t^{(j)};
\mu_t^{(j)},\Sigma_t^{(j)}
\right)
\)
and that the transition model is affine Gaussian,
\(
z_{t+1}^{(j)}
=
A_t^{(j)}z_t^{(j)}
+
b_t^{(j)}
+
\varepsilon_t^{(j)},
\varepsilon_t^{(j)}
\sim
\mathcal N\!\left(
0,Q_t^{(j)}
\right),
\)
where \(\varepsilon_t^{(j)}\) is independent of \(z_t^{(j)}\). Then the
propagated prior in Eq.~\eqref{eq:priorupdate} is Gaussian:
\[
\kappa_{t+1}\!\left(z_{t+1}^{(j)}\right)
=
\mathcal N\!\left(
z_{t+1}^{(j)};
A_t^{(j)}\mu_t^{(j)}+b_t^{(j)},
A_t^{(j)}\Sigma_t^{(j)}(A_t^{(j)})^\top+Q_t^{(j)}
\right).
\]

\end{styledcorollary}

\begin{proof}
An affine transformation of a Gaussian random variable is Gaussian. Since
\(\varepsilon_t^{(j)}\) is independent Gaussian noise, the resulting sum is
Gaussian. Its mean and covariance are obtained by linearity of expectation and
independence:
\[
\mathbb E[z_{t+1}^{(j)}]
=
A_t^{(j)}\mu_t^{(j)}+b_t^{(j)},
\]
and
\[
\operatorname{Cov}(z_{t+1}^{(j)})
=
A_t^{(j)}\Sigma_t^{(j)}(A_t^{(j)})^\top+Q_t^{(j)}.
\]
\end{proof}


\begin{styledtheorem}{Monotonicity of a threshold-based component policy}{app:monotone-policy}

Fix a component \(j\in\{D,H,M\}\). Suppose the admissible component actions are
totally ordered by intervention strength,
\(
c^{(0)} \preceq c^{(1)} \preceq \cdots \preceq c^{(L)},
\)
and the component policy \(\pi_\psi^{(j)}:[0,1]\to\mathcal C^{(j)}\) is
constant on each interval of an ordered threshold partition of $[0,1]$, with nondecreasing intervention tiers across successive intervals. Then, for any
\(0\leq\rho_1\leq\rho_2\leq1\),
\[
\pi_\psi^{(j)}(\rho_1)
\preceq
\pi_\psi^{(j)}(\rho_2).
\]
Thus, increasing the estimated risk rating cannot select a weaker regulatory
intervention.

\end{styledtheorem}

\begin{proof}
By hypothesis, the policy partitions \([0,1]\) into ordered intervals and assigns
a nondecreasing action tier to successive intervals. Since
\(\rho_1\leq\rho_2\), the interval containing \(\rho_2\) cannot precede the
interval containing \(\rho_1\). Therefore, the action selected at \(\rho_2\)
is at least as strong as the action selected at \(\rho_1\).
\end{proof}


While a finite-horizon chance constraint ensures high-confidence probabilistic bounds over small execution windows, long-horizon operation requires a corresponding bound on the persistence of risk-budget violations. The following result establishes ultimate boundedness of the risk-excess Lyapunov function under a dissipative closed loop.

\begin{styledtheorem}{Infinite-Horizon Ultimate Boundedness of Risk Violations}{app:asymptotic_risk_stability}

Let $\rho_t =
\left(\rho_t^{(D)},\rho_t^{(H)},\rho_t^{(M)}\right)^\top
\in[0,1]^3$ denote the vector of component-risk ratings at time $t$,
and let
$\bar{B} =
\left(\bar{B}^{(D)},\bar{B}^{(H)},\bar{B}^{(M)}\right)^\top$
be the vector of normalized component-risk budgets. Define the
quadratic budget-violation Lyapunov function
\[
V(\rho_t)
=
\sum_{j\in\{D,H,M\}}
\left(
\left(\rho_t^{(j)}-\bar{B}^{(j)}\right)_+
\right)^2,
\]
where $(x)_+=\max(0,x)$. Let
\(
\mathcal{F}_t=\sigma(y_{1:t},c_{1:t-1})
\)
denote the filtration through time $t$, and let
$\Pr_{\pi_\psi}$ be the probability measure induced by the regulatory
policy $c_t=\pi_\psi(\rho_t)$. Expectations under this law are denoted by
$\mathbb E_{\pi_\psi}$, with the subscript omitted when the law is clear.

Suppose that the closed-loop CENTRY system satisfies the following
conditions:
\begin{enumerate}[nosep]
    \item \textbf{Strict Lyapunov Dissipativity:}
    There exist constants $\alpha\in(0,1]$ and $\epsilon\geq0$ such
    that, for every $t\geq1$,
    \(
    \mathbb{E}_{\pi_\psi}
    \!\left[
    V(\rho_{t+1})-V(\rho_t)
    \,\middle|\,
    \mathcal{F}_t
    \right]
    \leq
    -\alpha V(\rho_t)+\epsilon.
    \)

    \item \textbf{Bounded Initial Risk:}
    The initial risk-rating distribution has finite expected Lyapunov
    energy,
    \(
    \mathbb{E}_{\pi_\psi}[V(\rho_1)]=V_0<\infty.
    \)
\end{enumerate}

Then the following asymptotic safety properties hold:
\begin{enumerate}[label=(\roman*), nosep]
    \item \textbf{Mean Ultimate Boundedness:}
    \(
    \limsup_{t\to\infty}
    \mathbb{E}_{\pi_\psi}[V(\rho_t)]
    \leq
    \frac{\epsilon}{\alpha}.
    \)

    \item \textbf{Asymptotic Probability Bound:}
    For any tolerance $\eta>0$,
    \(
    \limsup_{t\to\infty}
    \Pr_{\pi_\psi}\!\left(
    V(\rho_t)\geq\eta
    \right)
    \leq
    \frac{\epsilon}{\alpha\eta}.
    \)
\end{enumerate}

\end{styledtheorem}

\begin{proof}
A zeroth power is one, including when the base is zero.

Taking expectations of the conditional drift inequality and applying
the tower property gives
\[
\mathbb{E}_{\pi_\psi}[V(\rho_{t+1})]
\leq
(1-\alpha)
\mathbb{E}_{\pi_\psi}[V(\rho_t)]
+\epsilon.
\]
Unrolling this recurrence yields
\[
\mathbb{E}_{\pi_\psi}[V(\rho_T)]
\leq
(1-\alpha)^{T-1}V_0
+
\frac{\epsilon}{\alpha}
\left(
1-(1-\alpha)^{T-1}
\right).
\]
Since $\alpha\in(0,1]$,
\[
\lim_{T\to\infty}(1-\alpha)^{T-1}=0,
\]
and therefore
\[
\limsup_{T\to\infty}
\mathbb{E}_{\pi_\psi}[V(\rho_T)]
\leq
\frac{\epsilon}{\alpha},
\]
which proves (i).

For (ii), since $V(\rho_t)\geq0$, Markov's inequality gives, for
any $\eta>0$,
\[
\Pr_{\pi_\psi}\!\left(V(\rho_t)\geq\eta\right)
\leq
\frac{\mathbb{E}_{\pi_\psi}[V(\rho_t)]}{\eta}.
\]
Taking the limit superior and applying (i) gives
\[
\limsup_{t\to\infty}
\Pr_{\pi_\psi}\!\left(V(\rho_t)\geq\eta\right)
\leq
\frac{1}{\eta}
\limsup_{t\to\infty}
\mathbb{E}_{\pi_\psi}[V(\rho_t)]
\leq
\frac{\epsilon}{\alpha\eta}.
\]
This proves (ii).
\end{proof}

\noindent\textbf{Remark [Role of the contraction parameter $\alpha$].}
The containment and hitting-time guarantees require only $\alpha>0$
together with the stated contractive condition $\alpha\eta>\epsilon$.
By contrast, the geometric infinite-horizon bounds use the factor
$(1-\alpha)^t$ and therefore assume $\alpha\in(0,1]$. The stronger upper
bound $\alpha\le1$ is thus a requirement of that geometric guarantee,
not a necessary restriction on the basic negative-drift condition.


\begin{styledcorollary}{Almost-Sure Asymptotic Safety under a Summable Dissipativity Error}{app:almost_sure_safety}

Consider the setting of Theorem~\ref{thm:asymptotic_risk_stability}, but allow the
Lyapunov dissipativity error to vary with time. Suppose that, for an adapted
nonnegative sequence $\{\epsilon_t\}_{t\geq1}$,
\[
\mathbb{E}_{\pi_\psi}\!\left[
V(\rho_{t+1})-V(\rho_t)
\,\middle|\,
\mathcal{F}_t
\right]
\leq
-\alpha V(\rho_t)+\epsilon_t,
\qquad
\alpha\in(0,1],
\]
with $\epsilon_t$ being $\mathcal{F}_t$-measurable. Assume the integrability
condition
\[
\sum_{t=1}^{\infty}\mathbb{E}_{\pi_\psi}[\epsilon_t]<\infty.
\]
Then
\[
V(\rho_t)\longrightarrow0
\qquad\text{almost surely}.
\]
Equivalently, for every component $j\in\{D,H,M\}$,
\[
\left(\rho_t^{(j)}-\bar{B}^{(j)}\right)_+
\longrightarrow0
\qquad\text{almost surely}.
\]
Thus the magnitude of any component-wise budget violation vanishes
asymptotically almost surely. This is a pathwise asymptotic-safety guarantee;
it does not by itself imply that all violations cease after a finite time.

\end{styledcorollary}

\begin{proof}
The drift condition gives
\[
\mathbb{E}_{\pi_\psi}[V(\rho_{t+1})\mid\mathcal{F}_t]
\leq
V(\rho_t)-\alpha V(\rho_t)+\epsilon_t.
\]
Because $\epsilon_t\geq0$,
\[
\mathbb{E}_{\pi_\psi}[V(\rho_{t+1})\mid\mathcal{F}_t]
\leq
V(\rho_t)+\epsilon_t.
\]
Moreover, by Tonelli's theorem,
\[
\mathbb{E}_{\pi_\psi}\!\left[\sum_{t=1}^{\infty}\epsilon_t\right]
=
\sum_{t=1}^{\infty}\mathbb{E}_{\pi_\psi}[\epsilon_t]
<\infty,
\]
so $\sum_{t=1}^{\infty}\epsilon_t<\infty$ almost surely. The
Robbins--Siegmund almost-supermartingale convergence theorem therefore
applies with the generic Robbins--Siegmund sequences
$X_t=V(\rho_t)$, $a_t^{(\mathrm{RS})}=0$, $b_t^{(\mathrm{RS})}=\epsilon_t$, and
$c_t^{(\mathrm{RS})}=\alpha V(\rho_t)$, distinct from the system and regulatory actions. It follows that $V(\rho_t)$ converges almost surely
to a finite nonnegative random variable and that
\[
\sum_{t=1}^{\infty}\alpha V(\rho_t)<\infty
\qquad\text{almost surely}.
\]
Since $\alpha>0$, $\sum_tV(\rho_t)<\infty$ almost surely.
On the probability-one event where this summability and the finite-limit
conclusion both hold, the nonnegative sequence $V(\rho_t)$ must have
limit zero. Thus $V(\rho_t)\to0$ almost surely. Because $V$ is the sum of squared
positive-part violations, each component-wise positive-part violation
also converges to zero.
\end{proof}

\noindent\textbf{Remark [Connection to continual learning].}
Corollary~\ref{cor:vanishing_surprise} establishes vanishing population
predictive mismatch under its stated asymptotic-ELBO assumption, but vanishing
mismatch alone does \emph{not} imply the stronger summability condition
$\sum_t\epsilon_t<\infty$. Thus, almost-sure asymptotic safety follows when the
learning guarantees are strengthened so that the learning-induced dissipativity
errors are summable, for example through a summable upper bound on the
mismatch-to-dissipativity error. This distinction separates a statistical
convergence result from the stronger pathwise safety conclusion.

\subsection{Learning guarantees}
\label{app:subsec:Learning guarantees}
This subsection characterizes continual adaptation. It covers predictive-evidence improvement, the variational objective, online policy learning and regret, policy convergence, and the sample complexity of variational learning under a stationary strongly convex population objective.

\begin{styledtheorem}{Monotonic Improvement of the Predictive Evidence Bound}{app:predictive_evidence_bound}

Let
\(
\mathcal{F}_t^{-,y}=\sigma(y_{1:t-1},c_{1:t-1})
\)
denote the information available to CENTRY immediately before the
interaction $y_t$ is generated. Let
\(
p_{\mathrm{true}}(y_t\mid\mathcal{F}_t^{-,y})
\)
denote the true conditional distribution of the next interaction. Define
CENTRY's predictive distribution by
\[
p_{\mathrm{pred}}(y_t\mid\mathcal{F}_t^{-,y})
=
\int p_\theta(y_t\mid z_t)\,\kappa_t(z_t)\,dz_t,
\]
where $\kappa_t$ is the predictive latent prior. For $t>1$, this expression follows from Eq.~\eqref{eq:predictivedistribution} by substituting the propagated prior from Eqs.~\eqref{eq:jointpriorfactorization}--\eqref{eq:priorupdate}; at $t=1$, the initial prior is used. Define the realized
predictive surprise by
\[
\widehat S_t=-\log p_{\mathrm{pred}}(y_t\mid\mathcal{F}_t^{-,y}),
\]
and define the conditional population predictive mismatch by
\[
S_t^{\mathrm{(KL)}}
=
D_{\mathrm{KL}}\!\left(
 p_{\mathrm{true}}(\cdot\mid\mathcal{F}_t^{-,y})
 \,\middle\|
 p_{\mathrm{pred}}(\cdot\mid\mathcal{F}_t^{-,y})
\right).
\]
Assume that the conditional entropy of the true interaction distribution has
finite, constant expectation
\[
\mathbb E_{p_{\mathrm{true}}}
\left[
H\left(p_{\mathrm{true}}(\cdot\mid\mathcal F_t^{-,y})\right)
\right]=H^*<\infty
\]
for all $t$, that
$p_{\mathrm{true}}(\cdot\mid\mathcal F_t^{-,y})
\ll p_{\mathrm{pred}}(\cdot\mid\mathcal F_t^{-,y})$ almost surely,
and that all expectations below are finite.

Suppose further that the Learn stage uses the standard ($\beta=1$)
ELBO $\mathcal L_t(\theta_t,\phi_t)$ of Eq.~\eqref{eq:ELBO}, with
$\mathcal L_t$ a valid one-step predictive lower bound:
\[
\mathcal L_t(\theta_t,\phi_t)
\leq
\log p_{\mathrm{pred}}(y_t\mid\mathcal F_t^{-,y})
\qquad\text{a.s.}
\]
Finally, suppose the continual learning procedure is evaluated under a
common stationary environment phase and produces a non-decreasing
expected ELBO:
\[
\mathbb E[\mathcal L_{t+1}(\theta_{t+1},\phi_{t+1})]
\geq
\mathbb E[\mathcal L_t(\theta_t,\phi_t)]
\qquad\text{for every }t.
\]
Define
\[
U_t=-\mathbb E[\mathcal L_t(\theta_t,\phi_t)]-H^*.
\]
Then
\[
0\leq\mathbb E[S_t^{\mathrm{(KL)}}]\leq U_t,
\]
$\{U_t\}$ is non-increasing, and $U_t$ converges to a finite limit
$U_\infty\geq0$.
Moreover,
\[
\mathbb E[\widehat S_t]
=H^*+\mathbb E[S_t^{\mathrm{(KL)}}]
\leq H^*+U_t.
\]
Thus the negative expected ELBO bounds the expected predictive
cross-entropy and, after subtracting the irreducible true entropy $H^*$,
provides a monotone upper bound on population predictive mismatch.

\end{styledtheorem}

\begin{proof}
By the definition of conditional KL divergence,
\[
S_t^{\mathrm{(KL)}}
=
\mathbb E_{p_{\mathrm{true}}}
\left[
\left.
\log\frac{p_{\mathrm{true}}(y_t\mid\mathcal F_t^{-,y})}
{p_{\mathrm{pred}}(y_t\mid\mathcal F_t^{-,y})}
\right|\mathcal F_t^{-,y}
\right].
\]
Taking expectations and using the definition of conditional entropy gives
\[
\mathbb E[S_t^{\mathrm{(KL)}}]
=-H^*-
\mathbb E\left[\log p_{\mathrm{pred}}(y_t\mid\mathcal F_t^{-,y})\right].
\]
Equivalently, because
\(
\widehat S_t=-\log p_{\mathrm{pred}}(y_t\mid\mathcal F_t^{-,y})
\),
\[
\mathbb E[\widehat S_t]
=
H^*+\mathbb E[S_t^{\mathrm{(KL)}}].
\]
The one-step predictive ELBO property implies
\[
-\mathbb E[\log p_{\mathrm{pred}}(y_t\mid\mathcal F_t^{-,y})]
\leq -\mathbb E[\mathcal L_t],
\]
so
\[
0\leq\mathbb E[S_t^{\mathrm{(KL)}}]
\leq
-\mathbb E[\mathcal L_t]-H^*=U_t.
\]
The non-negativity follows from Gibbs' inequality.

By the assumed non-decreasing expected ELBO,
\[
U_{t+1}
=-\mathbb E[\mathcal L_{t+1}]-H^*
\leq
-\mathbb E[\mathcal L_t]-H^*=U_t.
\]
Thus $U_t$ is non-increasing. Since it is also non-negative, it converges
to a finite limit $U_\infty\geq0$.

Finally,
\[
\mathbb E[\widehat S_t]
=-\mathbb E[\log p_{\mathrm{pred}}(y_t\mid\mathcal F_t^{-,y})]
=H^*+\mathbb E[S_t^{\mathrm{(KL)}}]
\leq H^*+U_t,
\]
which proves the result.
\end{proof}


\begin{styledcorollary}{Vanishing Asymptotic Predictive Mismatch}{app:vanishing_surprise}

Under the assumptions of Theorem~\ref{thm:predictive_evidence_bound}, suppose additionally that
\[
\lim_{t\to\infty}\mathbb E[\mathcal L_t(\theta_t,\phi_t)]=-H^*.
\]
Then the population predictive mismatch vanishes:
\[
\lim_{t\to\infty}\mathbb E[S_t^{\mathrm{(KL)}}]=0.
\]
If, in addition, $H^*=0$ (i.e., the true conditional interaction distribution has zero entropy; in the usual discrete case this means it is almost surely deterministic), then
\[
\lim_{t\to\infty}\mathbb E[\widehat S_t]=0.
\]

\end{styledcorollary}

\begin{proof}
By Theorem~\ref{thm:predictive_evidence_bound},
\[
0\leq\mathbb E[S_t^{\mathrm{(KL)}}]
\leq U_t,
\qquad
U_t=-\mathbb E[\mathcal L_t]-H^*.
\]
The assumed convergence of the expected ELBO gives $U_t\to0$. The
squeeze theorem therefore yields
$\mathbb E[S_t^{\mathrm{(KL)}}]\to0$. If $H^*=0$, the identity
$\mathbb E[\widehat S_t]=H^*+\mathbb E[S_t^{\mathrm{(KL)}}]$
then gives $\mathbb E[\widehat S_t]\to0$.
\end{proof}

\noindent\textbf{Remark [Sufficient conditions].}
The condition $\mathbb E[\mathcal L_t]\to-H^*$ is guaranteed whenever
(i) the one-step variational bound becomes asymptotically tight and
(ii) the predictive model becomes asymptotically well-specified, so that
$\mathbb E[S_t^{\mathrm{(KL)}}]\to0$. The two conditions should be kept
conceptually distinct: variational tightness concerns the inference gap,
whereas predictive well-specification concerns the mismatch between the
learned predictive distribution and the true conditional distribution.


The following theorem establishes one-step improvement of the current variational objective. It is a statement about the optimization performed at a given learning step; it does not by itself imply convergence of parameters or monotonic improvement of objectives evaluated on changing future data.

\begin{styledtheorem}{Iterative Improvement of the Variational Objective}{app:elbo_per_iteration_improvement}

Under the Learn-stage updating rule in Eq.~\eqref{eq:learnoperator},
suppose that the parameter update is performed by exact maximization of
the current variational objective:
\(
(\theta_{t+1},\phi_{t+1})
=
\arg\max_{\theta,\phi}
\mathcal{L}_t(\theta,\phi).
\)
Then
\(
\mathcal{L}_t(\theta_{t+1},\phi_{t+1})
\geq
\mathcal{L}_t(\theta_t,\phi_t)
\)
for every $t$.

More generally, if the Learn stage produces an approximate update
whose optimization error is bounded by $\varepsilon_t\geq0$, such that
\(
\mathcal{L}_t(\theta_{t+1},\phi_{t+1})
\geq
\sup_{\theta,\phi}\mathcal{L}_t(\theta,\phi)
-\varepsilon_t,
\)
then
\(
\mathcal{L}_t(\theta_{t+1},\phi_{t+1})
\geq
\mathcal{L}_t(\theta_t,\phi_t)
-\varepsilon_t.
\)

\end{styledtheorem}

\begin{proof}
Under exact maximization,
\[
(\theta_{t+1},\phi_{t+1})
=
\arg\max_{\theta,\phi}
\mathcal{L}_t(\theta,\phi),
\]
so, by the definition of the maximum,
\[
\mathcal{L}_t(\theta_{t+1},\phi_{t+1})
=
\sup_{\theta,\phi}
\mathcal{L}_t(\theta,\phi)
\geq
\mathcal{L}_t(\theta_t,\phi_t).
\]
This proves the first claim.

For an approximate update satisfying
\[
\mathcal{L}_t(\theta_{t+1},\phi_{t+1})
\geq
\sup_{\theta,\phi}\mathcal{L}_t(\theta,\phi)
-\varepsilon_t,
\]
we have
\[
\begin{aligned}
\mathcal{L}_t(\theta_{t+1},\phi_{t+1})
&\geq
\sup_{\theta,\phi}\mathcal{L}_t(\theta,\phi)
-\varepsilon_t\\
&\geq
\mathcal{L}_t(\theta_t,\phi_t)
-\varepsilon_t.
\end{aligned}
\]
This proves the approximate-update result.
\end{proof}


In continual operation, CENTRY adapts its regulatory policy parameters $\psi$ online during the Learn stage without prior knowledge of future task trajectories or environment shifts. The following result gives a sub-linear regret bound relative to the best fixed regulatory policy in hindsight. Under time-varying objectives, this is a static-comparator guarantee rather than a dynamic-regret guarantee.

\begin{styledtheorem}{Sub-Linear Regret Bound for Adaptive Regulation}{app:sublinear_regret_bound}

Let $\Psi \subset \mathbb{R}^{d_\psi}$ be a nonempty, convex, compact parameter space with bounded diameter $D_\Psi = \sup_{\psi,\psi' \in \Psi} \|\psi - \psi'\|_2 < \infty$. At each time step $t$, let the candidate policy $\pi_\psi$ induce the regulatory action $c_t=\pi_\psi(\rho_t)$, and let $J_t(\psi)$ denote the model-based prospective regulatory loss function evaluated during the Learn stage:
\[
J_t(\psi) = \mathbb{E}_{p_{\theta_t}, \pi_\psi} \!\left[ \lambda_S \widehat S_{t+1} + \sum_{j \in \{D,H,M\}} \lambda_j \left( \rho_{t+1}^{(j)} - \bar{B}^{(j)} \right)_+ + \lambda_C C(c_t) \right],
\]
where $\lambda_S, \lambda_j, \lambda_C \ge 0$, and $C(c)\geq0$ for every admissible regulatory action $c$, with $C(\cdot)$ denoting the operational regulatory cost penalty.

Suppose that:
\begin{enumerate}[nosep]
    \item \textbf{Convexity and Continuity:} For each $t$, $J_t(\psi)$ is continuous and convex with respect to $\psi$ over $\Psi$.
    \item \textbf{Bounded Subgradients:} At each $\psi\in\Psi$ and $t\ge1$, a subgradient of $J_t(\psi)$ is selected and denoted by $\nabla J_t(\psi)$ (the ordinary gradient when differentiable). These selected subgradients are uniformly bounded in norm, i.e., $\|\nabla J_t(\psi)\|_2 \le G$ for all $\psi \in \Psi$ and all $t \ge 1$, for some $G>0$.
    \item \textbf{Online Gradient Update:} Assume $\psi_1\in\Psi$, so projection ensures $\psi_t\in\Psi$ for every $t\ge1$. Policy parameters are updated via projected online gradient descent:
    \(
    \psi_{t+1} = \Pi_\Psi \!\left( \psi_t - \eta_t \nabla J_t(\psi_t) \right),
    \)
    where $\Pi_\Psi(v) = \arg\min_{\psi \in \Psi} \|v - \psi\|_2$ denotes Euclidean projection onto $\Psi$, and $\eta_t = \frac{D_\Psi}{G \sqrt{t}}$ is the step-size schedule.
\end{enumerate}

Then, the cumulative regulatory regret over any finite operational horizon $T \in \mathbb{N}$, defined as
\[
\mathrm{Regret}_T(\Psi) = \sum_{t=1}^T J_t(\psi_t) - \min_{\psi^* \in \Psi} \sum_{t=1}^T J_t(\psi^*),
\]
satisfies the sub-linear upper bound:
\(
\mathrm{Regret}_T(\Psi) \le \frac{3}{2} G D_\Psi \sqrt{T} = \mathcal{O}(\sqrt{T}).
\)
Consequently, the upper bound on average per-step regret vanishes asymptotically, so the adaptive policy is asymptotically no worse on average than the best fixed policy chosen in hindsight.

\end{styledtheorem}

\begin{proof}
If $D_\Psi=0$, then the nonempty policy class $\Psi$ is a singleton and regret is identically zero. Hence assume $D_\Psi>0$ below.

Let $\psi^* = \arg\min_{\psi \in \Psi} \sum_{t=1}^T J_t(\psi)$ be the optimal fixed policy parameter vector in hindsight over horizon $T$. Consider the squared Euclidean distance between the updated parameters $\psi_{t+1}$ and $\psi^*$:
\[
\|\psi_{t+1} - \psi^*\|_2^2 = \|\Pi_\Psi(\psi_t - \eta_t \nabla J_t(\psi_t)) - \psi^*\|_2^2.
\]
Because projection onto a convex set $\Psi$ is non-expansive, we have $\|\Pi_\Psi(u) - v\|_2 \le \|u - v\|_2$ for any $v \in \Psi$. Therefore,
\begin{equation}
\begin{aligned}
\|\psi_{t+1} - \psi^*\|_2^2 &\le \|\psi_t - \eta_t \nabla J_t(\psi_t) - \psi^*\|_2^2 \\
&= \|\psi_t - \psi^*\|_2^2 - 2 \eta_t \langle \nabla J_t(\psi_t), \psi_t - \psi^* \rangle + \eta_t^2 \|\nabla J_t(\psi_t)\|_2^2.
\end{aligned}
\label{app:eq:projection_inequality}
\end{equation}

By convexity of $J_t(\psi)$, the subgradient inequality yields:
\[
J_t(\psi_t) - J_t(\psi^*) \le \langle \nabla J_t(\psi_t), \psi_t - \psi^* \rangle.
\]
Rearranging \eqref{app:eq:projection_inequality} and substituting the gradient bound $\|\nabla J_t(\psi_t)\|_2 \le G$:
\begin{equation}
J_t(\psi_t) - J_t(\psi^*) \le \frac{\|\psi_t - \psi^*\|_2^2 - \|\psi_{t+1} - \psi^*\|_2^2}{2 \eta_t} + \frac{\eta_t}{2} G^2.
\label{app:eq:per_step_regret_bound}
\end{equation}

Summing \eqref{app:eq:per_step_regret_bound} across all time steps $t = 1, \ldots, T$:
\begin{equation}
\mathrm{Regret}_T(\Psi) = \sum_{t=1}^T \left( J_t(\psi_t) - J_t(\psi^*) \right) \le \sum_{t=1}^T \frac{\|\psi_t - \psi^*\|_2^2 - \|\psi_{t+1} - \psi^*\|_2^2}{2 \eta_t} + \frac{G^2}{2} \sum_{t=1}^T \eta_t.
\label{app:eq:summed_regret}
\end{equation}

Regrouping the first summation in \eqref{app:eq:summed_regret} via Abel summation (telescoping):
\[
\begin{aligned}
\sum_{t=1}^T \frac{\|\psi_t - \psi^*\|_2^2 - \|\psi_{t+1} - \psi^*\|_2^2}{2 \eta_t} &= \frac{\|\psi_1 - \psi^*\|_2^2}{2 \eta_1} + \sum_{t=2}^T \|\psi_t - \psi^*\|_2^2 \left( \frac{1}{2 \eta_t} - \frac{1}{2 \eta_{t-1}} \right) - \frac{\|\psi_{T+1} - \psi^*\|_2^2}{2 \eta_T}.
\end{aligned}
\]
Since $\|\psi_t - \psi^*\|_2 \le D_\Psi$ for all $t$, and noting that $\eta_t$ is non-increasing (so $\frac{1}{\eta_t} - \frac{1}{\eta_{t-1}} \ge 0$), we drop the negative term $-\frac{\|\psi_{T+1} - \psi^*\|_2^2}{2 \eta_T} \le 0$:
\[
\sum_{t=1}^T \frac{\|\psi_t - \psi^*\|_2^2 - \|\psi_{t+1} - \psi^*\|_2^2}{2 \eta_t} \le \frac{D_\Psi^2}{2} \left( \frac{1}{\eta_1} + \sum_{t=2}^T \left( \frac{1}{\eta_t} - \frac{1}{\eta_{t-1}} \right) \right) = \frac{D_\Psi^2}{2 \eta_T}.
\]

Substituting $\eta_t = \frac{D_\Psi}{G \sqrt{t}}$ into both parts of the bound:
\[
\frac{D_\Psi^2}{2 \eta_T} = \frac{D_\Psi^2 G \sqrt{T}}{2 D_\Psi} = \frac{1}{2} G D_\Psi \sqrt{T}.
\]
For the second summation, bounding the integral $\sum_{t=1}^T \frac{1}{\sqrt{t}} \le \int_0^T \frac{1}{\sqrt{\tau}} d\tau = 2\sqrt{T}$:
\[
\frac{G^2}{2} \sum_{t=1}^T \eta_t = \frac{G^2 D_\Psi}{2 G} \sum_{t=1}^T \frac{1}{\sqrt{t}} \le G D_\Psi \sqrt{T}.
\]

Combining both bounds into \eqref{app:eq:summed_regret}:
\[
\mathrm{Regret}_T(\Psi) \le \frac{1}{2} G D_\Psi \sqrt{T} + G D_\Psi \sqrt{T} = \frac{3}{2} G D_\Psi \sqrt{T},
\]
which establishes the $\mathcal{O}(\sqrt{T})$ upper bound. Dividing by $T$ shows that the upper bound on average per-step regret vanishes asymptotically.
\end{proof}


\begin{styledcorollary}{High-Probability Regret under Bounded Gradient Noise}{app:highprob_regret}

Under the setting of Theorem~\ref{thm:sublinear_regret_bound}, let
\[
\mathcal F_t^{-,\psi}=\sigma(\psi_1,g_1,\ldots,g_{t-1},h_1,\ldots,h_t)
\]
denote the policy-learning filtration generated by the initial iterate, past stochastic gradients, and selected subgradients through time $t$. The iterate $\psi_t$ is measurable with respect to this filtration by the projected update, and the subgradient $h_t$ is selected before the noise $\xi_t$ is drawn. Suppose that the policy update uses
stochastic gradients of the form
\[
g_t=h_t+\xi_t,
\qquad
h_t\in\partial J_t(\psi_t),
\qquad
\|h_t\|_2\leq G,
\]
where the noise sequence satisfies
\[
\mathbb E[\xi_t\mid\mathcal F_t^{-,\psi}]=0,
\qquad
\|\xi_t\|_2\leq\sigma
\qquad\text{almost surely},
\]
for some $\sigma\geq0$ with $G+\sigma>0$. Let
\[
\eta_t
=
\frac{D_\Psi}{(G+\sigma)\sqrt{t}}.
\]
For any fixed (non-random) comparator $\psi^\star\in\Psi$, define
\[
\mathrm{Regret}_T(\psi^\star)
=
\sum_{t=1}^{T}\bigl(J_t(\psi_t)-J_t(\psi^\star)\bigr).
\]
Then, for every $\delta\in(0,1)$, with probability at least
$1-\delta$,
\[
\mathrm{Regret}_T(\psi^\star)
\leq
\frac{3}{2}(G+\sigma)D_\Psi\sqrt{T}
+
\sigma D_\Psi
\sqrt{2T\log\!\left(\frac{1}{\delta}\right)}.
\]
Thus, bounded martingale gradient noise preserves the
$O(\sqrt T)$ regret rate while providing an explicit high-probability
single-trajectory guarantee against any fixed noise-independent comparator.
A high-probability guarantee against the best comparator chosen in hindsight
requires an additional uniform-concentration argument and is not asserted here.

\end{styledcorollary}

\begin{proof}
If $D_\Psi=0$, then the nonempty policy class $\Psi$ is a singleton and regret is identically zero. Hence assume $D_\Psi>0$ below.

Fix $\psi^\star\in\Psi$. By convexity,
\[
J_t(\psi_t)-J_t(\psi^\star)
\leq
\langle h_t,\psi_t-\psi^\star\rangle.
\]
Since
\[
h_t=g_t-\xi_t,
\]
we obtain
\[
J_t(\psi_t)-J_t(\psi^\star)
\leq
\langle g_t,\psi_t-\psi^\star\rangle
-
\langle\xi_t,\psi_t-\psi^\star\rangle.
\]

Projection non-expansiveness gives
\[
\begin{aligned}
\|\psi_{t+1}-\psi^\star\|_2^2
&\leq
\|\psi_t-\eta_tg_t-\psi^\star\|_2^2 \\
&=
\|\psi_t-\psi^\star\|_2^2
-
2\eta_t
\langle g_t,\psi_t-\psi^\star\rangle
+
\eta_t^2\|g_t\|_2^2.
\end{aligned}
\]
Hence
\[
\langle g_t,\psi_t-\psi^\star\rangle
\leq
\frac{
\|\psi_t-\psi^\star\|_2^2
-
\|\psi_{t+1}-\psi^\star\|_2^2
}{2\eta_t}
+
\frac{\eta_t}{2}\|g_t\|_2^2.
\]
Because
\[
\|g_t\|_2
\leq
\|h_t\|_2+\|\xi_t\|_2
\leq
G+\sigma
\]
almost surely, summing gives
\[
\begin{aligned}
\mathrm{Regret}_T(\psi^\star)
&\leq
\sum_{t=1}^{T}
\frac{
\|\psi_t-\psi^\star\|_2^2
-
\|\psi_{t+1}-\psi^\star\|_2^2
}{2\eta_t} \\
&\quad+
\frac{(G+\sigma)^2}{2}
\sum_{t=1}^{T}\eta_t
-
\sum_{t=1}^{T}
\langle\xi_t,\psi_t-\psi^\star\rangle.
\end{aligned}
\]
The first two terms are bounded by the projected-OGD calculation in
Theorem~\ref{thm:sublinear_regret_bound}, with gradient bound $G+\sigma$:
\[
\sum_{t=1}^{T}
\frac{
\|\psi_t-\psi^\star\|_2^2
-
\|\psi_{t+1}-\psi^\star\|_2^2
}{2\eta_t}
+
\frac{(G+\sigma)^2}{2}
\sum_{t=1}^{T}\eta_t
\leq
\frac{3}{2}(G+\sigma)D_\Psi\sqrt{T}.
\]

Define
\[
X_t=-\langle\xi_t,\psi_t-\psi^\star\rangle.
\]
Because $\psi_t$ is $\mathcal F_t^{-,\psi}$-measurable and $\psi^\star$ is fixed,
\[
\mathbb E[X_t\mid\mathcal F_t^{-,\psi}]=0.
\]
Since $X_t$ is $\mathcal F_{t+1}^{-,\psi}$-measurable, it is a martingale difference with respect to the filtration indexed by $\mathcal F_{t+1}^{-,\psi}$. Moreover,
\[
|X_t|
\leq
\|\xi_t\|_2\,\|\psi_t-\psi^\star\|_2
\leq
\sigma D_\Psi
\]
almost surely. If $\sigma=0$, then $X_t=0$ almost surely for every $t$, and the claimed bound follows directly from the projected-OGD bound above. Assume $\sigma>0$ below. Azuma--Hoeffding therefore yields
\[
\Pr\!\left(
\sum_{t=1}^{T}X_t
\geq
\sigma D_\Psi
\sqrt{2T\log\!\left(\frac1\delta\right)}
\right)
\leq\delta.
\]
On the complementary event, combining this bound with the deterministic
projected-OGD bound proves the result.
\end{proof}


\begin{styledtheorem}{Dynamic Regret Bound for Non-Stationary Environments}{app:dynamic_regret_bound}

Under the convexity, bounded-subgradient, and feasible-initialization assumptions of
Theorem~\ref{thm:sublinear_regret_bound}, suppose that, for each
$t=1,\ldots,T$, the instantaneous objective $J_t$ has a minimizer
\[
\psi_t^\star
\in
\arg\min_{\psi\in\Psi}J_t(\psi).
\]
Define the path length of the sequence of instantaneous optimal policies
by
\[
P_T
=
\sum_{t=1}^{T-1}
\|\psi_{t+1}^\star-\psi_t^\star\|_2.
\]
Suppose that the policy update uses the horizon-tuned constant step size
\[
\eta
=
\frac{D_\Psi}{G\sqrt{T}}
\]
throughout the horizon. Then the dynamic regulatory regret
\[
\mathrm{Regret}_T^d
=
\sum_{t=1}^{T}J_t(\psi_t)
-
\sum_{t=1}^{T}J_t(\psi_t^\star)
\]
is identically zero if $D_\Psi=0$, since the policy class is then a singleton. For $D_\Psi>0$, it satisfies
\[
\mathrm{Regret}_T^d
\leq
\frac{D_\Psi^2}{2\eta}
+
\frac{\eta G^2T}{2}
+
\frac{D_\Psi}{\eta}P_T.
\]
Consequently,
\[
\mathrm{Regret}_T^d
\leq
GD_\Psi\sqrt{T}
+
G\sqrt{T}\,P_T
=O\!\left(\sqrt{T}(1+P_T)\right).
\]
If $P_T=o(\sqrt{T})$, then
\[
\frac{1}{T}\mathrm{Regret}_T^d\longrightarrow0.
\]
Thus, under sub-$\sqrt T$ cumulative drift of the instantaneous optimal
policy, CENTRY's online regulatory policy has vanishing average dynamic
regret relative to the sequence of instantaneous optimal policies.

\end{styledtheorem}

\begin{proof}
If $D_\Psi=0$, then the nonempty policy class $\Psi$ is a singleton and regret is identically zero. Hence assume $D_\Psi>0$ below.

By convexity of $J_t$,
\[
J_t(\psi_t)-J_t(\psi_t^\star)
\leq
\langle g_t,\psi_t-\psi_t^\star\rangle,
\qquad
 g_t\in\partial J_t(\psi_t).
\]
Projection non-expansiveness gives
\[
\begin{aligned}
\|\psi_{t+1}-\psi_t^\star\|_2^2
&\leq
\|\psi_t-\eta g_t-\psi_t^\star\|_2^2\\
&=
\|\psi_t-\psi_t^\star\|_2^2
-2\eta\langle g_t,\psi_t-\psi_t^\star\rangle
+\eta^2\|g_t\|_2^2.
\end{aligned}
\]
Therefore,
\[
J_t(\psi_t)-J_t(\psi_t^\star)
\leq
\frac{
\|\psi_t-\psi_t^\star\|_2^2
-\|\psi_{t+1}-\psi_t^\star\|_2^2
}{2\eta}
+
\frac{\eta G^2}{2}.
\]
Summing over $t=1,\ldots,T$ and adding and subtracting
$\|\psi_{t+1}-\psi_{t+1}^\star\|_2^2$ gives
\[
\begin{aligned}
\mathrm{Regret}_T^d
&\leq
\frac{\|\psi_1-\psi_1^\star\|_2^2}{2\eta}
+
\frac{\eta G^2T}{2}\\
&\quad+
\sum_{t=1}^{T-1}
\frac{
\|\psi_{t+1}-\psi_{t+1}^\star\|_2^2
-
\|\psi_{t+1}-\psi_t^\star\|_2^2
}{2\eta}.
\end{aligned}
\]
Since all policy vectors lie in $\Psi$,
$\|\psi_1-\psi_1^\star\|_2\leq D_\Psi$. Further, for any
$x,a,b\in\Psi$,
\[
\|x-a\|_2^2-\|x-b\|_2^2
\leq2D_\Psi\|a-b\|_2.
\]
Hence
\[
\mathrm{Regret}_T^d
\leq
\frac{D_\Psi^2}{2\eta}
+
\frac{\eta G^2T}{2}
+
\frac{D_\Psi}{\eta}P_T.
\]
Substituting $\eta=D_\Psi/(G\sqrt T)$ yields
\[
\mathrm{Regret}_T^d
\leq
\frac12GD_\Psi\sqrt T
+
\frac12GD_\Psi\sqrt T
+
G\sqrt T\,P_T
=
GD_\Psi\sqrt T+G\sqrt T\,P_T,
\]
which is the stated bound. Finally,
\[
\frac{\mathrm{Regret}_T^d}{T}
\leq
\frac{GD_\Psi}{\sqrt T}
+
G\frac{P_T}{\sqrt T},
\]
which converges to zero when $P_T=o(\sqrt T)$.
\end{proof}

\noindent\textbf{Remark [Horizon-dependent step size].}
The constant step size $\eta=D_\Psi/(G\sqrt{T})$ used in the analysis of
Theorem~\ref{thm:dynamic_regret_bound} is horizon-dependent and therefore
is not directly implementable online when the horizon $T$ is unknown. In
an online implementation, the time-varying step size
$\eta_t=D_\Psi/(G\sqrt{t})$ of Theorem~\ref{thm:sublinear_regret_bound}
can be used instead. A standard time-varying dynamic-regret analysis then
retains the sublinear dependence on $T$, with a bound of order
$O(\sqrt{T}(1+P_T))$ up to an additional logarithmic factor. The
horizon-tuned result above is therefore an analysis of the dynamic-regret
performance under a fixed horizon, rather than a prescription that
requires the online algorithm to know $T$ in advance.

\noindent\textbf{Remark [Alternative regret regimes].}
The regret guarantees in this section use different step-size schedules
because they analyze different structural regimes. The deterministic
static-regret result uses the time-varying schedule
$\eta_t=D_\Psi/(G\sqrt t)$, whereas the dynamic-regret result uses the
horizon-dependent constant schedule
$\eta=D_\Psi/(G\sqrt T)$, and the bounded-noise high-probability result
uses $\eta_t=D_\Psi/((G+\sigma)\sqrt t)$. The strongly convex results
constitute a separate regime with $\eta_t=1/(\mu t)$ or
$\eta_t=1/[\mu(t+1)]$. These schedules should not be interchanged while
retaining the stated constants. If a single schedule is to be deployed
across noisy and noiseless settings, its complete regret bound should be
derived under that common schedule rather than obtained by informal
substitution of constants.


\begin{styledcorollary}{Asymptotic Bound on Average Component-Risk Budget Excess}{app:average_risk_compliance}

Under the hypotheses of Theorem~\ref{thm:sublinear_regret_bound}, assume in addition that
$\widehat S_{t+1}\geq0$ almost surely under every candidate policy
$\pi_\psi$, $\psi\in\Psi$, for each $t$ (so that the surprise term in
$J_t$ is non-negative), and that $C(c)\geq0$ for every admissible
regulatory action $c$ (as in Theorem~\ref{thm:sublinear_regret_bound}), and let
$\psi^*=\arg\min_{\psi\in\Psi}\sum_{t=1}^T J_t(\psi)$. Then
\[
\frac1T\sum_{t=1}^T\sum_{j\in\{D,H,M\}}\lambda_j\,
\mathbb{E}_{p_{\theta_t},\pi_{\psi_t}}\!\left[\left(\rho_{t+1}^{(j)}-\bar{B}^{(j)}\right)_+\right]
\;\le\;
\frac1T\sum_{t=1}^T J_t(\psi^*)
\;+\;
\frac{3}{2}\,\frac{GD_\Psi}{\sqrt T}.
\]
As $T\to\infty$, the slack term vanishes. The average weighted component-risk budget excess under the online policy is thus bounded asymptotically by the average total regulatory objective of the best fixed policy in hindsight, not necessarily by its component-risk budget excess alone.

\end{styledcorollary}

\begin{proof}
Since $\lambda_S,\lambda_C\ge0$, $C(c)\ge0$ for every admissible action
by the standing nonnegativity assumption on the regulatory cost, and
$\widehat S_{t+1}\ge0$ under the additional hypothesis of the present
corollary, we have $\lambda_S\widehat S_{t+1}\ge0$ and
$\lambda_C C(c_t)\ge0$. Dropping these two nonnegative terms from
$J_t(\psi)$ therefore gives, for every $\psi\in\Psi$ and every $t$,
\[
\sum_{j\in\{D,H,M\}}\lambda_j\,\mathbb{E}_{p_{\theta_t},\pi_\psi}\!\left[\left(\rho_{t+1}^{(j)}-\bar{B}^{(j)}\right)_+\right]
\;\le\;
J_t(\psi).
\]
Applying this at $\psi=\psi_t$, summing over $t=1,\dots,T$, and using
Theorem~\ref{thm:sublinear_regret_bound},
\[
\sum_{t=1}^T\sum_{j}\lambda_j\,\mathbb{E}_{p_{\theta_t},\pi_{\psi_t}}\!\left[\left(\rho_{t+1}^{(j)}-\bar{B}^{(j)}\right)_+\right]
\;\le\;
\sum_{t=1}^T J_t(\psi_t)
\;\le\;
\sum_{t=1}^T J_t(\psi^*) + \frac32 GD_\Psi\sqrt T.
\]
Dividing by $T$ gives the claim.
\end{proof}


\begin{styledcorollary}{Strongly Convex Policy Convergence}{app:strongly_convex_policy_convergence}

Consider the CENTRY regulation policy update of
Eq.~\eqref{eq:policylearning} on a nonempty compact convex set
$\Psi\subset\mathbb{R}^d$. In each of the two learning runs below, assume $\psi_1\in\Psi$ and use projection onto $\Psi$ at every update.
\begin{enumerate}
    \item \textbf{Logarithmic Regret.} \label{app:item:SC-regret}
    Suppose
    \[
    \psi_{t+1}
    =
    \Pi_{\Psi}(\psi_t-\eta_tg_t),
    \qquad
    g_t\in\partial J_t(\psi_t),
    \qquad
    \eta_t=\frac{1}{\mu t},
    \]
    where each instantaneous objective $J_t:\Psi\to\mathbb R$ is
    $\mu$-strongly convex for some $\mu>0$, and
    $\|g_t\|_2\leq G$. Then the cumulative regret against the fixed
    best-in-hindsight policy
    \[
    \psi^\star
    \in
    \arg\min_{\psi\in\Psi}
    \sum_{t=1}^{T}J_t(\psi)
    \]
    satisfies
    \[
    \mathrm{Regret}_T
    \leq
    \frac{G^2}{2\mu}(1+\ln T)
    =O(\log T).
    \]

    \item \textbf{Stationary Parameter Convergence.} \label{app:item:SC-converge}
    For a separate stationary-objective learning run, suppose that there exists
    a differentiable, $\mu$-strongly convex stationary objective
    $J:\Psi\to\mathbb R$ with a unique minimizer
    \[
    \psi^\star=\arg\min_{\psi\in\Psi}J(\psi),
    \]
    that lies in the interior of $\Psi$. Thus $\psi^\star$ is an unconstrained
    local minimizer and satisfies $\nabla J(\psi^\star)=0$. Suppose further that
    the stochastic update directions satisfy
    \[
    \mathbb E[g_t\mid\mathcal F_t^{-,\psi}]=\nabla J(\psi_t),
    \qquad
    \mathbb E[\|g_t\|_2^2\mid\mathcal F_t^{-,\psi}]\leq G^2.
    \]
    The conditional moments are taken with respect to the policy-learning
    filtration of this run, $\mathcal F_t^{-,\psi}=\sigma(\psi_1,g_1,\ldots,g_{t-1})$. For this stationary
    setting, use the step size
    \[
    \eta_t=\frac{1}{\mu(t+1)}.
    \]
    Then, for every $t\geq2$,
    \[
    \mathbb E\left[\|\psi_t-\psi^\star\|_2^2\right]
    \leq
    \frac{G^2}{\mu^2(t+1)}
    =O\left(\frac1t\right),
    \]
    and consequently,
    \[
    \mathbb E\left[\|\psi_t-\psi^\star\|_2\right]
    \leq
    \frac{G}{\mu\sqrt{t+1}}
    =O\left(\frac1{\sqrt t}\right).
    \]
    Thus the learned policy parameters converge to the unique stationary
    risk-optimal policy $\psi^\star$ in mean square.
\end{enumerate}

\end{styledcorollary}

\begin{proof}
\textbf{Logarithmic regret.}
Let
$D_t=\|\psi_t-\psi^\star\|_2^2$. Since $J_t$ is
$\mu$-strongly convex, the standard projected strongly-convex OGD
argument gives
\[
J_t(\psi_t)-J_t(\psi^\star)
\leq
\frac{\mu}{2}
\bigl[(t-1)D_t-tD_{t+1}\bigr]
+
\frac{G^2}{2\mu t}.
\]
Summing from $t=1$ to $T$ and using
\[
\sum_{t=1}^{T}[(t-1)D_t-tD_{t+1}]=-TD_{T+1}\leq0
\]
yields
\[
\mathrm{Regret}_T
\leq
\frac{G^2}{2\mu}
\sum_{t=1}^{T}\frac1t
\leq
\frac{G^2}{2\mu}(1+\ln T).
\]

\medskip
\textbf{Stationary-objective convergence.}
Let
\[
D_t=\|\psi_t-\psi^\star\|_2^2.
\]
By projection non-expansiveness,
\[
D_{t+1}
\leq
D_t-2\eta_t\langle g_t,\psi_t-\psi^\star\rangle
+\eta_t^2\|g_t\|_2^2.
\]
Taking conditional expectations with respect to $\mathcal F_t^{-,\psi}$ and using
unbiasedness and the second-moment bound gives
\[
\mathbb E[D_{t+1}\mid\mathcal F_t^{-,\psi}]
\leq
D_t-2\eta_t\langle\nabla J(\psi_t),\psi_t-\psi^\star\rangle
+\eta_t^2G^2.
\]
Because $J$ is differentiable and $\mu$-strongly convex, its
gradient is $\mu$-strongly monotone:
\[
\left\langle
\nabla J(\psi_t)-\nabla J(\psi^\star),
\psi_t-\psi^\star\right\rangle
\geq
\mu\|\psi_t-\psi^\star\|_2^2.
\]
Since $\nabla J(\psi^\star)=0$ by the interior-minimizer assumption,
\[
\langle\nabla J(\psi_t),\psi_t-\psi^\star\rangle
\geq
\mu D_t.
\]
Therefore,
\[
\mathbb E[D_{t+1}\mid\mathcal F_t^{-,\psi}]
\leq
(1-2\mu\eta_t)D_t+\eta_t^2G^2.
\]
With $\eta_t=1/[\mu(t+1)]$,
\[
\mathbb E[D_{t+1}\mid\mathcal F_t^{-,\psi}]
\leq
\frac{t-1}{t+1}D_t
+
\frac{G^2}{\mu^2(t+1)^2}.
\]
Taking expectations and writing $c=G^2/\mu^2$ gives
\[
e_{t+1}
\leq
\frac{t-1}{t+1}e_t+
\frac{c}{(t+1)^2},
\qquad e_t=\mathbb E[D_t].
\]
For $t=1$, this yields
\[
e_2\leq\frac{c}{4}.
\]
We claim that $e_t\leq c/(t+1)$ for all $t\geq2$. The base case follows
from $c/4\leq c/3$. If the claim holds at some $t\geq2$, then
\[
\begin{aligned}
e_{t+1}
&\leq
\frac{t-1}{t+1}\frac{c}{t+1}
+
\frac{c}{(t+1)^2}\\
&=
\frac{ct}{(t+1)^2}
\leq
\frac{c}{t+2},
\end{aligned}
\]
where the last inequality follows from
$t(t+2)\leq(t+1)^2$. Thus
\[
\mathbb E[D_t]
\leq
\frac{G^2}{\mu^2(t+1)},
\qquad t\geq2.
\]
Finally, Jensen's inequality gives
\[
\mathbb E[\|\psi_t-\psi^\star\|_2]
\leq
\sqrt{\mathbb E[D_t]}
\leq
\frac{G}{\mu\sqrt{t+1}}.
\]
This proves the stated mean-square convergence.
\end{proof}

\noindent\textbf{Remark [Interpretation and scope of the convergence guarantee].}
\begin{itemize}[nosep]

\item \emph{Part~\ref{app:item:SC-regret} is deterministic.}
It requires only $\mu$-strong convexity of each instantaneous objective
$J_t$ and the uniform subgradient bound $\|g_t\|_2\leq G$; no stochastic
or stationarity assumption is required. It therefore strengthens the
$O(1/\sqrt{T})$ average-regret guarantee of
Theorem~\ref{thm:sublinear_regret_bound} to
$O(\log T/T)$ average regret.

\item \emph{Part~\ref{app:item:SC-converge} requires a stationary objective.}
In the adversarial online setting of Part~\ref{app:item:SC-regret}, the
objectives $J_t$ may change from round to round, so logarithmic regret
alone does not imply convergence of the policy parameters to a fixed
point. The additional assumption of a single strongly convex objective
$J$, together with unbiased stochastic gradients, provides the
contraction needed for convergence to its unique minimizer
$\psi^\star$.

\item \emph{The convergence is in mean square.}
Specifically,
\[
\mathbb{E}\!\left[
\|\psi_t-\psi^\star\|_2^2
\right]
=
O(1/t),
\]
which implies convergence in probability and
\[
\mathbb{E}\!\left[
\|\psi_t-\psi^\star\|_2
\right]
=
O(1/\sqrt t).
\]
The result does not, by itself, establish almost-sure convergence of
the realized policy iterates.

\item \emph{Interpretation for CENTRY.}
Within the stationary strongly convex regime, the result formalizes a
``regulator settles down'' property: the learned CENTRY policy
concentrates around the unique minimizer $\psi^\star$ of the stationary policy objective as
the number of learning iterations increases, even though the monitored
system may continue to generate observations and the latent-state
estimate may continue to be updated.
\end{itemize}


\begin{styledtheorem}{Sample Complexity of Variational Learning}{app:sample_complexity}

Under the stationary-environment assumptions of
Theorem~\ref{thm:predictive_evidence_bound}, let
$w=(\theta,\phi)\in\Xi$, where $\Xi\subset\mathbb R^d$ is closed and
convex. Let the population objective
\[
\mathcal L(w)=\mathbb E[\mathcal L_t(w)]
\]
be $\mu$-strongly concave. Equivalently, let
$f(w)=-\mathcal L(w)$ be $\mu$-strongly convex. Let
\[
\mathcal F_t^{-,w}=\sigma(w_1,g_1,\ldots,g_{t-1})
\]
denote the model-learning filtration containing the information available
immediately before $w_t$ is updated. Suppose the online Learn stage uses
projected stochastic-gradient updates
\[
w_{t+1}=\Pi_\Xi(w_t-\eta_tg_t),
\qquad
\eta_t=\frac{2}{\mu(t+1)},
\]
with
\[
\mathbb E[g_t\mid\mathcal F_t^{-,w}]=\nabla f(w_t),
\qquad
\mathbb E\!\left[\|g_t\|_2^2\mid\mathcal F_t^{-,w}\right]\le G^2.
\]
Define the polynomially weighted average
\[
\widetilde w_N
=
\frac{2}{N(N+1)}\sum_{t=1}^{N}t\,w_t,
\]
and let
\[
w^\star\in\arg\min_{w\in\Xi}f(w),
\qquad
\mathcal L^\star=\mathcal L(w^\star).
\]
Then
\[
\mathbb E\!\left[
\mathcal L^\star-\mathcal L(\widetilde w_N)
\right]
\leq
\frac{2G^2}{\mu(N+1)}.
\]
Thus the polynomially weighted online variational iterate has expected
ELBO gap $O(1/N)$.

If $f$ is additionally $L$-smooth and $w^\star$ is an interior
unconstrained optimum, then, for $N\geq2$, the last iterate satisfies
\[
\mathbb E\!\left[
\mathcal L^\star-\mathcal L(w_N)
\right]
\leq
\frac{2LG^2}{\mu^2(N+1)}.
\]

\end{styledtheorem}

\begin{proof}
Let $D_t=\|w_t-w^\star\|_2^2$. Projection non-expansiveness gives
\[
D_{t+1}
\leq
D_t-2\eta_t\langle g_t,w_t-w^\star\rangle+
\eta_t^2\|g_t\|_2^2.
\]
Taking conditional expectations and using unbiasedness gives
\[
\mathbb E[D_{t+1}\mid\mathcal F_t^{-,w}]
\leq
D_t-2\eta_t\langle\nabla f(w_t),w_t-w^\star\rangle+
\eta_t^2G^2.
\]
By $\mu$-strong convexity,
\[
\langle\nabla f(w_t),w_t-w^\star\rangle
\geq
f(w_t)-f(w^\star)+\frac{\mu}{2}D_t.
\]
With $\eta_t=2/[\mu(t+1)]$,
\[
\mathbb E[D_{t+1}\mid\mathcal F_t^{-,w}]
\leq
\frac{t-1}{t+1}D_t
-\frac{4}{\mu(t+1)}\bigl(f(w_t)-f(w^\star)\bigr)
+\frac{4G^2}{\mu^2(t+1)^2}.
\]
Multiplying by $\mu t(t+1)/4$ and rearranging yields
\[
t\bigl(f(w_t)-f(w^\star)\bigr)
\leq
\frac{\mu}{4}t(t-1)D_t
-\frac{\mu}{4}t(t+1)\mathbb E[D_{t+1}\mid\mathcal F_t^{-,w}]
+\frac{G^2t}{\mu(t+1)}.
\]
Taking expectations and summing from $t=1$ to $N$ gives the exact
potential cancellation
\[
\begin{aligned}
\sum_{t=1}^{N}t\,\mathbb E[f(w_t)-f(w^\star)]
&\leq
-\frac{\mu}{4}N(N+1)\mathbb E[D_{N+1}]
+\frac{G^2}{\mu}\sum_{t=1}^{N}\frac{t}{t+1}\\
&\leq
\frac{NG^2}{\mu}.
\end{aligned}
\]
By convexity of $f$,
\[
f(\widetilde w_N)
\leq
\frac{2}{N(N+1)}\sum_{t=1}^{N}t\,f(w_t).
\]
Therefore,
\[
\mathbb E[f(\widetilde w_N)-f(w^\star)]
\leq
\frac{2G^2}{\mu(N+1)},
\]
which is the weighted-average ELBO-gap bound because $f=-\mathcal L$.

For the last iterate, let
\[
e_t=\mathbb E[D_t],
\qquad
c=\frac{G^2}{\mu^2}.
\]
From the preceding recursion,
\[
e_{t+1}
\leq
\frac{t-1}{t+1}e_t
+
\frac{4c}{(t+1)^2}.
\]
For $t=1$, the recursion gives
\[
e_2\leq\frac{4c}{4}=c\leq\frac{4c}{3}.
\]
We claim that
\[
e_t\leq\frac{4c}{t+1},
\qquad t\geq2.
\]
The base case $t=2$ follows from the preceding bound. Suppose the claim
holds for some $t\geq2$. Then
\[
\begin{aligned}
e_{t+1}
&\leq
\frac{t-1}{t+1}\frac{4c}{t+1}
+
\frac{4c}{(t+1)^2}\\
&=
\frac{4ct}{(t+1)^2}
\leq
\frac{4c}{t+2},
\end{aligned}
\]
because
$t(t+2)\leq(t+1)^2$. Thus
\[
\mathbb E[D_N]
= e_N
\leq
\frac{4G^2}{\mu^2(N+1)}.
\]
Because $w^\star$ is an interior unconstrained optimum,
$\nabla f(w^\star)=0$. By $L$-smoothness,
\[
f(w_N)-f(w^\star)
\leq
\frac{L}{2}\|w_N-w^\star\|_2^2.
\]
Taking expectations proves
\[
\mathbb E[f(w_N)-f(w^\star)]
\leq
\frac{2LG^2}{\mu^2(N+1)}.
\]
Using $f=-\mathcal L$ completes the proof.
\end{proof}

\noindent\textbf{Remark [Averaging and smoothness].}
The $O(1/N)$ theorem concerns the polynomially weighted averaged
iterate and does not require $L$-smoothness; only strong convexity and
the stated stochastic second-moment bound are used. Smoothness is needed
only for the optional last-iterate refinement, together with the
interior-optimum condition.


The following theorem gives a one-step policy-improvement guarantee for the model-based prospective regulatory objective when the model parameters are held fixed. It does not by itself establish convergence across changing model parameters or changing objectives.

\begin{styledtheorem}{Policy Improvement Under the Model-Based Objective}{app:policy_improvement}

Let $\pi_{\psi_t}$ be the regulatory policy at time $t$, and let
$\pi_{\psi_{t+1}}$ be the policy obtained by the update in
Eq.~\eqref{eq:policylearning}. For each candidate policy $\pi$, let
$c_t=\pi(\rho_t)$ denote its induced regulatory action. Define the
model-based prospective regulatory objective
\[
J(\pi,\theta_t)
=
\mathbb{E}_{p_{\theta_t},\pi}
\left[
\lambda_S \widehat S_{t+1}
+
\sum_{j\in\{D,H,M\}}
\lambda_j
\left(
\rho_{t+1}^{(j)}-\bar{B}^{(j)}
\right)_+
+
\lambda_C C(c_t)
\right].
\]
If the policy update performs exact minimization over a candidate set containing the preceding policy (in particular, the preceding policy must be feasible for a constrained update),
\(
\psi_{t+1}
=
\arg\min_{\psi}J(\pi_\psi,\theta_t),
\)
then
\(
J(\pi_{\psi_{t+1}},\theta_t)
\leq
J(\pi_{\psi_t},\theta_t).
\)
Thus, the updated policy cannot have a larger model-based prospective
regulatory objective than the policy from the preceding iteration,
when both are evaluated under the same model parameters $\theta_t$.

\end{styledtheorem}

\begin{proof}
By definition, with the minimization taken over the stated candidate set containing $\psi_t$,
\[
\psi_{t+1}
=
\arg\min_{\psi}J(\pi_\psi,\theta_t).
\]
Therefore,
\[
J(\pi_{\psi_{t+1}},\theta_t)
=
\min_{\psi}J(\pi_\psi,\theta_t)
\leq
J(\pi_{\psi_t},\theta_t),
\]
which proves the result.
\end{proof}


\subsection{Inference guarantees}
\label{app:subsec:Inference guarantees}
This subsection addresses the statistical behavior of CENTRY's latent-state inference, including predictive surprise, temporal scale separation, robustness to prior initialization, rating consistency, component separation and identifiability, source attribution, and steady-state posterior uncertainty.

Throughout this subsection, the pre-interaction filtration is
\(
\mathcal F_t^{-,y}:=\sigma(y_{1:t-1},c_{1:t-1}),
\)
which contains the information available immediately before the interaction $y_t$ is generated.

The following theorem formalizes the conceptual argument developed in Section~\ref{subsec:Role of Predict and Surprise in CENTRY} regarding the role of Surprise as a computable certificate of how much inferential and adaptive work the current step requires.

\begin{styledtheorem}{Predictive Surprise and the Variational Objective}{app:surprise_elbo_relation}

Let $\kappa_t(z_t)$ be the predictive latent prior and define the
one-step predictive marginal by
\[
p_{\mathrm{pred}}(y_t\mid\mathcal F_t^{-,y})
=
\int p_\theta(y_t\mid z_t)\kappa_t(z_t)\,dz_t.
\]
Let
\[
\widehat S_t=-\log p_{\mathrm{pred}}(y_t\mid\mathcal F_t^{-,y})
\]
be the realized predictive negative log-likelihood, and let
$q_t(z_t)=q_\phi(z_t\mid y_{1:t},c_{1:t-1})$ be a variational posterior.
For $\beta>0$, define
\[
\mathcal L_{t,\beta}(\theta,\phi)
=
\mathbb E_{q_t}[\log p_\theta(y_t\mid z_t)]
-\beta D_{\mathrm{KL}}(q_t\|\kappa_t).
\]
Assume $q_t\ll\kappa_t$, $p_\theta(y_t\mid z_t)>0$
$\kappa_t$-almost everywhere on the support of $q_t$, and all displayed
quantities are finite. Then
\[
\mathcal L_{t,\beta}
=
\log p_{\mathrm{pred}}(y_t\mid\mathcal F_t^{-,y})
-D_{\mathrm{KL}}\!
\left(q_t\middle\|p_\theta(z_t\mid y_t,\mathcal F_t^{-,y})\right)
-(\beta-1)D_{\mathrm{KL}}(q_t\|\kappa_t).
\]
Consequently, for every $\beta\geq1$,
\[
\mathcal L_{t,\beta}\leq\log p_{\mathrm{pred}}(y_t\mid\mathcal F_t^{-,y})
=-\widehat S_t,
\]
so that
\[
-\mathcal L_{t,\beta}\geq\widehat S_t.
\]
The result is a pointwise variational statement about the realized
one-step predictive likelihood; it does not identify realized surprise
with a conditional KL divergence.

\end{styledtheorem}

\begin{proof}
Bayes' rule gives, $q_t$-almost everywhere,
\[
\log p_\theta(y_t\mid z_t)+\log\kappa_t(z_t)
=
\log p_{\mathrm{pred}}(y_t\mid\mathcal F_t^{-,y})
+\log p_\theta(z_t\mid y_t,\mathcal F_t^{-,y}).
\]
Taking expectation under $q_t$ and rearranging yields the standard
variational identity
\[
\mathbb E_{q_t}[\log p_\theta(y_t\mid z_t)]
-D_{\mathrm{KL}}(q_t\|\kappa_t)
=
\log p_{\mathrm{pred}}(y_t\mid\mathcal F_t^{-,y})
-D_{\mathrm{KL}}\left(q_t\middle\|p_\theta(z_t\mid y_t,\mathcal F_t^{-,y})\right).
\]
Subtracting an additional $(\beta-1)D_{\mathrm{KL}}(q_t\|\kappa_t)$
gives the boxed identity. For $\beta\geq1$, both KL divergences on the
right-hand side are non-negative and their coefficients are non-positive,
so
$\mathcal L_{t,\beta}\leq\log p_{\mathrm{pred}}=-\widehat S_t$.
\end{proof}


\begin{styledcorollary}{Surprise--ELBO Bound for General $\beta$}{app:surprise_elbo_general_beta}

Under the hypotheses of Theorem~\ref{thm:surprise_elbo_relation}, for arbitrary
$\beta>0$,
\[
\mathcal L_{t,\beta}
\leq
-\widehat S_t
+(1-\beta)D_{\mathrm{KL}}(q_t\|\kappa_t).
\]
For $\beta\geq1$ this reduces to
\[
\mathcal L_{t,\beta}\leq-\widehat S_t.
\]
Moreover, the exact identity is
\[
\mathcal L_{t,\beta}
=-\widehat S_t
-D_{\mathrm{KL}}\left(q_t\middle\|p_\theta(z_t\mid y_t,\mathcal F_t^{-,y})\right)
-(\beta-1)D_{\mathrm{KL}}(q_t\|\kappa_t).
\]
Equality in the first displayed upper bound occurs exactly when
$q_t=p_\theta(z_t\mid y_t,\mathcal F_t^{-,y})$ almost everywhere.
For $\beta=1$, under the same posterior-equality condition, this gives
$\mathcal L_{t,1}=-\widehat S_t$.
For $\beta>1$, equality in the stronger bound
$\mathcal L_{t,\beta}\leq-\widehat S_t$ additionally requires
$D_{\mathrm{KL}}(q_t\|\kappa_t)=0$.

\end{styledcorollary}

\begin{proof}
From the exact identity in Theorem~\ref{thm:surprise_elbo_relation},
non-negativity of the posterior KL divergence gives
\[
\mathcal L_{t,\beta}
\leq
-\widehat S_t
+(1-\beta)D_{\mathrm{KL}}(q_t\|\kappa_t).
\]
Equality in this first upper bound occurs exactly when
\(q_t=p_\theta(z_t\mid y_t,\mathcal F_t^{-,y})\) almost everywhere.
For $\beta=1$, the propagated-prior KL term vanishes, so this condition
also gives $\mathcal L_{t,1}=-\widehat S_t$. For $\beta>1$, obtaining
equality in the stronger bound $\mathcal L_{t,\beta}\leq-\widehat S_t$
additionally requires $(\beta-1)D_{\mathrm{KL}}(q_t\|\kappa_t)=0$,
and hence $D_{\mathrm{KL}}(q_t\|\kappa_t)=0$.
\end{proof}


\begin{styledcorollary}{Asymptotic Bound on the Rate of Surprise-Flagged Interactions}{app:surprise_flagging_rate}

Let $\widehat S_t = -\log p_{\mathrm{pred}}(y_t\mid y_{1:t-1},c_{1:t-1})$ denote the
operationally-computed, per-interaction surprise of
Eq.~\eqref{eq:surprisenll} (as opposed to the population-level $S_t^{\mathrm{(KL)}}$ of
Theorem~\ref{thm:predictive_evidence_bound}), and suppose
$\widehat S_t\ge0$ almost surely for every $t$ (which holds, in particular,
whenever the space of interactions is discrete, so that
$p_{\mathrm{pred}}(\cdot\mid\cdot)$ is a probability mass function).
Under the hypotheses of Theorem~\ref{thm:predictive_evidence_bound}
and Theorem~\ref{thm:surprise_elbo_relation} at $\beta=1$, and assuming
$\widehat S_t\geq0$ almost surely, for every
threshold $\tau>0$,
\(
\Pr\!\left(\widehat S_t\ge\tau\right)
\;\le\;
\frac{U_t+H^*}{\tau},
\text{ and hence }
\limsup_{t\to\infty}\Pr\!\left(\widehat S_t\ge\tau\right)
\;\le\;
\frac{U_\infty+H^*}{\tau}.
\)
That is, for any fixed Surprise-flagging threshold used to allocate
additional diagnostic, verification, or adaptation effort, the
long-run flagging rate is asymptotically bounded. The bound tightens with the predictive-ELBO bound
$U_t$, but an irreducible contribution from the true interaction entropy $H^*$
remains in the general stochastic case.

\end{styledcorollary}

\begin{proof}
By Theorem~\ref{thm:surprise_elbo_relation} at $\beta=1$,
$\mathcal{L}_t(\theta_t,\phi_t)\le -\widehat S_t$ pointwise. Taking
expectations and using the definition of $U_t$,
\[
\mathbb{E}_{p_{\mathrm{true}}}[\widehat S_t]
\le
-\mathbb{E}_{p_{\mathrm{true}}}[\mathcal{L}_t(\theta_t,\phi_t)]
=
U_t+H^*.
\]
Since $\widehat S_t\ge0$ a.s., Markov's inequality gives
$\Pr(\widehat S_t\ge\tau)\le \mathbb{E}_{p_{\mathrm{true}}}[\widehat S_t]/\tau\le(U_t+H^*)/\tau$.
Taking $\limsup_{t\to\infty}$ and using $U_t\to U_\infty$
(Theorem~\ref{thm:predictive_evidence_bound}) gives the result.
\end{proof}


\begin{styledtheorem}{Temporal Scale Separation under Linear-Gaussian Dynamics}{app:scale_separation}

Suppose that the fast interaction state and each slow component-risk
state follow stationary scalar linear-Gaussian autoregressive dynamics,
\(
z_{t+1}^{(F)}
=
\alpha_F z_t^{(F)}
+
\varepsilon_{t+1}^{(F)},
\)
and
\(
z_{t+1}^{(j)}
=
\alpha_j z_t^{(j)}
+
\varepsilon_{t+1}^{(j)}\),
\(
j\in\{D,H,M\},
\)
where
\(
|\alpha_F|<|\alpha_j|<1
\)
for every $j\in\{D,H,M\}$. Assume that the innovation processes
$\{\varepsilon_t^{(F)}\}$ and $\{\varepsilon_t^{(j)}\}$ have zero mean,
strictly positive finite variance for each component, are Gaussian, and are serially uncorrelated over time,
i.e.,
\(
\operatorname{Cov}
\!\left(
\varepsilon_s^{(\ell)},
\varepsilon_u^{(\ell)}
\right)
=
0\),
\(
s\neq u,
\)
for $\ell\in\{F,D,H,M\}$.

Then the autocorrelation functions satisfy
\(
r_F(k)=\alpha_F^k\),
\(
r_j(k)=\alpha_j^k,
\)
for every $k\geq0$ (a zeroth power is one, including when the base is zero), where $r_F$ and $r_j$ denote the
autocorrelation functions of $\{z_t^{(F)}\}$ and
$\{z_t^{(j)}\}$, respectively. Consequently, the magnitude of the
autocorrelation of the fast interaction state decays more rapidly than
that of every slow component-risk state. In particular, the
characteristic exponential correlation-decay times (for nonzero coefficients)
\(
\tau_F
=
-\frac{1}{\log|\alpha_F|}
\)
and
\(
\tau_j
=
-\frac{1}{\log|\alpha_j|}
\)
satisfy
\(
\tau_F<\tau_j\),
\(
j\in\{D,H,M\}.
\)
If $\alpha_F=0$, define $\tau_F:=0$ (the limiting decay time); the fast-state autocorrelation vanishes at every positive lag, and the same strict ordering holds. Thus, the fast interaction state evolves on a shorter characteristic
temporal scale than each component-risk state.

\end{styledtheorem}

\begin{proof}
Consider first a generic stationary scalar AR(1) process
\[
z_{t+1}
=
\alpha z_t+\varepsilon_{t+1},
\qquad
|\alpha|<1,
\]
where the innovation sequence $\{\varepsilon_t\}$ has zero mean,
strictly positive finite variance $\sigma_\varepsilon^2$, and is serially uncorrelated. A zeroth power is one, including when the base is zero.
The stationary solution has the causal representation
\[
z_t
=
\sum_{m=0}^{\infty}
\alpha^m\varepsilon_{t-m}.
\]

Let
\[
\gamma(k)
=
\operatorname{Cov}(z_{t+k},z_t)
\]
denote the autocovariance function. For $k\geq1$,
\[
\begin{aligned}
\gamma(k)
&=
\operatorname{Cov}
\!\left(
\alpha z_{t+k-1}+\varepsilon_{t+k},
z_t
\right)\\
&=
\alpha\gamma(k-1)
+
\operatorname{Cov}
\!\left(
\varepsilon_{t+k},z_t
\right).
\end{aligned}
\]
Using the causal representation of $z_t$,
\[
\begin{aligned}
\operatorname{Cov}
\!\left(
\varepsilon_{t+k},z_t
\right)
&=
\operatorname{Cov}
\!\left(
\varepsilon_{t+k},
\sum_{m=0}^{\infty}
\alpha^m\varepsilon_{t-m}
\right)\\
&=
\sum_{m=0}^{\infty}
\alpha^m
\operatorname{Cov}
\!\left(
\varepsilon_{t+k},
\varepsilon_{t-m}
\right)
=
0,
\end{aligned}
\]
because $t+k\neq t-m$ for every $m\geq0$ and the innovations are
serially uncorrelated. Hence,
\[
\gamma(k)=\alpha\gamma(k-1).
\]
Iterating this recursion gives
\[
\gamma(k)=\alpha^k\gamma(0).
\]
Since
\[
\gamma(0)
=
\operatorname{Var}(z_t)
=
\frac{\sigma_\varepsilon^2}{1-\alpha^2},
\]
the autocorrelation function is
\[
r(k)
=
\frac{\gamma(k)}{\gamma(0)}
=
\alpha^k.
\]

Applying this result to the fast interaction state and to each
component-risk state yields
\[
r_F(k)=\alpha_F^k
\]
and
\[
r_j(k)=\alpha_j^k,
\qquad
j\in\{D,H,M\}.
\]
Therefore,
\[
|r_F(k)|=|\alpha_F|^k
\]
and
\[
|r_j(k)|=|\alpha_j|^k.
\]
Because
\[
|\alpha_F|<|\alpha_j|<1,
\]
the magnitude of the autocorrelation of the fast interaction state
decays exponentially more rapidly with lag than that of every slow
component-risk state.

If $\alpha_F=0$, the fast-state autocorrelation is zero at every positive lag, so its decay time is defined as zero, while every slow-state decay time is strictly positive and finite. This proves the strict ordering in that case. Otherwise, the corresponding exponential correlation-decay times are
\[
\tau_F
=
-\frac{1}{\log|\alpha_F|}
\]
and
\[
\tau_j
=
-\frac{1}{\log|\alpha_j|}.
\]
Since
\[
|\alpha_F|<|\alpha_j|<1,
\]
we have
\[
-\log|\alpha_F|
>
-\log|\alpha_j|
>
0.
\]
Taking reciprocals of these positive quantities reverses the
inequality, giving
\[
\tau_F<\tau_j,
\qquad
j\in\{D,H,M\}.
\]
Thus, the fast interaction state has a shorter characteristic
correlation timescale than each component-risk state.
\end{proof}


\begin{styledtheorem}{Asymptotic Resilience to Prior Miscalibration}{app:prior_forgetting}

Let $q_t(z_t\mid y_{1:t},c_{1:t-1})$ and $\tilde q_t(z_t\mid y_{1:t},c_{1:t-1})$
denote the inferred joint posteriors at time $t$ under the SELP cycle,
initialized with two different joint priors $\kappa_1(z_1)$ and
$\tilde\kappa_1(z_1)$, respectively. Let $\kappa_t$ and $\tilde\kappa_t$
denote the corresponding predictive priors at time $t$.

Suppose that:
\begin{enumerate}[nosep]
    \item \textbf{Contractive transition:} The joint latent-state transition
    operator $P_t$ induced by $p_\theta(z_{t+1}\mid z_t,y_t,c_t)$ is uniformly
    contractive in the Wasserstein distance $\mathcal W$: there exists a
    constant $\gamma\in[0,1)$ such that, for every $t$ and every pair of
    probability distributions $\mu,\nu$ on the latent-state space,
    $\mathcal W(P_t\mu,P_t\nu)\leq\gamma\mathcal W(\mu,\nu)$.

    \item \textbf{Non-expansive measurement update:} For the common observation
    $y_t$, the Bayesian measurement-update operator induced by the likelihood
    $p_\theta(y_t\mid z_t)$ is non-expansive in the Wasserstein distance, i.e.,
    $\mathcal W(q_t,\tilde q_t)\leq\mathcal W(\kappa_t,\tilde\kappa_t)$ for every $t$.

    \item \textbf{Finite initial Wasserstein distance:}
    $\mathcal W(q_1,\tilde q_1)<\infty$.
\end{enumerate}

Then, under an identical stream of observable interactions and regulatory actions and the same model parameters in both runs,
\[
\mathcal W(q_t,\tilde q_t)\leq
\mathcal W(q_1,\tilde q_1)\,\gamma^{t-1},\qquad t\geq1.
\]
A zeroth power is one, including when the base is zero. Consequently, $\mathcal W(q_t,\tilde q_t)\xrightarrow[t\to\infty]{}0$.
Thus, the effect of the initial prior specification vanishes geometrically fast,
guaranteeing asymptotic resilience to prior miscalibration.

\end{styledtheorem}

\begin{proof}
We proceed by induction on $t$. For the base case $t=1$, the stated bound holds
with equality since $\gamma^0=1$.

Assume the bound holds for some $t\geq1$. At time $t+1$, the predictive priors
are obtained by applying the latent-state transition operator to the respective
posteriors from time $t$: $\kappa_{t+1}=P_tq_t$ and
$\tilde\kappa_{t+1}=P_t\tilde q_t$. By the uniform contractivity assumption,
\[
\mathcal W(\kappa_{t+1},\tilde\kappa_{t+1})
=\mathcal W(P_tq_t,P_t\tilde q_t)
\leq\gamma\mathcal W(q_t,\tilde q_t).
\]
Next, applying the non-expansive measurement-update assumption for the shared
observation $y_{t+1}$ gives
\[
\mathcal W(q_{t+1},\tilde q_{t+1})
\leq\mathcal W(\kappa_{t+1},\tilde\kappa_{t+1}).
\]
Combining these inequalities,
\[
\mathcal W(q_{t+1},\tilde q_{t+1})
\leq\gamma\mathcal W(q_t,\tilde q_t).
\]
Substituting the inductive hypothesis yields
\[
\mathcal W(q_{t+1},\tilde q_{t+1})
\leq\gamma\left(\mathcal W(q_1,\tilde q_1)\gamma^{t-1}\right)
=\mathcal W(q_1,\tilde q_1)\gamma^t.
\]
This completes the induction. Finally, because $\gamma\in[0,1)$ and the initial
distance is finite, $\gamma^{t-1}\to0$, and hence
$\mathcal W(q_t,\tilde q_t)\to0$.
\end{proof}

\noindent\textbf{Remark [Scope of the non-expansive measurement-update assumption].}
The non-expansiveness condition in Assumption~2 is a substantive
stability condition and should not be interpreted as a generic property
of Bayesian conditioning. In particular, even in multivariate
linear-Gaussian state-space models, conditioning on a common,
partially informative observation can increase the Wasserstein distance
between posterior distributions when the prior covariance contains
correlations across observed and unobserved directions. Thus, the
contractive transition in Assumption~1 does not by itself guarantee the
one-step posterior contraction required by the theorem.
To see why the scalar case can satisfy this condition automatically for
a useful class of prior differences, consider a common linear-Gaussian
observation $y=wz+\text{noise}$, with noise $\sim\mathcal N(0,\omega)$,
applied to two scalar priors $\mathcal N(\mu,\Sigma)$ and
$\mathcal N(\tilde\mu,\Sigma)$ that share the same variance $\Sigma$
but differ in mean. Since the priors share their variance,
$\mathcal W(\mathcal N(\mu,\Sigma),\mathcal N(\tilde\mu,\Sigma))=|\mu-\tilde\mu|$,
and the posteriors also share a variance. Their posterior means are
$\mu+K(y-w\mu)$ and $\tilde\mu+K(y-w\tilde\mu)$, where
$K=w\Sigma/(w^2\Sigma+\omega)$ is the shared Kalman gain. Hence the
posterior mean gap, and therefore the posterior Wasserstein distance,
contracts by the factor
\[
\frac{\omega}{w^2\Sigma+\omega}\in(0,1].
\]
Thus the scalar case provides a natural setting in which the
measurement-update condition is automatically satisfied for differences
in prior means with a common prior variance.

For higher-dimensional systems, non-expansiveness can instead be
established under additional structural conditions on the observation
operator and prior covariance, for example when the observation
directions are aligned with invariant eigenspaces of the covariance or
when the state and observation models have a compatible
component-wise or block-diagonal structure. Such conditions must be
verified for the particular CENTRY state-space model; they do not follow
from linear-Gaussianity alone.

Accordingly, Theorem~\ref{thm:prior_forgetting} should be interpreted as
a conditional resilience guarantee: whenever the latent transition is
uniformly contractive and the common-observation Bayesian update is
non-expansive in the chosen Wasserstein metric, initial prior
miscalibration is forgotten exponentially fast.


\begin{styledtheorem}{Consistency of Component Risk Ratings}{app:risk_rating_consistency}

Let $R_t^{(j)}$ denote the true component risk associated with component
$j\in\{D,H,M\}$, and let
\[
\rho_t^{(j)}
=
g^{(j)}\!\left(\mu_t^{(j)},\Sigma_t^{(j)}\right)
\]
be CENTRY's component risk rating at time $t$, where $g^{(j)}$ is the risk
functional of Eq.~\eqref{eq:riskfunctional}. Suppose that, for each
$j\in\{D,H,M\}$:

\begin{enumerate}[nosep]
\item \textbf{Risk-state identifiability:} There exists a continuous
function $h_j$ such that
\[
R_t^{(j)}
=
h_j\!\left(z_t^{(j),*}\right),
\]
where $z_t^{(j),*}$ denotes the true latent state of component $j$.
Moreover, $h_j$ is injective on the relevant support of the latent
state.

\item \textbf{Posterior mean consistency:} The variational posterior
mean asymptotically recovers the true latent state, in the sense that
\[
\mu_t^{(j)}-z_t^{(j),*}
\xrightarrow{p}
0
\qquad\text{as }t\to\infty.
\]

\item \textbf{Posterior covariance stabilization:} The variational
posterior covariance converges in probability to a limiting covariance
matrix,
\[
\Sigma_t^{(j)}
\xrightarrow{p}
\Sigma_\infty^{(j)},
\]
where $\Sigma_\infty^{(j)}$ need not be zero.

\item \textbf{Uniform continuity of the risk functional:} There exists
a fixed compact set $K_j$ such that
\[
z_t^{(j),*}\in K_j
\qquad\text{almost surely for all }t,
\]
and there exists an open neighborhood $\mathcal U_j$ of
\[
K_j\times\{\Sigma_\infty^{(j)}\}
\]
on which $g^{(j)}$ is uniformly continuous. Furthermore,
\[
g^{(j)}\!\left(z^{(j)},0\right)
=
h_j\!\left(z^{(j)}\right)
\]
for every $z^{(j)}$ in the relevant support of the latent state.
\end{enumerate}

Then, for every component $j\in\{D,H,M\}$,
\[
\rho_t^{(j)}
-
g^{(j)}\!\left(z_t^{(j),*},\Sigma_\infty^{(j)}\right)
\xrightarrow{p}
0
\qquad\text{as }t\to\infty.
\]

Equivalently,
\[
\rho_t^{(j)}
=
g^{(j)}\!\left(z_t^{(j),*},\Sigma_\infty^{(j)}\right)
+
o_p(1).
\]

Consequently, the difference between the CENTRY component risk rating
and the true component risk satisfies
\[
\rho_t^{(j)}-R_t^{(j)}
=
g^{(j)}\!\left(z_t^{(j),*},\Sigma_\infty^{(j)}\right)
-
h_j\!\left(z_t^{(j),*}\right)
+
o_p(1).
\]

Thus, exact risk-rating consistency,
\[
\rho_t^{(j)}-R_t^{(j)}
\xrightarrow{p}
0,
\]
is obtained whenever
\[
g^{(j)}\!\left(z,\Sigma_\infty^{(j)}\right)
=
h_j(z)
\]
throughout the relevant support of the latent state. In particular, exact
consistency holds if $\Sigma_\infty^{(j)}=0$, because assumption~4 then gives
$g^{(j)}(z,\Sigma_\infty^{(j)})=g^{(j)}(z,0)=h_j(z)$.

\end{styledtheorem}

\begin{proof}
Fix a component $j\in\{D,H,M\}$. We first establish
\[
g^{(j)}\!\left(\mu_t^{(j)},\Sigma_t^{(j)}\right)
-
g^{(j)}\!\left(z_t^{(j),*},\Sigma_\infty^{(j)}\right)
\xrightarrow{p}
0.
\]

Let
\[
\mathcal K_j
=
K_j\times\{\Sigma_\infty^{(j)}\}.
\]
By assumption~4, $\mathcal K_j$ is compact and is contained in the
open set $\mathcal U_j$. Therefore, there exists $\delta_0>0$ such that
every point whose distance from $\mathcal K_j$ is less than $\delta_0$
belongs to $\mathcal U_j$.

Since $g^{(j)}$ is uniformly continuous on $\mathcal U_j$, for every
$\varepsilon>0$ there exists $0<\delta<\delta_0$ such that, for any
$(u,S),(v,T)\in\mathcal U_j$,
\[
\|u-v\|+\|S-T\|<\delta
\quad\Longrightarrow\quad
\left|
g^{(j)}(u,S)-g^{(j)}(v,T)
\right|
<\varepsilon.
\]

Because
\[
z_t^{(j),*}\in K_j
\qquad\text{almost surely for all }t,
\]
and because
\[
\mu_t^{(j)}-z_t^{(j),*}\xrightarrow{p}0,
\qquad
\Sigma_t^{(j)}-\Sigma_\infty^{(j)}\xrightarrow{p}0,
\]
we have
\[
\Pr\!\left(
\|\mu_t^{(j)}-z_t^{(j),*}\|
+
\|\Sigma_t^{(j)}-\Sigma_\infty^{(j)}\|
\ge\delta
\right)
\longrightarrow0.
\]
Indeed, by the union bound,
\[
\begin{aligned}
&
\Pr\!\left(
\|\mu_t^{(j)}-z_t^{(j),*}\|
+
\|\Sigma_t^{(j)}-\Sigma_\infty^{(j)}\|
\ge\delta
\right)
\\
&\qquad\le
\Pr\!\left(
\|\mu_t^{(j)}-z_t^{(j),*}\|
\ge\frac{\delta}{2}
\right)
+
\Pr\!\left(
\|\Sigma_t^{(j)}-\Sigma_\infty^{(j)}\|
\ge\frac{\delta}{2}
\right),
\end{aligned}
\]
and both terms converge to zero.

Define the event
\[
A_t(\delta)
=
\left\{
\|\mu_t^{(j)}-z_t^{(j),*}\|
+
\|\Sigma_t^{(j)}-\Sigma_\infty^{(j)}\|
<\delta
\right\}.
\]
On $A_t(\delta)$, the point
\[
\left(\mu_t^{(j)},\Sigma_t^{(j)}\right)
\]
lies within distance $\delta<\delta_0$ of
\[
\left(z_t^{(j),*},\Sigma_\infty^{(j)}\right)
\in\mathcal K_j.
\]
Hence
\[
\left(\mu_t^{(j)},\Sigma_t^{(j)}\right)\in\mathcal U_j.
\]
Moreover, both points belong to $\mathcal U_j$, and their distance is
less than $\delta$. Uniform continuity therefore gives
\[
\left|
g^{(j)}\!\left(\mu_t^{(j)},\Sigma_t^{(j)}\right)
-
g^{(j)}\!\left(z_t^{(j),*},\Sigma_\infty^{(j)}\right)
\right|
<\varepsilon
\]
on $A_t(\delta)$.

Consequently,
\[
\begin{aligned}
&
\Pr\!\left(
\left|
g^{(j)}\!\left(\mu_t^{(j)},\Sigma_t^{(j)}\right)
-
g^{(j)}\!\left(z_t^{(j),*},\Sigma_\infty^{(j)}\right)
\right|
\ge\varepsilon
\right)
\\
&\qquad\le
\Pr\!\left(A_t(\delta)^c\right)
\\
&\qquad=
\Pr\!\left(
\|\mu_t^{(j)}-z_t^{(j),*}\|
+
\|\Sigma_t^{(j)}-\Sigma_\infty^{(j)}\|
\ge\delta
\right)
\longrightarrow0.
\end{aligned}
\]
Since $\varepsilon>0$ was arbitrary,
\[
g^{(j)}\!\left(\mu_t^{(j)},\Sigma_t^{(j)}\right)
-
g^{(j)}\!\left(z_t^{(j),*},\Sigma_\infty^{(j)}\right)
\xrightarrow{p}
0.
\]

Using
\[
\rho_t^{(j)}
=
g^{(j)}\!\left(\mu_t^{(j)},\Sigma_t^{(j)}\right),
\]
we obtain
\[
\rho_t^{(j)}
-
g^{(j)}\!\left(z_t^{(j),*},\Sigma_\infty^{(j)}\right)
\xrightarrow{p}
0,
\]
which proves the first claim and, equivalently,
\[
\rho_t^{(j)}
=
g^{(j)}\!\left(z_t^{(j),*},\Sigma_\infty^{(j)}\right)
+
o_p(1).
\]

Next, assumption~1 gives
\[
R_t^{(j)}
=
h_j\!\left(z_t^{(j),*}\right).
\]
Therefore,
\[
\begin{aligned}
\rho_t^{(j)}-R_t^{(j)}
&=
\rho_t^{(j)}
-
g^{(j)}\!\left(z_t^{(j),*},\Sigma_\infty^{(j)}\right)
\\
&\quad+
g^{(j)}\!\left(z_t^{(j),*},\Sigma_\infty^{(j)}\right)
-
h_j\!\left(z_t^{(j),*}\right).
\end{aligned}
\]
The first term converges to zero in probability, and hence
\[
\rho_t^{(j)}-R_t^{(j)}
=
g^{(j)}\!\left(z_t^{(j),*},\Sigma_\infty^{(j)}\right)
-
h_j\!\left(z_t^{(j),*}\right)
+
o_p(1).
\]

If
\[
g^{(j)}\!\left(z,\Sigma_\infty^{(j)}\right)=h_j(z)
\]
throughout the relevant support of the latent state, the remaining
state-dependent term is identically zero. Consequently,
\[
\rho_t^{(j)}-R_t^{(j)}
\xrightarrow{p}
0.
\]

In particular, if $\Sigma_\infty^{(j)}=0$, assumption~4 implies
\[
g^{(j)}\!\left(z_t^{(j),*},\Sigma_\infty^{(j)}\right)
=
g^{(j)}\!\left(z_t^{(j),*},0\right)
=
h_j\!\left(z_t^{(j),*}\right)
=
R_t^{(j)},
\]
and therefore
\[
\rho_t^{(j)}-R_t^{(j)}
\xrightarrow{p}
0.
\]

Since $j$ was arbitrary, the result holds for every
$j\in\{D,H,M\}$.
\end{proof}

\noindent\textbf{Remark [Posterior moment consistency and the uncertainty floor].}
For a time-varying latent state, posterior mean consistency and posterior
covariance stabilization are filtering properties rather than consequences
of Doob's posterior consistency theorem, which concerns static unknown
parameters. In particular, prior forgetting, as characterized by
Theorem~\ref{thm:prior_forgetting}, does not imply
$\Sigma_t^{(j)}\to0$. Prior forgetting concerns the asymptotic loss of
dependence of the filtering distribution on its initial prior; it does not
assert that the posterior distribution collapses to a point.

For a continually evolving, noise-driven state-space system, the natural
asymptotic regime is instead
\[
\Sigma_t^{(j)}
\longrightarrow
\Sigma_\infty^{(j)},
\]
where $\Sigma_\infty^{(j)}$ may be strictly positive. In the
linear-Gaussian setting, Theorem~\ref{thm:steady_state_riccati} establishes
such a steady-state posterior covariance under the usual detectability and
stabilizability conditions. Thus, prior forgetting and Riccati convergence
are compatible: the former describes forgetting of the initial prior,
whereas the latter characterizes the limiting posterior uncertainty.

\noindent\textbf{Remark [Uniformity over the latent-state trajectory].}
The compactness and uniform-continuity condition in assumption~4 is a
sufficient condition that makes the convergence argument valid despite
the fact that the evaluation point
$z_t^{(j),*}$ may change with $t$. Pointwise continuity of $g^{(j)}$ at each
fixed point
$(z,\Sigma_\infty^{(j)})$ would not, by itself, be sufficient, because
the sequence of true latent states may move through the state space and
encounter arbitrarily poor local moduli of continuity. Confining the
trajectory to the fixed compact set $K_j$ and requiring uniform
continuity on a neighborhood of
$K_j\times\{\Sigma_\infty^{(j)}\}$ provides the required uniform control.

A common sufficient condition is that $g^{(j)}$ be continuous on a compact
neighborhood of
$K_j\times\{\Sigma_\infty^{(j)}\}$; uniform continuity then follows from
the Heine--Cantor theorem.

\noindent\textbf{Remark [Interpretation of consistency].}
The theorem does not, in general, assert that the component risk rating
converges exactly to the true component risk when
$\Sigma_\infty^{(j)}\neq0$. Instead, it establishes the sharper statement
that the CENTRY rating asymptotically evaluates the risk functional at the
true latent state and the limiting posterior uncertainty:
\[
\rho_t^{(j)}
=
g^{(j)}\!\left(z_t^{(j),*},\Sigma_\infty^{(j)}\right)
+
o_p(1).
\]
Thus, any persistent asymptotic discrepancy from the true risk is
attributable to the residual uncertainty represented by
$\Sigma_\infty^{(j)}$. Exact risk-rating consistency requires that this
limiting uncertainty have no effect on the risk functional, namely
\[
g^{(j)}\!\left(z,\Sigma_\infty^{(j)}\right)=h_j(z)
\]
throughout the relevant support. The special case
$\Sigma_\infty^{(j)}=0$ guarantees this condition through assumption~4.


The following theorem formalizes a calibration bridge from CENTRY's
component risk ratings to the corresponding true operational-risk budgets.

\begin{styledtheorem}{Robust Transfer from Rating-Based to Operational-Risk Safety}{app:rating_to_operational_safety}

Fix a component $j\in\{D,H,M\}$ and a future time $t+k$. Let
$R_{t+k}^{(j)}$ be the true operational component risk and let
$\rho_{t+k}^{(j)}$ be CENTRY's component risk rating. Let $B^{(j)}$ be
the operational risk budget and $\bar{B}^{(j)}$ the corresponding rating
threshold. Suppose that, conditional on $\mathcal F_t$,
\[
\Pr\left(R_{t+k}^{(j)}>B^{(j)},
\rho_{t+k}^{(j)}\leq\bar{B}^{(j)} \middle|\mathcal F_t\right)
\leq\eta_k^{(j)}
\]
and
\[
\Pr\left(\rho_{t+k}^{(j)}>\bar{B}^{(j)} \middle|\mathcal F_t\right)
\leq\alpha_k^{(j)}.
\]
Then
\[
\Pr\left(R_{t+k}^{(j)}>B^{(j)} \middle|\mathcal F_t\right)
\leq\eta_k^{(j)}+\alpha_k^{(j)}.
\]
Consequently, if the regulatory policy controls the rating-violation
probability such that $\alpha_k^{(j)} \leq \delta_k^{(j)}-\eta_k^{(j)}$
for an allowable tolerance $\delta_k^{(j)}\geq\eta_k^{(j)}$, then the
true operational risk satisfies the chance constraint
\[
\Pr\left(R_{t+k}^{(j)}>B^{(j)} \middle|\mathcal F_t\right)
\leq\delta_k^{(j)}.
\]
Moreover, if this condition holds for all $j\in\{D,H,M\}$ and
$k=1,\ldots,T$, and
\[
\sum_{k=1}^{T}\sum_{j\in\{D,H,M\}}\delta_k^{(j)}\leq\delta,
\]
then
\[
\Pr\left(
R_{t+k}^{(j)}\leq B^{(j)},
\;\forall j\in\{D,H,M\},
\;\forall k=1,\ldots,T
\middle|\mathcal F_t\right)
\geq1-\delta.
\]
Thus a rating-based safety rule transfers to a true operational-risk
safety probability whenever the probability of rating underestimation and
the probability of rating exceedance are separately controlled.

\end{styledtheorem}

\begin{proof}
Define the operational budget violation event
$A=\left\{R_{t+k}^{(j)}>B^{(j)}\right\}$ and the rating threshold violation event
$B=\left\{\rho_{t+k}^{(j)}>\bar{B}^{(j)}\right\}$. By partitioning $A$ into
$A\cap B$ and $A\cap B^c$,
\[
\Pr(A\mid\mathcal F_t)
=
\Pr(A\cap B\mid\mathcal F_t) + \Pr(A\cap B^c\mid\mathcal F_t).
\]
Since $A\cap B \subseteq B$, we have $\Pr(A\cap B\mid\mathcal F_t) \le \Pr(B\mid\mathcal F_t) \leq \alpha_k^{(j)}$.
By the first condition,
\[
\Pr(A\cap B^c\mid\mathcal F_t)
=
\Pr\!\left(R_{t+k}^{(j)}>B^{(j)}, \, \rho_{t+k}^{(j)}\le\bar{B}^{(j)} \,\middle|\, \mathcal F_t\right)
\le
\eta_k^{(j)}.
\]
Summing these two bounds gives
\[
\Pr\!\left(R_{t+k}^{(j)}>B^{(j)} \,\middle|\, \mathcal F_t\right)
\le
\eta_k^{(j)} + \alpha_k^{(j)}.
\]
If the regulatory policy controls the rating violation rate such that $\alpha_k^{(j)} \leq \delta_k^{(j)} - \eta_k^{(j)}$, then
\[
\Pr\!\left(R_{t+k}^{(j)}>B^{(j)} \,\middle|\, \mathcal F_t\right)
\leq
\delta_k^{(j)}.
\]
Finally, applying the conditional union bound over all components $j\in\{D,H,M\}$ and horizon steps $k=1,\ldots,T$ (analogous to Theorem~\ref{thm:finite_horizon_risk_guarantee}) yields:
\[
\Pr\!\left(\bigcup_{k=1}^{T}\bigcup_{j\in\{D,H,M\}}\left\{R_{t+k}^{(j)}>B^{(j)}\right\} \,\middle|\, \mathcal F_t\right)
\le
\sum_{k=1}^{T}\sum_{j\in\{D,H,M\}}\delta_k^{(j)}
\leq
\delta.
\]
Taking the complement event proves that the true joint operational risk remains bounded by $B^{(j)}$ with conditional probability at least $1-\delta$.
\end{proof}

\noindent\textbf{Remark [Interpretation].}
This theorem is a decomposition-based calibration bridge. The first term
$\eta_k^{(j)}$ controls dangerous underestimation by the rating, while
$\alpha_k^{(j)}$ controls the probability that the rating itself exceeds
its threshold. It is complementary to the one-sided probability calibration
condition in Theorem~\ref{thm:finite_horizon_risk_guarantee}; the two
formulations may be used according to which calibration quantities are
directly estimable in an implementation.


\begin{styledtheorem}{Separation of Component Risk Contributions}{app:risk_decomposition_separation}

Let $R_t=R(\rho_t)$ denote the system-level expected risk at time $t$ as a
function of the component risk ratings
\(
\rho_t=
\left(
\rho_t^{(D)},\rho_t^{(H)},\rho_t^{(M)}
\right)
\in\mathbb{R}_{\geq 0}^3.
\)
Suppose that $R$ is continuously differentiable on the nonnegative orthant
(with derivatives at boundary points understood as continuous extensions
of the corresponding interior derivatives) and positively homogeneous
of degree one in the component ratings, i.e.,
\(
R(\lambda\rho)=\lambda R(\rho)\),
\(
\forall \lambda>0.
\)
Then the exact additive decomposition
\(
R(\rho_t)
=
\sum_{j\in\{D,H,M\}}
\frac{\partial R}{\partial \rho^{(j)}}(\rho_t)\,
\rho_t^{(j)}
\)
holds. 

If $R(\rho_t)>0$, define the component risk-contribution shares by
\(
s_t^{(j)}
=
\frac{
\frac{\partial R}{\partial \rho^{(j)}}(\rho_t)\,
\rho_t^{(j)}
}{
R(\rho_t)
}\),
\(
j\in\{D,H,M\}.
\)
Then
\(
\sum_{j\in\{D,H,M\}}s_t^{(j)}=1.
\)

If, in addition, $R$ is non-decreasing in each component rating, so that
\(
\frac{\partial R}{\partial\rho^{(j)}}(\rho)\geq 0\),
\(
j\in\{D,H,M\},
\)
then
\(
s_t^{(j)}\geq 0
\)
for all $j$.

Moreover, if $R$ is twice continuously differentiable (with the same
boundary convention for second derivatives) and submodular in
the component ratings, then, for every $i\neq j$,
\(
\frac{\partial^2 R}
{\partial\rho^{(i)}\partial\rho^{(j)}}(\rho_t)
\leq 0,
\)
so that the component ratings are weakly substitutable rather than
complementary in their effect on system-level risk.

\end{styledtheorem}

\begin{proof}
Because $R$ is differentiable and positively homogeneous of degree one,
Euler's homogeneous-function theorem applies in the interior of the
nonnegative orthant, and the identity extends to its boundary by
continuity of $R$ and its first derivatives. Thus,
\[
\sum_{j\in\{D,H,M\}}
\frac{\partial R}{\partial\rho^{(j)}}(\rho_t)\,
\rho_t^{(j)}
=
R(\rho_t).
\]
Thus,
\[
R(\rho_t)
=
\sum_{j\in\{D,H,M\}}
\frac{\partial R}{\partial\rho^{(j)}}(\rho_t)\,
\rho_t^{(j)},
\]
which establishes the exact additive decomposition.

Assume now that $R(\rho_t)>0$. Dividing the decomposition by
$R(\rho_t)$ gives
\[
\sum_{j\in\{D,H,M\}}
\frac{
\frac{\partial R}{\partial\rho^{(j)}}(\rho_t)\,
\rho_t^{(j)}
}{
R(\rho_t)
}
=1.
\]
By the definition of $s_t^{(j)}$, this is precisely
\[
\sum_{j\in\{D,H,M\}}s_t^{(j)}=1.
\]
Furthermore, since $\rho_t^{(j)}\geq0$ and
\[
\frac{\partial R}{\partial\rho^{(j)}}(\rho_t)\geq0,
\]
each numerator in the definition of $s_t^{(j)}$ is non-negative.
Because $R(\rho_t)>0$, it follows that
\[
s_t^{(j)}\geq0
\qquad
\text{for all }j.
\]
Hence the quantities $s_t^{(j)}$ form a valid set of non-negative
component risk-contribution shares summing to one.

Finally, suppose that $R$ is twice continuously differentiable and
submodular in the component ratings. For a twice continuously
differentiable function on the relevant domain, submodularity implies
that its mixed second partial derivatives satisfy (also at boundary
points by continuity)
\[
\frac{\partial^2 R}
{\partial\rho^{(i)}\partial\rho^{(j)}}(\rho_t)
\leq0,
\qquad i\neq j.
\]
Thus, an increase in one component rating does not increase the marginal
effect of another component rating on system-level risk. The component
ratings are therefore weakly substitutable rather than complementary in
their contribution to system-level risk. This completes the proof.
\end{proof}


\begin{styledtheorem}{Component Identifiability under Local Fisher-Information Distinguishability}{app:component_identifiability}

Fix a time $t$ and let
\(
\mathcal{J}=\{F,D,H,M\}
\)
denote the latent components. Write
\(
z_t
=
\bigl(
z_t^{(j)}
\bigr)_{j\in\mathcal{J}}\),
\(
z_t^{(j)}\in\mathbb{R}^{d_j}\),
\(
d=\sum_{j\in\mathcal{J}}d_j.
\)
Suppose the predictive prior factorizes across components as in
Eq.~\eqref{eq:jointpriorfactorization},
\(
\kappa_t(z_t)
=
\prod_{j\in\mathcal{J}}
\mathcal{N}
\!\left(
z_t^{(j)};
\mu_t^{(j)},\Sigma_t^{(j)}
\right),
\)
with
\(
\Sigma_t
=
\operatorname{blkdiag}
\!\left(
\Sigma_t^{(j)}
\right)_{j\in\mathcal{J}}
\succ0.
\)

Let
\(
f_\theta(z_t)
=
\mathbb{E}_\theta[y_t\mid z_t]
\in\mathbb{R}^{d_y}
\)
denote the decoder mean. Linearize the decoder locally around the
prior mean
\(
\mu_t
=
\bigl(
\mu_t^{(j)}
\bigr)_{j\in\mathcal{J}}
\)
and suppose that the resulting local likelihood is Gaussian:
\(
p_\theta(y_t\mid z_t)
\approx
\mathcal{N}
\!\left(
y_t;
f_\theta(\mu_t)
+
\sum_{j\in\mathcal{J}}
W_t^{(j)}
\bigl(
z_t^{(j)}-\mu_t^{(j)}
\bigr),
\Omega_t
\right),
\)
where
\(
W_t^{(j)}
=
\left.
\nabla_{z^{(j)}}f_\theta(z)
\right|_{z=\mu_t}
\in\mathbb{R}^{d_y\times d_j}\),
\(
\Omega_t\succ0.
\)
Define
\(
W_t
=
\bigl[
W_t^{(F)}
\;
W_t^{(D)}
\;
W_t^{(H)}
\;
W_t^{(M)}
\bigr]
\)
and the corresponding local Fisher information matrix
\(
\mathcal{I}_t
=
W_t^\top\Omega_t^{-1}W_t.
\)
Assume that
\(
\mathcal{I}_t\succ0.
\)
Equivalently, $W_t$ has full column rank, and hence necessarily
$d_y\geq d$.

Consider the corresponding local conjugate-Gaussian Bayesian
specialization of the Explain update. Define
\(
\widetilde z_t
=
z_t-\mu_t\),
\(
\widetilde y_t
=
y_t-f_\theta(\mu_t),
\)
and let
\(
\widehat{\Sigma}_t
=
\left(
\Sigma_t^{-1}+\mathcal{I}_t
\right)^{-1},
\)
\(
\widehat{\widetilde z}_t
=
\widehat{\Sigma}_t
W_t^\top\Omega_t^{-1}\widetilde y_t.
\)
Suppose that the true latent state differs from the prior mean in
exactly one component $j^\star\in\mathcal{J}$, by a nonzero anomaly
\(
\Delta\in\mathbb{R}^{d_{j^\star}}\setminus\{0\},
\)
so that
\(
\widetilde y_t
=
W_t^{(j^\star)}\Delta+\eta_t\),
\(
\eta_t\sim\mathcal{N}(0,\Omega_t).
\)
Let $\iota_{j^\star}(\Delta)\in\mathbb{R}^{d}$ denote the vector obtained
by placing $\Delta$ in component $j^\star$ and zeros in all other
components.

Then the following hold.

\textbf{(a) Bias bound.}
The expected posterior mean shift satisfies
\(
\mathbb{E}_{\eta_t}
\!\left[
\widehat{\widetilde z}_t
\right]
-
\iota_{j^\star}(\Delta)
=
-
\widehat{\Sigma}_t
\Sigma_t^{-1}
\iota_{j^\star}(\Delta),
\)
and therefore
\(
\left\|
\mathbb{E}_{\eta_t}
\!\left[
\widehat{\widetilde z}_t
\right]
-
\iota_{j^\star}(\Delta)
\right\|_2
\leq
\left\|
\mathcal{I}_t^{-1}
\right\|_{\mathrm{op}}
\left\|
\Sigma_t^{-1}
\right\|_{\mathrm{op}}
\|\Delta\|_2.
\)
Consequently, for every component $j\neq j^\star$, the corresponding
block of the expected posterior mean shift satisfies
\[
\mathbb{E}_{\eta_t}\!\left[
\widehat{\widetilde z}_t^{(j)}
\right]\longrightarrow 0
\]
whenever
\[
\left\|
\mathcal{I}_t^{-1}
\right\|_{\mathrm{op}}
\left\|
\Sigma_t^{-1}
\right\|_{\mathrm{op}}
\longrightarrow0.
\]
In particular, this occurs if either the local observational
information becomes arbitrarily large,
\(
\left\|\mathcal{I}_t^{-1}\right\|_{\mathrm{op}}\longrightarrow0,
\)
while the prior remains fixed, or the prior becomes arbitrarily
diffuse,
\(
\left\|\Sigma_t^{-1}\right\|_{\mathrm{op}}\longrightarrow0,
\)
while the observational information remains fixed.

\textbf{(b) Exact attribution under orthogonal component sensitivities.}
If, in addition,
\(
\left(W_t^{(j)}\right)^\top
\Omega_t^{-1}
W_t^{(k)}
=
0\),
\(
j\neq k,
\)
so that $\mathcal{I}_t$ is block-diagonal with respect to the
component decomposition, then
\(
\mathbb{E}_{\eta_t}
\!\left[
\widehat{\widetilde z}_t^{(j)}
\right]
=
0\),
\(
j\neq j^\star,
\)
for every positive-definite block-diagonal prior covariance
$\Sigma_t$ and every anomaly
$\Delta$.
Thus, under orthogonal local component sensitivities, an anomaly
originating in one component produces no systematic posterior
attribution to any other component in this local Gaussian model.

\textbf{(c) Strict posterior-uncertainty reduction.}
The posterior covariance satisfies
\[
\widehat{\Sigma}_t=\left(\Sigma_t^{-1}+\mathcal I_t\right)^{-1}
\prec\Sigma_t.
\]
Equivalently, for every nonzero $v\in\mathbb R^d$,
\[
v^\top\widehat{\Sigma}_t v<v^\top\Sigma_t v.
\]
Thus, under $\mathcal I_t\succ0$, the interaction provides strictly
positive local information in every latent direction and strictly reduces
posterior uncertainty in every nonzero joint latent direction.

\end{styledtheorem}

\begin{proof}
Under the local Gaussian likelihood,
\[
\widetilde y_t\mid\widetilde z_t
\sim
\mathcal{N}
\!\left(
W_t\widetilde z_t,\Omega_t
\right),
\]
while the centered prior is
\[
\widetilde z_t
\sim
\mathcal{N}(0,\Sigma_t).
\]
Completing the square in the joint log-density gives the Gaussian
posterior
\[
\widetilde z_t\mid\widetilde y_t
\sim
\mathcal{N}
\!\left(
\widehat{\widetilde z}_t,
\widehat{\Sigma}_t
\right),
\]
where
\[
\widehat{\Sigma}_t^{-1}
=
\Sigma_t^{-1}
+
W_t^\top\Omega_t^{-1}W_t
=
\Sigma_t^{-1}+\mathcal{I}_t
\]
and
\[
\widehat{\widetilde z}_t
=
\widehat{\Sigma}_t
W_t^\top
\Omega_t^{-1}
\widetilde y_t.
\]

For the local Gaussian likelihood,
\[
-\nabla_z^2
\log p_\theta(y_t\mid z_t)
=
W_t^\top\Omega_t^{-1}W_t
=
\mathcal{I}_t,
\]
so $\mathcal{I}_t$ is the local Fisher information matrix for the
latent state. The assumption $\mathcal{I}_t\succ0$ therefore implies
that all latent directions are locally distinguishable from the
observation.

\textbf{(a)}
Since
\[
\widetilde y_t
=
W_t^{(j^\star)}\Delta+\eta_t
\]
and
\[
\mathbb{E}[\eta_t]=0,
\]
we obtain
\[
\begin{aligned}
\mathbb{E}_{\eta_t}
\!\left[
\widehat{\widetilde z}_t
\right]
&=
\widehat{\Sigma}_t
W_t^\top
\Omega_t^{-1}
W_t^{(j^\star)}
\Delta\\
&=
\widehat{\Sigma}_t
\mathcal{I}_t
\iota_{j^\star}(\Delta).
\end{aligned}
\]
Because
\[
\mathcal{I}_t
=
\widehat{\Sigma}_t^{-1}
-
\Sigma_t^{-1},
\]
we have
\[
\widehat{\Sigma}_t\mathcal{I}_t
=
I_d-\widehat{\Sigma}_t\Sigma_t^{-1}.
\]
Therefore,
\[
\mathbb{E}_{\eta_t}
\!\left[
\widehat{\widetilde z}_t
\right]
-
\iota_{j^\star}(\Delta)
=
-
\widehat{\Sigma}_t
\Sigma_t^{-1}
\iota_{j^\star}(\Delta).
\]

Since
\[
\widehat{\Sigma}_t^{-1}
=
\Sigma_t^{-1}+\mathcal{I}_t
\succeq
\mathcal{I}_t
\succ0,
\]
the order-reversing property of matrix inversion gives
\[
\widehat{\Sigma}_t
\preceq
\mathcal{I}_t^{-1}.
\]
Consequently,
\[
\left\|
\widehat{\Sigma}_t
\right\|_{\mathrm{op}}
\leq
\left\|
\mathcal{I}_t^{-1}
\right\|_{\mathrm{op}}.
\]
Using submultiplicativity of the operator norm and the fact that
\[
\left\|
\iota_{j^\star}(\Delta)
\right\|_2
=
\|\Delta\|_2,
\]
we obtain
\[
\begin{aligned}
\left\|
\mathbb{E}_{\eta_t}
\!\left[
\widehat{\widetilde z}_t
\right]
-
\iota_{j^\star}(\Delta)
\right\|_2
&\leq
\left\|
\widehat{\Sigma}_t
\right\|_{\mathrm{op}}
\left\|
\Sigma_t^{-1}
\right\|_{\mathrm{op}}
\|\Delta\|_2\\
&\leq
\left\|
\mathcal{I}_t^{-1}
\right\|_{\mathrm{op}}
\left\|
\Sigma_t^{-1}
\right\|_{\mathrm{op}}
\|\Delta\|_2.
\end{aligned}
\]
Because the true anomaly has support only in component $j^\star$, the
corresponding blocks of
$\iota_{j^\star}(\Delta)$ are zero for $j\neq j^\star$. Hence the same
bound controls the expected attribution to every incorrect component,
and those expected attributions vanish whenever the product of the two
operator norms tends to zero.

\textbf{(b)}
Suppose now that $\mathcal{I}_t$ is block-diagonal. Since $\Sigma_t$ is
also block-diagonal by construction,
\[
\widehat{\Sigma}_t^{-1}
=
\Sigma_t^{-1}+\mathcal{I}_t
\]
is block-diagonal, and hence so are $\widehat{\Sigma}_t$ and
\[
\widehat{\Sigma}_t\Sigma_t^{-1}.
\]
Therefore,
\[
\left(
\widehat{\Sigma}_t
\Sigma_t^{-1}
\iota_{j^\star}(\Delta)
\right)^{(j)}
=
0,
\qquad
j\neq j^\star.
\]
The identity established in part (a) then gives
\[
\mathbb{E}_{\eta_t}
\!\left[
\widehat{\widetilde z}_t^{(j)}
\right]
=
0,
\qquad
j\neq j^\star.
\]
Thus the expected posterior shift contains no contribution from an
incorrect component. This proves exact component-wise attribution in
the sense of zero expected cross-component attribution.

\textbf{(c)} Since $\mathcal I_t\succ0$,
\[
\widehat{\Sigma}_t^{-1}=\Sigma_t^{-1}+\mathcal I_t\succ\Sigma_t^{-1}.
\]
The order-reversing property of matrix inversion for positive-definite
matrices gives $\widehat{\Sigma}_t\prec\Sigma_t$. Hence, for every
nonzero $v$,
\[
v^\top\widehat{\Sigma}_t v<v^\top\Sigma_t v,
\]
which establishes strict reduction of posterior uncertainty in every
nonzero latent direction.
\end{proof}

\noindent\textbf{Remark [Scope of the identifiability result].}
The theorem assumes that the true latent state differs from the prior mean in exactly one component. This is an idealized setting that isolates the source-attribution problem. In practice, an anomaly may affect multiple components simultaneously, in which case posterior attribution can be intrinsically ambiguous. The result should therefore be interpreted as a local, best-case attribution guarantee: under the stated component-wise separation and sufficient local Fisher information, systematic cross-component attribution is controlled, and under orthogonal component sensitivities it vanishes in expectation. When the prior precision $\Sigma_t^{-1}$ is large relative to the local information $\mathcal{I}_t$, the bias term in part~(a) need not be negligible relative to $\|\Delta\|_2$, so accurate source attribution is not guaranteed merely by observing a nonzero anomaly. The bound shows that attribution ambiguity is reduced by stronger local observational information and by less informative (more diffuse) priors. Part~(c) additionally establishes strict reduction of posterior uncertainty in every nonzero latent direction; this is a local information statement and should not be interpreted as a global identifiability guarantee beyond the local Gaussian approximation.


\begin{styledcorollary}{High-Probability Correct Source Attribution}{app:highprob_attribution}

Under the assumptions of part~(b) of
Theorem~\ref{thm:component_identifiability}, consider the following
source-attribution bound. The preceding uncertainty-reduction statement
is valid for the general vector-valued component blocks. For the following
source-attribution bound, additionally suppose that each latent component
is scalar, i.e., $d_j=1$ for every $j\in\mathcal{J}$. All component-level quantities in the attribution construction below are therefore scalar. Write
\(
W_t^{(j)}\in\mathbb{R}^{d_y}
\)
and
\(
\mathcal{I}_t^{(j)}
:=
\left(W_t^{(j)}\right)^\top
\Omega_t^{-1}
W_t^{(j)}
>0
\)
for the corresponding positive scalar diagonal block. Let
\(
\widehat{\Sigma}_t^{(j)}
=
\left(
(\Sigma_t^{(j)})^{-1}
+
\mathcal{I}_t^{(j)}
\right)^{-1}
\)
denote the scalar posterior variance of component $j$, and define
\(
\Xi_j
=
\left(
\widehat{\Sigma}_t^{(j)}
\right)^2
\mathcal{I}_t^{(j)}
\)
and the standardized posterior shift
\(
T_j
=
\frac{
\widehat{\widetilde z}_t^{(j)}
}{
\sqrt{\Xi_j}
}.
\)
Suppose the anomaly occurs in the unique component
$j^\star\in\mathcal{J}$ with magnitude
\(
\Delta\neq0.
\)
Define the source-attribution rule
\(
\widehat{\jmath}
=
\arg\max_{j\in\mathcal{J}}
|T_j|,
\)
with arbitrary deterministic tie-breaking.

Then
\(
\Pr
\!\left(
\widehat{\jmath}\neq j^\star
\right)
\leq
\min
\left\{
1,\,
\left(2|\mathcal{J}|-1\right)
\exp
\!\left(
-\frac{
\mathcal{I}_t^{(j^\star)}\Delta^2
}{8}
\right)
\right\}.
\)

Since $|\mathcal{J}|=4$ for CENTRY,
\(
\Pr
\!\left(
\widehat{\jmath}\neq j^\star
\right)
\leq
\min
\left\{
1,\,
7
\exp
\!\left(
-\frac{
\mathcal{I}_t^{(j^\star)}\Delta^2
}{8}
\right)
\right\}.
\)

Consequently,
\(
\Pr
\!\left(
\widehat{\jmath}\neq j^\star
\right)
\longrightarrow0
\)
as $|\Delta|\longrightarrow\infty$.
The resulting exponential bound depends on the prior covariance only
through a cancellation in the standardized statistic and is therefore
independent of $\Sigma_t$.

\end{styledcorollary}

\begin{proof}
Because $\mathcal{I}_t$ is block-diagonal, the standardized posterior
shift for any component $j\neq j^\star$ is
\[
\widehat{\widetilde z}_t^{(j)}
=
\widehat{\Sigma}_t^{(j)}
\left(W_t^{(j)}\right)^\top
\Omega_t^{-1}\eta_t.
\]
Since
\[
\eta_t\sim\mathcal{N}(0,\Omega_t),
\]
it follows that
\[
\widehat{\widetilde z}_t^{(j)}
\sim
\mathcal{N}
\!\left(
0,
\left(
\widehat{\Sigma}_t^{(j)}
\right)^2
\mathcal{I}_t^{(j)}
\right)
=
\mathcal{N}(0,\Xi_j).
\]
Therefore,
\[
T_j\sim\mathcal{N}(0,1),
\qquad
j\neq j^\star.
\]

For the true component,
\[
\begin{aligned}
\widehat{\widetilde z}_t^{(j^\star)}
&=
\widehat{\Sigma}_t^{(j^\star)}
\left(W_t^{(j^\star)}\right)^\top
\Omega_t^{-1}
\left(
W_t^{(j^\star)}\Delta+\eta_t
\right)\\
&=
\widehat{\Sigma}_t^{(j^\star)}
\mathcal{I}_t^{(j^\star)}
\Delta
+
\widehat{\Sigma}_t^{(j^\star)}
\left(W_t^{(j^\star)}\right)^\top
\Omega_t^{-1}\eta_t.
\end{aligned}
\]
Consequently,
\[
T_{j^\star}
\sim
\mathcal{N}(m_{j^\star},1),
\]
where
\[
m_{j^\star}
=
\frac{
\widehat{\Sigma}_t^{(j^\star)}
\mathcal{I}_t^{(j^\star)}
\Delta
}{
\sqrt{\Xi_{j^\star}}
}.
\]
Using
\[
\Xi_{j^\star}
=
\left(
\widehat{\Sigma}_t^{(j^\star)}
\right)^2
\mathcal{I}_t^{(j^\star)},
\]
we obtain
\[
m_{j^\star}
=
\operatorname{sgn}(\Delta)
\sqrt{
\mathcal{I}_t^{(j^\star)}
}
\,|\Delta|,
\]
and hence
\[
m_{j^\star}^2
=
\mathcal{I}_t^{(j^\star)}\Delta^2.
\]
Thus the standardized signal-to-noise ratio is independent of the
prior covariance.

For any $\tau>0$, the event
\[
\left\{
|T_{j^\star}|\geq\tau
\right\}
\cap
\bigcap_{j\neq j^\star}
\left\{
|T_j|<\tau
\right\}
\]
implies
\[
\widehat{\jmath}=j^\star.
\]
Therefore, by the union bound,
\[
\Pr
\!\left(
\widehat{\jmath}\neq j^\star
\right)
\leq
\Pr
\!\left(
|T_{j^\star}|<\tau
\right)
+
\sum_{j\neq j^\star}
\Pr
\!\left(
|T_j|\geq\tau
\right).
\]
No independence among the statistics $T_j$ is required for this step.

For each incorrect component,
\[
T_j\sim\mathcal{N}(0,1),
\]
so the standard Gaussian tail bound gives
\[
\Pr(|T_j|\geq\tau)
\leq
2e^{-\tau^2/2}.
\]
Hence,
\[
\sum_{j\neq j^\star}
\Pr(|T_j|\geq\tau)
\leq
2(|\mathcal{J}|-1)e^{-\tau^2/2}.
\]

The absolute-value decision rule is invariant under replacing $T_{j^\star}$ by $\operatorname{sgn}(m_{j^\star})T_{j^\star}\sim\mathcal N(|m_{j^\star}|,1)$ when $m_{j^\star}\ne0$. After this sign change, we use $T_{j^\star}$ for the sign-adjusted statistic and $m_{j^\star}$ for its positive mean $|m_{j^\star}|$ (the bound is trivial for zero mean). For $0<\tau\leq m_{j^\star}$,
\[
\begin{aligned}
\Pr(|T_{j^\star}|<\tau)
&\leq
\Pr(T_{j^\star}<\tau)\\
&=
\Phi(\tau-m_{j^\star})\\
&=
\Phi\!\left(
-(m_{j^\star}-\tau)
\right)\\
&\leq
\exp
\!\left(
-\frac{(m_{j^\star}-\tau)^2}{2}
\right).
\end{aligned}
\]
Choose
\[
\tau=\frac{m_{j^\star}}{2}>0.
\]
Then
\[
\Pr(|T_{j^\star}|<\tau)
\leq
e^{-m_{j^\star}^2/8},
\]
while
\[
\sum_{j\neq j^\star}
\Pr(|T_j|\geq\tau)
\leq
2(|\mathcal{J}|-1)e^{-m_{j^\star}^2/8}.
\]
Consequently,
\[
\Pr
\!\left(
\widehat{\jmath}\neq j^\star
\right)
\leq
\left(
2|\mathcal{J}|-1
\right)
e^{-m_{j^\star}^2/8}.
\]
Using
\[
m_{j^\star}^2
=
\mathcal{I}_t^{(j^\star)}\Delta^2
\]
gives
\[
\Pr
\!\left(
\widehat{\jmath}\neq j^\star
\right)
\leq
\left(
2|\mathcal{J}|-1
\right)
\exp
\!\left(
-\frac{
\mathcal{I}_t^{(j^\star)}\Delta^2
}{8}
\right).
\]
Since probabilities cannot exceed one, the stated
$\min\{1,\cdot\}$ form follows. Finally, because
$\mathcal{I}_t^{(j^\star)}>0$,
the exponential term converges to zero as
$|\Delta|\to\infty$.
\end{proof}


\begin{styledcorollary}{High-Probability Recovery of the Shifted-Component Set}{app:multi_attribution}

Adopt the notation and the block-diagonal Fisher-information assumption
of part~(b) of Theorem~\ref{thm:component_identifiability}. Suppose
additionally that each latent component is scalar, i.e., $d_j=1$ for
every $j\in\mathcal J$. Suppose that, instead of a single anomalous
component, a subset
$\mathcal K\subseteq\mathcal J$ of components experiences simultaneous
mean shifts, with $\mathcal K\neq\varnothing$, so that
\[
\widetilde y_t
=
\sum_{k\in\mathcal K}W_t^{(k)}\Delta_k+\eta_t,
\qquad
\eta_t\sim\mathcal N(0,\Omega_t),
\]
for nonzero shift magnitudes $\{\Delta_k\}_{k\in\mathcal K}$. Define, for
each $j\in\mathcal J$, the standardized posterior shift
$T_j:=\widehat{\widetilde z}_t^{(j)}/\sqrt{\Xi_j}$ as in
Corollary~\ref{cor:highprob_attribution}, where
$\widehat{\widetilde z}_t^{(j)}$ denotes the $j$-th block of the local
posterior mean shift $\widehat{\widetilde z}_t$ defined in
Theorem~\ref{thm:component_identifiability}, set
$m_k:=\Delta_k\sqrt{\mathcal I_t^{(k)}}$ for $k\in\mathcal K$ and
$m_{\min}:=\min_{k\in\mathcal K}|m_k|$, and let
\[
\widehat{\mathcal K}:=\{j\in\mathcal J:|T_j|\ge m_{\min}/2\}.
\]
Then
\[
\Pr\!\left(\widehat{\mathcal K}\ne\mathcal K\right)
\leq
\min\!\left\{
1,
\left(2|\mathcal J|-|\mathcal K|\right)
\exp\!\left(-\frac{m_{\min}^2}{8}\right)
\right\}.
\]

\end{styledcorollary}

\begin{proof}
By the block-diagonal assumption,
$(W_t^{(j)})^\top\Omega_t^{-1}W_t^{(k)}=0$ for $j\ne k$, so, exactly as
in the proof of part~(b) of Theorem~\ref{thm:component_identifiability},
every cross term $(W_t^{(j)})^\top\Omega_t^{-1}W_t^{(k)}\Delta_k$ with
$k\ne j$ vanishes, giving
\[
\widehat{\widetilde z}_t^{(j)}
=
\begin{cases}
\widehat\Sigma_t^{(j)}\mathcal I_t^{(j)}\Delta_j
+\widehat\Sigma_t^{(j)}(W_t^{(j)})^\top\Omega_t^{-1}\eta_t,
& j\in\mathcal K,\\[4pt]
\widehat\Sigma_t^{(j)}(W_t^{(j)})^\top\Omega_t^{-1}\eta_t,
& j\notin\mathcal K.
\end{cases}
\]
As in the proof of Corollary~\ref{cor:highprob_attribution}, each
$(W_t^{(j)})^\top\Omega_t^{-1}\eta_t$ is a linear function of the single
Gaussian vector $\eta_t$, so $\{T_j\}_{j\in\mathcal J}$ is jointly
Gaussian, and block-diagonality makes any two of them uncorrelated,
hence independent. Consequently,
\[
T_j\sim\mathcal N(0,1),\quad j\notin\mathcal K,
\qquad
T_k\sim\mathcal N(m_k,1),\quad k\in\mathcal K,
\]
exactly as if each shifted component were the only one present: shifts
in the other components of $\mathcal K$ contribute nothing to $T_j$ or
$T_k$ by orthogonality. Set $\tau:=m_{\min}/2$.

For $j\notin\mathcal K$, the Gaussian tail bound gives
\[
\Pr(|T_j|\ge\tau)\le2e^{-\tau^2/2}.
\]

For $k\in\mathcal K$: since $|T_k|=|-T_k|$ and
$-T_k\sim\mathcal N(-m_k,1)$, the distribution of $|T_k|$ depends on
$m_k$ only through $|m_k|\ge m_{\min}\ge2\tau$. Arguing as in the proof of
Corollary~\ref{cor:highprob_attribution} (with $|m_k|$ in place of
$m_{j^\star}$),
\[
\Pr(|T_k|<\tau)
\le
\exp\!\left(-\frac{(|m_k|-\tau)^2}{2}\right)
\le
\exp\!\left(-\frac{m_{\min}^2}{8}\right).
\]

The event $\{\widehat{\mathcal K}\ne\mathcal K\}$ implies that either
some $k\in\mathcal K$ has $|T_k|<\tau$ or some $j\notin\mathcal K$ has
$|T_j|\ge\tau$. By the union bound (independence is not required for
this step),
\[
\Pr(\widehat{\mathcal K}\ne\mathcal K)
\leq
|\mathcal K|e^{-m_{\min}^2/8}
+
2(|\mathcal J|-|\mathcal K|)e^{-m_{\min}^2/8},
\]
which is the stated bound. When $|\mathcal K|=1$, writing
$\mathcal K=\{j^\star\}$, this becomes
\[
\Pr(\widehat{\mathcal K}\ne\{j^\star\})
\leq
(2|\mathcal J|-1)e^{-m_{j^\star}^2/8},
\]
matching the numerical bound of Corollary~\ref{cor:highprob_attribution}:
both $\widehat{\mathcal K}=\{j^\star\}$ and
$\hat\jmath=j^\star$ hold whenever $|T_{j^\star}|\ge\tau$ and
$|T_j|<\tau$ for every $j\ne j^\star$, so the two proofs bound the same
event.
\end{proof}

\noindent\textbf{Remark [Threshold and attribution rules].}
The threshold-based rule $\widehat{\mathcal K}$ and the argmax rule
$\hat\jmath=\arg\max_j|T_j|$ are genuinely different procedures when
$|\mathcal K|>1$; the corollary does not claim that the argmax rule
extends to multi-component recovery. When $|\mathcal K|=1$, they are
linked through the event that the shifted component exceeds the same
threshold while all unshifted components remain below it, which is why
the numerical error bound agrees with Corollary~\ref{cor:highprob_attribution}.
The threshold $m_{\min}/2$ is an oracle threshold because $m_{\min}$ is
unknown in practice; this is the same idealized signal-strength
assumption used by the preceding single-component attribution result.

\begin{styledcorollary}{Sample Complexity of Component Discrimination}{app:sample_complexity_discrimination}

Under the assumptions of part~(b) of Theorem~\ref{thm:component_identifiability},
suppose that each latent component is scalar and that the local Fisher
information for component $j$ is $\mathcal I_t^{(j)}>0$. Consider two
competing hypotheses $H_{j_1}$ and $H_{j_2}$ about which single component
$j\in\{j_1,j_2\}$ is responsible for the observed elevated risk rating,
with $j_1\neq j_2$. Suppose that under $H_{j_k}$, the $k$-th component
experiences a nonzero mean shift of magnitude $\Delta_k$, while all other
components are unshifted. Define
\[
\Delta_{\min}:=
\min\{|\Delta_1|,|\Delta_2|\},
\qquad
\mathcal I_{\min}:=
\min_{j\in\{j_1,j_2\}}\mathcal I_t^{(j)},
\]
and let
\[
m_{\min}:=\sqrt{\mathcal I_{\min}}\,\Delta_{\min}.
\]
For $\delta\in(0,1)$ and integer $N\ge1$, suppose that $N$ conditionally independent observations with the same local model, information matrices, and shift magnitudes under each hypothesis are available and,
for each observation $n$, let
$T_{j,n}=\widehat{\widetilde z}_{t,n}^{(j)}/\sqrt{\Xi_{j,n}}$ be the
standardized posterior shift, with $\Xi_{j,n}=\Xi_j$ as defined in Corollary~\ref{cor:highprob_attribution}, since the local model and information matrices are the same across observations. Define the normalized aggregate statistic
\[
\overline T_j^{(N)}:=
\frac{1}{\sqrt N}\sum_{n=1}^{N}T_{j,n},
\]
and use the threshold-based attribution rule that selects $j_k$ when
$|\overline T_{j_k}^{(N)}|\ge m_{\min}\sqrt N/2$ and the other candidate
component does not cross this threshold; if neither or both candidate
components cross the threshold, the decision is declared a failure.
Then a sufficient sample size for correct discrimination with probability
at least $1-\delta$ is
\[
N
\ge
\frac{8}{\mathcal I_{\min}\Delta_{\min}^2}
\log\!\left(\frac{2|\mathcal J|-1}{\delta}\right).
\]
In particular, for $|\mathcal J|=4$,
\[
N
\ge
\frac{8}{\mathcal I_{\min}\Delta_{\min}^2}
\log\!\left(\frac{7}{\delta}\right).
\]
Thus, under the stated conditional-independence and Gaussian assumptions,
the Explain stage requires
$O\!\left((\mathcal I_{\min}\Delta_{\min}^2)^{-1}\log(1/\delta)\right)$
effectively independent observations to discriminate between the two
component-risk hypotheses with probability at least $1-\delta$.

\end{styledcorollary}

\begin{proof}
Under hypothesis $H_{j_k}$, Corollary~\ref{cor:highprob_attribution}
gives, for each observation $n$,
\[
T_{j_k,n}\sim\mathcal N(m_{j_k},1),
\qquad
T_{j,n}\sim\mathcal N(0,1),
\quad j\neq j_k,
\]
where
\[
|m_{j_k}|=|\Delta_k|\sqrt{\mathcal I_t^{(j_k)}}\geq m_{\min}.
\]
The information matrices and shift magnitudes are fixed across the observations under each hypothesis, so these marginal distributions are identical across $n$. By conditional independence across observations, the normalized sums
satisfy
\[
\overline T_{j_k}^{(N)}\sim
\mathcal N(\sqrt N\,m_{j_k},1),
\qquad
\overline T_j^{(N)}\sim\mathcal N(0,1),
\quad j\neq j_k.
\]
Set
\[
\tau_N:=\frac{m_{\min}\sqrt N}{2}.
\]
For the truly shifted component, the Gaussian tail bound gives
\[
\Pr\!\left(
|\overline T_{j_k}^{(N)}|<\tau_N
\middle|H_{j_k}\right)
\leq
\exp\!\left[-\frac{(\sqrt N\,|m_{j_k}|-\tau_N)^2}{2}\right]
\leq
\exp\!\left(-\frac{N m_{\min}^2}{8}\right).
\]
For each unshifted component $j\neq j_k$,
\[
\Pr\!\left(|\overline T_j^{(N)}|\geq\tau_N
\middle|H_{j_k}\right)
\leq
2\exp\!\left(-\frac{\tau_N^2}{2}\right)
=
2\exp\!\left(-\frac{N m_{\min}^2}{8}\right).
\]
Let $\widehat\jmath_N$ denote the selected candidate, or a failure symbol when neither or both candidates cross the threshold. Failure implies that the shifted component falls below the threshold or at least one of the remaining $|\mathcal J|-1$ components crosses it. Hence, by the union bound,
\[
\Pr\!\left(\widehat\jmath_N\neq j_k\mid H_{j_k}\right)
\leq
(2|\mathcal J|-1)
\exp\!\left(-\frac{N m_{\min}^2}{8}\right).
\]
Since
\[
m_{\min}^2=\mathcal I_{\min}\Delta_{\min}^2,
\]
this error probability is at most $\delta$ whenever
\[
N
\geq
\frac{8}{\mathcal I_{\min}\Delta_{\min}^2}
\log\!\left(\frac{2|\mathcal J|-1}{\delta}\right).
\]
The same bound holds under either $H_{j_1}$ or $H_{j_2}$, completing the
proof.
\end{proof}

\noindent\textbf{Remark [Interpretation and scope].}
This corollary is a sample-complexity refinement of the attribution
guarantees of Corollaries~\ref{cor:highprob_attribution} and
\ref{cor:multi_attribution}. It quantifies how many conditionally
independent interactions the Explain stage needs to reliably distinguish
between two competing single-component risk hypotheses, in terms of local
Fisher information and latent-space separation. The result uses an
aggregate Gaussian statistic across observations; it does not rely on a
majority-vote amplification argument. The threshold is idealized because
$m_{\min}$ is generally unknown in practice and would need to be replaced
by a conservative, data-driven lower bound or a prespecified design
threshold. For temporally correlated observations, the stated bound does
not apply without an additional dependence analysis; an effective-sample-size
interpretation is valid only when such an approximation is justified.

\noindent\textbf{Remark [Diagnostic role of the latent factor $F$].}
The factor $F$ is an inference-level latent diagnostic rather than an
independently regulated component. When attribution identifies $F$, any
operational consequence is mediated through its influence on one or more
regulated components $D,H,M$. An independent operational control channel
for $F$ would require additional $F$-specific risk budgets, constraints,
and corresponding safety guarantees.


\begin{styledtheorem}{Steady-State Posterior Uncertainty}{app:steady_state_riccati}

Consider a time-invariant linear-Gaussian component model with Gaussian initial state, fixed transition and observation matrices, and mutually independent Gaussian process and observation noise sequences, independent also of the initial state:
\[
z_{t+1}^{(j)}=A^{(j)}z_t^{(j)}+b^{(j)}+\varepsilon_{t+1}^{(j)},\qquad
 y_{t+1}^{(j)}=W^{(j)}z_{t+1}^{(j)}+\nu_{t+1}^{(j)},
\]
where $\varepsilon_{t+1}^{(j)}\sim\mathcal N(0,Q^{(j)})$ and $\nu_{t+1}^{(j)}\sim\mathcal N(0,\Omega^{(j)})$. Assume the component is filtered using this correctly specified model, and let
$\Sigma_t^{(j)}$ denote the posterior covariance of
$z_t^{(j)}$ given the observations through time $t$.
Assume that the process-noise covariance satisfies
\(Q^{(j)}\succeq0\) and that the observation-noise covariance satisfies
\(\Omega^{(j)}\succ0\).

\textbf{(a) Exact filtering recursion.}
The posterior covariance is independent of the realized observations
and satisfies the discrete-time Riccati recursion
\(
\Sigma_{t+1}^{(j)}
=
\mathcal R^{(j)}\!\left(\Sigma_t^{(j)}\right),
\)
where
\(
\bar\Sigma_{t+1}^{(j)}
=
A^{(j)}\Sigma_t^{(j)}(A^{(j)})^\top
+
Q^{(j)}
\)
is the one-step predicted covariance and
\[
\mathcal R^{(j)}(P)
=
\bar P
-
\bar P(W^{(j)})^\top
\left(
W^{(j)}\bar P(W^{(j)})^\top
+
\Omega^{(j)}
\right)^{-1}
W^{(j)}\bar P,
\]
with
\(
\bar P
=
A^{(j)}P(A^{(j)})^\top+Q^{(j)}.
\)
Equivalently, whenever $\bar P\succ0$,
\[
\mathcal R^{(j)}(P)
=
\left[
\bar P^{-1}
+
(W^{(j)})^\top
(\Omega^{(j)})^{-1}
W^{(j)}
\right]^{-1}.
\]

\textbf{(b) Steady-state covariance.}
Suppose that the pair
$(A^{(j)},W^{(j)})$ is detectable and that the pair
$\bigl(A^{(j)},(Q^{(j)})^{1/2}\bigr)$ is stabilizable. That is,
for every $\lambda\in\mathbb C$ with $|\lambda|\ge1$,
\(
\operatorname{rank}
\begin{bmatrix}
A^{(j)}-\lambda I\\
W^{(j)}
\end{bmatrix}
=
d_j,
\)
and
\(
\operatorname{rank}
\begin{bmatrix}
A^{(j)}-\lambda I &
(Q^{(j)})^{1/2}
\end{bmatrix}
=
d_j.
\)
Then the discrete-time algebraic Riccati equation
\(
\Sigma
=
\mathcal R^{(j)}(\Sigma)
\)
has a unique stabilizing solution
\(
\Sigma_\infty^{(j)}\in\mathbb S_+^{d_j}.
\)
If
\[
K_\infty^{(j)}
=
\bar\Sigma_\infty^{(j)}
(W^{(j)})^\top
\left(
W^{(j)}\bar\Sigma_\infty^{(j)}
(W^{(j)})^\top
+
\Omega^{(j)}
\right)^{-1},
\]
where
\(
\bar\Sigma_\infty^{(j)}
=
A^{(j)}\Sigma_\infty^{(j)}
(A^{(j)})^\top+Q^{(j)},
\)
then
\(
A^{(j)}
\left(I-K_\infty^{(j)}W^{(j)}\right)
\)
is Schur stable, and
\(
\Sigma_t^{(j)}
\longrightarrow
\Sigma_\infty^{(j)}
\) as $t\to\infty$
for every initial covariance
$\Sigma_1^{(j)}\in\mathbb S_+^{d_j}$.
Moreover, the convergence is geometric.

\end{styledtheorem}

\begin{proof}
\textbf{(a) Exact filtering recursion.}
Suppose that, after incorporating observations through time $t$,
the posterior distribution of $z_t^{(j)}$ is Gaussian with covariance
$\Sigma_t^{(j)}$. By the affine-Gaussian transition model,
the one-step predicted distribution is Gaussian with covariance
\[
\bar\Sigma_{t+1}^{(j)}
=
A^{(j)}\Sigma_t^{(j)}(A^{(j)})^\top
+
Q^{(j)}.
\]
The corresponding mean is
\[
\bar\mu_{t+1}^{(j)}
=
A^{(j)}\mu_t^{(j)}+b^{(j)}.
\]

Conditioning this Gaussian prior on
\[
y_{t+1}^{(j)}
=
W^{(j)}z_{t+1}^{(j)}
+
\nu_{t+1}^{(j)},
\qquad
\nu_{t+1}^{(j)}
\sim
\mathcal N(0,\Omega^{(j)}),
\]
gives the standard linear-Gaussian posterior covariance
\[
\Sigma_{t+1}^{(j)}
=
\bar\Sigma_{t+1}^{(j)}
-
\bar\Sigma_{t+1}^{(j)}(W^{(j)})^\top
\left(
W^{(j)}\bar\Sigma_{t+1}^{(j)}(W^{(j)})^\top
+
\Omega^{(j)}
\right)^{-1}
W^{(j)}\bar\Sigma_{t+1}^{(j)}.
\]
The matrix being inverted is positive definite because
$\Omega^{(j)}\succ0$. Thus the recursion is well-defined for every
$\Sigma_t^{(j)}\succeq0$.

Neither the propagation covariance nor the Gaussian conditioning
formula depends on the realized observation value. Hence
$\Sigma_{t+1}^{(j)}$ is a deterministic function of
$\Sigma_t^{(j)}$, and therefore of the initial covariance
$\Sigma_1^{(j)}$ alone.

When $\bar\Sigma_{t+1}^{(j)}\succ0$, the
Woodbury matrix identity gives the equivalent information-form
expression
\[
\left(\Sigma_{t+1}^{(j)}\right)^{-1}
=
\left(\bar\Sigma_{t+1}^{(j)}\right)^{-1}
+
(W^{(j)})^\top
(\Omega^{(j)})^{-1}
W^{(j)}.
\]

\textbf{(b) Steady-state convergence.}
Under detectability of $(A^{(j)},W^{(j)})$ and stabilizability of
$\bigl(A^{(j)},(Q^{(j)})^{1/2}\bigr)$, standard discrete-time algebraic
Riccati theory for the Kalman filtering problem \cite{lancaster1995algebraic} guarantees the
existence and uniqueness of a stabilizing non-negative solution
$\Sigma_\infty^{(j)}$ of
\[
\Sigma=\mathcal R^{(j)}(\Sigma).
\]
The associated steady-state gain is
\[
K_\infty^{(j)}
=
\bar\Sigma_\infty^{(j)}
(W^{(j)})^\top
\left(
W^{(j)}\bar\Sigma_\infty^{(j)}
(W^{(j)})^\top
+
\Omega^{(j)}
\right)^{-1},
\]
and the stabilizing property means that
\[
A^{(j)}
\left(I-K_\infty^{(j)}W^{(j)}\right)
\]
is Schur stable.

The same standard Riccati convergence theorem implies that the
iterates of the Riccati map generated from every initial covariance
$\Sigma_1^{(j)}\succeq0$ converge to this stabilizing solution:
\[
\Sigma_t^{(j)}
\longrightarrow
\Sigma_\infty^{(j)}.
\]
Under the stated hypotheses, this convergence is geometric.

Thus the long-run posterior uncertainty of component $j$ is
characterized by the unique stabilizing solution of the associated
discrete-time algebraic Riccati equation.
\end{proof}


\begin{styledcorollary}{Closed-Form Steady-State Variance for a Scalar Component}{app:scalar_steady_state_variance}

Suppose that $d_j=1$ and write
\(
a:=A^{(j)},
\)
\(
q:=Q^{(j)}>0,
\)
\(
w:=W^{(j)}\neq0,
\)
and
\(
\omega:=\Omega^{(j)}>0.
\)
Then
\[
\Sigma_{t+1}^{(j)}
=
\frac{\omega\left(a^2\Sigma_t^{(j)}+q\right)}
{w^2\left(a^2\Sigma_t^{(j)}+q\right)+\omega},
\]
and
\[
\Sigma_t^{(j)}
\longrightarrow
\Sigma_\infty^{(j)}
\qquad\text{as }t\to\infty
\]
for every $\Sigma_1^{(j)}\ge0$.

The limiting variance is the unique non-negative solution of
\[
w^2a^2
\left(\Sigma_\infty^{(j)}\right)^2
+
\left(
w^2q+\omega-\omega a^2
\right)
\Sigma_\infty^{(j)}
-
\omega q
=
0.
\]
For $a\neq0$, it is given by
\[
\Sigma_\infty^{(j)}
=
\frac{
\omega a^2-w^2q-\omega
+
\sqrt{
\left(\omega a^2-w^2q-\omega\right)^2
+
4a^2w^2q\omega
}
}
{2a^2w^2}.
\]
For $a=0$,
\[
\Sigma_\infty^{(j)}
=
\frac{q\omega}{w^2q+\omega}.
\]
In particular,
\[
0<\Sigma_\infty^{(j)}<\infty.
\]

\end{styledcorollary}

\begin{proof}
The scalar Riccati recursion follows directly from
Theorem~\ref{thm:steady_state_riccati}:
\[
\Sigma_{t+1}
=
R(\Sigma_t)
:=
\frac{\omega(a^2\Sigma_t+q)}
{w^2(a^2\Sigma_t+q)+\omega}.
\]

A steady-state variance satisfies
\(
\Sigma_\infty=R(\Sigma_\infty).
\)
Multiplying by the positive denominator gives
\[
\Sigma_\infty
\left[
w^2(a^2\Sigma_\infty+q)+\omega
\right]
=
\omega(a^2\Sigma_\infty+q),
\]
and hence
\[
w^2a^2\Sigma_\infty^2
+
(w^2q+\omega-\omega a^2)\Sigma_\infty
-
\omega q
=
0.
\]

If $a\neq0$, the product of the two roots is
\[
-\frac{\omega q}{w^2a^2}<0.
\]
Thus exactly one root is positive and the other is negative. The
non-negative root is therefore necessarily
\[
\Sigma_\infty
=
\frac{
\omega a^2-w^2q-\omega
+
\sqrt{
(\omega a^2-w^2q-\omega)^2
+
4a^2w^2q\omega
}
}
{2a^2w^2}.
\]
In particular, this establishes both existence and uniqueness of the
non-negative fixed point.

If $a=0$, the recursion becomes
\[
\Sigma_{t+1}
=
\frac{\omega q}{w^2q+\omega},
\]
so the covariance reaches its steady-state value after one update:
\[
\Sigma_\infty
=
\frac{\omega q}{w^2q+\omega}.
\]

It remains to establish convergence for $a\neq0$. Define
\[
M=
\begin{pmatrix}
\omega a^2 & \omega q\\
w^2a^2 & w^2q+\omega
\end{pmatrix}.
\]
Writing
\[
\Sigma_t=\frac{u_t}{v_t}
\]
gives the projective iteration
\[
\begin{pmatrix}
u_{t+1}\\
v_{t+1}
\end{pmatrix}
=
M
\begin{pmatrix}
u_t\\
v_t
\end{pmatrix},
\qquad
(u_1,v_1)=(\Sigma_1,1).
\]
The trace and determinant are
\[
\operatorname{tr}(M)
=
\omega a^2+w^2q+\omega,
\qquad
\det(M)=\omega^2a^2.
\]
The discriminant is
\[
\begin{aligned}
\operatorname{tr}(M)^2-4\det(M)
&=
(\omega a^2+w^2q+\omega)^2-4\omega^2a^2\\
&=
(\omega a^2-w^2q-\omega)^2
+
4\omega a^2w^2q
>0.
\end{aligned}
\]
Hence $M$ has two distinct real eigenvalues
\[
\lambda_+>\lambda_-\ge0.
\]
Since their product is
\(
\omega^2a^2>0
\)
and their sum is positive, both eigenvalues are in fact positive:
\[
0<\lambda_-<\lambda_+.
\]
Consequently,
\[
0<\frac{\lambda_-}{\lambda_+}<1.
\]

Let $v_+$ and $v_-$ be eigenvectors associated with
$\lambda_+$ and $\lambda_-$, respectively. The projective iteration
converges to the direction of $v_+$ provided that the initial vector
$(\Sigma_1,1)^\top$ has a nonzero component in the $v_+$ direction.
The eigenvector corresponding to $\lambda_-$, however, represents the
negative fixed point $\Sigma_-<0$ of the Riccati map, whereas every
valid initial condition satisfies $\Sigma_1\ge0$. Therefore
$(\Sigma_1,1)^\top$ cannot be aligned with $v_-$, because such alignment
would require $\Sigma_1=\Sigma_-<0$. Thus every valid initial vector has
a nonzero component in the dominant $v_+$ direction.

It follows from the spectral decomposition of $M$ that
\[
M^{t-1}
\begin{pmatrix}
\Sigma_1\\
1
\end{pmatrix}
=
c_+\lambda_+^{t-1}v_+
+
c_-\lambda_-^{t-1}v_-,
\qquad c_+\neq0.
\]
After division by the second coordinate, the contribution of the
subdominant eigenvector is proportional to
\[
\left(\frac{\lambda_-}{\lambda_+}\right)^{t-1},
\]
which converges geometrically to zero. Hence the projective state
converges geometrically to the direction of $v_+$.

The projective fixed point associated with $v_+$ is the unique
non-negative fixed point of $R$, namely $\Sigma_\infty$. Therefore
\[
\Sigma_t\longrightarrow\Sigma_\infty
\]
for every $\Sigma_1\ge0$, with geometric convergence.
\end{proof}

\noindent\textbf{Remark [Unobservable component].}
The scalar result also illustrates the role of observability. If
$w=0$, the component does not enter the observation channel and the
Riccati recursion reduces to
\[
\Sigma_{t+1}=a^2\Sigma_t+q.
\]
Thus, for $q>0$,
\[
\Sigma_t\to\frac{q}{1-a^2}
\quad\text{if }|a|<1,
\qquad\text{whereas}\qquad
\Sigma_t\to\infty
\quad\text{if }|a|\ge1.
\]
Hence a component that is not observed can maintain bounded uncertainty
only when its own dynamics are stable.

\noindent\textbf{Remark [Positive steady-state uncertainty].}
When $q>0$, the steady-state variance remains strictly positive even
though the component is observed. Thus, in a stochastic
state-space model, continual observations generally do not imply
posterior covariance convergence to zero. Rather, the covariance
converges to a positive uncertainty floor determined jointly by the
process noise, observation noise, system dynamics, and observation
coefficient. This is the scalar case underlying the nonzero
steady-state covariance used in the component-risk consistency result.


\subsection{Operational guarantees}
\label{app:subsec:Operational guarantees}
This subsection gives explicit operational and computational guarantees for monitoring and regulation: expected risk reduction, detection delay, high-confidence detection horizons, false-alarm control, containment time, cumulative excursion exposure, stabilization under persistent regulation, one-step Lyapunov drift, and per-cycle computational complexity.

\noindent\textbf{Remark [Learning-to-regulation interface].}
The learning guarantees above do not, by themselves, establish the
effective-regulation condition required by the operational guarantees.
In particular, low cumulative or average policy regret, predictive
improvement, or convergence of policy parameters does not imply a
per-step causal probability of effective regulation, nor does it imply
a bound on dangerous false-negative rates. A formal bridge from learning
performance to operational safety would require additional assumptions
linking policy behavior to true risk and risk ratings, together with an
admissible comparator policy that satisfies the required effective-
regulation properties. Accordingly, the operational results below are
conditional on their stated effective-regulation assumptions.

\begin{styledlemma}{Foster--Lyapunov Bound on Expected Hitting Time}{app:foster_hitting_time}

Let $\{X_k\}_{k\ge0}$ be adapted to a filtration
$\{\mathcal F_k\}$, let $\mathcal T$ be a measurable set, and define
\(
\tau:=\inf\{k\ge0:X_k\in\mathcal T\}.
\)
Let $V\ge0$ be measurable with $\mathbb E[V(X_0)]<\infty$. If there exists a constant $c>0$ such that
\(
\mathbb E\!\left[
V(X_{k+1})\mid\mathcal F_k
\right]
\le
V(X_k)-c
\)
whenever \(X_k\notin\mathcal T,
\)
then
\(
\mathbb E[\tau]
\le
\frac{\mathbb E[V(X_0)]}{c}.
\)

\end{styledlemma}

\begin{proof}
Define the stopped process
\[
M_k:=V(X_{k\wedge\tau})+c(k\wedge\tau).
\]
If $k<\tau$, then $X_k\notin\mathcal T$ and therefore
\[
\mathbb E[M_{k+1}-M_k\mid\mathcal F_k]
=
\mathbb E[V(X_{k+1})-V(X_k)\mid\mathcal F_k]+c
\le0.
\]
If $k\ge\tau$, then $M_{k+1}=M_k$. Hence $\{M_k\}$ is a
supermartingale and
\[
\mathbb E[M_k]\le\mathbb E[M_0]
=
\mathbb E[V(X_0)].
\]
Since $V\ge0$,
\[
c\,\mathbb E[k\wedge\tau]
\le
\mathbb E[M_k]
\le
\mathbb E[V(X_0)].
\]
Letting $k\to\infty$ and applying monotone convergence yields
\[
\mathbb E[\tau]
\le
\frac{\mathbb E[V(X_0)]}{c}.
\]
\end{proof}


\begin{styledtheorem}{Expected Risk Reduction Bound}{app:risk_reduction_bound}

Let $\mathcal{F}_t$ denote the information available at time $t$, including
the current component-risk ratings $\rho_t$ and the history of observations
and regulatory actions. Let $c_t$ be the regulatory action selected at time
$t$ according to $\pi_\psi(\rho_t)$, and let
\(
\Delta R_t^{(j)}
=
R_{t+1}^{(j)}-R_t^{(j)}
\)
denote the one-step change in the risk of component
$j\in\{D,H,M\}$. Define the set of components whose current risk ratings
exceed their budgets by
\(
\mathcal{S}_t
:=
\left\{
j\in\{D,H,M\}:
\rho_t^{(j)}>\bar{B}^{(j)}
\right\}.
\)

Suppose that, conditional on $\mathcal{F}_t$:

\begin{enumerate}[nosep]
    \item \textbf{Effective regulation.}
    With probability at least $p_{\mathrm{eff}}\in(0,1]$, the regulatory
    action $c_t$ is effective. More precisely, on an event
    $E_t\in\mathcal{F}_{t+1}$ with
    \(
    \Pr(E_t\mid\mathcal{F}_t)\geq p_{\mathrm{eff}},
    \)
    the following hold almost surely:
    \(
    \Delta R_t^{(j)}<0\),
    \(j\in\mathcal{S}_t,
    \)
    and
    \(
    \Delta R_t^{(j)}\leq 0\),
    \(j\notin\mathcal{S}_t.
    \)

    \item \textbf{Guaranteed reduction when effective.}
    There exist $\eta_j>0$ such that, conditional on $E_t$ and
    $\mathcal{F}_t$,
    \(
    \mathbb{E}\!\left[
        \Delta R_t^{(j)}
        \mid E_t,\mathcal{F}_t
    \right]
    \leq -\eta_j\),
    \(j\in\mathcal{S}_t.
    \)

    \item \textbf{Bounded risk growth.}
    There exist finite constants $\zeta_j\geq0$ such that
    \(
    \Delta R_t^{(j)}\leq\zeta_j
    \) almost surely, for every $j\in\{D,H,M\}$.
\end{enumerate}

Then the conditional expected one-step change in total component risk
satisfies
\[
\mathbb{E}\!\left[
    \left.
    \sum_{j\in\{D,H,M\}}\Delta R_t^{(j)}
    \,\right|\,\mathcal{F}_t
\right]
\leq
-p_{\mathrm{eff}}
\sum_{j\in\mathcal{S}_t}\eta_j
+
(1-p_{\mathrm{eff}})
\sum_{j\in\{D,H,M\}}\zeta_j.
\]

Furthermore, let
\(
A:=\sum_{j\in\{D,H,M\}}\zeta_j\),
\(
B_t:=\sum_{j\in\mathcal{S}_t}\eta_j.
\)
Whenever $A+B_t>0$, if
\(
p_{\mathrm{eff}}
\geq
\frac{A}{A+B_t},
\)
then
\[
\mathbb{E}\!\left[
    \left.
    \sum_{j\in\{D,H,M\}}\Delta R_t^{(j)}
    \,\right|\,\mathcal{F}_t
\right]
\leq 0.
\]
If the inequality is strict, i.e.,
\(
p_{\mathrm{eff}}
>
\frac{A}{A+B_t},
\)
then the conditional expected net change in total risk is strictly
negative.

\end{styledtheorem}

\begin{proof}
Fix $t$ and condition throughout on the information
$\mathcal F_t$. For notational simplicity, suppress the explicit
conditioning on $\mathcal F_t$ in intermediate expressions.

By the law of total expectation,
\[
\begin{aligned}
\mathbb{E}\!\left[\left.\sum_j\Delta R_t^{(j)}\,\right|\,\mathcal F_t\right]
&=
\Pr(E_t\mid\mathcal F_t)
\mathbb{E}\!\left[\left.\sum_j\Delta R_t^{(j)}\,\right|\,E_t,\mathcal F_t\right]
\\
&\quad+
\Pr(E_t^c\mid\mathcal F_t)
\mathbb{E}\!\left[\left.\sum_j\Delta R_t^{(j)}\,\right|\,E_t^c,\mathcal F_t\right].
\end{aligned}
\]

On the effective event $E_t$, Assumption~1 gives
\[
\Delta R_t^{(j)}\leq0,
\qquad j\notin\mathcal S_t,
\]
almost surely, while Assumption~2 gives
\[
\mathbb{E}\!\left[
    \left.\Delta R_t^{(j)}\,\right|\,E_t,\mathcal F_t
\right]
\leq-\eta_j,
\qquad j\in\mathcal S_t.
\]
Consequently,
\[
\mathbb{E}\!\left[
    \left.\sum_j\Delta R_t^{(j)}\,\right|\,E_t,\mathcal F_t
\right]
\leq
-\sum_{j\in\mathcal S_t}\eta_j.
\]

On the ineffective event $E_t^c$, Assumption~3 gives the almost-sure
upper bound
\[
\Delta R_t^{(j)}\leq\zeta_j
\]
for every component. Therefore,
\[
\mathbb{E}\!\left[
    \left.\sum_j\Delta R_t^{(j)}\,\right|\,E_t^c,\mathcal F_t
\right]
\leq
\sum_j\zeta_j.
\]

Let
\[
B_t:=\sum_{j\in\mathcal S_t}\eta_j,
\qquad
A:=\sum_j\zeta_j,
\qquad
p_t:=\Pr(E_t\mid\mathcal F_t).
\]
Combining the preceding bounds gives
\[
\begin{aligned}
\mathbb{E}\!\left[\left.\sum_j\Delta R_t^{(j)}\,\right|\,\mathcal F_t\right]
&\leq
-p_tB_t+(1-p_t)A.
\end{aligned}
\]
The right-hand side is non-increasing in $p_t$ because
\[
\frac{\partial}{\partial p_t}
\left[-p_tB_t+(1-p_t)A\right]
=-(A+B_t)\leq0.
\]
By Assumption~1,
$p_t\geq p_{\mathrm{eff}}$, and hence
\[
\mathbb{E}\!\left[\left.\sum_j\Delta R_t^{(j)}\,\right|\,\mathcal F_t\right]
\leq
-p_{\mathrm{eff}}B_t
+(1-p_{\mathrm{eff}})A,
\]
which is the stated bound.

For the second claim, assume $A+B_t>0$. The condition
\[
p_{\mathrm{eff}}
\geq
\frac{A}{A+B_t}
\]
is equivalent to
\[
p_{\mathrm{eff}}B_t
\geq
(1-p_{\mathrm{eff}})A.
\]
Hence,
\[
-p_{\mathrm{eff}}B_t
+(1-p_{\mathrm{eff}})A
\leq0.
\]
Substitution into the established conditional risk-change bound gives
\[
\mathbb{E}\!\left[\left.\sum_j\Delta R_t^{(j)}\,\right|\,\mathcal F_t\right]
\leq0.
\]
If
\[
p_{\mathrm{eff}}
>
\frac{A}{A+B_t},
\]
then the preceding inequality is strict, and therefore
\[
\mathbb{E}\!\left[\left.\sum_j\Delta R_t^{(j)}\,\right|\,\mathcal F_t\right]<0.
\]
Thus regulation produces a strictly negative conditional expected
one-step change in total component risk under the stated effectiveness
condition.
\end{proof}


\begin{styledtheorem}{Bounded Expected Detection Delay}{app:detection_delay}

Fix a component $j\in\{D,H,M\}$ and suppress the superscript $(j)$
where unambiguous. Suppose the drift term $b_t^{(j)}$ in the
affine-Gaussian transition of Corollary~\ref{cor:gaussian-propagation}
is piecewise constant in time, with $A_t^{(j)}=a$, $|a|<1$, held fixed
throughout, and with $b_t^{(j)}$ equal to some baseline value for
$t<t_0$ and equal to a new value $b_{\mathrm{new}}$ for all $t\ge t_0$,
this new value being known to, and correctly used by, the filter from
$t_0$ onward. Assume the scalar linear-Gaussian observation model and independent Gaussian noises of Theorem~\ref{thm:steady_state_riccati}, with a nonzero scalar observation coefficient $w$ and the correctly specified steady-state filter. Let
\[
z^\star:=\frac{b_{\mathrm{new}}}{1-a}
\]
denote the new equilibrium latent state induced by this shift, let
$\mu_t:=\mathbb E[z_t\mid y_{1:t},c_{1:t-1}]$ denote the scalar
steady-state Kalman-filter posterior mean, and let $\Sigma_\infty$
denote the associated steady-state posterior variance of
Theorem~\ref{thm:steady_state_riccati}, already reached by time $t_0$. For this scalar specialization, assume that the Gaussian filter is correctly specified for the full information $\mathcal F_t$: conditional on $\mathcal F_t$, the filtering error has distribution $\mathcal N(0,\Sigma_\infty)$, and the next process and observation noises are mutually independent and jointly independent of the current latent state and $\mathcal F_t$.
Let $\bar b:=\inf\{\mu:g^{(j)}(\mu,\Sigma_\infty)\ge\bar{B}^{(j)}\}$ be the
latent-scale threshold corresponding to the risk-rating budget
$\bar{B}^{(j)}$, and suppose $g^{(j)}(\cdot,\Sigma_\infty)$ is
non-decreasing, with the threshold attained (that is, the risk rating at $\bar b$ is at least the budget), so that
\[
\mu_t\ge\bar b\Longrightarrow\rho_t^{(j)}=g^{(j)}(\mu_t,\Sigma_\infty)\ge\bar{B}^{(j)}.
\]

Assume:
\begin{enumerate}[label=(D\arabic*),ref=(D\arabic*)]
\item\label{app:ass:detection-nonneg} \emph{(Non-negative persistence.)} $0\le a<1$.
\item\label{app:ass:detection-margin} \emph{(Positive equilibrium margin.)} $\gamma:=z^\star-\bar b>0$.
\item\label{app:ass:detection-onset} \emph{(Sub-threshold onset.)} $\bar z:=\mu_{t_0}<\bar b$. This is a condition on the observed onset history. The statements below apply on the $\mathcal F_{t_0}$-measurable event $\{\bar z<\bar b\}$; averages over onset histories are taken conditional on this event when it has positive probability.
\item\label{app:ass:detection-responsive} \emph{(Responsiveness.)} With
$\sigma_\mu^2:=q-(1-a^2)\Sigma_\infty>0$ the innovation variance of the
posterior-mean process ($q$ the process-noise variance of $z_t$), where
$\xi_t\sim\mathcal{N}(0,\sigma_\mu^2)$ and
$\mathbb{E}[|\xi_t|]=\sigma_\mu\sqrt{2/\pi}$, and
$c:=(1-a)\gamma-\sigma_\mu\sqrt{2/\pi}>0$.
\end{enumerate}

Define the latent-mean detection time
$\tau_{\mathrm{detect}}:=\inf\{t\ge t_0:\mu_t\ge\bar b\}$ and the
rating detection time $\tau_\rho:=\inf\{t\ge t_0:\rho_t^{(j)}\ge\bar{B}^{(j)}\}$. Then
$\tau_\rho\le\tau_{\mathrm{detect}}$ almost surely, and, with
$\Delta:=z^\star-\bar z>\gamma>0$,
\[
\mathbb E[\tau_{\mathrm{detect}}-t_0\mid\mathcal F_{t_0}]\le\frac{\Delta}{c},
\qquad\text{hence}\qquad
\mathbb E[\tau_\rho-t_0\mid\mathcal F_{t_0}]\le\frac{\Delta}{c}.
\]
Thus, under a responsive filter and a sufficiently strong regime shift, CENTRY detects the onset of a sustained elevated-risk regime and crosses the corresponding component risk-rating threshold within an expected number of cycles, conditional on the onset information, bounded by $\Delta/c$. Averaging over sub-threshold onset histories, both expected delays are bounded by $\mathbb E[\Delta\mid\bar z<\bar b]/c$ when $\Delta$ is integrable under that conditional law.

\end{styledtheorem}

\begin{proof}
For notational simplicity, suppress the component superscript $(j)$
throughout the proof.

\emph{Step 1: posterior-mean dynamics.}
At steady-state filtering covariance, the scalar Kalman update can be
written using the dimensionless gain $\lambda=\bar\Sigma_\infty w^2/(\bar\Sigma_\infty w^2+\Omega^{(j)})$, where $\bar\Sigma_\infty=a^2\Sigma_\infty+q$, as
\[
\mu_{t+1}
=
a\mu_t+b_{\mathrm{new}}
+
\lambda
\left[
\frac{y_{t+1}}{w}
-
(a\mu_t+b_{\mathrm{new}})
\right].
\]
Define the filtering error
\[
e_t:=z_t-\mu_t.
\]
Since
\[
z_{t+1}
=
az_t+b_{\mathrm{new}}+\varepsilon_{t+1}
\]
and
\[
y_{t+1}
=
wz_{t+1}+\nu_{t+1},
\]
substitution gives
\[
\begin{aligned}
\mu_{t+1}
&=
a\mu_t+b_{\mathrm{new}}
+
\lambda
\left[
a(z_t-\mu_t)
+\varepsilon_{t+1}
+\frac{\nu_{t+1}}{w}
\right]\\
&=
a\mu_t+b_{\mathrm{new}}+\xi_{t+1},
\end{aligned}
\]
where
\[
\xi_{t+1}
:=
a\lambda e_t
+
\lambda\varepsilon_{t+1}
+
\frac{\lambda}{w}\nu_{t+1}.
\]

Under the stated filtering assumptions, the conditional law of $e_t$ given $\mathcal F_t$ is $\mathcal N(0,\Sigma_\infty)$ and does not depend on $\mathcal F_t$. Thus $e_t$ is independent of $\mathcal F_t$. The next process and measurement noises are mutually independent and jointly independent of $e_t$ and $\mathcal F_t$, since $\mu_t$ is $\mathcal F_t$-measurable. Consequently, $\xi_{t+1}$ is zero-mean Gaussian and independent of $\mathcal F_t$. Thus $\{\mu_t\}$ admits the Markov representation
\[
\mu_{t+1}
=
a\mu_t+b_{\mathrm{new}}+\xi_{t+1}.
\]

\emph{Step 2: steady-state variance of the filtering innovation.}
At steady-state covariance the conditional variance of the observation innovation is $w^2\bar\Sigma_\infty+\Omega^{(j)}$. Using the independence just established and the scalar Kalman covariance update $\Sigma_\infty=\bar\Sigma_\infty(1-\lambda)$ gives
\[
\operatorname{Var}(\xi_{t+1}\mid\mathcal F_t)
=\lambda\bar\Sigma_\infty
=\bar\Sigma_\infty-\Sigma_\infty
=q-(1-a^2)\Sigma_\infty=\sigma_\mu^2.
\]
This calculation requires steady-state filtering covariance, not stationarity of the posterior mean after onset. Hence
\[
\mathbb E[|\xi_{t+1}|\mid\mathcal F_t]=\sigma_\mu\sqrt{2/\pi}.
\]

\emph{Step 3: Lyapunov drift toward the rating threshold.}
Define
\[
V(\mu)
:=
(z^\star-\mu)_+
\]
and
\[
\mathcal T
:=
[\bar b,\infty).
\]
For $\mu<\bar b$, we have $\mu<z^\star$ and hence
\[
V(\mu)=z^\star-\mu.
\]
Using
\[
z^\star
=
az^\star+b_{\mathrm{new}},
\]
the posterior-mean recursion gives
\[
\begin{aligned}
V(\mu_{t+1})
&=
\left(
z^\star-a\mu_t-b_{\mathrm{new}}-\xi_{t+1}
\right)_+\\
&=
\left(
a(z^\star-\mu_t)-\xi_{t+1}
\right)_+.
\end{aligned}
\]
For $u\ge0$ and arbitrary $v$,
\[
(u-v)_+\le u+|v|.
\]
Here $u=a(z^\star-\mu_t)$ is non-negative by
Assumption~\ref{app:ass:detection-nonneg}, since
$z^\star-\mu_t=V(\mu_t)>0$ on this event.
Therefore,
\[
V(\mu_{t+1})
\le
aV(\mu_t)+|\xi_{t+1}|.
\]
Taking conditional expectation gives
\[
\mathbb E[
V(\mu_{t+1})\mid\mathcal F_t
]
\le
aV(\mu_t)
+
\sigma_\mu\sqrt{\frac{2}{\pi}}.
\]
Whenever $\mu_t<\bar b$,
\[
V(\mu_t)
=
z^\star-\mu_t
>
z^\star-\bar b
=
\gamma.
\]
Hence
\[
aV(\mu_t)
=
V(\mu_t)-(1-a)V(\mu_t)
<
V(\mu_t)-(1-a)\gamma.
\]
It follows that
\[
\mathbb E[
V(\mu_{t+1})\mid\mathcal F_t
]
\le
V(\mu_t)-c,
\]
where
\[
c
=
(1-a)\gamma
-
\sigma_\mu\sqrt{\frac{2}{\pi}}.
\]
By assumption, $c>0$.

\emph{Step 4: expected hitting time.}
Apply Lemma~\ref{lem:foster_hitting_time} conditionally on $\mathcal F_{t_0}$ with
\[
X_k=\mu_{t_0+k},
\qquad
\mathcal T=[\bar b,\infty).
\]
Because
\[
\mu_{t_0}=\bar z,
\]
we have
\[
V(\mu_{t_0})
=
z^\star-\bar z
=
\Delta.
\]
This equality is valid by Assumption~\ref{app:ass:detection-onset}.
Therefore,
\[
\mathbb E[
\tau_{\mathrm{detect}}-t_0\mid\mathcal F_{t_0}
]
\le
\frac{\Delta}{c}.
\]

Finally, by the threshold-compatibility property stated above,
\[
\mu_t\ge\bar b
\quad\Longrightarrow\quad
\rho_t^{(j)}\ge\bar{B}^{(j)}.
\]
Consequently,
\[
\tau_\rho\le\tau_{\mathrm{detect}}
\]
almost surely, which gives
\[
\mathbb E[\tau_\rho-t_0\mid\mathcal F_{t_0}]
\le
\mathbb E[\tau_{\mathrm{detect}}-t_0\mid\mathcal F_{t_0}]
\le
\frac{\Delta}{c}.
\]
\end{proof}

\noindent\textbf{Remark [Relation to the steady-state uncertainty result].}
Theorem~\ref{thm:steady_state_riccati} establishes convergence of the
filtering covariance to $\Sigma_\infty^{(j)}$. The present theorem then
uses that steady-state filter to quantify the expected time required
for the posterior mean to reach a prescribed risk-rating threshold.
Thus the two results characterize complementary aspects of monitoring:
the former bounds long-run estimation uncertainty, whereas the latter
bounds the response time to a sustained risk excursion.

\noindent\textbf{Remark [Scope of the detection-delay result].}
The theorem is deliberately a fixed-model filtering result. It assumes
that the transition and observation parameters of the elevated regime
are already correctly specified at the onset of the excursion. It
therefore measures the delay attributable to the \emph{Explain} stage
and does not include any additional delay required by the \emph{Learn}
stage to identify or adapt to a changed regime. A joint bound on
parameter-learning and filtering delay would require an additional
analysis of the parameter-update dynamics.

\noindent\textbf{Remark [Sustained excursions].}
The result concerns a persistent change in the underlying risk regime,
rather than an isolated noisy threshold crossing. Under Gaussian
process noise, even when $z^\star>\bar b^{(j)}$, individual realizations of
$z_t^{(j)}$ may temporarily fall below $\bar b^{(j)}$. Thus the theorem should
be interpreted as a bound on the filtering response to a sustained
elevation of the latent risk process, not as a guarantee that the
instantaneous true risk remains above its budget at every subsequent
time.


\begin{styledcorollary}{High-Confidence Detection Within a Finite Horizon}{app:detection_horizon}

Under the hypotheses of
Theorem~\ref{thm:detection_delay}, for every $\delta\in(0,1)$, almost surely on $\{\bar z<\bar b\}$,
\(
\Pr\left(
\tau_{\mathrm{detect}}-t_0
\le
\frac{\Delta}{c\delta}\mid\mathcal F_{t_0}
\right)
\ge
1-\delta.
\)
Moreover, for the previously defined risk-rating crossing time
\(
\tau_\rho
=
\inf\left\{
t\ge t_0:
\rho_t^{(j)}\ge\bar{B}^{(j)}
\right\},
\)
one also has
\(
\Pr\left(
\tau_\rho-t_0
\le
\frac{\Delta}{c\delta}\mid\mathcal F_{t_0}
\right)
\ge
1-\delta.
\)

Both probability bounds also hold conditional on the sub-threshold onset event $\{\bar z<\bar b\}$, when it has positive probability, by the tower property.
\end{styledcorollary}

\begin{proof}
By Theorem~\ref{thm:detection_delay}, on the onset event $\{\bar z<\bar b\}$,
\[
\mathbb E[
\tau_{\mathrm{detect}}-t_0\mid\mathcal F_{t_0}
]
\le
\frac{\Delta}{c}.
\]
Since $\tau_{\mathrm{detect}}-t_0\ge0$, conditional Markov's inequality gives, for any positive onset-measurable $T$,
\[
\Pr\left(
\tau_{\mathrm{detect}}-t_0>T\mid\mathcal F_{t_0}
\right)
\le
\frac{\Delta}{cT}.
\]
Setting
\[
T=\frac{\Delta}{c\delta}
\]
which is strictly positive on the onset event because $\Delta>\gamma>0$, yields
\[
\Pr\left(
\tau_{\mathrm{detect}}-t_0
\le
\frac{\Delta}{c\delta}\mid\mathcal F_{t_0}
\right)
\ge
1-\delta.
\]
Since
\[
\tau_\rho\le\tau_{\mathrm{detect}}
\qquad\text{almost surely},
\]
the same probability bound holds for $\tau_\rho$.

The bounds conditional on the sub-threshold onset event follow by the tower property when that event has positive probability.
\end{proof}

\noindent\textbf{Remark [Uncertainty vs. variance].}
The quantity $\Sigma_\infty^{(j)}$ measures uncertainty in the latent
state conditional on the observations, whereas $\sigma_\mu^2$ measures
the innovation variance of the posterior mean process. These quantities
therefore have different interpretations. In particular,
\[
\sigma_\mu^2
=
q-(1-a^2)\Sigma_\infty^{(j)},
\]
so reducing posterior uncertainty does not imply a reduction in the
variance of the posterior mean. Rather, improved observability transfers
more of the latent process variation from posterior uncertainty into
the observable posterior-mean trajectory.


\begin{styledtheorem}{False-Alarm Rate under the Null}{app:false_alarm_rate}

Fix a monitored component $j\in\{D,H,M\}$ and consider the baseline
(no-excursion) regime. Let the latent state have stationary mean
\(
\bar z^{(j)}
=
\mathbb{E}\!\left[z_t^{(j)}\mid\text{baseline}\right].
\)
Assume:
\begin{enumerate}[label=(A\arabic*),ref=(A\arabic*)]
  \item\label{app:ass:steady-state}
  \textbf{Steady-state filter.}
  The filter is initialized in its invariant Kalman regime, so that
  \(
  \Sigma_t^{(j)}=\Sigma_\infty^{(j)}
  \)
  for all $t$, and the component risk rating of
  Eq.~(\ref{eq:riskfunctional}) reduces to
  \(
  \rho_t^{(j)}
  =
  g^{(j)}\!\left(\mu_t^{(j)},\Sigma_\infty^{(j)}\right).
  \)

  \item\label{app:ass:gauss-ar1}
  \textbf{Gaussian AR(1) posterior mean.}
  Under the baseline regime, the steady-state posterior mean is a
  stationary Gaussian AR(1) process,
  \(
  \mu_t^{(j)}-\bar z^{(j)}
  =
  a^{(j)}
  \bigl(\mu_{t-1}^{(j)}-\bar z^{(j)}\bigr)
  +
  \xi_t^{(j)}\),
  \(
  |a^{(j)}|<1,
  \)
  where the innovations are independent and identically distributed as
  \(
  \xi_t^{(j)}
  \sim
  \mathcal{N}\!\left(0,\sigma_{\mu}^{2,(j)}\right)\),
  \(
  \sigma_{\mu}^{2,(j)}>0.
  \)
  Consequently, the stationary marginal distribution is
  \(
  \mu_t^{(j)}
  \sim
  \mathcal{N}\!\left(
      \bar z^{(j)},v^{(j)}
  \right)\),
  \(
  v^{(j)}
  =
  \frac{\sigma_{\mu}^{2,(j)}}
       {1-(a^{(j)})^2}.
  \)

  \item\label{app:ass:monotone-g}
  \textbf{Monotone rating function.}
  The map
  \(
  \mu
  \longmapsto
  g^{(j)}\!\left(\mu,\Sigma_\infty^{(j)}\right)
  \)
  is non-decreasing.
\end{enumerate}

For a rating threshold $\bar{B}^{(j)}\in[0,1]$, define the corresponding
latent-state threshold by
\[
\bar b^{(j)}
=
\inf
\left\{
\mu\in\mathbb{R}:
g^{(j)}\!\left(\mu,\Sigma_\infty^{(j)}\right)
\geq
\bar{B}^{(j)}
\right\}.
\]
If the upper level set is empty, then $P_{\mathrm{FA}}^{(j)}=0$ trivially. Otherwise, assume that $\bar b^{(j)}\in\mathbb{R}$.

Whenever
\(
\bar b^{(j)}\geq\bar z^{(j)},
\)
the steady-state false-alarm probability
\(
P_{\mathrm{FA}}^{(j)}
:=
\Pr\!\left(
\rho_t^{(j)}\geq\bar{B}^{(j)}
\mid\text{baseline}
\right)
\)
satisfies
\[
P_{\mathrm{FA}}^{(j)}
=
\Pr\!\left(
\mu_t^{(j)}\geq\bar b^{(j)}
\mid\text{baseline}
\right)
\leq
\frac{1}{2}
\exp\!\left(
-\frac{
\bigl(\bar b^{(j)}-\bar z^{(j)}\bigr)^2
}{
2v^{(j)}
}
\right)
=
\frac{1}{2}
\exp\!\left(
-\frac{
\bigl(\bar b^{(j)}-\bar z^{(j)}\bigr)^2
\bigl(1-(a^{(j)})^2\bigr)
}{
2\sigma_{\mu}^{2,(j)}
}
\right).
\]
Moreover, the long-run empirical false-alarm rate converges almost
surely to $P_{\mathrm{FA}}^{(j)}$:
\(
\lim_{T\to\infty}
\frac{1}{T}
\sum_{t=1}^{T}
\mathbf{1}
\!\left\{
\rho_t^{(j)}\geq\bar{B}^{(j)}
\right\}
=
P_{\mathrm{FA}}^{(j)}
\) a.s. and therefore
\[
\lim_{T\to\infty}
\frac{1}{T}
\sum_{t=1}^{T}
\mathbf{1}
\!\left\{
\rho_t^{(j)}\geq\bar{B}^{(j)}
\right\}
\leq
\frac{1}{2}
\exp\!\left(
-\frac{
\bigl(\bar b^{(j)}-\bar z^{(j)}\bigr)^2
\bigl(1-(a^{(j)})^2\bigr)
}{
2\sigma_{\mu}^{2,(j)}
}
\right)
\qquad \text{a.s.}
\]

\end{styledtheorem}

\begin{proof}
\textbf{Step 1 (reduction to a Gaussian tail event).}
By Assumption~\ref{app:ass:monotone-g}, the function
\[
\mu\mapsto
g^{(j)}\!\left(\mu,\Sigma_\infty^{(j)}\right)
\]
is non-decreasing. Hence its upper level set
\[
S^{(j)}
=
\left\{
\mu\in\mathbb{R}:
g^{(j)}\!\left(\mu,\Sigma_\infty^{(j)}\right)
\geq\bar{B}^{(j)}
\right\}
\]
is upward closed. Since $S^{(j)}$ is nonempty and
$\bar b^{(j)}=\inf S^{(j)}\in\mathbb{R}$, it has the form
\[
S^{(j)}
=
[\bar b^{(j)},\infty)
\qquad\text{or}\qquad
S^{(j)}
=
(\bar b^{(j)},\infty).
\]
By Assumption~\ref{app:ass:gauss-ar1}, $\mu_t^{(j)}$ has a non-degenerate
Gaussian marginal distribution and therefore has no atoms. In particular,
\[
\Pr\!\left(
\mu_t^{(j)}=\bar b^{(j)}
\mid\text{baseline}
\right)=0.
\]
Consequently,
\[
P_{\mathrm{FA}}^{(j)}
=
\Pr\!\left(
\mu_t^{(j)}\geq\bar b^{(j)}
\mid\text{baseline}
\right).
\]

\textbf{Step 2 (Gaussian tail bound).}
By Assumption~\ref{app:ass:gauss-ar1},
\[
\mu_t^{(j)}
\sim
\mathcal{N}\!\left(
\bar z^{(j)},v^{(j)}
\right),
\qquad
v^{(j)}
=
\frac{\sigma_{\mu}^{2,(j)}}
     {1-(a^{(j)})^2}.
\]
Since $\sigma_\mu^{2,(j)}>0$ and $|a^{(j)}|<1$, we have $v^{(j)}>0$. Therefore,
\[
P_{\mathrm{FA}}^{(j)}
=
Q\!\left(
\frac{\bar b^{(j)}-\bar z^{(j)}}
     {\sqrt{v^{(j)}}}
\right),
\]
where
\(
Q(x)=\Pr(Z\geq x)
\),
\(
Z\sim\mathcal{N}(0,1)
\).
Since $\bar b^{(j)}\geq\bar z^{(j)}$, the argument of $Q$ is
nonnegative. Using the standard one-sided Gaussian tail inequality
\[
Q(x)\leq\frac{1}{2}e^{-x^2/2},
\qquad x\geq0,
\]
we obtain
\[
P_{\mathrm{FA}}^{(j)}
\leq
\frac{1}{2}
\exp\!\left(
-\frac{
\bigl(\bar b^{(j)}-\bar z^{(j)}\bigr)^2
}{
2v^{(j)}
}
\right)
=
\frac{1}{2}
\exp\!\left(
-\frac{
\bigl(\bar b^{(j)}-\bar z^{(j)}\bigr)^2
\bigl(1-(a^{(j)})^2\bigr)
}{
2\sigma_{\mu}^{2,(j)}
}
\right).
\]

\textbf{Step 3 (long-run false-alarm rate).}
Under Assumption~\ref{app:ass:gauss-ar1}, the process
$\{\mu_t^{(j)}\}_{t\geq1}$ is a stationary Gaussian AR(1) process with
$|a^{(j)}|<1$. Since $|a^{(j)}|<1$, the process is stationary and ergodic
(it admits a causal moving-average representation with absolutely summable
coefficients). Since
\[
\mathbf{1}
\!\left\{
\rho_t^{(j)}\geq\bar{B}^{(j)}
\right\}
=
\mathbf{1}
\!\left\{
\mu_t^{(j)}\geq\bar b^{(j)}
\right\}
\quad\text{a.s.},
\]
Birkhoff's ergodic theorem gives
\[
\lim_{T\to\infty}
\frac{1}{T}
\sum_{t=1}^{T}
\mathbf{1}
\!\left\{
\rho_t^{(j)}\geq\bar{B}^{(j)}
\right\}
=
\mathbb{E}
\left[
\mathbf{1}
\!\left\{
\mu_t^{(j)}\geq\bar b^{(j)}
\right\}
\right]
=
P_{\mathrm{FA}}^{(j)}
\qquad\text{a.s.}
\]
Combining this almost-sure equality with the Gaussian tail bound from
Step~2 establishes the stated result.
\end{proof}

\noindent\textbf{Remark [Sensitivity/specificity duality and the Explain stage].}
\begin{itemize}[nosep]
  \item \emph{Specificity complement.}
  Theorem~\ref{thm:detection_delay} and
  Corollary~\ref{cor:detection_horizon} characterize the detection
  delay---the sensitivity of the Explain stage, namely how rapidly a
  genuine excursion is flagged. The present theorem provides the
  corresponding specificity characterization: under the null
  (no-excursion) regime, the probability of a spurious flag is bounded
  by a Gaussian tail determined by the gap
  \[
  \bar b^{(j)}-\bar z^{(j)}
  \]
  between the latent-state threshold implied by the rating threshold
  $\bar{B}^{(j)}$ and the baseline mean. Thus, the two results jointly
  characterize the sensitivity/specificity trade-off of the Explain
  stage. In particular, increasing the rating threshold
  $\bar{B}^{(j)}$ cannot decrease the corresponding latent threshold
  $\bar b^{(j)}$ when $g^{(j)}$ is non-decreasing; whenever it strictly
  increases $\bar b^{(j)}$, the present bound decreases exponentially,
  while Corollary~\ref{cor:detection_horizon} characterizes the
  corresponding effect on detection delay.

  \item \emph{Gaussian steady-state assumption.}
  Assumption~\ref{app:ass:gauss-ar1} provides the Gaussian stationary
  marginal needed for the false-alarm bound. In a scalar
  linear-Gaussian state-space model, the Kalman filter produces
  Gaussian posterior means and a white Gaussian innovation sequence
  under the correctly specified model. The precise AR(1) coefficient
  and innovation variance, however, depend on the particular
  state-transition and observation equations. Accordingly, the
  parameters $a^{(j)}$ and $\sigma_{\mu}^{2,(j)}$ should be identified
  from the steady-state Kalman recursion established for the model
  underlying Theorem~\ref{thm:detection_delay}.

  \item \emph{Component-level to system-level.}
  Applying the union bound over the three monitored components gives
  \[
  \begin{aligned}
  \Pr\!\left(
  \exists\,j\in\{D,H,M\}:
  \rho_t^{(j)}\geq\bar{B}^{(j)}
  \mid\text{baseline}
  \right)
  \leq
  \min\!\left\{
  1,\,
  \frac{1}{2}
  \sum_{j\in\{D,H,M\}}
  \exp\!\left(
  -\frac{
  \bigl(\bar b^{(j)}-\bar z^{(j)}\bigr)^2
  }{
  2v^{(j)}
  }
  \right)
  \right\}.
  \end{aligned}
  \]
  Thus, the system-level false-alarm probability is controlled by the
  sum of the component-level Gaussian tails, paralleling the
  system-level risk aggregation of
  Corollary~\ref{cor:system_level_risk_guarantee}.
  
  \item \emph{Connection to scalar steady-state results.}
The parameters $a^{(j)}$ and $\sigma_{\mu}^{2,(j)}$ appearing in 
Theorem~\ref{thm:false_alarm_rate} can be identified from the steady-state 
Kalman recursion established in Corollary~\ref{cor:scalar_steady_state_variance} 
when the component dynamics are scalar and linear-Gaussian. Specifically, 
for a scalar component, $a^{(j)}$ is the AR(1) coefficient of the latent 
state and $\sigma_{\mu}^{2,(j)}$ is the innovation variance of the posterior 
mean process derived in the proof of Theorem~\ref{thm:detection_delay}.

\end{itemize}


\begin{styledtheorem}{Expected Time-to-Containment}{app:time_to_containment}

Let an excursion of component $j^{\star}\in\{D,H,M\}$ begin at time
$t_{0}$, and let
\(
  \tau_{\rho}
  \;=\;
  \inf\!\left\{
    t\ge t_{0}:
    \rho_{t}^{(j^{\star})}\ge\bar{B}^{(j^{\star})}
  \right\}
\)
be the rating-crossing (detection) time. Define the containment time by
\(
  \tau_{\mathrm{safe}}
  \;=\;
  \inf\!\left\{
    t\ge\tau_{\rho}:
    V(\rho_t)\le\eta
  \right\},
\)
where
\(
  V(\rho)
  =
  \sum_{j\in\{D,H,M\}}
  \bigl(\rho^{(j)}-\bar{B}^{(j)}\bigr)_{+}^{2}
\)
is the Lyapunov risk-excess function and $\eta>0$ is the safety
tolerance.

Assume:
\begin{enumerate}[label=(A\arabic*),ref=(A\arabic*)]

  \item\label{app:ass:detection-delay}
  \emph{(Detection.)}
  The rating-crossing time $\tau_{\rho}$ is a stopping time and its
  expected detection delay is finite, with
  \(
    \mathbb{E}\!\left[\tau_{\rho}-t_{0}\right] \le \frac{\Delta}{c}, \Delta>0, c>0.
  \)
  Here $\Delta$ and $c$ are finite deterministic constants defining an unconditional bound, not a random onset-dependent gap. To obtain this bound from Theorem~\ref{thm:detection_delay} for the same component and onset time, assume its model and responsiveness conditions hold, with fixed $c>0$. Let $A=\{\bar z<\bar b\}$ be the sub-threshold onset event. On $A^c$, the threshold is already reached at onset, so $\tau_\rho=t_0$. If $\Pr(A)=0$, (A1) holds for any $\Delta>0$. Otherwise, choose a finite deterministic $\Delta>0$ with $\Delta\ge\mathbb E[z^\star-\bar z\mid A]$, provided that conditional mean is finite. By the detection-delay theorem and the tower property, $\mathbb E[\tau_\rho-t_0]=\Pr(A)\mathbb E[\tau_\rho-t_0\mid A]\le\Delta/c$. Expectations and probabilities without explicit conditioning in this theorem and its end-to-end corollary are unconditional.

  \item\label{app:ass:active-regulation}
  \emph{(Active and effective regulation from detection onward.)}
  From time $\tau_{\rho}$ onward, the CENTRY regulation policy $c_t$
  is active and satisfies the effective-regulation condition of
  Theorem~\ref{thm:risk_reduction_bound} throughout the containment
  phase. Assume, in addition, that the Lyapunov dissipativity condition of
  Theorem~\ref{thm:asymptotic_risk_stability} holds for every
  $t\ge\tau_{\rho}$:
  \(
    \mathbb{E}\!\left[
      V(\rho_{t+1})-V(\rho_t)
      \,\middle|\,
      \mathcal{F}_t
    \right]
    \le
    -\alpha V(\rho_t)+\epsilon,
    \alpha>0, 
    \epsilon\ge0.
  \)

  \item\label{app:ass:safety-regime}
  \emph{(Contractive safety regime.)}
  The safety tolerance satisfies
  \(
    \alpha\eta>\epsilon.
  \)

\end{enumerate}

Then $\tau_{\mathrm{safe}}$ is finite almost surely and, conditional on
the information available at detection,
\[
  \mathbb{E}\!\left[
    \tau_{\mathrm{safe}}-\tau_{\rho}
    \,\middle|\,
    \mathcal{F}_{\tau_{\rho}}
  \right]
  \le
  \frac{
    V(\rho_{\tau_{\rho}})
  }{
    \alpha\eta-\epsilon
  }
  \qquad\text{a.s.}
\]
Consequently,
\(
  \mathbb{E}\!\left[
    \tau_{\mathrm{safe}}-\tau_{\rho}
  \right]
  \le
  \frac{
    \mathbb{E}\!\left[V(\rho_{\tau_{\rho}})\right]
  }{
    \alpha\eta-\epsilon
  },
\)
and the end-to-end excursion-to-containment time satisfies
\[
  \mathbb{E}\!\left[
    \tau_{\mathrm{safe}}-t_{0}
  \right]
  \le
  \frac{\Delta}{c}
  +
  \frac{
    \mathbb{E}\!\left[V(\rho_{\tau_{\rho}})\right]
  }{
    \alpha\eta-\epsilon
  }.
\]
Thus the expected time from excursion onset to containment decomposes
into a detection term and a post-detection regulation term.

\end{styledtheorem}

\begin{proof}
The detection term follows directly from
Assumption~\ref{app:ass:detection-delay}. It remains to establish the
post-detection containment bound.

\medskip
\noindent
\textbf{Step 1 (re-index time from the detection time).}
Because $\rho_t$ is adapted and $\tau_\rho$ is a stopping time, the first entrance time $\tau_{\mathrm{safe}}$ into the measurable set $\{\rho:V(\rho)\le\eta\}$ at or after $\tau_\rho$ is also a stopping time in discrete time. Let
\[
  \sigma
  =
  \tau_{\mathrm{safe}}-\tau_{\rho}
  \ge0
\]
and define the shifted filtration
\[
  \mathcal{G}_s
  =
  \mathcal{F}_{\tau_{\rho}+s},
  \qquad s\ge0.
\]
Then $\sigma$ is a stopping time for $\{\mathcal G_s\}$. Define the stopped Lyapunov process
\[
  W_s
  =
  V\!\left(
    \rho_{(\tau_{\rho}+s)\wedge\tau_{\mathrm{safe}}}
  \right),
  \qquad s\ge0.
\]
Then
\[
  \sigma
  =
  \inf\{s\ge0:W_s\le\eta\}.
\]
For $s<\sigma$,
\[
  W_s
  =
  V(\rho_{\tau_{\rho}+s})
  >
  \eta,
\]
whereas for $s\ge\sigma$,
\[
  W_s
  =
  V(\rho_{\tau_{\mathrm{safe}}})
  \le\eta.
\]

Under the standing model assumption
$\rho^{(j)},\bar{B}^{(j)}\in[0,1]$, one has
\[
  0\le V(\rho)\le3,
\]
so the stopped process $W_s$ is bounded and therefore integrable.

\medskip
\noindent
\textbf{Step 2 (negative drift outside the safe set).}
For $s<\sigma$, the stopping has not yet occurred, and hence
\[
  W_s
  =
  V(\rho_{\tau_{\rho}+s}).
\]
By Assumption~\ref{app:ass:active-regulation},
\[
\begin{aligned}
\mathbb{E}\!\left[
  W_{s+1}-W_s
  \,\middle|\,
  \mathcal{G}_s
\right]
&=
\mathbb{E}\!\left[
  V(\rho_{\tau_{\rho}+s+1})
  -
  V(\rho_{\tau_{\rho}+s})
  \,\middle|\,
  \mathcal{F}_{\tau_{\rho}+s}
\right]
\\
&\le
-\alpha W_s+\epsilon.
\end{aligned}
\]
Since $W_s>\eta$ for $s<\sigma$,
\[
  -\alpha W_s+\epsilon
  \le
  -\alpha\eta+\epsilon
  =
  -(\alpha\eta-\epsilon).
\]
Define
\[
  \gamma_{\mathrm{drift}}
  :=
  \alpha\eta-\epsilon
  >
  0.
\]
Thus, before containment,
\[
  \mathbb{E}\!\left[
    W_{s+1}-W_s
    \,\middle|\,
    \mathcal{G}_s
  \right]
  \le-\gamma_{\mathrm{drift}}.
\]

\medskip
\noindent
\textbf{Step 3 (construction of a supermartingale).}
Define
\[
  M_s
  =
  W_s+\gamma_{\mathrm{drift}}(s\wedge\sigma).
\]
For $s<\sigma$,
\[
\begin{aligned}
\mathbb{E}\!\left[
  M_{s+1}-M_s
  \,\middle|\,
  \mathcal{G}_s
\right]
&=
\mathbb{E}\!\left[
  W_{s+1}-W_s
  \,\middle|\,
  \mathcal{G}_s
\right]
+\gamma_{\mathrm{drift}}
\\
&\le
-\alpha W_s+\epsilon+\alpha\eta-\epsilon
\\
&=
-\alpha(W_s-\eta)
\\
&\le0.
\end{aligned}
\]
For $s\ge\sigma$, both $W_s$ and $s\wedge\sigma$ remain constant, and
therefore
\[
  M_{s+1}-M_s=0.
\]
Hence $\{M_s\}_{s\ge0}$ is a nonnegative
$\{\mathcal{G}_s\}$-supermartingale.

\medskip
\noindent
\textbf{Step 4 (optional stopping).}
For any integer $N\ge1$, the stopping time $\sigma\wedge N$ is bounded.
Therefore, the optional stopping theorem gives
\[
  \mathbb{E}\!\left[
    M_{\sigma\wedge N}
    \,\middle|\,
    \mathcal{F}_{\tau_{\rho}}
  \right]
  \le
  M_0
  =
  W_0
  =
  V(\rho_{\tau_{\rho}})
  \qquad\text{a.s.}
\]
Since
\[
  M_{\sigma\wedge N}
  =
  W_{\sigma\wedge N}
  +
  \gamma_{\mathrm{drift}}(\sigma\wedge N)
\]
and $W_{\sigma\wedge N}\ge0$, it follows that
\[
  \gamma_{\mathrm{drift}}\,
  \mathbb{E}\!\left[
    \sigma\wedge N
    \,\middle|\,
    \mathcal{F}_{\tau_{\rho}}
  \right]
  \le
  V(\rho_{\tau_{\rho}})
  \qquad\text{a.s.}
\]
Because $\sigma\wedge N\uparrow\sigma$ as $N\to\infty$, the conditional
monotone convergence theorem gives
\[
  \mathbb{E}\!\left[
    \sigma
    \,\middle|\,
    \mathcal{F}_{\tau_{\rho}}
  \right]
  \le
  \frac{
    V(\rho_{\tau_{\rho}})
  }{\gamma_{\mathrm{drift}}}
  =
  \frac{
    V(\rho_{\tau_{\rho}})
  }{\alpha\eta-\epsilon}
  \qquad\text{a.s.}
\]
The right-hand side is finite. Hence
\[
  \mathbb{E}\!\left[
    \sigma
    \,\middle|\,
    \mathcal{F}_{\tau_{\rho}}
  \right]
  <\infty
  \qquad\text{a.s.},
\]
which implies
\[
  \sigma<\infty
  \qquad\text{a.s.}
\]
and therefore
\[
  \tau_{\mathrm{safe}}<\infty
  \qquad\text{a.s.}
\]
Taking expectations yields
\[
  \mathbb{E}[\sigma]
  \le
  \frac{
    \mathbb{E}[V(\rho_{\tau_{\rho}})]
  }{
    \alpha\eta-\epsilon
  }.
\]
Since
$\sigma=\tau_{\mathrm{safe}}-\tau_{\rho}$,
\[
  \mathbb{E}\!\left[
    \tau_{\mathrm{safe}}-\tau_{\rho}
  \right]
  \le
  \frac{
    \mathbb{E}[V(\rho_{\tau_{\rho}})]
  }{
    \alpha\eta-\epsilon
  }.
\]

\medskip
\noindent
\textbf{Step 5 (combine detection and containment).}
Finally,
\[
\begin{aligned}
\mathbb{E}\!\left[
  \tau_{\mathrm{safe}}-t_0
\right]
&=
\mathbb{E}\!\left[
  \tau_{\rho}-t_0
\right]
+
\mathbb{E}\!\left[
  \tau_{\mathrm{safe}}-\tau_{\rho}
\right]
\\
&\le
\frac{\Delta}{c}
+
\frac{
  \mathbb{E}[V(\rho_{\tau_{\rho}})]
}{
  \alpha\eta-\epsilon
}.
\end{aligned}
\]
This proves the stated end-to-end excursion-to-containment bound.
\end{proof}

\noindent\textbf{Remark [Interpretation and scope of the containment guarantee].}
\begin{itemize}[nosep]
  \item \emph{Active regulation is essential.}
  The dissipativity condition
  \[
    \mathbb{E}\!\left[
      V(\rho_{t+1})-V(\rho_t)
      \mid\mathcal{F}_t
    \right]
    \le-\alpha V(\rho_t)+\epsilon
  \]
  describes the regulated closed loop. Its application from
  $t=\tau_{\rho}$ onward therefore requires the CENTRY regulation policy
  $c_t$ to be active and effective after the excursion has been detected.
  This is precisely where the effective-regulation condition of
  Theorem~\ref{thm:risk_reduction_bound} enters. Detection by itself does
  not imply containment.

  \item \emph{Conditional bound is the sharper result.}
  The quantity
  $V(\rho_{\tau_{\rho}})$ is random and is determined by the risk state
  at the moment of detection. The conditional inequality
  \[
    \mathbb{E}\!\left[
      \tau_{\mathrm{safe}}-\tau_{\rho}
      \mid\mathcal{F}_{\tau_{\rho}}
    \right]
    \le
    \frac{V(\rho_{\tau_{\rho}})}
         {\alpha\eta-\epsilon}
  \]
  therefore gives a state-dependent containment guarantee. The
  unconditional bound follows immediately by taking expectations.

  \item \emph{Uniform bound on the detection-time risk excess.}
  If, in addition,
  \[
    V(\rho_{\tau_{\rho}})\le V_{\max}
    \qquad\text{a.s.},
  \]
  then
  \[
    \mathbb{E}\!\left[
      \tau_{\mathrm{safe}}-t_0
    \right]
    \le
    \frac{\Delta}{c}
    +
    \frac{V_{\max}}
         {\alpha\eta-\epsilon}.
  \]
  In particular, if
  $\rho^{(j)},\bar{B}^{(j)}\in[0,1]$ for all three monitored components,
  then $V(\rho)\le3$, and the crude universal bound
  \[
    \mathbb{E}\!\left[
      \tau_{\mathrm{safe}}-t_0
    \right]
    \le
    \frac{\Delta}{c}
    +
    \frac{3}{\alpha\eta-\epsilon}
  \]
  follows.

  \item \emph{Detection--regulation decomposition.}
  The first term,
  $\Delta/c$, quantifies the expected time from excursion onset to
  rating crossing. The second term,
  \[
    \frac{\mathbb{E}[V(\rho_{\tau_{\rho}})]}
         {\alpha\eta-\epsilon},
  \]
  quantifies the expected additional time required for effective
  regulation to drive the system into the safety set. Thus the theorem
  provides a single end-to-end expected excursion-to-containment
  guarantee linking detection with subsequent regulation.

  \item \emph{Role of the safety tolerance.}
  The condition
  \[
    \alpha\eta>\epsilon
  \]
  ensures that the Lyapunov drift is uniformly negative whenever
  $V(\rho)>\eta$. Within this Foster--Lyapunov argument, it is therefore
  the condition that yields a finite expected hitting-time bound.
  It should not be interpreted as a necessary condition for finite-time
  containment under every possible regulated process.

  \item \emph{Relation to asymptotic stability.}
  The theorem strengthens the asymptotic risk-stability result by
  converting the negative-drift condition into an explicit expected
  hitting-time guarantee for a prescribed safety set. The result thus
  quantifies not merely whether the regulated risk remains stable, but
  how long, in expectation, the system takes to enter the specified
  safety region after an excursion has been detected.
\end{itemize}


\begin{styledcorollary}{High-Confidence End-to-End Excursion Resolution Time}{app:end_to_end_confidence}

Under the assumptions of Theorem~\ref{thm:time_to_containment}, for every
$\delta_2\in(0,1)$,
\[
\Pr\!\left(
\tau_{\mathrm{safe}}-\tau_\rho
\le
\frac{V(\rho_{\tau_\rho})}{(\alpha\eta-\epsilon)\delta_2}
\;\middle|\;
\mathcal F_{\tau_\rho}
\right)
\ge1-\delta_2
\qquad\text{a.s.}
\]
Consequently, for every $\delta_1,\delta_2\in(0,1)$ with
$\delta:=\delta_1+\delta_2<1$,
\[
\Pr\!\left(
\tau_{\mathrm{safe}}-t_0
\le
\frac{\Delta}{c\delta_1}
+
\frac{V(\rho_{\tau_\rho})}{(\alpha\eta-\epsilon)\delta_2}
\right)
\ge1-\delta.
\]

\end{styledcorollary}

\begin{proof}
By Theorem~\ref{thm:time_to_containment},
\[
\mathbb E\!\left[\tau_{\mathrm{safe}}-\tau_\rho\mid\mathcal F_{\tau_\rho}\right]
\le
\frac{V(\rho_{\tau_\rho})}{\alpha\eta-\epsilon}
\qquad\text{a.s.}
\]
If $V(\rho_{\tau_\rho})=0$, then
$\tau_{\mathrm{safe}}=\tau_\rho$ by the definition of
$\tau_{\mathrm{safe}}$, so the first claim holds trivially. Otherwise,
$V(\rho_{\tau_\rho})>0$, and the conditional Markov inequality gives, for
any strictly positive $\mathcal F_{\tau_\rho}$-measurable threshold $T$ on this event,
\[
\Pr\!\left(
\tau_{\mathrm{safe}}-\tau_\rho>T
\mid\mathcal F_{\tau_\rho}
\right)
\le
\frac{V(\rho_{\tau_\rho})}
{(\alpha\eta-\epsilon)T}
\qquad\text{a.s.}
\]
Setting
\[
T=\frac{V(\rho_{\tau_\rho})}
{(\alpha\eta-\epsilon)\delta_2}
\]
yields the first claim.

For the second claim, let
\[
A:=\left\{\tau_\rho-t_0>\frac{\Delta}{c\delta_1}\right\}
\]
and
\[
B:=\left\{
\tau_{\mathrm{safe}}-\tau_\rho>
\frac{V(\rho_{\tau_\rho})}
{(\alpha\eta-\epsilon)\delta_2}
\right\}.
\]
Since $\tau_\rho-t_0\ge0$ and $\Delta$ and $c$ are positive deterministic constants under Assumption~\ref{app:ass:detection-delay}, Markov's inequality gives $\Pr(A)\le\frac{c\delta_1}{\Delta}\,\mathbb E[\tau_\rho-t_0]\le\delta_1$. By the
first claim and the tower property,
$\Pr(B)=\mathbb E[\Pr(B\mid\mathcal F_{\tau_\rho})]\le\delta_2$.
By the union bound,
$\Pr(A\cup B)\le\delta_1+\delta_2=\delta$. On
$(A\cup B)^c$,
\[
\tau_{\mathrm{safe}}-t_0
=
(\tau_\rho-t_0)+(\tau_{\mathrm{safe}}-\tau_\rho)
\le
\frac{\Delta}{c\delta_1}
+
\frac{V(\rho_{\tau_\rho})}
{(\alpha\eta-\epsilon)\delta_2},
\]
which gives the claim since $\Pr((A\cup B)^c)\ge1-\delta$.
\end{proof}

\noindent\textbf{Remark [Confidence allocation].}
The total confidence level can be allocated arbitrarily between detection and
post-detection containment. A simple symmetric choice is
$\delta_1=\delta_2=\delta/2$, which yields a single overall confidence
level $1-\delta$ with equal failure-probability allocation to the two phases.

\noindent\textbf{Remark [Composition of conditional confidence guarantees].}
When several high-probability guarantees are combined, their failure
probabilities may be allocated by the union bound, without requiring
independence, provided the component events are stated under compatible
conditioning information (or are first converted to a common conditioning
framework using the tower property). The quantitative confidence cost is
result-dependent: attribution sample complexity is logarithmic in
$1/\delta$, whereas Markov-type detection or containment bounds can scale
inverse-linearly in the allocated failure probability. Thus a composite
confidence budget should preserve the regime and conditioning structure
of each constituent guarantee rather than treating all confidence costs
as interchangeable.


\begin{styledcorollary}{Expected Cumulative Excess-Risk Exposure During Containment}{app:cumulative_exposure}

Under the assumptions of Theorem~\ref{thm:time_to_containment}, the
Lyapunov risk-excess accumulated over the containment window satisfies,
conditional on the information available at detection,
\[
\mathbb E\!\left[\sum_{t=\tau_\rho}^{\tau_{\mathrm{safe}}-1}V(\rho_t)\,\middle|\,\mathcal F_{\tau_\rho}\right]
\leq\frac{\eta\,V(\rho_{\tau_\rho})}{\alpha\eta-\epsilon}
\qquad\text{a.s.},
\]
and hence
\[
\mathbb E\!\left[\sum_{t=\tau_\rho}^{\tau_{\mathrm{safe}}-1}V(\rho_t)\right]
\leq\frac{\eta\,\mathbb E[V(\rho_{\tau_\rho})]}{\alpha\eta-\epsilon}.
\]
This quantity provides an ``excess-risk-time'' measure that combines the
magnitude and duration of a risk excursion.

\end{styledcorollary}

\begin{proof}
Adopt the notation of the proof of Theorem~\ref{thm:time_to_containment}:
$\sigma:=\tau_{\mathrm{safe}}-\tau_\rho$, the shifted filtration
$\mathcal G_s:=\mathcal F_{\tau_\rho+s}$, and
$W_s:=V(\rho_{(\tau_\rho+s)\wedge\tau_{\mathrm{safe}}})$, so
$W_s=V(\rho_{\tau_\rho+s})$ for $s<\sigma$. For $s<\sigma$, the
same Lyapunov drift condition used in that proof gives
\[
\mathbb E[W_{s+1}-W_s\mid\mathcal G_s]\leq-\alpha W_s+\epsilon.
\]
Define
\[
M_s:=W_s+\alpha\sum_{k=0}^{(s\wedge\sigma)-1}W_k-\epsilon(s\wedge\sigma).
\]
For $s<\sigma$,
\[
\mathbb E[M_{s+1}-M_s\mid\mathcal G_s]
=\mathbb E[W_{s+1}-W_s\mid\mathcal G_s]+\alpha W_s-\epsilon\leq0,
\]
while for $s\geq\sigma$, $M_{s+1}=M_s$. Hence $\{M_s\}$ is a
$\{\mathcal G_s\}$-supermartingale.

For every $N$, the stopping time $\sigma\wedge N$ is bounded. Moreover,
$0\leq W_s\leq3$ under $\rho^{(j)},\bar{B}^{(j)}\in[0,1]$, so the
optional stopping theorem gives
\[
\mathbb E[M_{\sigma\wedge N}\mid\mathcal G_0]
\leq M_0=W_0=V(\rho_{\tau_\rho}).
\]
Since $W_{\sigma\wedge N}\geq0$,
\[
\alpha\,\mathbb E\!\left[\sum_{k=0}^{(\sigma\wedge N)-1}W_k\,\middle|\,\mathcal G_0\right]
\leq V(\rho_{\tau_\rho})+\epsilon\,\mathbb E[\sigma\wedge N\mid\mathcal G_0].
\]
Letting $N\to\infty$ and using conditional monotone convergence together
with Theorem~\ref{thm:time_to_containment},
\[
\mathbb E\!\left[\sum_{k=0}^{\sigma-1}W_k\,\middle|\,\mathcal F_{\tau_\rho}\right]
\leq\frac{V(\rho_{\tau_\rho})}{\alpha}
+\frac{\epsilon}{\alpha}\frac{V(\rho_{\tau_\rho})}{\alpha\eta-\epsilon}
=\frac{\eta\,V(\rho_{\tau_\rho})}{\alpha\eta-\epsilon}.
\]
Since the sum on the left is exactly
$\sum_{t=\tau_\rho}^{\tau_{\mathrm{safe}}-1}V(\rho_t)$, the result follows.
Taking expectations yields the unconditional bound.
\end{proof}

\noindent\textbf{Remark [Interpretation of cumulative exposure].}
The bound is exactly $\eta$ times the conditional duration bound of
Theorem~\ref{thm:time_to_containment}. It therefore has the natural
interpretation of an ``excess-risk-time'' exposure measure: under the
Lyapunov drift assumptions, the cumulative quadratic risk violation is
bounded as though the violation level were held at the containment
threshold $\eta$ for the duration allowed by the expected containment
bound. It complements duration by accounting for excursion severity.


\begin{styledtheorem}{Mean-Square Stabilization under Persistent Regulation}{app:posterior_stabilization}

Fix a component \(j\in\{D,H,M\}\). Suppose that from a fixed finite time
\(\tau\) onward an effective regulatory action remains continuously active
and that the resulting latent-state dynamics are
\(
z_{t+1}^{(j)}
=
A^{(j)}z_t^{(j)}+b^{(j)}+\varepsilon_{t+1}^{(j)}\),
\(t\ge\tau.
\)
For this analysis, let $\mathcal G_t$ denote the filtration enlarged to include the latent states through time $t$ (not information available to CENTRY). Assume that
\(
\|A^{(j)}\|_2\le\gamma_j<1,
\)
and \(\{\varepsilon_t^{(j)}\}_{t\ge\tau+1}\) is a martingale-difference
sequence satisfying
\(
\mathbb E[\varepsilon_{t+1}^{(j)}\mid\mathcal G_t]=0\),
\(
\mathbb E[\|\varepsilon_{t+1}^{(j)}\|_2^2\mid\mathcal G_t]\le\sigma_j^2<\infty.
\)
Assume also $\mathbb E\|z_\tau^{(j)}-(I-A^{(j)})^{-1}b^{(j)}\|_2^2<\infty$. A zeroth power is interpreted as one, including for $\gamma_j=0$. Then the regulated dynamics have the unique equilibrium mean
\(
\mu_\infty^{(j)}=(I-A^{(j)})^{-1}b^{(j)},
\)
and, for every \(t\ge\tau\),
\[
\mathbb E\!\left[
\|z_t^{(j)}-\mu_\infty^{(j)}\|_2^2
\,\middle|\,
\mathcal G_\tau
\right]
\le
\gamma_j^{2(t-\tau)}
\|z_\tau^{(j)}-\mu_\infty^{(j)}\|_2^2
+
\sigma_j^2
\sum_{k=0}^{t-\tau-1}\gamma_j^{2k}.
\]
Consequently,
\(
\limsup_{t\to\infty}
\mathbb E\!\left[
\|z_t^{(j)}-\mu_\infty^{(j)}\|_2^2
\,\middle|\,
\mathcal G_\tau
\right]
\le
\frac{\sigma_j^2}{1-\gamma_j^2}.
\)
Thus persistent effective regulation makes the slow latent state
mean-square stable: the effect of the initial condition decays
geometrically, while persistent process noise leaves a bounded residual
dispersion.

If, in addition, the posterior mean is mean-square consistent for the
latent state,
\[
\mathbb E\!\left[
\|\mu_t^{(j)}-z_t^{(j)}\|_2^2
\right]\to0,
\]
then, since
\[
\|\mu_t^{(j)}-\mu_\infty^{(j)}\|_2^2\leq
2\|\mu_t^{(j)}-z_t^{(j)}\|_2^2
+2\|z_t^{(j)}-\mu_\infty^{(j)}\|_2^2,
\]
we have
\[
\limsup_{t\to\infty}
\mathbb E\!\left[
\|\mu_t^{(j)}-\mu_\infty^{(j)}\|_2^2
\right]
\leq
\frac{2\sigma_j^2}{1-\gamma_j^2}.
\]
\end{styledtheorem}

\begin{proof}
Since $\|A^{(j)}\|_2<1$, the spectral radius of $A^{(j)}$ is strictly less than one, so $I-A^{(j)}$ is invertible and the equilibrium mean is uniquely defined. Let
\(
e_t^{(j)}=z_t^{(j)}-\mu_\infty^{(j)}.
\)
Since \((I-A^{(j)})\mu_\infty^{(j)}=b^{(j)}\),
\(
e_{t+1}^{(j)}
=
A^{(j)}e_t^{(j)}+\varepsilon_{t+1}^{(j)}.
\)
The martingale-difference property eliminates the conditional
cross term, giving
\(
\mathbb E[\|e_{t+1}^{(j)}\|_2^2\mid\mathcal G_t]
\le
\gamma_j^2\|e_t^{(j)}\|_2^2+\sigma_j^2.
\)
Iterating conditionally on $\mathcal G_\tau$ yields the finite-time bound. The geometric-series bound gives the limiting conditional inequality; taking expectations in the finite-time bound and using the finite initial second moment also gives the unconditional bound needed below. Finally,
\[
\|\mu_t^{(j)}-\mu_\infty^{(j)}\|_2
\le
\|\mu_t^{(j)}-z_t^{(j)}\|_2
+
\|z_t^{(j)}-\mu_\infty^{(j)}\|_2,
\]
and squaring using $(u+v)^2\le 2(u^2+v^2)$, taking expectations, and applying the additional posterior-consistency assumption transfers the mean-square stabilization statement to the posterior mean.
\end{proof}

\noindent\textbf{Remark [Interpretation].}
In the presence of process noise, persistent regulation should not be
interpreted as convergence of the latent state to a deterministic point.
Rather, it removes dependence on the initial condition geometrically and
confines the regulated state to a stationary neighborhood of the
equilibrium. Posterior concentration around the realized latent state is
a separate property governed by the observation model.


\begin{styledcorollary}{One-Step Contraction of the Risk-Rating Lyapunov Function}{app:one_step_lyapunov_contraction}

Adopt the notation of Theorem~\ref{thm:risk_reduction_bound}, and let
\(
V(\rho_t)=
\sum_{j\in\{D,H,M\}}
\bigl(\rho_t^{(j)}-\bar{B}^{(j)}\bigr)_+^2.
\)
Suppose that, whenever \(V(\rho_t)>0\), the effective-regulation analysis
yields the direct conditional Lyapunov-drift bound
\(
\mathbb{E}\!\left[
V(\rho_{t+1})-V(\rho_t)
\,\middle|\,
\mathcal F_t
\right]
\le
-p_{\mathrm{eff}}B_t+(1-p_{\mathrm{eff}})A,
\)
where
\(
A=\sum_{j\in\{D,H,M\}}\zeta_j\),
\(
B_t=\sum_{j\in\mathcal S_t}\eta_j,
\)
and
\(
\mathcal S_t
=
\{j:\rho_t^{(j)}>\bar{B}^{(j)}\}.
\)
If \(A+B_t>0\) and
\(
p_{\mathrm{eff}}
>
\frac{A}{A+B_t},
\)
then
\(
\mathbb{E}\!\left[
V(\rho_{t+1})
\,\middle|\,
\mathcal F_t
\right]
<
V(\rho_t)
\)
on the event \(\{V(\rho_t)>0\}\).

\end{styledcorollary}

\begin{proof}
On \(\{V(\rho_t)>0\}\), \(\mathcal S_t\neq\emptyset\), so \(B_t>0\).
The assumed drift bound gives
\[
\mathbb{E}\!\left[
V(\rho_{t+1})-V(\rho_t)
\,\middle|\,
\mathcal F_t
\right]
\le
-p_{\mathrm{eff}}B_t+(1-p_{\mathrm{eff}})A
=
A-p_{\mathrm{eff}}(A+B_t).
\]
The stated strict inequality on \(p_{\mathrm{eff}}\) is equivalent to
\(
A-p_{\mathrm{eff}}(A+B_t)<0.
\)
Hence the conditional Lyapunov drift is strictly negative.
\end{proof}

\noindent\textbf{Remark [Relation to the expected-risk-reduction theorem].}
Theorem~\ref{thm:risk_reduction_bound} controls the conditional expected
change in the \emph{sum of component risks}. A negative expected change
in that linear aggregate does not, by itself, imply a negative expected
change in the quadratic positive-part function \(V\). The present
corollary therefore states explicitly the additional Lyapunov-drift
condition needed for the contraction conclusion. This distinction is
important because a component can move while remaining below its budget,
and the squaring and positive-part operations are nonlinear.


\begin{styledcorollary}{Per-Cycle Computational Complexity of SELP}{app:selp_complexity}

Consider the sequential variational implementation described in
Section~\ref{sec:Implementation}. Let
\(
d=\sum_{j\in\{F,D,H,M\}}d_j
\)
denote the total latent dimension, let $h$ denote the hidden dimension
of the inference network, let $d_y$ denote the dimension of the current
interaction $y_t$, and let $\lvert\mathcal C\rvert$ denote the cardinality
of the finite admissible regulatory-action set. Assume that:

\begin{enumerate}[label=(i),nosep]
  \item the inference network maintains its recurrent state rather
  than recomputing the complete interaction history;
  \item the latent covariance is represented with a fixed-dimensional
  representation whose propagation costs $\mathcal O(d^2)$;
  \item the decoder and inference-network evaluations have complexity
  $\mathcal O(hd_y+d^2)$ and $\mathcal O(h^2+hd_y)$, respectively; and
  \item the number of optimization steps performed during a single
  \textsc{Learn} operation is bounded by a fixed positive integer $K\ge1$, with each step having the cost of a constant number of forward/backward evaluations; risk-rating extraction is $\mathcal O(d)$ and action evaluation costs $\mathcal O(1)$ per candidate. Any additional multi-step trajectory or chance-constraint evaluation costs are excluded.
\end{enumerate}

Then one complete SELP cycle has arithmetic complexity
\(
\mathcal O\!\left(
K(h^2+hd_y+d^2)
+
|\mathcal C|
+
d
\right).
\)
For fixed $K$ this can be written as
\(
\mathcal O\!\left(
h^2+hd_y+d^2+|\mathcal C|
\right).
\)
In particular, the per-cycle complexity does not grow with the length
of the interaction history when the recurrent state is maintained
incrementally.

\end{styledcorollary}

\begin{proof}
The \textsc{Predict} operation evaluates the generative model and
requires $\mathcal O(hd_y+d^2)$ arithmetic operations under the stated
architectural assumptions. The \textsc{Surprise} calculation is
$\mathcal O(d_y)$ for a scalar likelihood or KL evaluation and is
therefore dominated by the other terms.

The \textsc{Explain} operation evaluates the inference network once,
requiring $\mathcal O(h^2+hd_y)$ operations. Risk-rating extraction
requires $\mathcal O(d)$ operations.

Evaluation of the admissible regulatory actions requires
$\mathcal O(|\mathcal C|)$ operations. Finally, each optimization step
during \textsc{Learn} has the same asymptotic order as a constant
number of forward and backward evaluations, namely
$\mathcal O(h^2+hd_y+d^2)$, and there are at most $K$ such steps.
Posterior propagation contributes $\mathcal O(d^2)$.

Since $K\ge1$, the costs independent of the number of Learn steps can be absorbed into the $K$-scaled term. Summing these contributions gives
\[
\mathcal O\!\left(
K(h^2+hd_y+d^2)+|\mathcal C|+d
\right),
\]
which reduces to the stated expression when $K$ is a fixed constant.
None of these operations requires revisiting the full interaction
history, so the per-cycle complexity is independent of the history
length.
\end{proof}

\noindent\textbf{Remark [Interpretation].}
The result is a statement about the \emph{incremental per-cycle}
complexity of SELP, not about the total computational cost of running
SELP for $T$ cycles. Under fixed architectural dimensions and fixed
$K$, the latter scales linearly as $\mathcal O(T)$.


\subsection{System-level guarantees}
\label{app:subsec:System-level guarantees}
This subsection combines component-level monitoring and control properties into system-level results, including robust transfer from rating-based to operational-risk safety, policy optimality, Pareto regulation, almost-sure asymptotic budget compliance, robust recursive feasibility under bounded prediction error, eventual feasibility under an explicit margin and absorbing-set condition, and invariance under a properly specified no-intervention policy.


\begin{styledtheorem}{Pareto-Optimal Regulation Frontier}{app:pareto_regulation}

Let
\[
\mathcal{J}:=\{D,H,M\},
\]
and let
\(\Psi\subset\mathbb{R}^{d_\psi}\) be a non-empty compact parameter
space for the regulatory policy \(\pi_\psi\). For each component
\(j\in\mathcal{J}\), define the component risk-excess function
\[
V_j(r)
:=
\left(r-\bar{B}^{(j)}\right)_+^2.
\]
For \(\psi\in\Psi\), let the induced regulatory action be
$c_t=\pi_\psi(\rho_t)$ and define the three expected performance objectives
\[
S(\psi)
:=
\mathbb{E}_\psi[\widehat S_t],
\]
\[
V(\psi)
:=
\sum_{j\in\mathcal{J}}
\mathbb{E}_\psi
\!\left[
V_j\!\left(\rho_t^{(j)}\right)
\right],
\]
and
\[
C(\psi)
:=
\mathbb{E}_\psi[C(c_t)].
\]
Thus, smaller values of all three objectives are preferred. Define the
achievable objective set
\[
\Omega
:=
\left\{
\bigl(S(\psi),V(\psi),C(\psi)\bigr):
\psi\in\Psi
\right\}
\subset\mathbb{R}^3,
\]
and let \(\mathcal{P}\subseteq\Omega\) denote its Pareto frontier:
\[
\mathcal{P}
:=
\left\{
\omega\in\Omega:
\nexists\,\omega'\in\Omega
\text{ such that }
\omega'_k\leq\omega_k\ \forall k,
\quad
\omega'_k<\omega_k
\text{ for some }k
\right\}.
\]

Assume that the mapping
\[
F:\Psi\to\mathbb{R}^3,
\qquad
F(\psi)
=
\bigl(S(\psi),V(\psi),C(\psi)\bigr),
\]
is continuous. Then the following hold.

\begin{enumerate}[nosep]

\item \textbf{Existence of Pareto-optimal operating points.}
The achievable objective set \(\Omega=F(\Psi)\) is non-empty and compact,
and the Pareto frontier \(\mathcal{P}\) is non-empty.

\item \textbf{Scalarization yields Pareto-optimal operating points.}
Let
\[
\lambda_S>0,
\qquad
\lambda_V>0,
\qquad
\lambda_C>0,
\]
and define the weighted-sum objective
\[
J(\psi)
:=
\lambda_S S(\psi)
+
\lambda_V V(\psi)
+
\lambda_C C(\psi).
\]
Because \(\Psi\) is compact and \(J\) is continuous, a global minimizer
\[
\psi^\star
\in
\arg\min_{\psi\in\Psi}J(\psi)
\]
exists. Every such global minimizer induces a Pareto-optimal operating
point
\[
\omega^\star
=
\bigl(
S(\psi^\star),
V(\psi^\star),
C(\psi^\star)
\bigr)
\in\mathcal{P}.
\]
This scalarization result does not imply that every Pareto-optimal
operating point can be obtained by a positive weighted sum; such a
converse generally requires additional structural assumptions, such as
convexity of the achievable objective set.

\item \textbf{First-order stationarity.}
If \(S,V,C\) are each differentiable in a neighborhood of a global minimizer
\(\psi^\star\) and
\[
\psi^\star\in\operatorname{int}(\Psi),
\]
then
\[
\nabla_\psi J(\psi^\star)=0,
\]
that is,
\[
\left.
\left[
\lambda_S\nabla_\psi S(\psi)
+
\lambda_V\nabla_\psi V(\psi)
+
\lambda_C\nabla_\psi C(\psi)
\right]
\right|_{\psi=\psi^\star}
=0.
\]
If \(\psi^\star\) lies on the boundary of \(\Psi\), and if \(J\) is differentiable and, locally around
\(\psi^\star\), the feasible parameter set admits a differentiable
constraint representation
\[
\Psi
=
\left\{
\psi:
g_r(\psi)\leq0,\ r=1,\ldots,m,
\quad
h_s(\psi)=0,\ s=1,\ldots,\ell
\right\},
\]
with an appropriate constraint qualification holding at
\(\psi^\star\), then the Karush--Kuhn--Tucker (KKT) necessary conditions
hold: there exist multipliers
\[
\alpha_r\geq0,
\qquad
\beta_s\in\mathbb{R},
\]
such that
\[
\nabla_\psi J(\psi^\star)
+
\sum_{r=1}^{m}
\alpha_r\nabla g_r(\psi^\star)
+
\sum_{s=1}^{\ell}
\beta_s\nabla h_s(\psi^\star)
=0,
\]
together with
\[
g_r(\psi^\star)\leq0,
\qquad
h_s(\psi^\star)=0,
\]
and
\[
\alpha_r g_r(\psi^\star)=0,
\qquad
r=1,\ldots,m.
\]

\end{enumerate}

\end{styledtheorem}

\begin{proof}
\textbf{Part 1 (existence of Pareto-optimal operating points).}
Since \(\Psi\) is non-empty and compact and \(F\) is continuous,
\[
\Omega=F(\Psi)
\]
is non-empty and compact.

Let
\[
\lambda=(\lambda_1,\lambda_2,\lambda_3)
\in\mathbb{R}_{++}^3
\]
be any vector with strictly positive components and consider
\[
L(\omega):=\lambda^\top\omega,
\qquad
\omega\in\Omega.
\]
Because \(\Omega\) is non-empty and compact and \(L\) is continuous,
there exists a global minimizer
\[
\omega^\star\in\Omega.
\]
Suppose, for contradiction, that
\(\omega^\star\notin\mathcal{P}\). Then there exists
\(\omega'\in\Omega\) such that
\[
\omega'_k\leq\omega^\star_k
\qquad
\forall k\in\{1,2,3\},
\]
with strict inequality for at least one component. Since all components
of \(\lambda\) are strictly positive,
\[
\lambda^\top\omega'
<
\lambda^\top\omega^\star,
\]
contradicting the global minimality of \(\omega^\star\). Hence
\[
\omega^\star\in\mathcal{P},
\]
and therefore
\[
\mathcal{P}\neq\varnothing.
\]

\medskip
\noindent
\textbf{Part 2 (scalarization yields Pareto-optimal operating points).}
Because \(J\) is continuous on the compact set \(\Psi\), a global
minimizer \(\psi^\star\) exists. Define
\[
\omega^\star
=
\bigl(
S(\psi^\star),
V(\psi^\star),
C(\psi^\star)
\bigr).
\]
Suppose, for contradiction, that \(\omega^\star\) is not Pareto-optimal.
Then there exists \(\omega'\in\Omega\) such that
\[
\omega'_k\leq\omega^\star_k
\qquad
\forall k\in\{1,2,3\},
\]
with strict inequality for at least one component. By the definition of
\(\Omega\), there exists \(\psi'\in\Psi\) such that
\[
\omega'
=
\bigl(
S(\psi'),
V(\psi'),
C(\psi')
\bigr).
\]
Since
\(\lambda_S,\lambda_V,\lambda_C>0\),
the Pareto domination implies
\[
\begin{aligned}
J(\psi')
&=
\lambda_S S(\psi')
+
\lambda_V V(\psi')
+
\lambda_C C(\psi')
\\
&<
\lambda_S S(\psi^\star)
+
\lambda_V V(\psi^\star)
+
\lambda_C C(\psi^\star)
\\
&=
J(\psi^\star),
\end{aligned}
\]
which contradicts the global optimality of \(\psi^\star\). Hence
\[
\omega^\star\in\mathcal P.
\]

\medskip
\noindent
\textbf{Part 3 (first-order stationarity).}
Since \(\psi^\star\) is a global minimizer of \(J\) over \(\Psi\), it is
also a local minimizer. If
\[
\psi^\star\in\operatorname{int}(\Psi)
\]
and \(J\) is differentiable in a neighborhood of \(\psi^\star\), the
standard first-order necessary condition gives
\[
\nabla_\psi J(\psi^\star)=0.
\]
Substituting the definition of \(J\) gives
\[
\left.
\left[
\lambda_S\nabla_\psi S(\psi)
+
\lambda_V\nabla_\psi V(\psi)
+
\lambda_C\nabla_\psi C(\psi)
\right]
\right|_{\psi=\psi^\star}
=0.
\]

If \(\psi^\star\) lies on the boundary of \(\Psi\), suppose that,
locally around \(\psi^\star\), the feasible parameter set has the
differentiable representation
\[
\Psi
=
\left\{
\psi:
g_r(\psi)\leq0,\ r=1,\ldots,m,
\quad
h_s(\psi)=0,\ s=1,\ldots,\ell
\right\},
\]
and that an appropriate constraint qualification holds at
\(\psi^\star\). The KKT necessary conditions then imply the existence
of multipliers
\[
\alpha_r\geq0,
\qquad
\beta_s\in\mathbb{R},
\]
such that
\[
\nabla_\psi J(\psi^\star)
+
\sum_{r=1}^{m}
\alpha_r\nabla g_r(\psi^\star)
+
\sum_{s=1}^{\ell}
\beta_s\nabla h_s(\psi^\star)
=0,
\]
together with primal feasibility and complementary slackness.
This completes the proof.

\end{proof}


\begin{styledtheorem}{Almost-Sure Asymptotic Budget Compliance}{app:almost_sure_compliance}

Adopt the setting of Theorem~\ref{thm:asymptotic_risk_stability} and
suppose that the dissipativity condition holds in the homogeneous form
\[
\mathbb{E}\!\left[
  V(\rho_{t+1})
  \,\middle|\,
  \mathcal F_t
\right]
\leq
(1-\alpha)V(\rho_t)
\]
a.s. for every \(t\geq1\), for some \(\alpha\in(0,1]\). In this theorem and its corollaries, a zeroth power is one, including when the base is zero. Suppose also that
\[
V_0:=\mathbb E[V(\rho_1)]<\infty.
\]
Then
\[
\mathbb E[V(\rho_t)]
\leq
(1-\alpha)^{t-1}V_0,
\qquad
t\geq1,
\]
and
\[
V(\rho_t)\longrightarrow0
\qquad\text{almost surely}.
\]
Consequently,
\[
\bigl(\rho_t^{(j)}-\bar{B}^{(j)}\bigr)_+
\longrightarrow0
\qquad\text{a.s.}
\]
for every \(j\in\{D,H,M\}\).

Thus homogeneous dissipativity guarantees asymptotic budget
compliance in the sense that any budget violation converges to zero.
Finite-time permanent entry into the feasible region requires an
additional assumption.

\end{styledtheorem}

\begin{proof}
Taking expectations in the conditional drift inequality gives
\[
\mathbb E[V(\rho_{t+1})]
\leq
(1-\alpha)\mathbb E[V(\rho_t)].
\]
Iteration therefore yields
\[
\mathbb E[V(\rho_t)]
\leq
(1-\alpha)^{t-1}V_0,
\qquad
t\geq1.
\]
Hence
\[
\sum_{t=1}^{\infty}\mathbb E[V(\rho_t)]
\leq
V_0
\sum_{t=1}^{\infty}(1-\alpha)^{t-1}
=
\frac{V_0}{\alpha}
<\infty.
\]
By Tonelli's theorem,
\[
\mathbb E\!\left[
  \sum_{t=1}^{\infty}V(\rho_t)
\right]
=
\sum_{t=1}^{\infty}\mathbb E[V(\rho_t)]
<\infty.
\]
Hence
\[
\sum_{t=1}^{\infty}V(\rho_t)<\infty
\qquad\text{a.s.},
\]
which implies
\[
V(\rho_t)\longrightarrow0
\qquad\text{a.s.}
\]
Since
\[
V(\rho_t)
=
\sum_{j\in\{D,H,M\}}
\bigl(\rho_t^{(j)}-\bar{B}^{(j)}\bigr)_+^2,
\]
each nonnegative summand must converge to zero almost surely.
Therefore
\[
\bigl(\rho_t^{(j)}-\bar{B}^{(j)}\bigr)_+
\longrightarrow0
\qquad\text{a.s.}
\]
for every monitored component.
\end{proof}


\begin{styledcorollary}{L1 Convergence and Uniform Integrability of the Risk-Rating Violation}{app:uniform_integrability}

Under the assumptions of Theorem~\ref{thm:almost_sure_compliance},
\(V(\rho_t)\to0\) almost surely and
\(
\mathbb E[V(\rho_t)]\leq(1-\alpha)^{t-1}V_0\to0
\).
Consequently,
\[
V(\rho_t)\longrightarrow0
\qquad\text{in }L^1,
\]
and the family
\(\{V(\rho_t):t\geq1\}\) is uniformly integrable. In particular,
\[
\lim_{t\to\infty}\mathbb E[V(\rho_t)]
=
\mathbb E\left[\lim_{t\to\infty}V(\rho_t)\right]
=0.
\]
Thus the almost-sure and mean budget-compliance conclusions are
compatible with interchange of the limit and expectation.

\end{styledcorollary}

\begin{proof}
Theorem~\ref{thm:almost_sure_compliance} gives both almost-sure
convergence of $V(\rho_t)$ to zero and convergence of
$\mathbb E[V(\rho_t)]$ to zero. Hence $V(\rho_t)\to0$ in $L^1$.
Every sequence converging in $L^1$ is uniformly integrable, and the
stated interchange of limit and expectation follows immediately.
\end{proof}


\begin{styledcorollary}{Tail Probability of Future Risk-Budget Violations}{app:future_violation_tail}

Under the assumptions of
Theorem~\ref{thm:almost_sure_compliance}, let
\[
V_0:=\mathbb E[V(\rho_1)]<\infty.
\]
Then, for every \(\eta>0\) and every integer \(n\ge1\),
\[
\Pr\!\left(
\sup_{t\ge n}V(\rho_t)\ge\eta
\right)
\leq
\min\!\left\{
1,\,
\frac{V_0}{\eta}
(1-\alpha)^{n-1}
\right\}.
\]
Consequently,
\[
\Pr\!\left(
\sup_{t\ge n}
\max_{j\in\{D,H,M\}}
\bigl(\rho_t^{(j)}-\bar{B}^{(j)}\bigr)_+
\ge\sqrt{\eta}
\right)
\leq
\min\!\left\{
1,\,
\frac{V_0}{\eta}
(1-\alpha)^{n-1}
\right\}.
\]
Thus, under homogeneous dissipativity, the probability of experiencing
any future risk-budget violation exceeding a prescribed tolerance
decays geometrically with the time elapsed under regulation.

\end{styledcorollary}

\begin{proof}
Under the homogeneous dissipativity assumption,
\[
\mathbb E[
V(\rho_{t+1})
\mid
\mathcal F_t
]
\leq
(1-\alpha)V(\rho_t)
\leq
V(\rho_t),
\]
so \(\{V(\rho_t)\}_{t\ge1}\) is a nonnegative supermartingale.

For any finite \(N\ge n\), Doob's maximal inequality gives
\[
\Pr\!\left(
\max_{n\le t\le N}V(\rho_t)\ge\eta
\right)
\leq
\frac{\mathbb E[V(\rho_n)]}{\eta}.
\]
By Theorem~\ref{thm:almost_sure_compliance},
\[
\mathbb E[V(\rho_n)]
\leq
(1-\alpha)^{n-1}V_0.
\]
Therefore,
\[
\Pr\!\left(
\max_{n\le t\le N}V(\rho_t)\ge\eta
\right)
\leq
\frac{V_0}{\eta}
(1-\alpha)^{n-1}.
\]
Since $V(\rho_t)\to0$ almost surely and $\eta>0$, on the probability-one
convergence event any supremum of $V(\rho_t)$ over $t\ge n$ that is at
least $\eta$ is attained at a finite time. Thus, letting \(N\to\infty\) gives
\[
\Pr\!\left(
\sup_{t\ge n}V(\rho_t)\ge\eta
\right)
\leq
\frac{V_0}{\eta}
(1-\alpha)^{n-1}.
\]
Since probabilities cannot exceed one, the stated
\(\min\{1,\cdot\}\) form follows.

Finally,
\[
V(\rho_t)
=
\sum_j
\bigl(
\rho_t^{(j)}-\bar{B}^{(j)}
\bigr)_+^2
\]
implies
\[
\max_j
\bigl(
\rho_t^{(j)}-\bar{B}^{(j)}
\bigr)_+^2
\leq
V(\rho_t).
\]
Hence
\[
\left\{
\sup_{t\ge n}
\max_j
(\rho_t^{(j)}-\bar{B}^{(j)})_+
\ge\sqrt{\eta}
\right\}
\subseteq
\left\{
\sup_{t\ge n}V(\rho_t)\ge\eta
\right\},
\]
which establishes the second inequality.
\end{proof}


\begin{styledcorollary}{High-Probability Eventual Risk-Budget Tolerance}{app:highprob_eventual_tolerance}

Under the assumptions of
Corollary~\ref{cor:future_violation_tail}, let
\(\eta>0\) and \(\delta\in(0,1)\). If \(V_0=0\), the conclusions below hold with $n=1$. Otherwise, if \(0<\alpha<1\) and
\[
n
\ge
1+
\frac{
\log\!\left(
V_0/(\eta\delta)
\right)
}{
-\log(1-\alpha)
},
\]
then
\[
\Pr\!\left(
V(\rho_t)<\eta
\quad
\text{for all }t\ge n
\right)
\ge
1-\delta.
\]
Consequently,
\[
\Pr\!\left(
\bigl(\rho_t^{(j)}-\bar{B}^{(j)}\bigr)_+
<
\sqrt{\eta},
\quad
\forall j\in\{D,H,M\},
\quad
\forall t\ge n
\right)
\ge
1-\delta.
\]

For \(\alpha=1\), the homogeneous dissipativity condition implies
\(V(\rho_t)=0\) almost surely for every \(t\ge2\).

Thus, for any prescribed violation tolerance and confidence level,
CENTRY admits an explicit finite burn-in horizon after which the
risk-budget violation remains below that tolerance for all future
times with probability at least \(1-\delta\).

\end{styledcorollary}

\begin{proof}
If $V_0=0$, the nonnegative violation energy has zero expectation at every time by Theorem~\ref{thm:almost_sure_compliance}, so it is zero almost surely at every time (by a countable intersection). Both claims follow for $n=1$. For $V_0>0$ and $0<\alpha<1$, Corollary~\ref{cor:future_violation_tail} gives
\[
\Pr\!\left(
\sup_{t\ge n}V(\rho_t)\ge\eta
\right)
\leq
\frac{V_0}{\eta}(1-\alpha)^{n-1}.
\]
The stated lower bound on \(n\) is equivalent to
\[
\frac{V_0}{\eta}(1-\alpha)^{n-1}
\leq
\delta.
\]
Therefore,
\[
\Pr\!\left(
\sup_{t\ge n}V(\rho_t)<\eta
\right)
\geq
1-\delta.
\]
Since
\[
\max_j
\bigl(
\rho_t^{(j)}-\bar{B}^{(j)}
\bigr)_+^2
\leq
V(\rho_t),
\]
the component-wise statement follows.

If \(\alpha=1\), then
\[
\mathbb E[
V(\rho_{t+1})\mid\mathcal F_t
]
\leq0.
\]
Because \(V(\rho_{t+1})\ge0\), this implies
\(V(\rho_{t+1})=0\) almost surely for every \(t\ge1\), and hence
\(V(\rho_t)=0\) almost surely for every \(t\ge2\).
\end{proof}


\begin{styledtheorem}{Robust Recursive Feasibility under Bounded Risk-Prediction Error}{app:robust_recursive_feasibility}

Let
\[
\mathcal F_t=\sigma(y_{1:t},c_{1:t-1})
\]
and let $\mathcal C$ denote the admissible regulatory-action set. Fix a
prediction horizon $T\geq1$. Suppose that CENTRY predicts normalized
component risk ratings
\[
\widehat\rho_{t+k\mid t}^{(j)},
\qquad
j\in\{D,H,M\},
\qquad k=1,\ldots,T,
\]
for each candidate action sequence (the dependence of the predictions on that sequence is suppressed in the notation). Assume that the future ratings under application of each candidate sequence satisfy the bounded prediction-error
condition
\[
\rho_{t+k}^{(j)}
\leq
\widehat\rho_{t+k\mid t}^{(j)}+e_k^{(j)}
\]
almost surely, where each known margin $e_k^{(j)}\geq0$ is $\mathcal F_t$-measurable for the candidate sequence. The predictions and margins are evaluated for the particular sequence appearing in the feasible-set definition. The error bound is assumed valid for every admissible history-measurable selection of a candidate sequence. Define
\[
\mathcal A_t^{\mathrm{rob}}
=
\left\{
(c_t,\ldots,c_{t+T-1})\in\mathcal C^T:
\widehat\rho_{t+k\mid t}^{(j)}+e_k^{(j)}
\leq\bar{B}^{(j)},
\ \forall j\in\{D,H,M\},\ \forall k=1,\ldots,T
\right\}.
\]
Suppose that:
\begin{enumerate}[nosep]
\item \textbf{Initial robust feasibility:}
$\mathcal A_1^{\mathrm{rob}}\neq\varnothing$.
\item \textbf{Robust shift-and-append:} whenever
$(c_t,\ldots,c_{t+T-1})\in\mathcal A_t^{\mathrm{rob}}$, after observing
$y_{t+1}$ and updating the posterior there exists a terminal action
$c_{t+T}\in\mathcal C$ such that
$(c_{t+1},\ldots,c_{t+T})\in\mathcal A_{t+1}^{\mathrm{rob}}$
almost surely.
\item \textbf{Receding-horizon implementation:} at each time $t$, CENTRY
selects a sequence in $\mathcal A_t^{\mathrm{rob}}$, applies its first
action, observes $y_{t+1}$, updates its posterior, and resolves the
finite-horizon problem.
\end{enumerate}
Then
\[
\mathcal A_t^{\mathrm{rob}}\neq\varnothing
\qquad\text{almost surely for every }t\geq1,
\]
and along the receding-horizon closed loop every future rating satisfies
\[
\rho_{t+k}^{(j)}\leq\bar{B}^{(j)}
\qquad
j\in\{D,H,M\},\quad k=1,\ldots,T,
\]
almost surely.

\end{styledtheorem}

\begin{proof}
For any sequence in $\mathcal A_t^{\mathrm{rob}}$, the robust constraint gives
\[
\widehat\rho_{t+k\mid t}^{(j)}+e_k^{(j)}\leq\bar{B}^{(j)}.
\]
The bounded prediction-error condition therefore implies
\[
\rho_{t+k}^{(j)}
\leq
\widehat\rho_{t+k\mid t}^{(j)}+e_k^{(j)}
\leq\bar{B}^{(j)}
\]
almost surely for every component and predicted time under execution of the selected sequence.

It remains to establish recursive feasibility. Initial robust feasibility
provides $\mathcal A_1^{\mathrm{rob}}\neq\varnothing$. Suppose that
$\mathcal A_t^{\mathrm{rob}}\neq\varnothing$. CENTRY selects a sequence
in this set and applies its first action. By the robust shift-and-append
assumption, after observing $y_{t+1}$ and updating the posterior, the
shifted sequence augmented by a terminal action belongs to
$\mathcal A_{t+1}^{\mathrm{rob}}$ almost surely. Hence
$\mathcal A_{t+1}^{\mathrm{rob}}\neq\varnothing$ almost surely.
Induction establishes feasibility at all times. For the actual receding-horizon trajectory, apply the one-step error bound to each action actually executed; the countable intersection of these almost-sure one-step safety events gives safety at every subsequent time, hence over every finite future horizon.
\end{proof}
\noindent\textbf{Remark [Indexing of the robust feasibility guarantee].}
The robust recursive-feasibility conclusion is a future-step guarantee:
from a feasible base time $t=1$, it yields
$V(\rho_t)=0$ for $t\ge2$. A conclusion beginning at $t=1$ additionally
requires initial feasibility, $V(\rho_1)=0$. This robust result is
stronger in its safety conclusion than the almost-sure eventual-feasibility
result, but it rests on different worst-case prediction-error and
recursive-feasibility assumptions; it is not a consequence of the
eventual-feasibility theorem alone.

\noindent\textbf{Remark [Calibration of the robust error margin].}
The margins $e_k^{(j)}$ are treated here as known, almost-sure prediction
error bounds. This is an idealized robust-control assumption: in practice,
the margins must themselves be estimated or learned from validation data,
calibration procedures, conformal or other uncertainty-quantification
methods, or conservative domain knowledge. The recursive-feasibility and
safety conclusions are therefore conditional on the stated margins being
valid bounds; underestimation of $e_k^{(j)}$ can invalidate the guarantee.
Extending the result to uncertain or estimated error margins would require
an additional analysis accounting for both prediction error and uncertainty
in the error bound itself.


\begin{styledcorollary}{Operational-Risk Safety Under Robust Recursive Feasibility}{app:robust_operational_safety}

Adopt the setting and conclusion of Theorem~\ref{thm:robust_recursive_feasibility},
so that a robustly feasible action sequence is selected at every $t$ and
\[
\rho_{t+k}^{(j)}\leq\bar{B}^{(j)}
\qquad\text{almost surely}
\]
along the receding-horizon closed loop for every fixed $t\ge1$, $j\in\{D,H,M\}$, and $k=1,\ldots,T$. Fix such a time $t$. Suppose in addition, for each
$j,k$, that the dangerous-false-negative condition of
Theorem~\ref{thm:rating_to_operational_safety} holds:
\[
\Pr\!\left(
R_{t+k}^{(j)}>B^{(j)},
\rho_{t+k}^{(j)}\leq\bar{B}^{(j)}\mid\mathcal F_t
\right)
\leq\eta_k^{(j)}.
\]
Then
\[
\Pr\!\left(R_{t+k}^{(j)}>B^{(j)}\mid\mathcal F_t\right)
\leq\eta_k^{(j)}
\qquad\text{almost surely},
\]
with no contribution from a rating-overshoot term. Consequently, if in
addition
\[
\sum_{k=1}^T\sum_{j\in\{D,H,M\}}\eta_k^{(j)}\leq\delta,
\]
then
\[
\Pr\!\left(
\bigcap_{k=1}^T\bigcap_{j\in\{D,H,M\}}
\left\{R_{t+k}^{(j)}\leq B^{(j)}\right\}
\ \middle|\ \mathcal F_t
\right)
\geq1-\delta.
\]
Thus, robust recursive feasibility removes the rating-overshoot contribution
from the rating-to-operational-risk transfer bound; the remaining safety
uncertainty is entirely due to the calibrated dangerous-false-negative rate.

\end{styledcorollary}

\begin{proof}
By Theorem~\ref{thm:robust_recursive_feasibility},
$\rho_{t+k}^{(j)}\leq\bar{B}^{(j)}$ almost surely, so the event
\[
A:=\left\{\rho_{t+k}^{(j)}>\bar{B}^{(j)}\right\}
\]
has conditional probability zero given $\mathcal F_t$ almost surely. Let
$E:=\{R_{t+k}^{(j)}>B^{(j)}\}$. The tautological inclusion
$E\subseteq A\cup(E\cap A^c)$ gives
\[
\begin{aligned}
\Pr(E\mid\mathcal F_t)
&\leq
\Pr(A\mid\mathcal F_t)
+\Pr(E\cap A^c\mid\mathcal F_t)\\
&=0+\Pr(E\cap A^c\mid\mathcal F_t)\\
&\leq\eta_k^{(j)},
\end{aligned}
\]
where the last inequality is the dangerous-false-negative condition of
Theorem~\ref{thm:rating_to_operational_safety}. This proves the first claim.

For the joint claim, define
\[
E_k^{(j)}=\left\{R_{t+k}^{(j)}>B^{(j)}\right\}.
\]
The complement of the joint safety event is
\[
\bigcup_{k=1}^T\bigcup_{j\in\{D,H,M\}}E_k^{(j)}.
\]
The conditional union bound therefore gives
\[
\begin{aligned}
\Pr\!\left(
\bigcup_{k=1}^T\bigcup_{j\in\{D,H,M\}}E_k^{(j)}
\ \middle|\ \mathcal F_t
\right)
&\leq
\sum_{k=1}^T\sum_{j\in\{D,H,M\}}
\Pr(E_k^{(j)}\mid\mathcal F_t)\\
&\leq
\sum_{k=1}^T\sum_{j\in\{D,H,M\}}\eta_k^{(j)}\\
&\leq\delta.
\end{aligned}
\]
Taking complements proves the stated joint finite-horizon safety guarantee.
\end{proof}


\begin{styledcorollary}{Almost-Sure Eventual Feasibility}{app:eventual_feasibility}

Under the assumptions of Theorem~\ref{thm:almost_sure_compliance}, suppose
in addition that there exists a constant $\varepsilon_{\mathrm{feas}}>0$ such that every
violation satisfies
\[
V(\rho_t)>0
\quad\Longrightarrow\quad
V(\rho_t)\geq\varepsilon_{\mathrm{feas}}
\qquad\text{a.s.}
\]
Then there exists a finite random time
\(T_{\mathrm{safe}}\) such that
\[
V(\rho_t)=0
\qquad
\text{for all }t\geq T_{\mathrm{safe}}
\]
almost surely. Equivalently,
\[
\rho_t^{(j)}
\leq
\bar{B}^{(j)},
\qquad
\forall j\in\{D,H,M\},
\quad
\forall t\geq T_{\mathrm{safe}},
\]
almost surely.

Thus, under a positive minimum violation margin $\varepsilon_{\mathrm{feas}}$, homogeneous
Lyapunov dissipativity guarantees not merely asymptotic decay of the
risk excess but finite-time permanent entry into the feasible region
almost surely.

\end{styledcorollary}

\begin{proof}
By Theorem~\ref{thm:almost_sure_compliance},
\[
V(\rho_t)\longrightarrow0
\qquad\text{a.s.}
\]
The convergence event and the violation-margin events for all integer times have a probability-one intersection. On this event, let $T_{\mathrm{safe}}$ be the first integer time from which $V(\rho_t)<\varepsilon_{\mathrm{feas}}$ holds permanently; it is finite by convergence and measurable by countability of the time index (define it arbitrarily on the null complement). Thus
\[
V(\rho_t)<\varepsilon_{\mathrm{feas}}
\qquad
\text{for all }t\geq T_{\mathrm{safe}}.
\]
By the assumed violation-margin condition, \(V(\rho_t)>0\) would imply
\(V(\rho_t)\geq\varepsilon_{\mathrm{feas}}\), which is impossible once
\(V(\rho_t)<\varepsilon_{\mathrm{feas}}\). Therefore
\[
V(\rho_t)=0
\qquad
\text{for all }t\geq T_{\mathrm{safe}},
\]
almost surely. This is equivalent to
\[
\rho_t^{(j)}\leq\bar{B}^{(j)}
\]
for every monitored component and every
\(t\geq T_{\mathrm{safe}}\).
\end{proof}


\begin{styledtheorem}{Invariance of the Feasible Design Region under Deterministic No Intervention}{app:feasible_invariance}

Let
\[
\mathcal F_{\mathrm{feas}}
=
\left\{
\rho\in[0,1]^3:
\rho^{(j)}\le\bar{B}^{(j)}
\ \text{for all }j\in\{D,H,M\}
\right\}
\]
denote the feasible design region. Suppose that at time $t_0$,
\(
\rho_{t_0}\in\mathcal F_{\mathrm{feas}}.
\)
Assume that the no-intervention action
\(
c_t=c^{(0)}
\)
for all $t\ge t_0$ retains the current configuration and, in the setting of this theorem, induces a deterministic risk-rating transition
\(
\rho_{t+1}
=
\Phi^{(0)}(\rho_t),
\)
and that the feasible region is invariant under this transition:
\(
\Phi^{(0)}(\mathcal F_{\mathrm{feas}})\subseteq\mathcal F_{\mathrm{feas}}.
\)
Then
\(
\rho_t\in\mathcal F_{\mathrm{feas}}
\) for all $t\ge t_0$.
Equivalently,
\(
\rho_t^{(j)}\le\bar{B}^{(j)}
\)
for all $j\in\{D,H,M\}$
and all $t\ge t_0$.
Thus, once entered, the feasible design region is invariant under the
pure no-intervention policy whenever the induced deterministic
risk-rating dynamics preserve that region.

\end{styledtheorem}

\begin{proof}
The assumption $\rho_{t_0}\in\mathcal F_{\mathrm{feas}}$ establishes the claim at
$t=t_0$. Suppose inductively that $\rho_t\in\mathcal F_{\mathrm{feas}}$. By the
invariance assumption,
\[
\rho_{t+1}
=
\Phi^{(0)}(\rho_t)
\in
\Phi^{(0)}(\mathcal F_{\mathrm{feas}})
\subseteq
\mathcal F_{\mathrm{feas}}.
\]
Induction therefore gives $\rho_t\in\mathcal F_{\mathrm{feas}}$ for every
$t\ge t_0$.
\end{proof}


\begin{styledcorollary}{Sufficient Condition for No-Intervention Invariance}{app:no_intervention_invariance_sufficient}

Suppose that $\rho_{t_0}\in\mathcal F_{\mathrm{feas}}$ and $c_t=c^{(0)}$ for all $t\ge t_0$, and suppose
that under the no-intervention action the posterior mean and posterior
covariance of every monitored component remain unchanged for every $t\geq t_0$, where $t_0$ is the activation time of the no-intervention policy:
\[
\mu_{t+1}^{(j)}
=
\mu_t^{(j)},
\qquad
\Sigma_{t+1}^{(j)}
=
\Sigma_t^{(j)},
\qquad
j\in\{D,H,M\}.
\]
Then
\[
\rho_{t+1}^{(j)}
=
\rho_t^{(j)},
\qquad
j\in\{D,H,M\},
\]
and hence $\rho_t\in\mathcal F_{\mathrm{feas}}$ for all $t\ge t_0$; that is, the trajectory remains in the feasible design region under the no-intervention policy.

\end{styledcorollary}

\begin{proof}
By Eq.~\eqref{eq:riskfunctional},
\[
\rho_{t+1}^{(j)}
=
g^{(j)}\!\left(
\mu_{t+1}^{(j)},\Sigma_{t+1}^{(j)}
\right).
\]
Under the stated assumptions,
\[
\begin{aligned}
\rho_{t+1}^{(j)}
&=
g^{(j)}\!\left(
\mu_t^{(j)},\Sigma_t^{(j)}
\right)
\\
&=
\rho_t^{(j)}.
\end{aligned}
\]
Therefore
\[
\rho_{t+1}=\rho_t.
\]
Consequently, if
\(\rho_{t_0}\in\mathcal F_{\mathrm{feas}}\), then
\(\rho_t\in\mathcal F_{\mathrm{feas}}\) for every \(t\geq t_0\), as required.
\end{proof}


\newpage
\section*{About the Authors}

\begin{wrapfigure}{l}{0.25\textwidth}
\vspace{-0.4cm}
\includegraphics[width=1\linewidth]{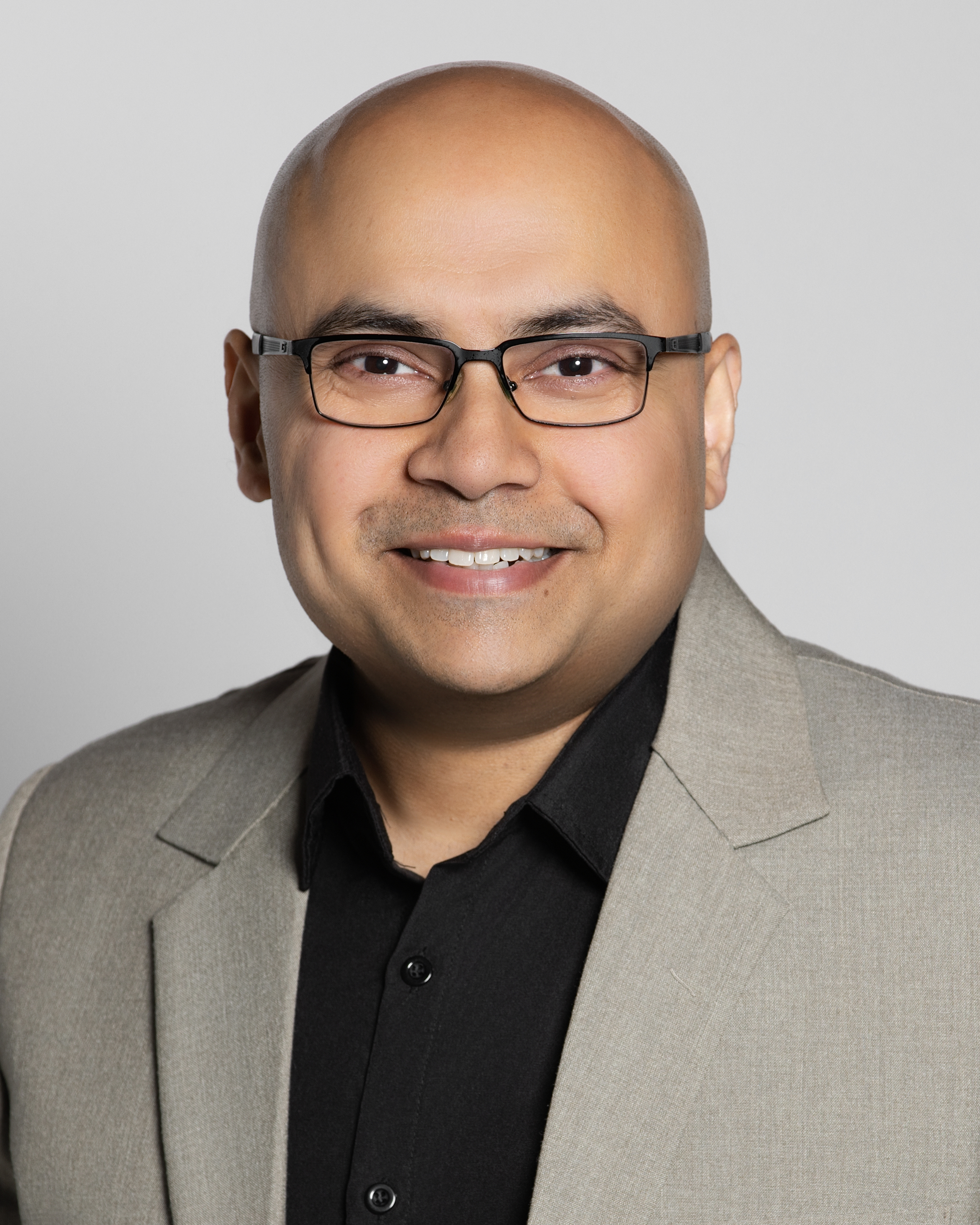} 
\end{wrapfigure}
\textbf{Bonny Banerjee} received M.S. in Electrical Engineering and Ph.D. in Computer Science (major AI, minor Cognitive Science) from the Ohio State University, USA. Just after graduating with Ph.D., he spent 3.5 years (2007-2011, through the Great Recession) leading the research at a startup, which resulted in a number of patents, substantial investor funding, and launch of a commercial product for the end-user which received wide media coverage. The intellectual property was acquired by the leading company in the field. For the last 15 years, he has served as a faculty at institutions of higher education, and as a consultant for small and large businesses in the U.S. and abroad. Dr. Banerjee holds 8 patents and has published over 75 peer-reviewed articles in reputed journals and conference proceedings on AI agents and related areas. He received the Best Paper Award at a NeurIPS 2023 Workshop. He has served as the Principal Investigator of research projects funded by the U.S. National Science Foundation, Department of Homeland Security, Army, St. Jude Children's Research Hospital, and startup investors. He serves on the editorial boards of \textit{IEEE Transactions on Cybernetics} (IEEE), \textit{Neural Networks} (Elsevier), and \textit{Brain Theory and Intelligence Science} (Springer Nature) journals. 
\\

\begin{wrapfigure}{l}{0.25\textwidth}
\vspace{-0.4cm}
\includegraphics[width=1\linewidth]{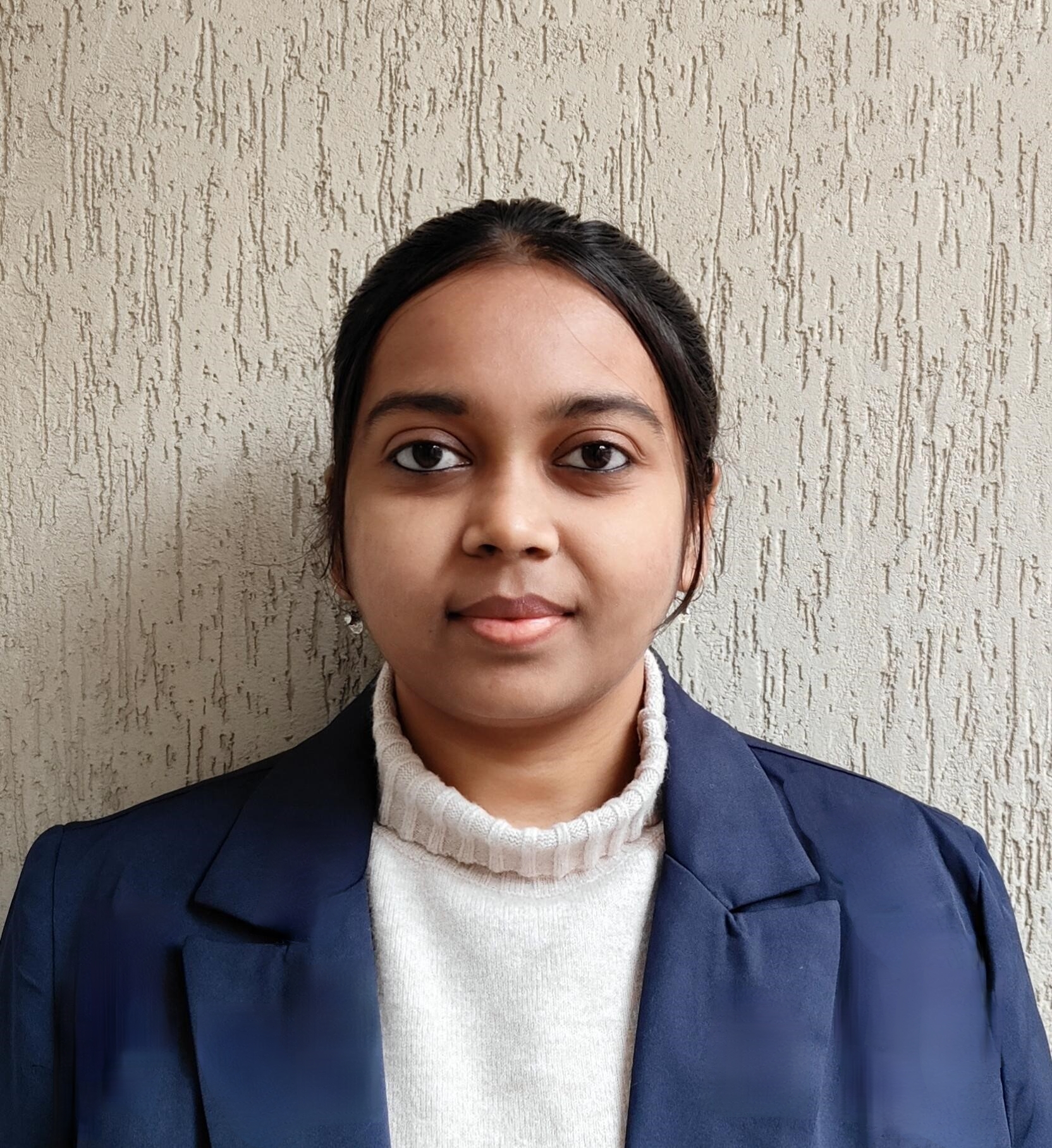} 
\end{wrapfigure}
\noindent\textbf{Geetha Raju} is the Principal Policy Analyst for AI and Data Science at the Centre for Responsible AI (CeRAI), Indian Institute of Technology (IIT) Madras, where she leads research, policy advisory, and strategic initiatives at the intersection of AI, society, and policy. Holding a Ph.D. in Information and Communication Engineering, specialising in AI and User Data Privacy, from Anna University, Dr. Geetha brings a strong technical foundation and an interdisciplinary socio-techno-legal perspective to AI governance. Building on her experience in AI research and technology implementation, she has transitioned into AI policy research, working to bridge the gap between technological advancements and policy and governance frameworks. Her work focuses on AI safety evaluations, impact assessments, and human-centric and responsible AI governance, bringing together technical research, socio-technical analysis, and policy approaches. She works on assessing AI risks and harms, understanding their implications for individuals and society, strengthening accountability, and developing practical governance tools and frameworks for the safe and responsible adoption of AI in India and the Global South. She has contributed to international AI policy discourse as a research group member at the Centre for AI and Digital Policy (CAIDP), engaging with global AI governance principles, standards, and policy developments.

Prior to joining CeRAI, Dr. Geetha held key roles in state and central government initiatives, including the Tamil Nadu e-Governance Agency (TNeGA) and the IndiaAI Mission at the Ministry of Electronics and Information Technology (MeitY). She contributed to shaping AI strategies, advancing emerging technology initiatives, and developing AI-powered e-governance applications to improve public service delivery and governance. Recognised for her contributions to the field, Dr. Geetha was featured in the INDIAai Expert Talk series and is a speaker at national and international forums on responsible AI, AI governance, and the societal implications of emerging technologies. She also has a U.S. patent granted, published journal articles, thought leadership pieces, and discussion papers in the AI governance space, contributing to research and policy discourse on the responsible development, deployment, and governance of AI.

\end{document}